\documentclass{article}

\PassOptionsToPackage{sort, numbers, compress}{natbib}
\usepackage[final]{neurips_2021}

\usepackage[dvipsnames]{xcolor}
\definecolor{darkb}{rgb}{0.2,0.2,0.8}
\usepackage[utf8]{inputenc} % allow utf-8 input
\usepackage[T1]{fontenc}    % use 8-bit T1 fonts
\usepackage[hidelinks,colorlinks = true,
            linkcolor = darkb,
            urlcolor  = darkb,
            citecolor = darkb,
            anchorcolor = darkb]{hyperref}
\usepackage{url}            % simple URL typesetting
\usepackage{booktabs}       % professional-quality tables
\usepackage{amsfonts}       % blackboard math symbols
\usepackage{nicefrac}       % compact symbols for 1/2, etc.
\usepackage{microtype}      % microtypography
\usepackage{xcolor}         % colors

\title{Bayesian Image Reconstruction using Deep Generative Models}

\author{%
  Razvan V.~Marinescu\\%\thanks{Use footnote for providing further information about author (webpage, alternative address)---\emph{not} for acknowledging funding agencies.} \\
  MIT CSAIL\\
  \texttt{razvan@csail.mit.edu} \\
  \And
  Daniel Moyer \\
  MIT CSAIL \\
  \texttt{dmoyer@csail.mit.edu} \\
  \AND
  Polina Golland \\
  MIT CSAIL \\
  \texttt{polina@csail.mit.edu} \\
}

\usepackage{filecontents}

\usepackage{tabu} % different font size for different table rows
\usepackage{amsmath}
\usepackage{amssymb}
\usepackage{mathtools}
\usepackage{colortbl} 
\usepackage[english]{babel}
\usepackage[utf8]{inputenc}
\usepackage{latexsym,amstext,amsfonts,graphicx,setspace,bm}
\usepackage[colorinlistoftodos]{todonotes}
\usepackage{enumerate} 
\usepackage{paralist}
\usepackage{float}
\usepackage{subcaption}
\usepackage{array}
\usepackage{comment}
\usepackage{mathtools}
\usepackage[utf8]{inputenc}
\usepackage[T1]{fontenc}
\usepackage[thinc]{esdiff}

\usepackage{mathtools}
\usepackage[utf8]{inputenc}
\usepackage[T1]{fontenc}
\usepackage[thinc]{esdiff}
\usepackage{anyfontsize}
\usepackage{placeins}

\usepackage{enumitem}

\newcommand\numberthis{\addtocounter{equation}{1}\tag{\theequation}}

\DeclarePairedDelimiterX{\infdivx}[2]{(}{)}{%
  #1\;\delimsize\|\;#2%
}

\DeclarePairedDelimiter{\norm}{\lVert}{\rVert}

\usepackage{tikz}
\usetikzlibrary{bayesnet}
\usetikzlibrary{arrows, positioning}
\usetikzlibrary{fit,positioning}
\usetikzlibrary{calc}

\tikzset{>=latex}

\usepackage{pgfplots}

\usetikzlibrary{intersections, pgfplots.fillbetween}
\definecolor{nice_blue}{RGB}{65, 105, 225}
\definecolor{nice_red}{RGB}{168, 34, 34}
\definecolor{IGCBlue}{HTML}{16197A}
\definecolor{dark_green}{RGB}{20, 110, 10}

\definecolor{graph_blue}{RGB}{144, 195, 212}
\definecolor{graph_purple}{RGB}{195, 144, 212}
\definecolor{graph_green}{RGB}{161, 212, 144}
\definecolor{graph_orred}{RGB}{212, 161, 144}

\DeclareMathOperator*{\argmin}{arg\,min}
\DeclareMathOperator*{\argmax}{arg\,max}

\definecolor{dark_green}{RGB}{20, 110, 10}
\definecolor{dark_blue}{RGB}{20, 10, 120}

\definecolor{LightGray}{rgb}{0.85,0.85,0.85}
\newcommand{\fld}{results}
\newcommand{\valid}{validation}

\newcommand{\inc}[1]{\raisebox{-.4\height}{\includegraphics[height=\w]{#1}}}

\newcommand{\w}{2.5cm}

\newcommand{\mc}[1]{\mathbb{#1}}

\newcommand{\ci}[1]{\circ{#1}}

\newcommand{\modifcolor}{black}

\newcommand{\tc}[1]{\textcolor{\modifcolor}{#1}}

\newcommand{\pulse}{PULSE \cite{menon2020pulse}}
\newcommand{\esrgan}{ESRGAN \cite{wang2018esrgan}}
\newcommand{\srfbn}{SRFBN \cite{li2019feedback}}
\newcommand{\patchgan}{SN-PatchGAN \cite{yu2019free}}

\newcommand{\brgm}{\textbf{BRGM}}
\newcommand{\dip}{Deep Image Prior \cite{ulyanov2018deep}} 

\newcommand{\wplus}{$\mathcal{W}^{+}$ }
\newcommand{\loss}{\mathcal{L}}

\begin{document}

\maketitle

\begin{abstract}
Machine learning models are commonly trained end-to-end and in a supervised setting, using paired (input, output) data. Examples include recent super-resolution methods that train on pairs of (low-resolution, high-resolution) images. However, these end-to-end approaches require re-training every time there is a distribution shift in the inputs (e.g., night images vs daylight) or relevant latent variables (e.g., camera blur or hand motion). In this work, we leverage state-of-the-art (SOTA) generative models (here StyleGAN2) for building powerful image priors, which enable application of Bayes' theorem for many downstream reconstruction tasks. Our method, \emph{Bayesian Reconstruction through Generative Models} (BRGM), uses a single pre-trained generator model to solve different image restoration tasks, i.e., super-resolution and in-painting, by combining it with different forward corruption models. We keep the weights of the generator model fixed, and reconstruct the image by estimating the Bayesian maximum a-posteriori (MAP) estimate over the input latent vector that generated the reconstructed image. We further use Variational Inference to approximate the posterior distribution over the latent vectors, from which we sample multiple solutions. We demonstrate BRGM on three large and diverse datasets: (i) 60,000 images from the Flick Faces High Quality dataset \cite{karras2019style} (ii) 240,000 chest X-rays from MIMIC III \cite{johnson2016mimic} and (iii) a combined collection of 5 brain MRI datasets with 7,329 scans \cite{dalca2018anatomical}. Across all three datasets and without any dataset-specific hyperparameter tuning, our simple approach yields performance competitive with current task-specific state-of-the-art methods on super-resolution and in-painting, while being more generalisable and without requiring any training. Our source code and pre-trained models are available online: \url{https://razvanmarinescu.github.io/brgm/}
\end{abstract}

\section{Introduction}

% end-to-end learning problems: disitributon shifts, and combinatorial explosion.
While end-to-end supervised learning is currently the most popular paradigm in the research community, it suffers from several problems. First, distribution shifts in the inputs often require re-training, as well as the effort of collecting an updated dataset. In some settings, such shifts can occur often (hospital scanners are upgraded) and even continuously (population is slowly aging due to improved healthcare).  Secondly, current state-of-the-art machine learning (ML) models often require prohibitive computational resources, which are only available in a select number of companies and research centers. Therefore, the ability to leverage \emph{pre-trained} models for solving downstream prediction or reconstruction tasks becomes crucial. %For example, instead of training a method to perform super-resolution on human faces, one can use a pre-trained face generator combined with a downsampling corruption model to create a super-resolution method that does not require any further training or fine-tuning. 

\newcommand{\diagfld}{images/diagram}
\newcommand{\fstikz}[1]{\footnotesize{#1}}

\newcommand{\figscl}{0.8}

\begin{figure*}
 \centering
     \begin{subfigure}[t]{0.35\textwidth}
         \centering
\begin{tikzpicture}[scale=\figscl, every node/.style={scale=\figscl}]

\def\rightimgX{4.5}

%top image center
\node (cxr_img) at (1, 3.5) {\includegraphics[width=2cm]{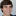}};
\node (cxr_text) at (1,4.75) {\footnotesize{Low Res.}};

%top image left
\node (blur_img) at (\rightimgX, 3.5) {\includegraphics[width=2cm]{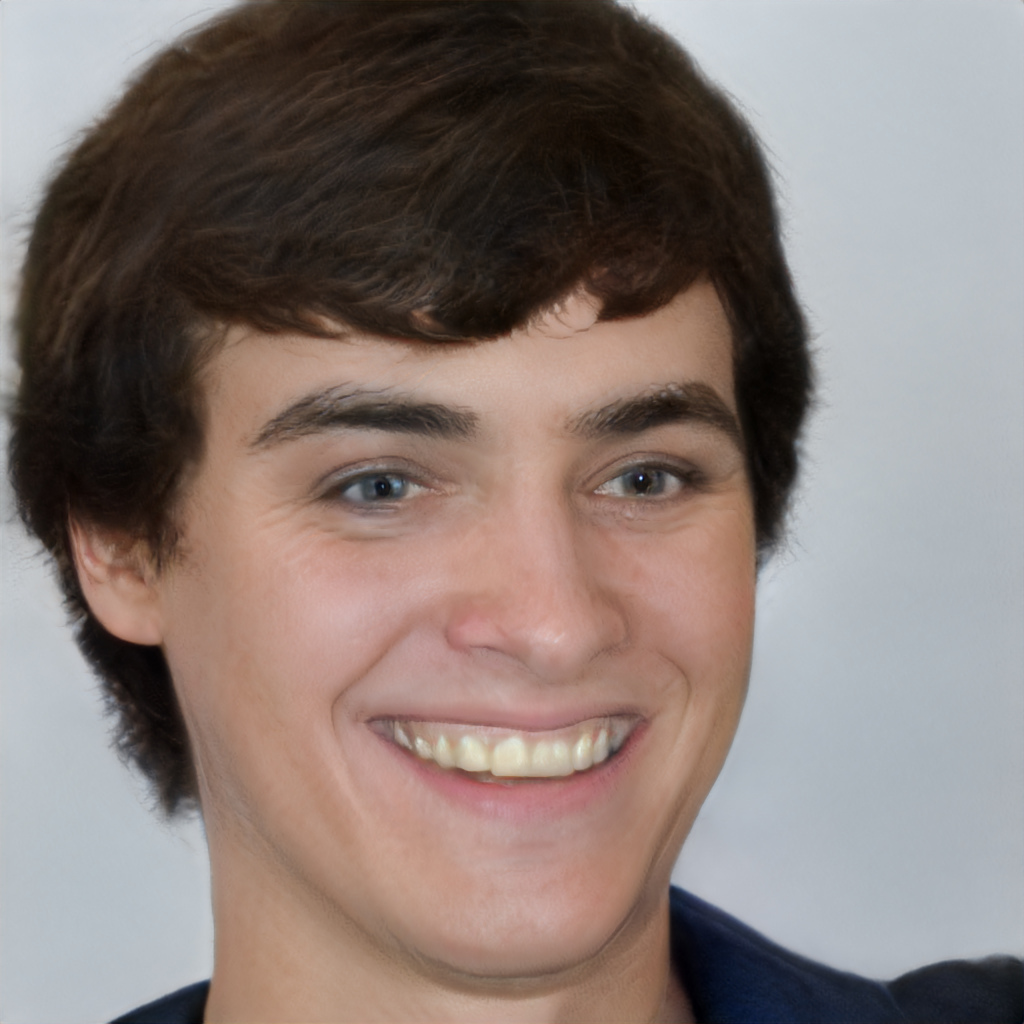}};
\node (blur_text) at (\rightimgX,4.75) {\footnotesize{High Res.}};

%top image center
\node (cxr_img) at (1, 1) {\includegraphics[width=2cm]{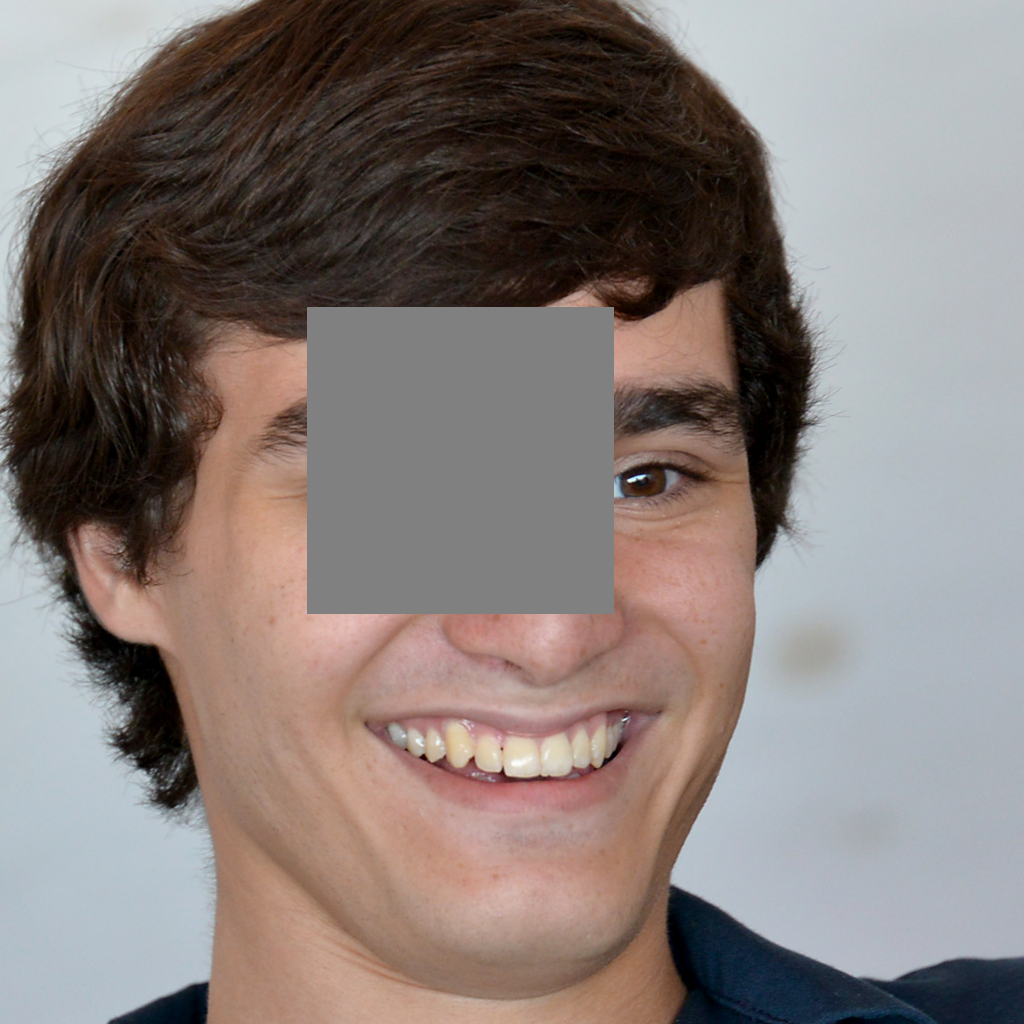}};
%top image left
\node (blur_img) at (\rightimgX, 1) {\includegraphics[width=2cm]{\diagfld/original.png}};

\node (left_img_top_out) at (2,3.5) {};
\node (right_img_top_in) at (\rightimgX - 1, 3.5) {};

\draw[->, thick]
(left_img_top_out)
to
(right_img_top_in) ;

\node (left_img_top_out_2) at (2,1) {};
\node (right_img_top_in_2) at (\rightimgX - 1, 1) {};

\draw[->, thick]
(left_img_top_out_2)
to
(right_img_top_in_2) ;

\node (left_img_top_in) at (2,2.5) {};
\node (right_img_top_out) at (\rightimgX - 1, 2.5) {};

\node (rng_text) at (2.75,3.75) {$f_1^{-1}(I)$};
\node (rng_text) at (2.75,3.25) {\fstikz{Learned}};
\node (rng_text) at (2.75,3) {\fstikz{Inverse}};
\node (rng_text) at (2.75,2.75) {\fstikz{Corruption}};

\node (rng_text) at (2.75,1.25) {$f_2^{-1}(I)$};

\end{tikzpicture}
         \caption{Task-specific methods}
         \label{diagram-prev}
     \end{subfigure}
     \begin{subfigure}[t]{0.6\textwidth}
         \centering
\begin{tikzpicture}[scale=\figscl, every node/.style={scale=\figscl}]

\def\rightimgX{4.5}

%Top left
\filldraw[fill=gray!50!white, draw=black, label={Latent}] (-2,2.5) rectangle (-1.5,4.5);
\node (rng_text) at (-1.75,4.75) {\footnotesize{Latent}};
\node (rng_text) at (-1.75,3.5) {$w$};

%top image center
\node (cxr_img) at (1, 3.5) {\includegraphics[width=2cm]{\diagfld/original.png}};
\node (cxr_text) at (1,4.75) {\footnotesize{High Res.}};

%top image right
\node (blur_img) at (\rightimgX, 3.5) {\includegraphics[width=2cm]{\diagfld/blur.png}};
\node (blur_text) at (\rightimgX,4.75) {\footnotesize{Low Res.}};

\node (blur_img) at (\rightimgX+2.3, 3.5) {\includegraphics[width=2cm]{\diagfld/blur.png}};
\node (blur_text) at (\rightimgX+2.3,4.75) {\footnotesize{Input}};

\draw[decoration={brace,mirror,raise=5pt},decorate]
  (4.5,2.5) -- node[below=6pt] {loss} (6.8,2.5);

\node (rng_top_out) at (-1.5,3.5) {};
\node (center_img_left_in) at (0, 3.5) {};

\draw[->, thick]
(rng_top_out)
to
(center_img_left_in) ;

\node (left_img_top_out) at (2,3.5) {};
\node (right_img_top_in) at (\rightimgX - 1, 3.5) {};

\draw[->, thick]
(left_img_top_out)
to
(right_img_top_in) ;

\node (left_img_top_in) at (2,2.5) {};
\node (right_img_top_out) at (\rightimgX - 1, 2.5) {};

% \draw[->, dashed, blue, thick]
% (right_img_top_out)
% to [in=-60, out=-120]
% (left_img_top_in) ;

\node (left_img_bot_out) at (0,2.5) {};
\node (rng_bot_out) at (-1.5, 2.5) {};
\draw[->, dashed, red, thick]
(rng_bot_out)
to [in=-120, out=-60]
(left_img_bot_out);

\node (rng_text) at (-0.75,3.75) {$G(w)$};
\node (rng_text) at (-0.75,3.25) {\fstikz{Image}};
\node (rng_text) at (-0.75,3) {\fstikz{Generation}};

\node (rng_text) at (2.75,3.75) {$f_1$};
\node (rng_text) at (2.75,3.25) {\fstikz{Known}};
\node (rng_text) at (2.75,3) {\fstikz{Corruption}};

\node (rng_text) at (-0.75,1.75) {\fstikz{Learned a-priori}};
\node (rng_text) at (-0.75,1.5) {\fstikz{given High Res Data}};
% \node (rng_text) at (-0.75,1.25) {\fstikz{given High Res Data}};

\node (rng_text) at (-0.5,0.75) {\fstikz{Can be redone}};
\node (rng_text) at (-0.5,0.5) {\fstikz{using the same $G(w)$}};
\node (rng_text) at (-0.5,0.25) {\fstikz{for multiple $f_i$}};

% \node (rng_text) at (2.75,1.75) {\fstikz{Optimized at}};
% \node (rng_text) at (2.75,1.5) {\fstikz{Test Time}};

\node (left_img_center_out) at (1,2.5) {};
\node (right_bot_out_extra) at (3.9 - 1, 1) {};
\draw[->, thick]
(left_img_center_out)
to [in=90, out=-90]
(1,1) 
to [out=0, in=180]
(right_bot_out_extra);

\node (right_bot_out_extra_2) at (3.9 - 1, 0.5) {};
\draw[->, thick]
(left_img_center_out)
to [in=90, out=-90]
(1,0.5) 
to [out=0, in=180]
(right_bot_out_extra_2);

\node (rng_text) at (1.75,1.25) {$f_2$};
\node (rng_text) at (1.75,0.75) {$f_3$};
\node (rng_text) at (1.75,0.25) {$\vdots$};

\node (blur_img) at (3.5+0.5, 0.5) {\includegraphics[width=1cm]{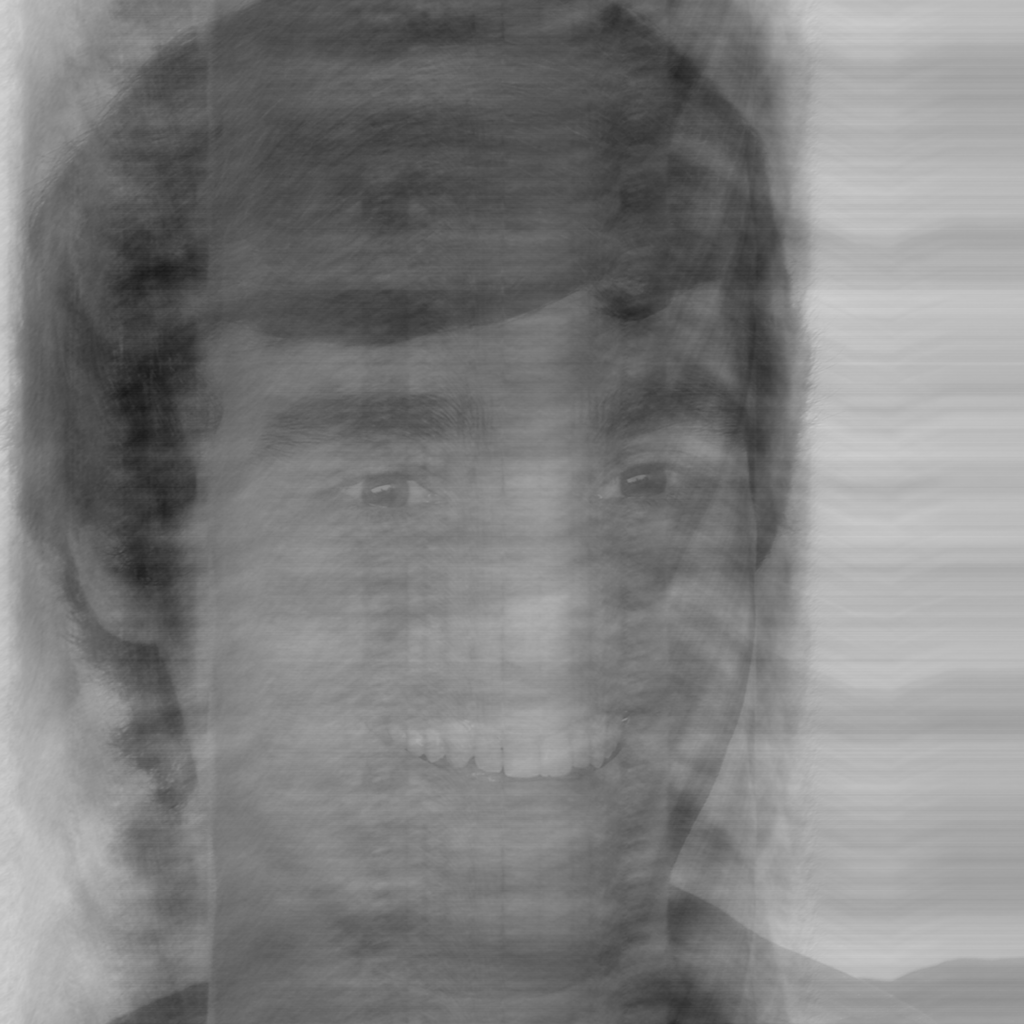}};
\node (blur_img) at (3.5, 1.0) {\includegraphics[width=1cm]{\diagfld/crop.png}};

\end{tikzpicture}

         \caption{Generative approach}
         \label{diagram-ours}
\end{subfigure}
\caption{(a) Classical deep-learning methods for image reconstruction learn to invert specific corruption models such as downsampling with a specific kernel or in-painting with rectangular masks. (b) We use a generative approach that can handle arbitrary corruption processes, such as downsampling or in-painting with an arbitrary mask, by optimizing for it on-the-fly at inference time. Given a latent vector $w$, we use generator $G$ to generate clean images $G(w)$, followed by a corruption model $f$ to generate a corrupted image $f \ci G(w)$. Given an input image $I$, we find the latent $w^*$ that \emph{generated the input image} using the Bayesian MAP estimate $w^* = \argmax_w p(w)p(I|f \ci G(w))$, and we use Variational Inference to sample from the posterior $p(w|I)$. This can be repeated for other corruption processes ($f_2$, $f_3$) such as masking, motion, to-grayscale, as well as for other parametrisations of the process (e.g., super-resolution with different kernels or factors).}
\label{diagram}
\end{figure*}

% Generative models as models emulating the data generating process. State-of-the-art. GANs, Flow, autoregressive. 
Deep generative models have recently obtained state-of-the-art results in simulating high-quality images from a variety of computer vision datasets. Generative Adversarial Networks (GANs) such as StyleGAN2 \cite{karras2020analyzing} and StyleGAN-ADA \cite{karras2020training} have been demonstrated for unconditional image generation, while BigGAN has shown impressive performance in class-conditional image generation \cite{brock2018large}. Similarly, Variational Autoencoder-based methods such as VQ-VAE \cite{van2017neural} and $\beta$-VAE \cite{higgins2016beta} have also been competitive in several image generation tasks. Other lines of research in deep generative models are auto-regressive models such as PixelCNN \cite{van2016conditional} and PixelRNN \cite{oord2016pixel}, as well as invertible flow models such as NeuralODE \cite{chen2018neural}, Glow \cite{kingma2018glow} and RealNVP \cite{dinh2016density}. While these models generate high-quality images that are similar to the training distribution, they are not directly applicable for solving more complex tasks such as image reconstruction. 

% Image reconstruction/inverse problems + prior literature
A particularly important application domain for generative models are inverse problems, which aim to reconstruct an image that has undergone a \emph{corruption process} such as blurring. Previous work has focused on regularizing the inversion process using smoothness \cite{tikhonov1963solution} or sparsity \cite{figueiredo2005bound,mairal2009online,aharon2006k} priors. However, such priors often result in blurry images, and do not enable hallucination of features, which is essential for such ill-posed problems. More recent deep-learning approaches \cite{wang2018esrgan,li2019feedback,yu2018generative,yu2019free} address this challenge using training data made of pairs of (low-resolution, high-resolution) images. However, one fundamental limitation is that they compute the pixelwise or perceptual loss in the high-resolution/in-painted space, which leads to the so-called \emph{averaging effect} \cite{ledig2017photo}: since multiple high-resolution images map to the same low-resolution image $I$, the loss function minimizes the \emph{average} of all such solutions, resulting in a blurry image. Some methods \cite{ledig2017photo,pathak2016context} address this through adversarial losses, which force the model to output a solution that lies on the image manifold. However, even with adversarial losses, it is not clear which solution image is retrieved, and how to sample multiple solutions from the posterior distribution.

% Classical Bayesian methods
To overcome the averaging of all possible solutions to ill-posed problems, one can build methods that estimate by design \emph{all potential solutions}, or a distribution of solutions, for which a Bayesian framework is a natural choice. Bayesian solutions for image reconstruction problems include: Markov Random Field (MRF) priors for denoising and in-painting \cite{roth2005fields}, generative models of photosensor responses from surfaces and illuminants \cite{brainard1997bayesian}, MRF models that leverage global statistics for in-painting \cite{levin2003learning}, Bayesian quantification of the distribution of scene parameters and light direction for inference of shape, surface properties and motion \cite{freeman1994generic}, and sparse derivative priors for super-resolution and image demosaicing \cite{russell2003exploiting}. The seminal work of \cite{geman1984stochastic} formulated the image reconstruction task as the Bayesian optimization over an energy function in a lattice-like physical system.

% non-bayesian: pan2020, asim2020, hand2018phase, kelkar2021prior, jalal2021robust, jalal2021instance
Starting with \cite{bora2017compressed}, several works proposed solving inverse problems using \emph{deep generative priors} \cite{pan2020exploiting,asim2020invertible,hand2018phase,kelkar2021prior}, which constrain the solution to belong to the learned image manifold of a deep generative model. While all these methods focus on deriving a single point estimate, some recent methods further derive a \emph{distribution} of potential solutions \cite{jalal2021robust,jalal2021instance} through Langevin dynamics. However, Langevin dynamics has slow mixing, and additionally, their loss function is \emph{only loosely} based on a Bayesian MAP estimate. A fully-Bayesian model allows optimizing  \emph{distributional losses} instead of single point-estimate losses, and allows the derivation of more complex uncertainty measures, such as $\alpha$-level confidence intervals. While \cite{lugmayr2020srflow,whang2021composing} solve the inverse problem in a Bayesian setting, they only demonstrate it for invertible flow models. It is thus still unclear how to use models other than flows, such as GANs or VAEs, to perform this reconstruction in a Bayesian framework, and how to use GANs to capture \emph{a distribution} of solutions for a given corrupted image.

In this work, we propose a Bayesian method to perform reconstruction through deep generative priors built using state-of-the-art GAN models. Given a pre-trained generator $G$ (here StyleGAN2), a known corruption model $f$, and a corrupted image $I$ to be restored, we minimize at inference time the Bayesian MAP estimate $w^* = \argmax_w p(w)p(I|f \ci G(w))$, where $w$ is latent vector given as input to $G$. We further adapt previous Variational Inference (VI) methods \cite{kingma2013auto,blundell2015weight} to approximate the posterior $p(w|I)$ with a Gaussian distribution, thus enabling sampling of multiple solutions. Our key theoretical contributions are (i) the formulation of image reconstruction in a principled Bayesian framework using deep generative priors and (ii) the adaptation of previous deep-learning based VI methods to sample multiple reconstructions, while on the application side, we demonstrate our method on three datasets, including two challenging medical datasets, and show its competitive performance against four state-of-the-art methods.

%We demonstrate our method on three different datasets and show its competitive performance against four state-of-the-art methods on unseen test images. 

%We demonstrate BRGM on three different datasets: (1) 60,000 images of human faces from the Flickr Faces High Quality (FFHQ) dataset \cite{karras2019style}, (2) $\approx$ 240,000 chest X-ray images from MIMIC III \cite{johnson2016mimic}, and (3) 7,329 brain MRI slices from a collection of 5 neuroimaging datasets \cite{dalca2018anatomical}. We evaluate our model on unseen test images against previous state-of-the-art task-specific approaches \cite{menon2020pulse,li2019feedback,wang2018esrgan,yu2019free}, where our model performs favorably on all three datasets, without any dataset-specific fine-tuning. %The contributions of our work are:
% \begin{itemize}
%  \item We demonstrate a framework of creating powerful baselines for various inverse problems, by combining different prior generative models with various corruption processes.
%  %\item At test time, we show how use back-propagation to estimate the unknown parameters of the corruption process
%  \item We demonstrate our method on a variety of \emph{cross-validated} datasets (FFHQ, X-rays, brain MRIs), with a combination of several corruption processes, each allowing us to perform different image restoration tasks, such as super-resolution and in-painting.
%  \item We evaluate our method against state-of-the-art super-resolution and inpainting methods. 
% % \item We provide pre-trained prior models, corruption processes, and source code: \url{anonymized}.
% \end{itemize}

\subsection{Related work}
 
\textbf{Deep Generative Prior (DGP) methods:} A related work to ours is \pulse, which uses pre-trained StyleGAN models for face super-resolution. Other DGP methods \cite{bora2017compressed,pan2020exploiting,asim2020invertible,hand2018phase,kelkar2021prior} 
 are close to our work, especially those that derive a distribution of potential solutions either through (i) direct sampling from conditionalized flow models \cite{lugmayr2020srflow}, (ii) Langevin dynamics \cite{jalal2021robust,jalal2021instance} or (iii) Variational Inference (VI) \cite{whang2021composing}, each having several advantages and disadvantages: (i) conditionalized flow models generate independent samples from the exact posterior, but use more restricted transformations due to the need to have computable Jacobian determinants, (ii) Langevin dynamics also sample the exact posterior, but generate highly correlated samples and can get stuck in distributional modes while (iii) VI methods can generate a large number of independent samples from an approximate posterior, but require fitting the variational parameters. Our sampling method that uses VI is closest to \cite{whang2021composing}, but we demonstrate it for GAN models such as StyleGAN2 instead of normalizing flows as in \cite{whang2021composing}.

\textbf{Encoder methods:} Other approaches attempt to invert generative models by estimating encoders that map the input images directly into the generator's latent space \cite{richardson2020encoding} in an end-to-end framework. Encoder methods are complimentary to our work, as they can be used to obtain a fast initial estimate of the latent $w$, followed by a slightly longer optimisation process such as ours that can give more accurate reconstructions.

% inverse problems - other similar approaches: Deep Bayesian Inversion, Noise2Noise, AmbientGAN, Deep Image Prior, 
\textbf{Other inverse problems methods:} Approaches similar to ours have also been discussed in inverse problems research. Deep Bayesian Inversion \cite{adler2018deep} performs image reconstruction using a supervised learning approximation. AmbientGAN \cite{bora2018ambientgan} builds a GAN model of clean images given noisy observations only, for a specified corruption model. Deep Image Prior (DIP) \cite{ulyanov2018deep} has shown that the structure of deep convolutional networks captures texture-level statistics that can be used for zero-shot image reconstruction. MimicGAN \cite{anirudh2020mimicgan} has shown how to optimise the parameters of the unknown corruption model. Image2StyleGAN \cite{abdal2019image2stylegan} finds a projection of a given clean image in the latent space of StyleGAN, while the more updated Image2StyleGAN++ \cite{abdal2020image2stylegan} uses the estimated latent projections to demonstrate image manipulations, as well as in-painting. Recent neural radiance fields (NeRF) \cite{mildenhall2020nerf} achieved state-of-the-art results for synthesizing novel views.

%asim2020invertible,hand2018phase,whang2021composing,kelkar2021prior,
%jalal2021robust,lugmayr2020srflow,prakash2021removing,ohayon2021high

%A mathematical analysis of compressed sensing with generative models has also been performed by \cite{bora2017compressed}.

\section{Method}

% Method overview + diagram
An overview of our method is shown in Fig. \ref{diagram}. We assume a given generator $G$ can model the distribution of clean images in a given dataset (e.g., human faces), then use a pre-defined forward model $f$ that corrupts the clean image. Given a corrupted input image $I$, we reconstruct it as $G(w^*)$, where $w^*$ is the Bayesian MAP estimate over the latent vector $w$ of $G$. The graphical model is given in Supp. Fig. \ref{modeldiagram}.

% \subsection{Bayesian Image Reconstruction}

Given an input corrupted image $I$, we aim to reconstruct the clean image $I_{CLN}$. In practice, there could be a distribution $p(I_{CLN}|I)$ of such clean images given a particular input image $I$, which is estimated using Bayes' theorem as $p(I_{CLN}|I) \propto p(I_{CLN}) p(I|I_{CLN})$. The prior term $p(I_{CLN})$ describes the manifold of clean images, restricting the possible reconstructions $I_{CLN}$ to \emph{realistic} images. In our context, the likelihood term $p(I|I_{CLN})$ describes the corruption process $f$, which takes a clean image and produces a corrupted image. 

% \begin{equation}
% \label{bpo}
% p(I_{CLN}|I) \propto p(I_{CLN}) p(I|I_{CLN})
% \end{equation}  

% x32 factor 
\newcommand{\ei}[1]{\fld/00653-recon-real-imagesffhq_test-super-resolution-orig/image#1-target} % Input
\newcommand{\eo}[1]{\fld/00653-recon-real-imagesffhq_test-super-resolution-orig/image#1-clean-step5000.jpg} % original StyleGAN2 inversion
\newcommand{\en}[1]{\fld/00654-recon-real-imagesffhq_test-super-resolution-nonoise/image#1-clean-step5000.jpg} % +no noise
\newcommand{\ew}[1]{\fld/00655-recon-real-imagesffhq_test-super-resolution-wplus/image#1-clean-step5000.jpg} % + W+
\newcommand{\el}[1]{\fld/00656-recon-real-imagesffhq_test-super-resolution-l2/image#1-clean-step5000.jpg} % + L2
\newcommand{\ep}[1]{\fld/00657-recon-real-imagesffhq_test-super-resolution-priorw/image#1-clean-step5000.jpg} % + prior_w
\newcommand{\ec}[1]{\fld/00658-recon-real-imagesffhq_test-super-resolution/image#1-clean-step5000.jpg} % + cosine
\newcommand{\eh}[1]{\fld/00653-recon-real-imagesffhq_test-super-resolution-orig/image#1-true.jpg} % HQ

% evolution in in-painting
\newcommand{\eii}[1]{\fld/00700-recon-real-imagesffhq_test-inpaint-eval/image#1-target} % Input
\newcommand{\eio}[1]{\fld/00705-recon-real-imagesffhq_test-inpaint-eval-orig/image#1-clean-step5000.jpg} % original StyleGAN2 inversion
\newcommand{\ein}[1]{\fld/00704-recon-real-imagesffhq_test-inpaint-eval-nonoise/image#1-clean-step5000.jpg} % +no noise
\newcommand{\eiw}[1]{\fld/00703-recon-real-imagesffhq_test-inpaint-eval-wplus/image#1-clean-step5000.jpg} % + W+
\newcommand{\eil}[1]{\fld/00702-recon-real-imagesffhq_test-inpaint-eval-l2/image#1-clean-step5000.jpg} % + L2
\newcommand{\eip}[1]{\fld/00701-recon-real-imagesffhq_test-inpaint-eval-priorw/image#1-clean-step5000.jpg} % + prior_w
\newcommand{\eic}[1]{\fld/00700-recon-real-imagesffhq_test-inpaint-eval/image#1-clean-step5000.jpg} % + cosine
\newcommand{\eih}[1]{\fld/00700-recon-real-imagesffhq_test-inpaint-eval/image#1-true.jpg} % HQ

% 14, 5, 12, 0 and 1 look v good, but include one bad example also: 
\newcommand{\nrE}{0014}
\newcommand{\nrEtwo}{0005}
\newcommand{\nrEthree}{0012}
\newcommand{\nrEfour}{0014}

\newcommand{\nrEI}{0000}

% need to regenerate them, as they are at x4 resolution. Maybe x8 or x16

\renewcommand{\w}{1.75cm}

\newcommand{\fnt}{\fontsize{7}{9}\selectfont}

\fboxsep=-0.3pt

\begin{figure*}
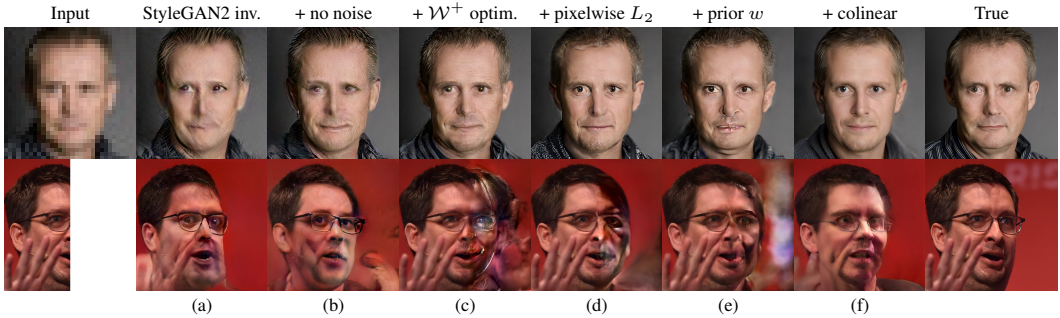

\centering
\setlength{\tabcolsep}{0pt}
\begin{tabu}{cccccccc}
\rowfont{\fnt} Input & StyleGAN2 inv. & + no noise & + \wplus optim. & + pixelwise $L_2$ & + prior $w$ & + colinear & True\\
 \inc{\ei{\nrE}} &  \inc{\eo{\nrE}} & \inc{\en{\nrE}} & \inc{\ew{\nrE}} & \inc{\el{\nrE}} & \inc{\ep{\nrE}} & \inc{\ec{\nrE}} & \inc{\eh{\nrE}}\\
\inc{\eii{\nrEI}} &  \inc{\eio{\nrEI}} & \inc{\ein{\nrEI}} & \inc{\eiw{\nrEI}} & \inc{\eil{\nrEI}} & \inc{\eip{\nrEI}} & \inc{\eic{\nrEI}} & \inc{\eih{\nrEI}}\\
\rowfont{\fnt} &  (a) & (b) & (c) & (d) & (e) & (f) & \\
\end{tabu}
\caption{Reconstructions as the loss function evolves from the original StyleGAN2 inversion to our proposed method. Top row shows super resolution, while bottom row shows in-painting. We start from (a) the original StyleGAN2 inversion, and (b) remove noise optimisation, (c) extend optimisation to full \wplus space, (d) add pixelwise $L_2$ term, (e) add prior on $w$ latent variables and (f) add colinear loss term for $w$. }
\label{evolution}
\end{figure*}
 
\subsection{The image prior term}  
\label{prior}
% actual instantiation: G as StyleGAN2, corruption models
The prior model $p(I_{CLN})$ has been trained \emph{a-priori}, before the corruption task is known, hence satisfying the \emph{principle of independent mechanisms} from causal modelling \cite{peters2017elements}. In our experiments, $I_{CLN} = G(w)$, where $w = [w_1, \dots, w_{18}] \in \mc{R}^{512 \times 18}$ is the latent vector of StyleGAN2 (18 vectors for each resolution level), $G:\mc{R}^{512 \times 18} \to \mc{R}^{n_G \times n_G}$ is the deterministic function given by the StyleGAN2 synthesis network, and $n_G \times n_G$ is the output resolution of StyleGAN2, in our case 1024x1024 (FFHQ, X-Rays) or 256x256 (brains). Our framework is not specific to StyleGAN2: other generator models such as invertible flows \cite{dinh2016density,kingma2018glow} or VAEs \cite{kingma2013auto} can be used, as long as one can flow gradients through the model.

We use the change of variables to express the probability density function over clean images:
\begin{equation}
 \label{changeofvar}
 p(I_{CLN}) \coloneqq p(G(w)) = p(w) \left|\diffp{G(w)}{w} \right|^{-1}
\end{equation}
While the traditional change of variables formula assumes that the function $G$ is invertible, it can be extended to non-invertible\footnote{The supp. section of \cite{cvitkovic2019minimal} presents an excellent introduction to the generalized change of variable theorem.} mappings\cite{cvitkovic2019minimal,krantz2008geometric}. In addition, we assume that the Jacobian determinant $\left|\diffp{G(w)}{w}\right|$ is constant for all $w$, which is a reasonable assumption in StyleGAN2 due to its path length regularization (see Eq. 4 in \cite{karras2020analyzing}).

We now seek to instantiate $p(w)$. Since the latent space of StyleGAN2 consists of many vectors $w = [w_1, \dots, w_L]$, where $L=18$ (one for each layer), we need to set meaningful priors for them. While StyleGAN2 assumed that all vectors $w_i$ are equal, we slightly relax that assumption but set two priors: (i) a cosine similarity prior similar to PULSE \cite{menon2020pulse} that ensures every pair $w_i$ and $w_j$ are roughly colinear, and (ii) another prior $\mathcal{N}(w_i|\mu, \sigma^2)$ that ensures the $w$ vectors lie in the same region as the vectors used during training. We use the following distribution for $p(w)$:
\begin{equation}
\label{priordef}
p(w) = \prod_i \mathcal{N}\left(w_i\middle|\mu, \sigma^2\right) \prod_{i,j} \mathcal{M}\left(cos^{-1}\frac{w_iw_j^T}{|w_i| |w_j|}\middle|0, \kappa\right) 
\end{equation}
where $cos^{-1}\frac{w_iw_j^T}{|w_i| |w_j|}$ is the angle between vectors $w_i$ and $w_j$, and $\mathcal{M}(.|0, \kappa)$ is the \emph{von Mises distribution} with mean zero and scale parameter $\kappa$ which ensures that vectors $w_i$ are aligned. This distribution is analogous to a Gaussian distribution over angles in $[0, 2\pi]$. We compute $\mu$ and $\sigma$ as the mean and standard deviation of 10,000 latent variables passed through the mapping network, like the original StyleGAN2 inversion \cite{karras2020analyzing}.

\subsection{The image likelihood term}
\label{likelihood}

We instantiate the likelihood term $p(I|I_{CLN})$ with a potentially probabilistic forward corruption process $f(I_{CLN}; \psi)$, parameterized\footnote{Since in our experiments $\psi$ is fixed, we drop the notation of $\psi$ in subsequent derivations.} by $\psi$. We study two types of corruption processes $f$ as follows:
\begin{itemize}
\item Super-resolution: $f_{SR}$ is defined as the forward operator that performs downsampling parameterized by a given kernel $k$. For a high-resolution image $I_{CLN}$, this produces a low-resolution (corrupted) image $I_{COR} = (I_{CLN} \circledast k)\downarrow^s$, where $\circledast$ denotes convolution and $\downarrow^s$ denotes downsampling operator by a factor $s$. The parameters are $\psi = \{k, s\}$
\item In-painting with arbitrary mask: $f_{IN}$ is implemented as an operator that performs pixelwise multiplication with a mask $M$. For a given clean image $I_{CLN}$ and a 2D binary mask $M$, it produces a cropped-out (corrupted) image $I_{COR} = I_{CLN} \odot M$, where $\odot$ is the Hadamard product. The parameters of this corruption process are $\psi = \{M\}$ where $M \in \{0,1\}^{H \times W}$, where $H$ and $W$ are the height and width of the image.
\end{itemize}

The likelihood model becomes:
\begin{equation}
\begin{aligned}
 \label{likelihoodDefJac}
p(I|I_{CLN}) = p(I|G(w)) = p(I|f \ci G(w)) |J_f\left(G(w)\right)|^{-1}
\end{aligned}
 \end{equation}
 where $ J_f\left(G(w)\right) = \frac{\partial f \ci G(w)}{\partial G(w)}$ is the Jacobian matrix of $f$ evaluated at $G(w)$, and is again assumed constant. For the noise model in $p(I|f \ci G(w))$, we consider two types of noise distributions: pixelwise independent Gaussian noise, as well as ``perceptual noise'', i.e. independent Gaussian noise in the perceptual VGG embedding space. This yields the following model\footnote{\tc{Model is equivalent to $p(I|f \ci G(w)) = \mathcal{N}\left(\begin{bmatrix} I\\ \phi(I) \end{bmatrix} \middle| \begin{bmatrix} f \ci G(w) \\ \phi \ci f \ci G(w) \end{bmatrix}, \begin{bmatrix} \sigma_{pixel}^2 \mc{I}_{n_{f}^2} & 0 \\ 0 & \sigma_{percept}^2 \mc{I}_{n_{\phi}^2} \end{bmatrix} \right)$}}:
\begin{equation}
\begin{aligned}
 \label{likelihoodDef}
  p(I|f \ci G(w)) = &\ \mathcal{N}(I|f \ci G(w), \sigma_{pixel}^2 \mc{I}_{n_f^2})\ \mathcal{N}(\phi(I)|\phi \ci f \ci G(w), \sigma_{percept}^2 \mc{I}_{n_{\phi}^2})
\end{aligned}
 \end{equation}
where $\phi: \mc{R}^{n_f \times n_f} \to \mc{R}^{n_\phi \times n_\phi}$ is the VGG network, $\sigma_{pixel}^2 \mc{I}_{n_f^2}$ and $\sigma_{percept}^2 \mc{I}_{n_{\phi}^2}$ are diagonal covariance matrices, $n_f \times n_f$ and $n_\phi \times n_\phi$ are the resolutions of the corrupted images $f \ci G(w)$ as well as perceptual embeddings $\phi \ci f \ci G(w)$. Images $I$, $\phi(I)$, $f \ci G(w)$ and $\phi \ci f \ci G(w)$ are flattened to 1D vectors, while covariance matrices $\mc{I}_{n_f^2}$ and $\mc{I}_{n_{\phi}^2}$ are of dimensions $n_f^2 \times n_f^2$ and $n_\phi^2 \times n_\phi^2$.

%%%%%%%%% Qualitative results %%%%%%%%%%%%%

% FFHQ

% 16x16
\newcommand{\srFL}[1]{\fld/00607-recon-real-imagesffhq_test-super-resolution/image#1-target} % LR
\newcommand{\srFB}[1]{\fld/00607-recon-real-imagesffhq_test-super-resolution/bicubic-x4/image#1-target} % Bicubic
\newcommand{\srFE}[1]{\valid/ESRGAN/results/ffhq_LLR/image#1-target_rlt} % ESRGAN
\newcommand{\srFS}[1]{\valid/SRFBN_CVPR19/results/SR/ffhq_LLR/SRFBN/x4/image#1-target} % SRFBN
\newcommand{\srFP}[1]{\valid/pulse/runs/ffhq_LLR/image#1-target} % Pulse
\newcommand{\srFO}[1]{\fld/00607-recon-real-imagesffhq_test-super-resolution/clean_64/image#1-clean-step5000} % Ours x4
\newcommand{\srFOF}[1]{\fld/00607-recon-real-imagesffhq_test-super-resolution/image#1-clean-step5000} % Ours full
\newcommand{\srFH}[1]{\fld/00607-recon-real-imagesffhq_test-super-resolution/image#1-true} % HR

% 32x32
\newcommand{\srFLtwo}[1]{\fld/00620-recon-real-imagesffhq_test-super-resolution/image#1-target} % LR
\newcommand{\srFBtwo}[1]{\fld/00620-recon-real-imagesffhq_test-super-resolution/bicubic-x4/image#1-target} % Bicubic
\newcommand{\srFEtwo}[1]{\valid/ESRGAN/results/ffhq_LLR32x32/image#1-target_rlt} % ESRGAN
\newcommand{\srFStwo}[1]{\valid/SRFBN_CVPR19/results/SR/ffhq_LLR32x32/SRFBN/x4/image#1-target} % SRFBN
\newcommand{\srFPtwo}[1]{\valid/pulse/runs/ffhq_LLR32x32/image#1-target} % Pulse
\newcommand{\srFOtwo}[1]{\fld/00620-recon-real-imagesffhq_test-super-resolution/clean_128/image#1-clean-step5000} % Ours x4 
\newcommand{\srFOFtwo}[1]{\fld/00620-recon-real-imagesffhq_test-super-resolution/image#1-clean-step5000} % Ours full
\newcommand{\srFHtwo}[1]{\fld/00620-recon-real-imagesffhq_test-super-resolution/image#1-true} % HR

% 64x64
\newcommand{\srFLthree}[1]{\fld/00624-recon-real-imagesffhq_test-super-resolution/image#1-target} % LR
\newcommand{\srFBthree}[1]{\fld/00624-recon-real-imagesffhq_test-super-resolution/bicubic-x4/image#1-target} % Bicubic
\newcommand{\srFEthree}[1]{\valid/ESRGAN/results/ffhq_LLR64x64/image#1-target_rlt} % ESRGAN
\newcommand{\srFSthree}[1]{\valid/SRFBN_CVPR19/results/SR/ffhq_LLR64x64/SRFBN/x4/image#1-target} % SRFBN
\newcommand{\srFPthree}[1]{\valid/pulse/runs/ffhq_LLR64x64/image#1-target} % Pulse
\newcommand{\srFOthree}[1]{\fld/00624-recon-real-imagesffhq_test-super-resolution/clean_256/image#1-clean-step5000} % Ours x4
\newcommand{\srFOFthree}[1]{\fld/00624-recon-real-imagesffhq_test-super-resolution/image#1-clean-step5000} % Ours full
\newcommand{\srFHthree}[1]{\fld/00624-recon-real-imagesffhq_test-super-resolution/image#1-true} % HR

% 128x128
\newcommand{\srFLfour}[1]{\fld/00598-recon-real-imagesffhq_test-super-resolution/image#1-target} % LR
\newcommand{\srFBfour}[1]{\fld/00598-recon-real-imagesffhq_test-super-resolution/bicubic-x4/image#1-target} % Bicubic
\newcommand{\srFEfour}[1]{\valid/ESRGAN/results/ffhq_LR/image#1-target_rlt} % ESRGAN
\newcommand{\srFSfour}[1]{\valid/SRFBN_CVPR19/results/SR/ffhq_LR/SRFBN/x4/image#1-target} % SRFBN
\newcommand{\srFPfour}[1]{\valid/pulse/runs/ffhq_LR/512x512/image#1-target} % Pulse
\newcommand{\srFOfour}[1]{\fld/00598-recon-real-imagesffhq_test-super-resolution/clean_512/image#1-clean-step5000} % Ours x4
\newcommand{\srFOFfour}[1]{\fld/00598-recon-real-imagesffhq_test-super-resolution/image#1-clean-step5000} % Ours full
\newcommand{\srFHfour}[1]{\fld/00598-recon-real-imagesffhq_test-super-resolution/image#1-true} % HR

\newcommand{\nrF}{0011} % 1,5 was good
\newcommand{\nrFtwo}{0001}
\newcommand{\nrFthree}{0002}
\newcommand{\nrFfour}{0003}

\renewcommand{\w}{1.66cm}

\renewcommand{\fnt}{\fontsize{8}{10}\selectfont}

\begin{figure*}
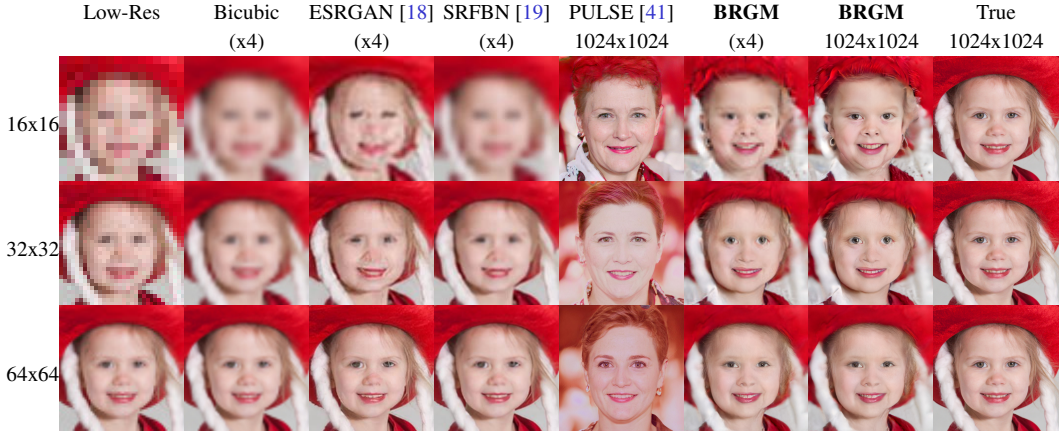

\begin{center}
%\begin{table}[]
%\centering
\setlength{\tabcolsep}{0pt}
\begin{tabu}{ccccccccc}\arrayrulecolor{LightGray}
\rowfont{\fnt} & Low-Res & Bicubic & \esrgan & \srfbn & \pulse  & \brgm      & \brgm      & True \\
\rowfont{\fnt} &    & (x4)    & (x4)   & (x4)  & 1024x1024 & (x4)      & 1024x1024 & 1024x1024 \\
\fnt 16x16 &
\inc{\srFL{\nrF}}&
\inc{\srFB{\nrF}}&
\inc{\srFE{\nrF}}&
\inc{\srFS{\nrF}}&
\inc{\srFP{\nrF}}&
\inc{\srFO{\nrF}}&
\inc{\srFOF{\nrF}}&
\inc{\srFH{\nrF}} \\

\fnt 32x32 &
\inc{\srFLtwo{\nrF}}&
\inc{\srFBtwo{\nrF}}&
\inc{\srFEtwo{\nrF}}&
\inc{\srFStwo{\nrF}}&
\inc{\srFPtwo{\nrF}}&
\inc{\srFOtwo{\nrF}}&
\inc{\srFOFtwo{\nrF}}&
\inc{\srFHtwo{\nrF}} \\

\fnt 64x64 &
\inc{\srFLthree{\nrF}}&
\inc{\srFBthree{\nrF}}&
\inc{\srFEthree{\nrF}}&
\inc{\srFSthree{\nrF}}&
\inc{\srFPthree{\nrF}}&
\inc{\srFOthree{\nrF}}&
\inc{\srFOFthree{\nrF}}&
\inc{\srFHthree{\nrF}} \\

% 128x128 &
% \inc{\srFLfour{\nrF}}&
% \inc{\srFBfour{\nrF}}&
% \inc{\srFEfour{\nrF}}&
% \inc{\srFSfour{\nrF}}&
% \inc{\srFPfour{\nrF}}&
% \inc{\srFOfour{\nrF}}&
% \inc{\srFOFfour{\nrF}}&
% \inc{\srFHfour{\nrF}} \\

% \inc{\srFL{\nrFtwo}}&
% \inc{\srFB{\nrFtwo}}&
% \inc{\srFE{\nrFtwo}}&
% \inc{\srFS{\nrFtwo}}&
% \inc{\srFP{\nrFtwo}}&
% \inc{\srFO{\nrFtwo}}&
% \inc{\srFH{\nrFtwo}} \\
% 
% \inc{\srFL{\nrFthree}}&
% \inc{\srFB{\nrFthree}}&
% \inc{\srFE{\nrFthree}}&
% \inc{\srFS{\nrFthree}}&
% \inc{\srFP{\nrFthree}}&
% \inc{\srFO{\nrFthree}}&
% \inc{\srFH{\nrFthree}} \\

\end{tabu}
\caption{Qualitative evaluation on FFHQ at different input resolutions. Left column shows low resolution inputs, while right column shows true high-quality images. ESRGAN and SRFBN show clear distortion and blurriness, while PULSE does not recover the true image due to strong priors. BRGM shows significant improvements, especially at low resolutions.}
\label{ffhq-super-resolution}
\end{center}
\end{figure*}

\subsection{Image restoration as Bayesian MAP estimate}
\label{bayesianmap}
The restoration of the optimal clean image $I_{CLN}^*$ given a noisy input image $I$ can be performed through the Bayesian maximum a-posteriori (MAP) estimate:
\begin{equation}
 I_{CLN}^* = \argmax_{I_{CLN}} p(I_{CLN}|I) = \argmax_{I_{CLN}} p(I_{CLN}) p(I|I_{CLN})
\end{equation}

We now instantiate the prior $p(I_{CLN})$ and the likelihood $p(I|I_{CLN})$ with formulas from Eq. \ref{priordef} and Eq. \ref{likelihoodDef}, and recast the problem as an optimisation over $w$: $ w^* = \argmax_w p(w) p(I|w)$. This can be simplified to the following loss function (see Supplementary section \ref{supderiv} for full derivation):
\begin{equation}
\label{fullloss}
\resizebox{\columnwidth}{!}{%
 $w^* = \argmin_w \underbrace{\sum_i \left(\frac{w_i-\mu}{\sigma_i}\right)^2}_\text{$\loss_w$} - 2\kappa \underbrace{\sum_{i,j} \frac{w_iw_j^T}{|w_i| |w_j|}}_\text{$\loss_{colin}$} + \sigma_{pixel}^{-2} \underbrace{\norm{I - f \ci G(w)}_2^2}_\text{$\loss_{pixel}$} + \sigma_{percept}^{-2} \underbrace{\norm{I - \phi \ci f \ci G(w)}_2^2}_\text{$\loss_{percept}$}$
 }
\end{equation}
which can be succinctly written as a weighted sum of four loss terms: $w^* = \argmin_w  \loss_{w} + \lambda_{colin} \loss_{colin}  +  \lambda_{pixel} \loss_{pixel} + \lambda_{percept} \loss_{percept}$, where $\loss_w$ is the prior loss over $w$, $\loss_{colin}$ is the colinearity loss on $w$, $\loss_{pixel}$ is the pixelwise loss on the corrupted images, and $\loss_{percept}$ is the perceptual loss, $\lambda_{colin} = -2\kappa$, $\lambda_{pixel} = \sigma_{pixel}^{-2}$ and $\lambda_{percept} = \sigma_{percept}^{-2}$. Given the Bayesian MAP solution $w^*$, the clean image is returned as $I_{CLN}^* = G(w^*)$

\subsection{Sampling multiple reconstructions using Variational Inference}
\label{samplingmethod}

To sample multiple image reconstructions from the posterior distribution $p(w|I)$, we use Variational Inference. We use an approach similar to the Variational Auto-encoder \cite{kingma2013auto,rezende2014stochastic}, where for each data-point we estimate a Gaussian distribution of latent vectors, with the main difference that we do not use an encoder network, but instead optimise the mean and covariance directly. Thus, our approach is also similar to Bayes-by-Backprop (BBB) \cite{blundell2015weight}, but we estimate a Gaussian distribution over the \emph{latent vector}, instead of the network weights as in their case.  

Variational inference (Hinton and Van Camp 1993, Graves 2011) aims to find a parametric approximation $q(w|\theta)$, where $\theta$ are parameters to be learned, to the true posterior $p(w|I)$ over the latent inputs $w$ to the generator network. We seek to minimize:
\begin{align*}
\theta^* & = \argmin_{\theta} KL\left[q(w|\theta)||p(w|I)\right] = \argmin_{\theta} \int q(w|\theta)\ log\ \frac{q(w|\theta)}{p(w)p(I|w)} dw
%& = \argmin_{\theta} KL [q(w|\theta) || p(w)]- \mathbb{E}_{q(w|\theta)}[log\  p(I|w)]
\end{align*}
Using the same approach as in Bayes-by-Backprop \cite{blundell2015weight}, we approximate the expected value over $q(w|\theta)$ using Monte Carlo samples $w^{(i)}$ taken from $q(w|\theta)$:
\begin{equation}
\theta^* = \argmin_{\theta} \sum_{i=1}^n log\ q(w^{(i)}|\theta) - log\ p(w^{(i)}) -  log\  p(I|w^{(i)})
\end{equation}
We parameterize $q(w|\theta)$ as a Gaussian distribution, although in practice we can choose any parametric form for $q$ (e.g. mixture of Gaussians) due to the Monte Carlo approximation. We sample the Gaussian by first sampling unit Gaussian noise $\epsilon$, and then shifting it by the variational mean $\mu_v$ and variational standard deviation $\sigma_v$. To ensure $\sigma_v$ is always positive, we re-parameterize it as $\sigma_v = log(1+exp(\rho_v))$. The variational posterior parameters are $\theta = [\mu_v, \rho_v]$. The prior and likelihood models, $p(w^{(i)})$ and $p(I|w^{(i)})$ are as defined in Eq. \ref{priordef} and Eq. \ref{likelihoodDef}.

While the role of the entropy term $\sum_{i=1}^n q(w^{(i)}|\theta)$ is to regularize the variance of $q$ and ensure there is no mode-collapse, we found it useful to also add a prior over the variational parameter $\sigma_v$, to give the samples more variability. We therefore optimise the following:
\begin{align*}
\theta^* = \argmin_{\theta} & -log\ p(\theta) + \sum_{i=1}^n log\ q(w^{(i)}|\theta) - log\ p(w^{(i)}) - log\  p(I|w^{(i)}) \numberthis \label{variationalloss}
\end{align*} 
where $p(\theta) = f(\sigma_v; \alpha, \beta)$ is an inverse gamma distribution on the variational parameter $\sigma_v$, with concentration $\alpha$ and rate $\beta$. This prior, although optional, encourages larger standard deviations, which ensure that as much of the posterior as possible is covered. Note that, even with this prior, we don't optimise $\sigma_v$ directly, rather we optimise $\rho_v$. We compute the gradients using the same approach as in Bayes-by-Backprop \cite{blundell2015weight}.

To sample the posterior $p(w|I)$, we also tried Stochastic Gradient Langevin Dynamics (SGLD) \cite{welling2011bayesian} and Variational Adam \cite{khan2018fast}, which is equivalent to Variational Online Gauss-Newton (VOGN) \cite{khan2018fast} in our case when the batch size is 1. However, we could not get these methods to work in our setup: SGLD was adding noise of too-high magnitude and the optimisation quickly diverged, while Variational Adam produced little variability between the samples.

%%%% Medical datasets %%%%%

\renewcommand{\w}{1.87cm}

% Xray
% 16x16
\newcommand{\srXL}[1]{\fld/00608-recon-real-imagesxray_frontal_test-super-resolution/image#1-target} % LR
\newcommand{\srXB}[1]{\fld/00608-recon-real-imagesxray_frontal_test-super-resolution/bicubic-x4/image#1-target} % Bicubic
\newcommand{\srXE}[1]{\valid/ESRGAN/results/xray_LLR/image#1-target_rlt} % ESRGAN
\newcommand{\srXS}[1]{\valid/SRFBN_CVPR19/results/SR/xray_LLR/SRFBN/x4/image#1-target} % SRFBN
\newcommand{\srXO}[1]{\fld/00608-recon-real-imagesxray_frontal_test-super-resolution/clean_64/image#1-clean-step5000} % Ours x4
\newcommand{\srXOF}[1]{\fld/00608-recon-real-imagesxray_frontal_test-super-resolution/image#1-clean-step5000} % Ours full
\newcommand{\srXH}[1]{\fld/00608-recon-real-imagesxray_frontal_test-super-resolution/image#1-true} % HR

% 32x32
\newcommand{\srXLtwo}[1]{\fld/00622-recon-real-imagesxray_frontal_test-super-resolution/image#1-target} % LR
\newcommand{\srXBtwo}[1]{\fld/00622-recon-real-imagesxray_frontal_test-super-resolution/bicubic-x4/image#1-target} % Bicubic
\newcommand{\srXEtwo}[1]{\valid/ESRGAN/results/xray_LLR32x32/image#1-target_rlt} % ESRGAN
\newcommand{\srXStwo}[1]{\valid/SRFBN_CVPR19/results/SR/xray_LLR32x32/SRFBN/x4/image#1-target} % SRFBN
\newcommand{\srXOtwo}[1]{\fld/00622-recon-real-imagesxray_frontal_test-super-resolution/clean_128/image#1-clean-step5000} % Ours x4
\newcommand{\srXOFtwo}[1]{\fld/00622-recon-real-imagesxray_frontal_test-super-resolution/image#1-clean-step5000} % Ours full
\newcommand{\srXHtwo}[1]{\fld/00622-recon-real-imagesxray_frontal_test-super-resolution/image#1-true} % HR

% 64x64
\newcommand{\srXLthree}[1]{\fld/00625-recon-real-imagesxray_frontal_test-super-resolution/image#1-target} % LR
\newcommand{\srXBthree}[1]{\fld/00625-recon-real-imagesxray_frontal_test-super-resolution/bicubic-x4/image#1-target} % Bicubic
\newcommand{\srXEthree}[1]{\valid/ESRGAN/results/xray_LLR64x64/image#1-target_rlt} % ESRGAN
\newcommand{\srXSthree}[1]{\valid/SRFBN_CVPR19/results/SR/xray_LLR64x64/SRFBN/x4/image#1-target} % SRFBN
\newcommand{\srXOthree}[1]{\fld/00625-recon-real-imagesxray_frontal_test-super-resolution/clean_256/image#1-clean-step5000} % Ours x4
\newcommand{\srXOFthree}[1]{\fld/00625-recon-real-imagesxray_frontal_test-super-resolution/image#1-clean-step5000} % Ours full
\newcommand{\srXHthree}[1]{\fld/00625-recon-real-imagesxray_frontal_test-super-resolution/image#1-true} % HR

% 128x128
\newcommand{\srXLfour}[1]{\fld/00601-recon-real-imagesxray_frontal_test-super-resolution/image#1-target} % LR
\newcommand{\srXBfour}[1]{\fld/00601-recon-real-imagesxray_frontal_test-super-resolution/bicubic-x4/image#1-target} % Bicubic
\newcommand{\srXEfour}[1]{\valid/ESRGAN/results/xray_LR/image#1-target_rlt} % ESRGAN
\newcommand{\srXSfour}[1]{\valid/SRFBN_CVPR19/results/SR/xray_LR/SRFBN/x4/image#1-target} % SRFBN
\newcommand{\srXOfour}[1]{\fld/00601-recon-real-imagesxray_frontal_test-super-resolution/clean_512/image#1-clean-step5000} % Ours x4
\newcommand{\srXOFfour}[1]{\fld/00601-recon-real-imagesxray_frontal_test-super-resolution/image#1-clean-step5000} % Ours full
\newcommand{\srXHfour}[1]{\fld/00601-recon-real-imagesxray_frontal_test-super-resolution/image#1-true} % HR

% Brains
% 16x16
\newcommand{\srBL}[1]{\fld/00626-recon-real-imagesbrains_test_mono-super-resolution/image#1-target} % LR
\newcommand{\srBB}[1]{\fld/00626-recon-real-imagesbrains_test_mono-super-resolution/bicubic-x4/image#1-target} % Bicubic
\newcommand{\srBE}[1]{\valid/ESRGAN/results/brains_LLR16x16/image#1-target_rlt} % ESRGAN
\newcommand{\srBS}[1]{\valid/SRFBN_CVPR19/results/SR/brains_LLR16x16/SRFBN/x4/image#1-target} % SRFBN
\newcommand{\srBO}[1]{\fld/00626-recon-real-imagesbrains_test_mono-super-resolution/clean_64/image#1-clean-step5000} % Ours x4
\newcommand{\srBOF}[1]{\fld/00626-recon-real-imagesbrains_test_mono-super-resolution/image#1-clean-step5000} % Ours full
\newcommand{\srBH}[1]{\fld/00626-recon-real-imagesbrains_test_mono-super-resolution/image#1-true} % HR

% 32x32
\newcommand{\srBLtwo}[1]{\fld/00604-recon-real-imagesbrains_test_mono-super-resolution/image#1-target} % LR
\newcommand{\srBBtwo}[1]{\fld/00604-recon-real-imagesbrains_test_mono-super-resolution/bicubic-x4/image#1-target} % Bicubic
\newcommand{\srBEtwo}[1]{\valid/ESRGAN/results/brains_LR/image#1-target_rlt} % ESRGAN
\newcommand{\srBStwo}[1]{\valid/SRFBN_CVPR19/results/SR/brains_LR/SRFBN/x4/image#1-target} % SRFBN
\newcommand{\srBOtwo}[1]{\fld/00604-recon-real-imagesbrains_test_mono-super-resolution/clean_128/image#1-clean-step5000} % Ours x4
\newcommand{\srBOFtwo}[1]{\fld/00604-recon-real-imagesbrains_test_mono-super-resolution/image#1-clean-step5000} % Ours full
\newcommand{\srBHtwo}[1]{\fld/00604-recon-real-imagesbrains_test_mono-super-resolution/image#1-true} % HR

% 64x64 don't show brains64x64, because original images were actually 192x168, not really 256x256.
% \newcommand{\srBLtwo}[1]{\fld/00627-recon-real-imagesbrains_test_mono-super-resolution/image#1-target} % LR
% \newcommand{\srBBtwo}[1]{\fld/00627-recon-real-imagesbrains_test_mono-super-resolution/bicubic-x4/image#1-target} % Bicubic
% \newcommand{\srBEtwo}[1]{\valid/ESRGAN/results/brains_LR/image#1-target_rlt} % ESRGAN
% \newcommand{\srBStwo}[1]{\valid/SRFBN_CVPR19/results/SR/brains_LR/SRFBN/x4/image#1-target} % SRFBN
% \newcommand{\srBPtwo}[1]{\valid/pulse/runs/brains_LR/512x512/image#1-target} % Pulse
% \newcommand{\srBOtwo}[1]{\fld/00604-recon-real-imagesbrains_test_mono-super-resolution/image#1-clean-step5000} % Ours
% \newcommand{\srBHtwo}[1]{\fld/00604-recon-real-imagesbrains_test_mono-super-resolution/image#1-true} % HR

\newcommand{\nrX}{0000}
\newcommand{\nrXtwo}{0000}

\newcommand{\nrB}{0001}
\newcommand{\nrBtwo}{0001}

\renewcommand{\fnt}{\fontsize{8}{10}\selectfont}

\begin{figure*}
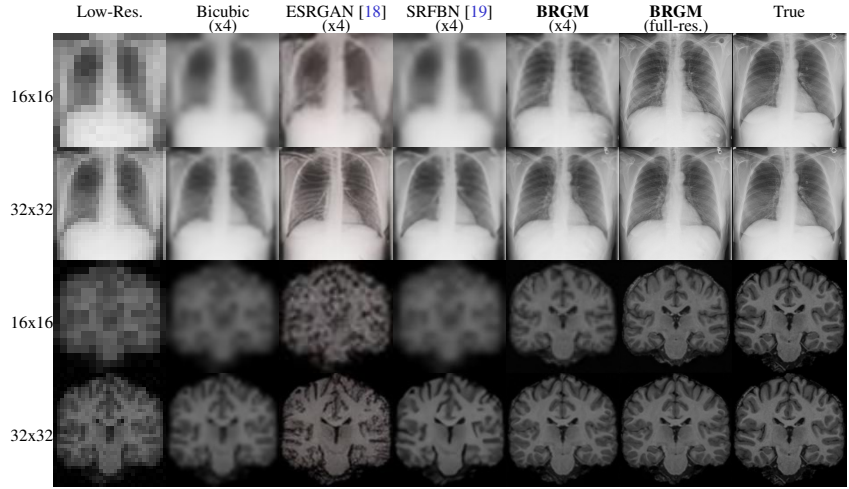

\begin{center}
%\begin{table}[]
%\centering
\setlength{\tabcolsep}{0pt}
\renewcommand{\arraystretch}{0}% Spread rows out...
\resizebox{0.8\columnwidth}{!}{%
\begin{tabu}{cccccccc}\arrayrulecolor{LightGray}
\rowfont{\fnt} & Low-Res. & Bicubic & \esrgan & \srfbn & \brgm & \brgm & True \\
\rowfont{\fnt} &    &  (x4)   &  (x4)  &  (x4) & (x4) & (full-res.) & \\   

\fnt 16x16 &
\inc{\srXL{\nrX}}&
\inc{\srXB{\nrX}}&
\inc{\srXE{\nrX}}&
\inc{\srXS{\nrX}}&
\inc{\srXO{\nrX}}&
\inc{\srXOF{\nrX}}&
\inc{\srXH{\nrX}} \\

\fnt 32x32 &
\inc{\srXLtwo{\nrX}}&
\inc{\srXBtwo{\nrX}}&
\inc{\srXEtwo{\nrX}}&
\inc{\srXStwo{\nrX}}&
\inc{\srXOtwo{\nrX}}&
\inc{\srXOFtwo{\nrX}}&
\inc{\srXHtwo{\nrX}} \\

% \fnt 64x64 &
% \inc{\srXLthree{\nrX}}&
% \inc{\srXBthree{\nrX}}&
% \inc{\srXEthree{\nrX}}&
% \inc{\srXSthree{\nrX}}&
% \inc{\srXOthree{\nrX}}&
% \inc{\srXOFthree{\nrX}}&
% \inc{\srXHthree{\nrX}} \\

% 128x128 &
% \inc{\srXLfour{\nrXtwo}}&
% \inc{\srXBfour{\nrXtwo}}&
% \inc{\srXEfour{\nrXtwo}}&
% \inc{\srXSfour{\nrXtwo}}&
% \inc{\srXOfour{\nrXtwo}}&
% \inc{\srXHfour{\nrXtwo}} \\

\fnt 16x16 &
\inc{\srBL{\nrB}}&
\inc{\srBB{\nrB}}&
\inc{\srBE{\nrB}}&
\inc{\srBS{\nrB}}&
\inc{\srBO{\nrB}}&
\inc{\srBOF{\nrB}}&
\inc{\srBH{\nrB}} \\

\fnt 32x32 &
\inc{\srBLtwo{\nrBtwo}}&
\inc{\srBBtwo{\nrBtwo}}&
\inc{\srBEtwo{\nrBtwo}}&
\inc{\srBStwo{\nrBtwo}}&
\inc{\srBOtwo{\nrBtwo}}&
\inc{\srBOFtwo{\nrBtwo}}&
\inc{\srBHtwo{\nrBtwo}} \\

\end{tabu}
}
\caption{Qualitative evaluation on medical datasets at different resolutions. The left column shows input images, while the right column shows the true high-quality images. BRGM shows improved quality of reconstructions across all resolution levels and datasets. We used the exact same setup as in FFHQ in Fig. \ref{ffhq-super-resolution}, without any dataset-specific parameter tuning.}
\label{medical-super-resolution}
\end{center}
\end{figure*}

\subsection{Model Optimisation}
\label{modeloptim}
% for the x4 resolution increase, one should stop at L2. For the x32 and x64 factor, one should also add +prior_w and +cosine

We optimise the loss in Eq. \ref{fullloss} using Adam \cite{kingma2014adam} with learning rate of 0.001, while fixing $\lambda_{colin}$, $\lambda_{pixel}$ and $\lambda_{percept}$, $\alpha$ and $\beta$ a-priori. On our datasets, we found the following values to give good results: $\lambda_{colin} = 0.03$, $\lambda_{pixel} = 10^{-5}$, $\lambda_{percept} = 0.01$, $\alpha = 0.1$ and $\beta = 0.95$. In Fig \ref{evolution}, we show image super-resolution and in-painting starting from the original StyleGAN2 inversion, and gradually modify the loss function and optimisation until we arrive at our proposed solution. The original StyleGAN2 inversion results in line artifacts for super-resolution, while for in-painting it cannot reconstruct well. After removing the optimisation of noise layers from the original StyleGAN2 inversion\cite{karras2020analyzing} and switching to the extended latent space \wplus, where each resolution-specific vector $w_1$, ... , $w_{18}$ is independent, the image quality improves for super-resolution, while for in-painting the existing image is recovered well, but the reconstructed part gets even worse. More improvements are observed by adding the pixelwise $L_2$ loss, mostly because the perceptual loss only operates at 256x256 resolution. Adding the prior on $w$ and the cosine loss produces smoother reconstructions with less artifacts, especially for in-painting. %While \imagestylegan showed that any image can be ``recovered'' with the StyleGAN2, we note this is not possible in our case for image reconstruction, as we cannot fit the noise variables and $w_i$ latent variables belonging the high-resolution layers.

\subsection{Model training and evaluation}
\label{trainingmain}

% datasets
We train our model on data from three datasets: (i) 70,000 images from FFHQ \cite{karras2019style} at $1024^2$ resolution, 240,000 frontal-view chest X-ray image from MIMIC III \cite{johnson2016mimic} at $1024^2$ resolution, as well as 7,329 middle coronal 2D slices from a collection of 5 brain datasets: ADNI \cite{jack2008alzheimer}, OASIS \cite{marcus2010open}, PPMI \cite{marek2018parkinson}, AIBL \cite{ellis2009australian} and ABIDE \cite{heinsfeld2018identification}. We obtained ethical approval for all data used. All brain images were pre-registered rigidly. For all experiments, we trained the generator, in our case StyleGAN2, on 90\% of the data, and left the remaining 10\% for testing. We did not use the pre-trained StyleGAN2 on FFHQ as it was trained on the full FFHQ. Training was performed on 4 Titan-Xp GPUs using StyleGAN2 config-e, and was performed for 20,000,000 images shown to the discriminator (20,000 kimg), which took almost 2 weeks on our hardware. For a description of the generator training on all three datasets, see Supp. Section \ref{training}. For a description of the inference times of BRGM MAP and VI, see Supp. Section \ref{times}.

% For super-resolution, we compared our approach to \pulse, \esrgan\ and \srfbn, while for in-painting, we compared to \patchgan.
We compare our method with the state-of-the-art methods in super-resolution (\esrgan\ and \srfbn) as well as in-painting (\patchgan). We additionally compared with \pulse\ due to its closeness to our method, as well as the use of the state-of-the-art StyleGAN model. For these methods, we downloaded the pre-trained models. We could not compare with NeRF \cite{mildenhall2020nerf} as it requires multiple views of the same object. Since DIP \cite{ulyanov2018deep} uses statistics in the input image only and cannot handle large masks or large super-resolution factors, we did not include it in the performance evaluation, although we show results with DIP in the supplementary material. For \pulse, we only tested it on FFHQ, as for the medical datasets it required re-training of the StyleGAN2 generator in their own PyTorch implementation, that was different from the official StyleGAN2 implementation. We release our code with the CC-BY license.

% other algorithms
\section{Results}

% \subsection{Super-resolution}

We applied BRGM and the other models on super-resolution at different resolution levels (Fig. \ref{ffhq-super-resolution} and Fig. \ref{medical-super-resolution}). On all three datasets, our method performs considerably better than other models, in particular at lower input resolutions: ESRGAN yields jittery artifacts, SRFBN gives smoothed-out results, while PULSE generates very high-resolution images that don't match the true image, likely due to the hard projection of their optimized latent to $S^{d-1}$, the unit sphere in $d$-dimensions, as opposed to a soft prior term such as $\loss_w$ in our case. Moreover, as opposed to ESRGAN and SRFBN, both our model as well as PULSE can perform more than x4 super-resolution, going up to 1024x1024. Without changing any hyper-parameters, we observe similar trends on the other two medical datasets.

% inpainting

\newcommand{\inFH}[1]{\fld/00636-recon-real-imagesffhq_test-inpaint-arbitrary/image#1-true} % HQ
\newcommand{\inFM}[1]{masks/image#1-merged} % Mask
\newcommand{\inFP}[1]{\valid/inpaint-sn-patchgan/output/ffhq/image#1} % SN-PatchGAN
\newcommand{\inFD}[1]{validation/dip/output/ffhq_arbitrary/image#1_6000} % Deep Image Prior
\newcommand{\inFO}[1]{\fld/00636-recon-real-imagesffhq_test-inpaint-arbitrary/inpaint/image#1} % Ours

\newcommand{\inFEH}[1]{\fld/00700-recon-real-imagesffhq_test-inpaint-eval/image#1-true} % HQ
\newcommand{\inFEM}[1]{\fld/00700-recon-real-imagesffhq_test-inpaint-eval/image#1-target} % Mask
\newcommand{\inFEP}[1]{\valid/inpaint-sn-patchgan/output/ffhq_eval/image#1} % SN-PatchGAN
\newcommand{\inFED}[1]{validation/dip/output/ffhq/image#1_6000} % Deep Image Prior
\newcommand{\inFEO}[1]{\fld/00700-recon-real-imagesffhq_test-inpaint-eval/inpaint/image#1} % Ours

\newcommand{\inXH}[1]{\fld/00699-recon-real-imagesxray_frontal_test-inpaint-eval/image#1-true} % HQ
\newcommand{\inXM}[1]{\fld/00699-recon-real-imagesxray_frontal_test-inpaint-eval/image#1-target} % Mask
\newcommand{\inXP}[1]{\valid/inpaint-sn-patchgan/output/xray_eval/image#1} % PatchGAN
\newcommand{\inXD}[1]{validation/dip/output/xray/image#1_6000} % Deep Image Prior
\newcommand{\inXO}[1]{\fld/00699-recon-real-imagesxray_frontal_test-inpaint-eval/inpaint/image#1} % Ours

\newcommand{\inBH}[1]{\fld/00707-recon-real-imagesbrains_test_mono-inpaint-eval/image#1-true} % HQ
\newcommand{\inBM}[1]{\fld/00707-recon-real-imagesbrains_test_mono-inpaint-eval/image#1-target} % Mask
\newcommand{\inBP}[1]{\valid/inpaint-sn-patchgan/output/brains_eval/image#1} % PatchGAN
\newcommand{\inBD}[1]{validation/dip/output/brains/image#1_6000} % Deep Image Prior
\newcommand{\inBO}[1]{\fld/00707-recon-real-imagesbrains_test_mono-inpaint-eval/inpaint/image#1} % Ours

\newcommand{\nrI}{0002} % main
\newcommand{\nrItwo}{0014} % main % 5 is v good
\newcommand{\nrIthree}{0000} % supplementary
\newcommand{\nrIfour}{0001} % supplementary
\newcommand{\nrIfive}{0008} % supplementary
\newcommand{\nrIsix}{0009} % supplementary

\newcommand{\nrIM}{0000} % main
\newcommand{\nrIMtwo}{0001} % supplementary
\newcommand{\nrIMthree}{0003} % supplementary
\newcommand{\nrIMfour}{0004} % supplementary

\newcommand{\nrIMB}{0001} % main
\newcommand{\nrIMBtwo}{0002} % supplementary
\newcommand{\nrIMBthree}{0003} % supplementary
\newcommand{\nrIMBfour}{0004} % supplementary
\newcommand{\nrIMBfive}{0003} % supplementary

\renewcommand{\w}{2.05cm}

\begin{figure}
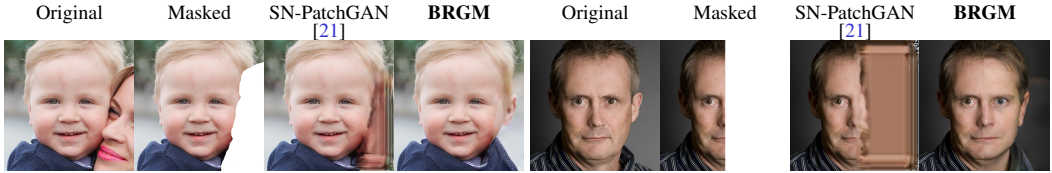

\begin{center}
%\begin{table}[]
%\centering
\setlength{\tabcolsep}{0pt}
\renewcommand{\arraystretch}{0.5}% Spread rows out...
\resizebox{\columnwidth}{!}{%
\begin{tabu}{cccc||cccc}\arrayrulecolor{white}
\rowfont{\footnotesize} Original & Masked & SN-PatchGAN & \brgm & Original & Masked & SN-PatchGAN & \brgm\\
\rowfont{\footnotesize} & & \cite{yu2019free} & & & & \cite{yu2019free} & \\

\inc{\inFH{\nrI}}&
\inc{\inFM{\nrI}}&
\inc{\inFP{\nrI}}&
\inc{\inFO{\nrI}}& 
\inc{\inFEH{\nrItwo}}&
\inc{\inFEM{\nrItwo}}&
\inc{\inFEP{\nrItwo}}&
\inc{\inFEO{\nrItwo}}\\

% \inc{\inXH{\nrIM}}&
% \inc{\inXM{\nrIM}}&
% \inc{\inXP{\nrIM}}&
% \inc{\inXO{\nrIM}}\\
% 
% \inc{\inBH{\nrIMB}}&
% \inc{\inBM{\nrIMB}}&
% \inc{\inBP{\nrIMB}}&
% \inc{\inBO{\nrIMB}}\\

\end{tabu}
}
\end{center}
\caption{Comparison between our method and \patchgan\ on in-painting. SN-PatchGAN fails on large masks, while our method can still recover the high-level structure.}
\label{inpainting}
\end{figure}

Fig. \ref{inpainting} illustrates our method's performance on in-painting with arbitrary as well as rectangular masks, as compared to to the leading in-painting model \patchgan. Our method produces considerably better results than SN-PatchGAN. In particular, SN-PatchGAN lacks high-level semantics in the reconstruction, and cannot handle large masks. For example, in the first figure, when the mother is cropped out, SN-PatchGAN is unable to reconstruct the ear. Our method on the other hand is able to reconstruct the ear and the jawline. One reason for the lower performance of SN-PatchGAN could be that it was trained on CelebA, which has lower variation than FFHQ. In Supplementary Figs. \ref{ffhq-inpainting}, \ref{xray-inpainting} and \ref{brain-inpainting}, we show further in-painting examples with our method as well as \patchgan, on all three datasets, and for different types of arbitrary masks.

\newcommand{\flds}{results/samFFHQ}

\newcommand{\stepth}{step1100}

\newcommand{\is}{15489}
\newcommand{\step}{step460}

\newcommand{\fldst}{results/samFFHQinp}
\newcommand{\ist}{22706}
\newcommand{\stept}{step290}

\begin{figure*}
\renewcommand{\w}{1.98cm}
% \fboxsep=0pt
\centering
\setlength{\tabcolsep}{0pt}
\renewcommand{\arraystretch}{1.5}% Spread rows out...
\begin{tabular}{cc||ccccc}\arrayrulecolor{white}
Input & True & Est. Mean & Sample 1 & Sample 2 & Sample 3 & Sample 4\\
\inc{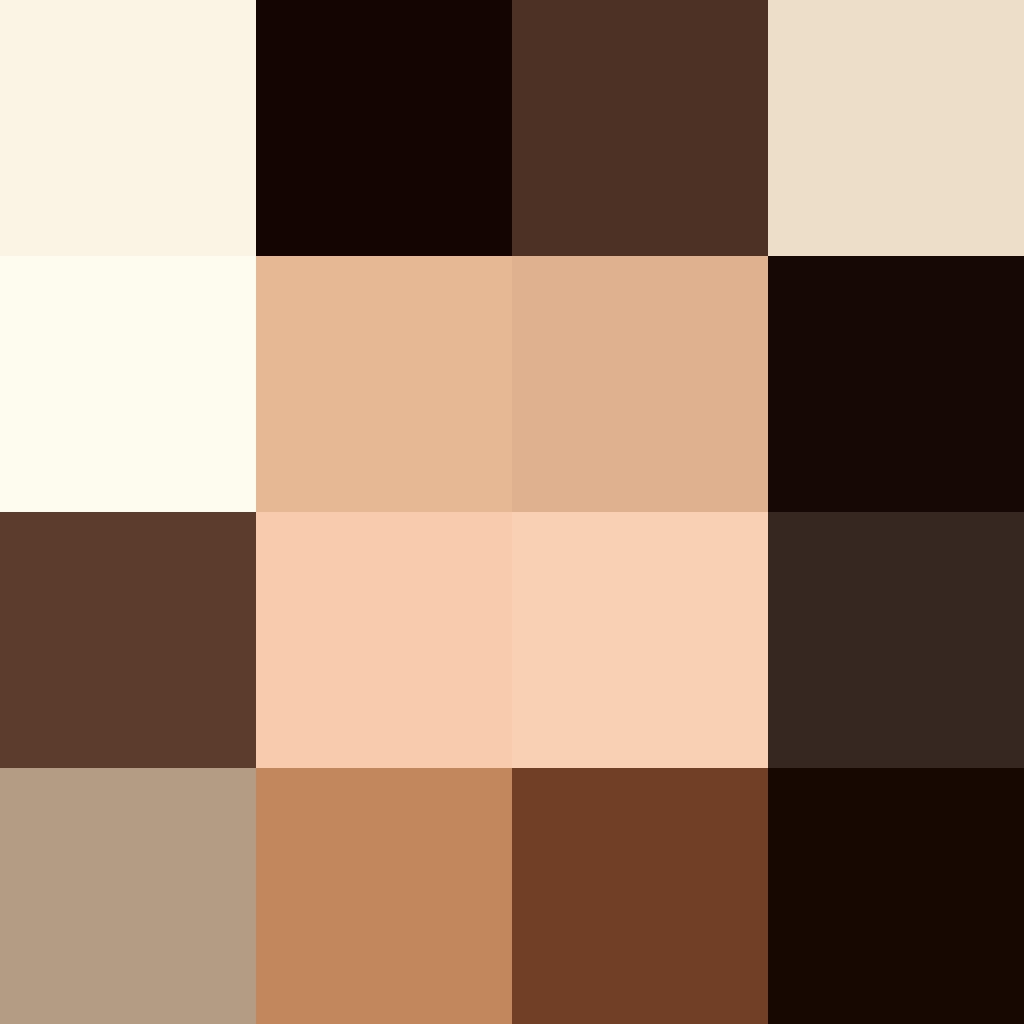} & \inc{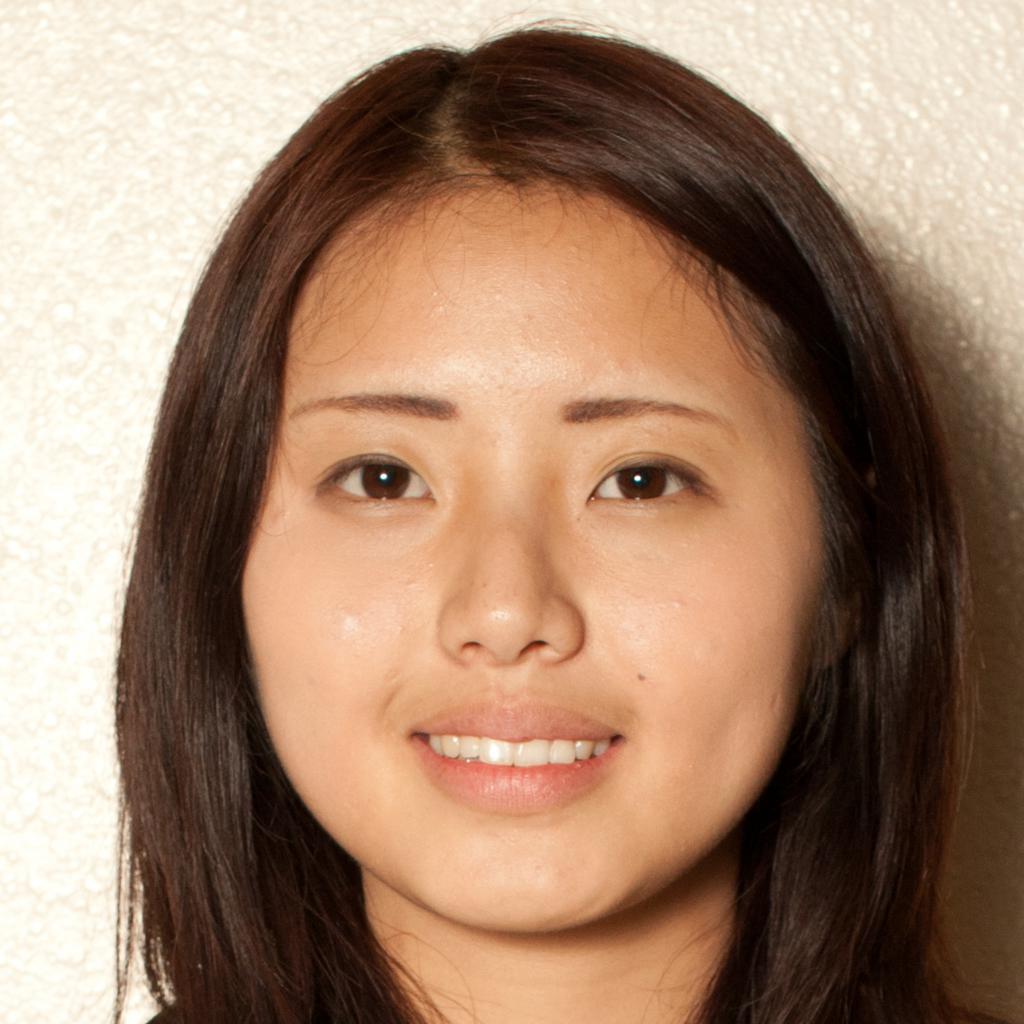} & \inc{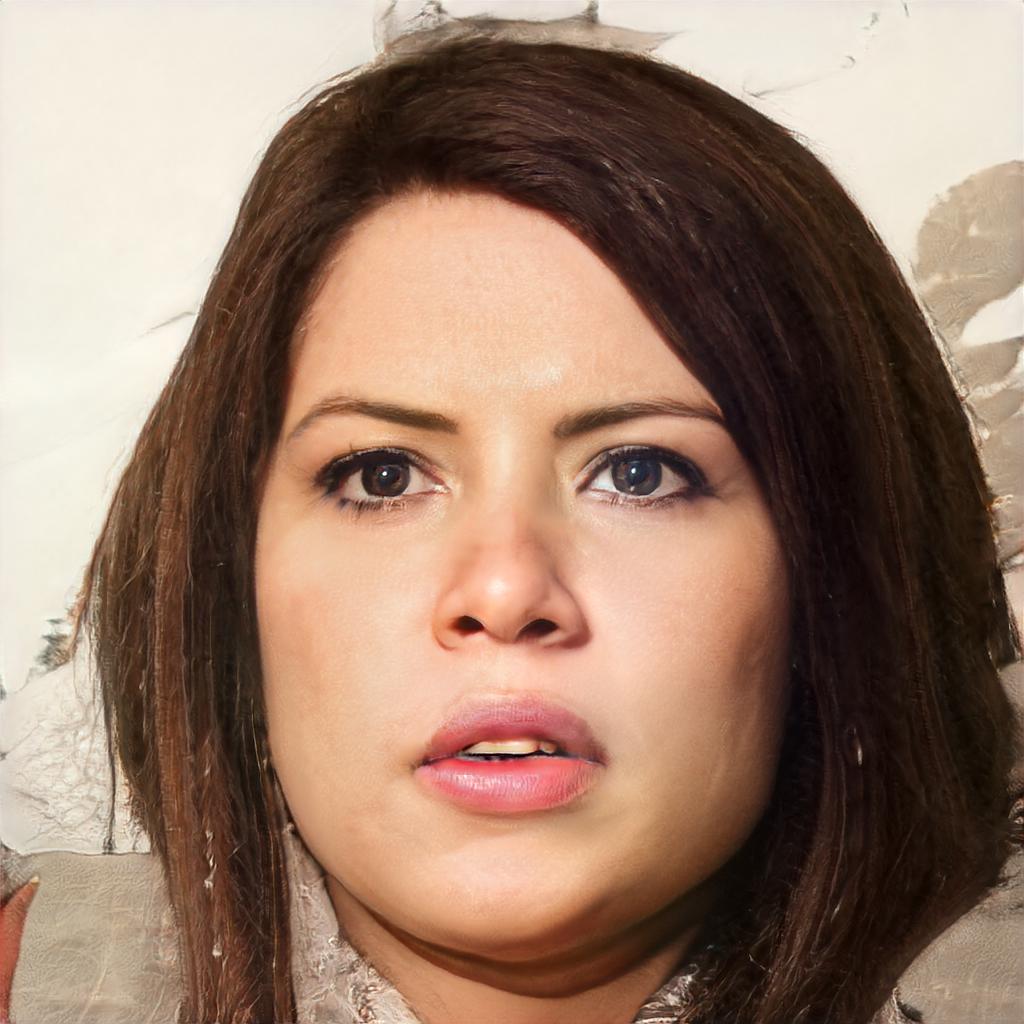} & \inc{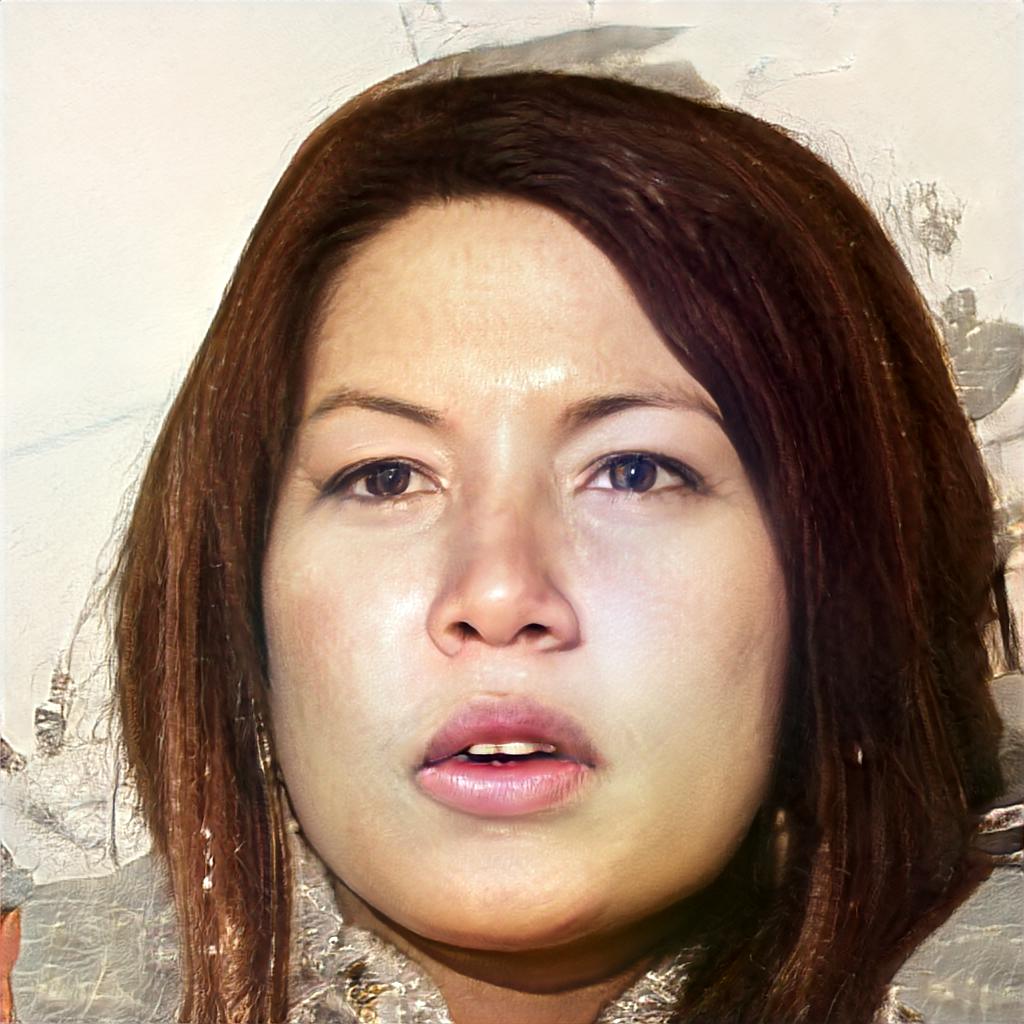} & \inc{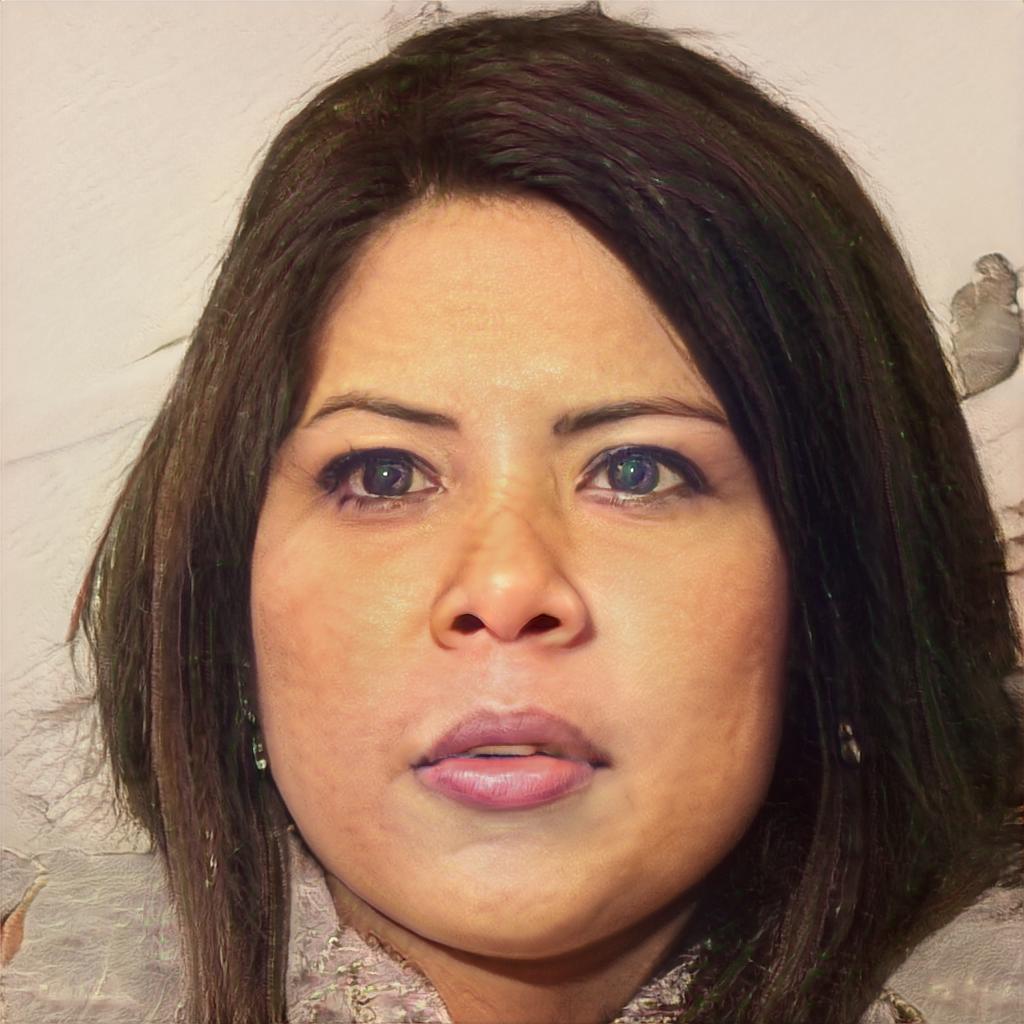} & \inc{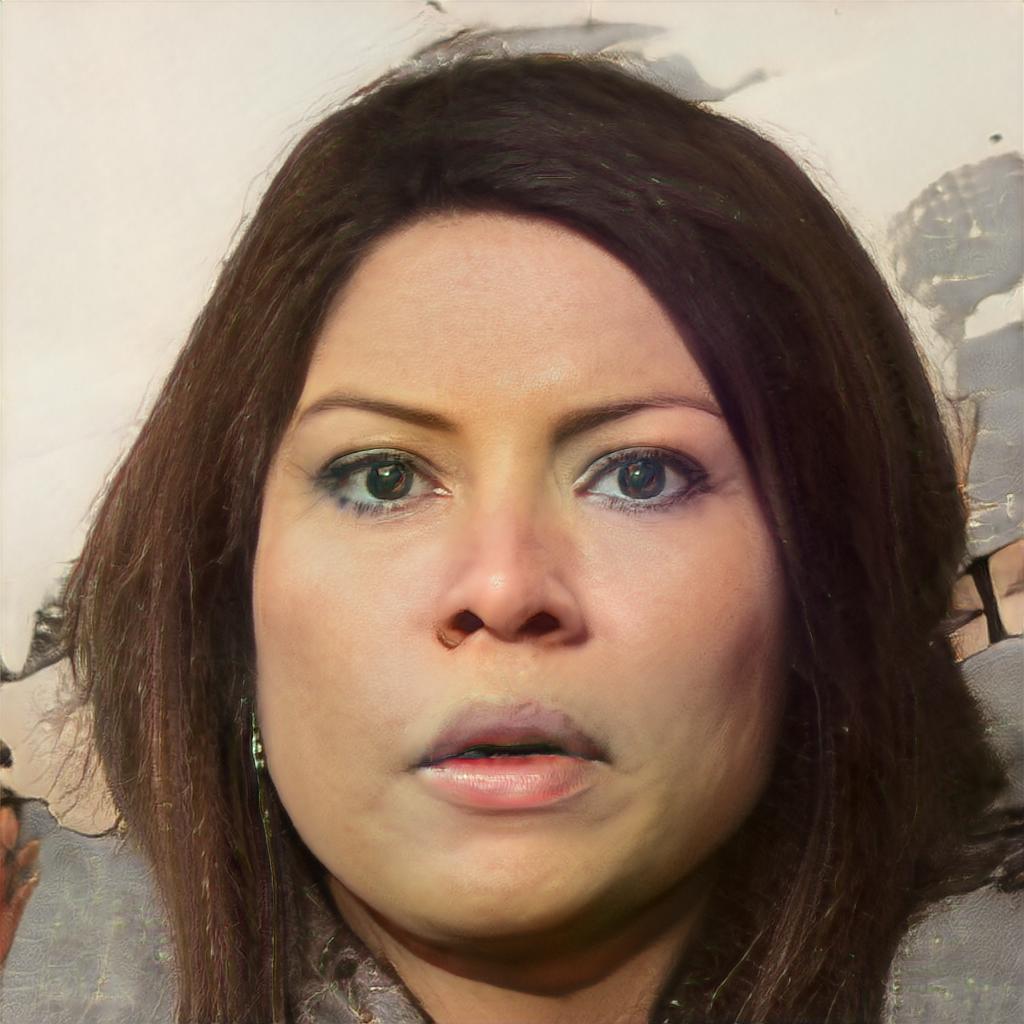} & \inc{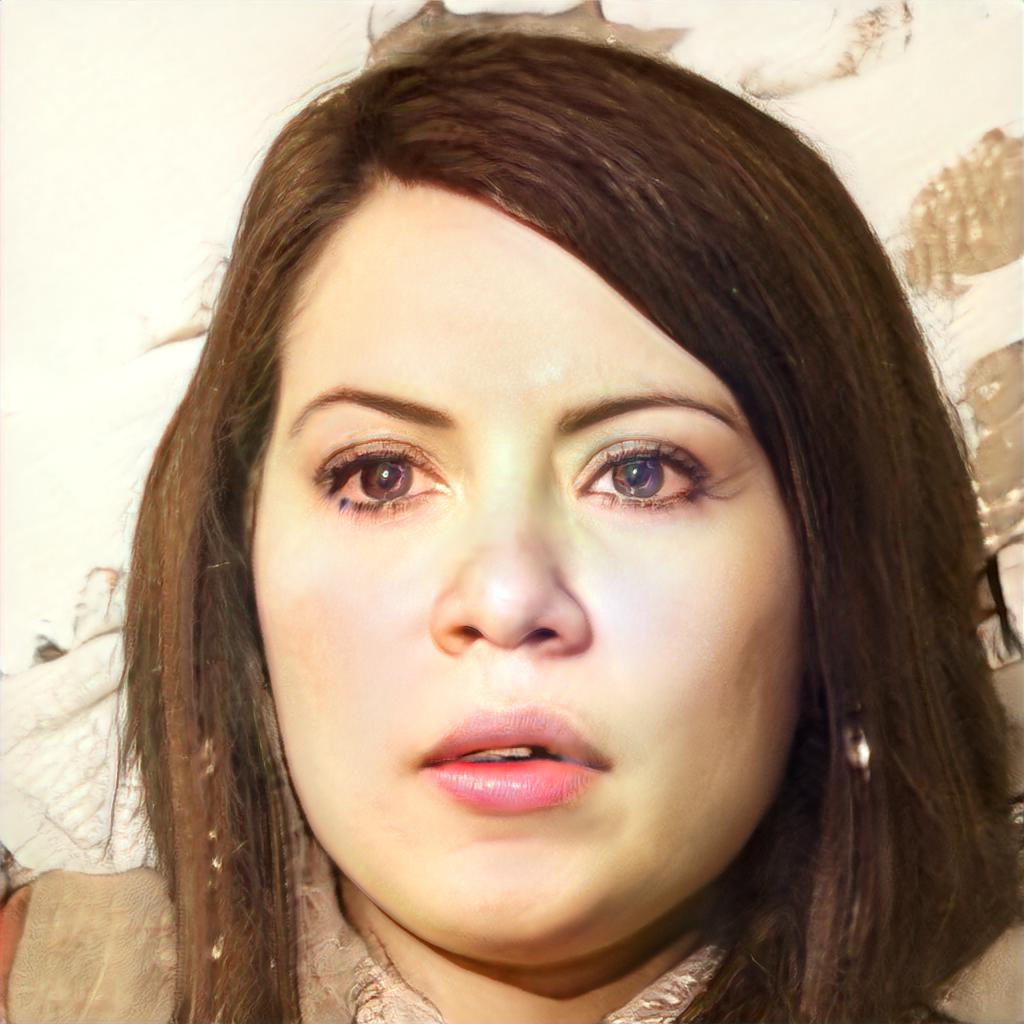}\\
 
% LR & \inc{\flds/\is_target.jpg} & \inc{\flds/\is_corrupted_\step.jpg} & \inc{\flds/\is_corrsample1_\step.jpg} & \inc{\flds/\is_corrsample2_\step.jpg} & \inc{\flds/\is_corrsample3_\step.jpg} & \inc{\flds/\is_corrsample4_\step.jpg} & \inc{\flds/\is_corrsample5_\step.jpg}\\

\inc{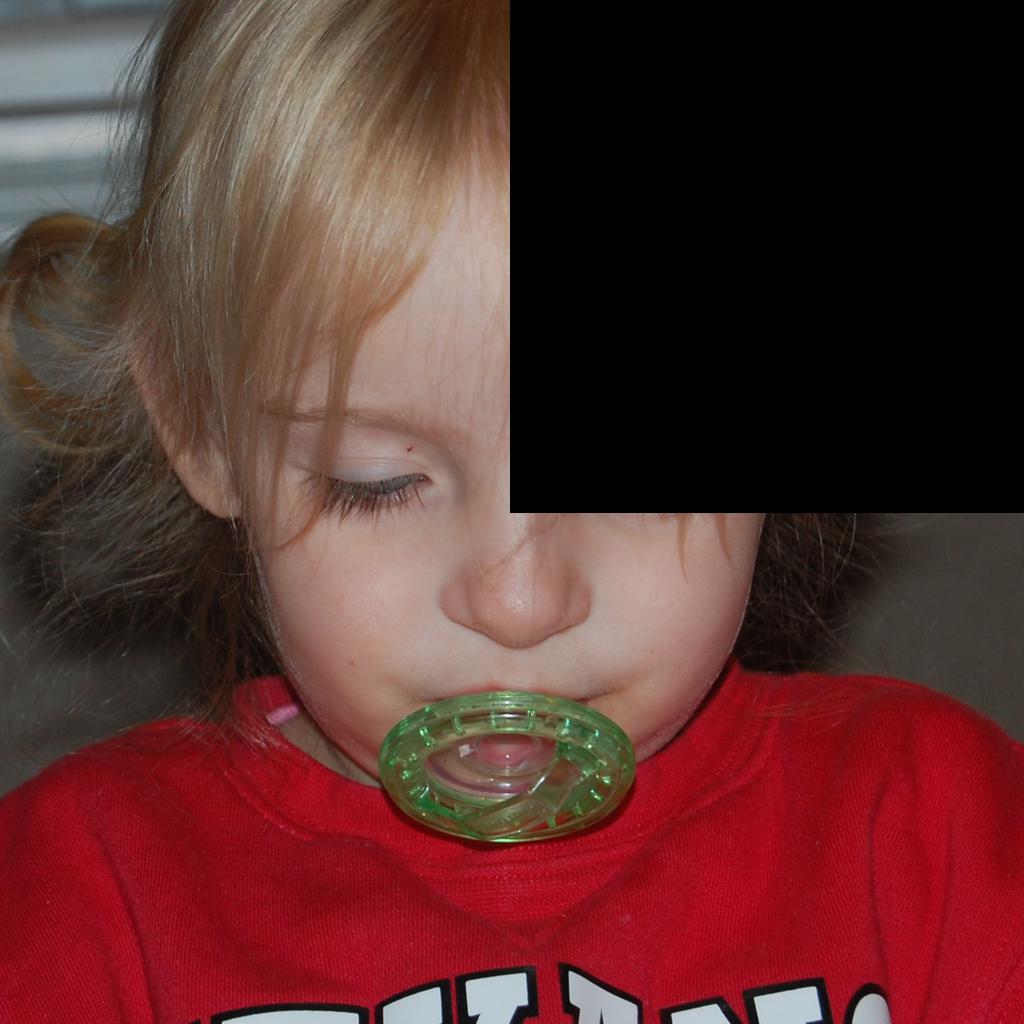} & \inc{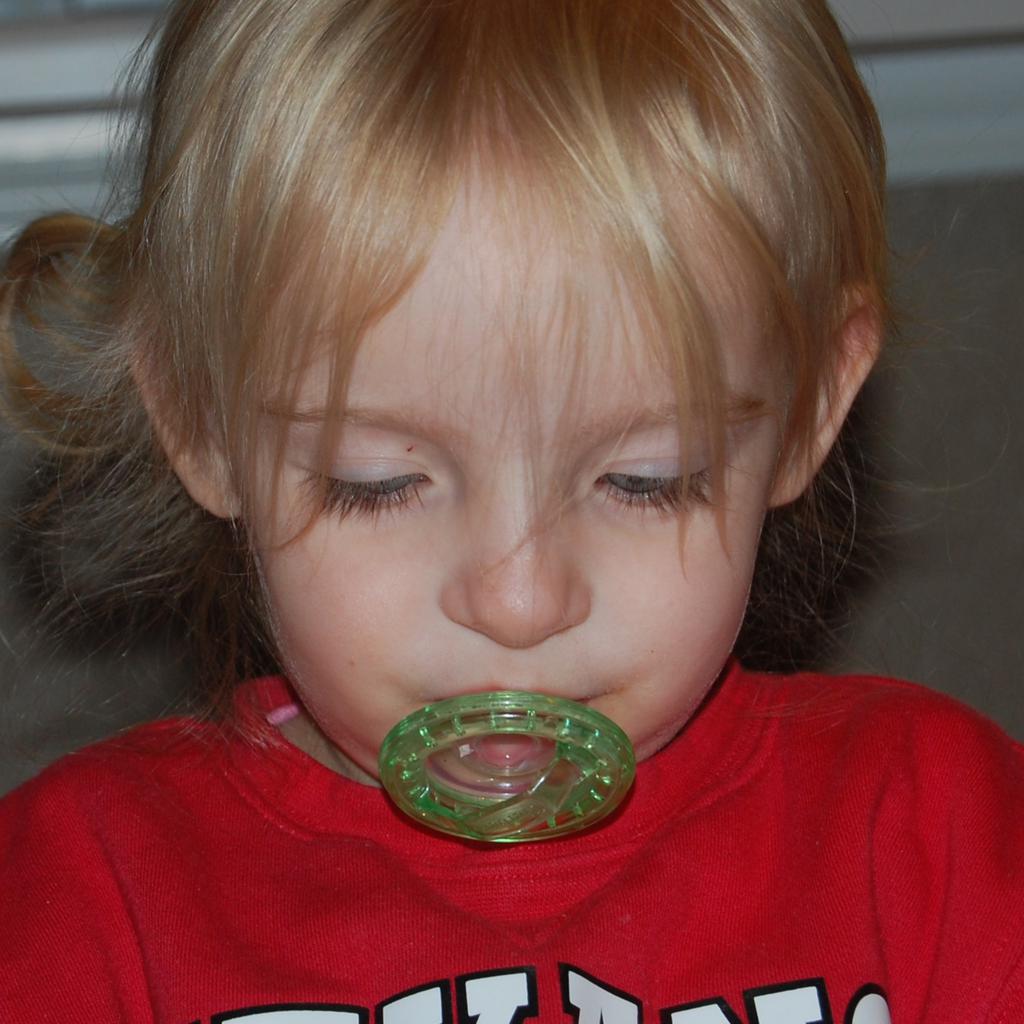} & \inc{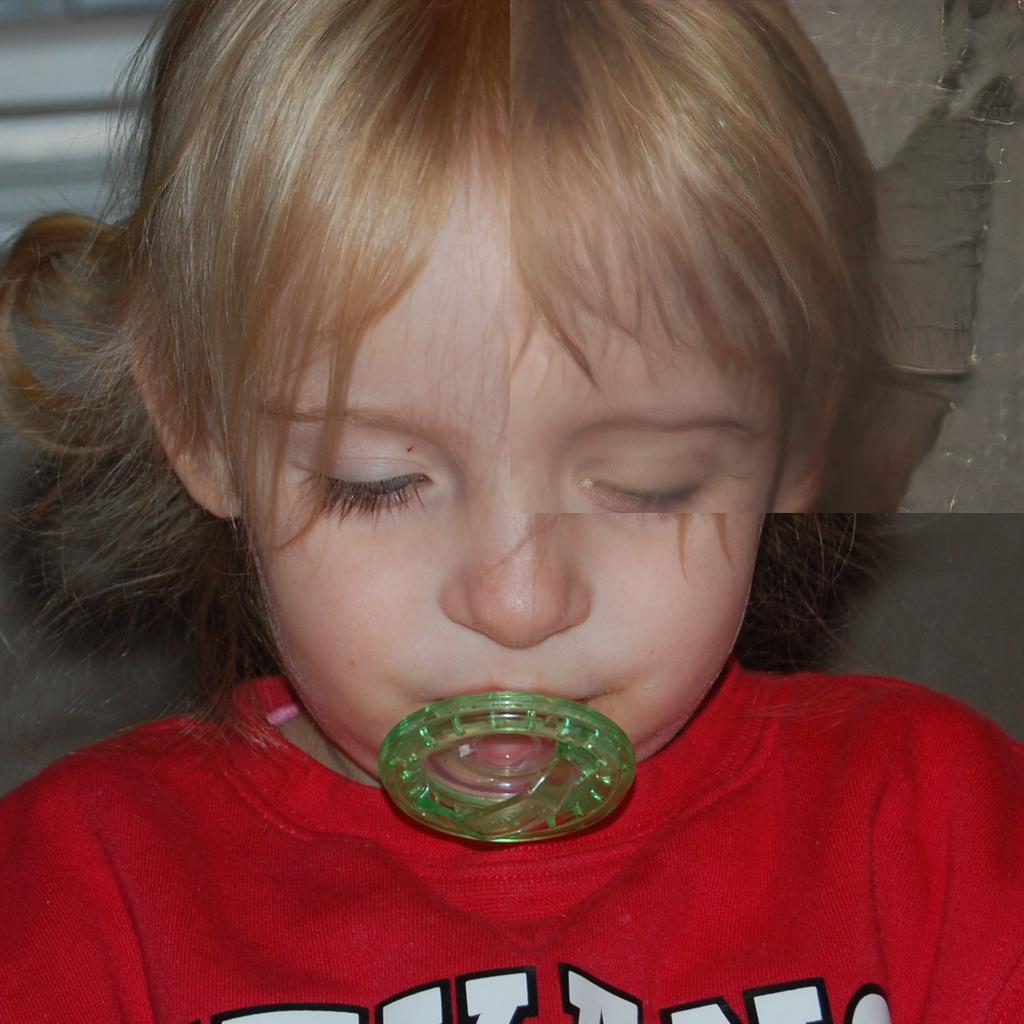} & \inc{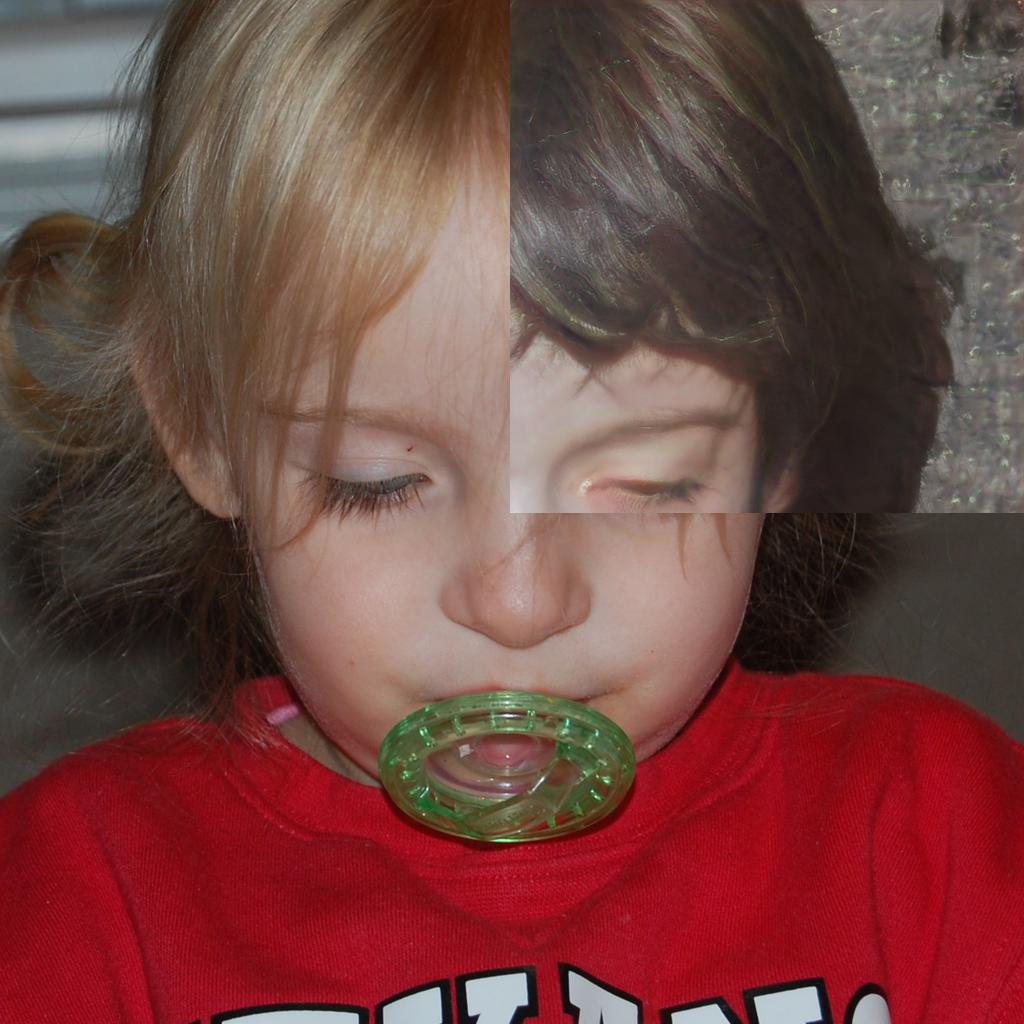} & \inc{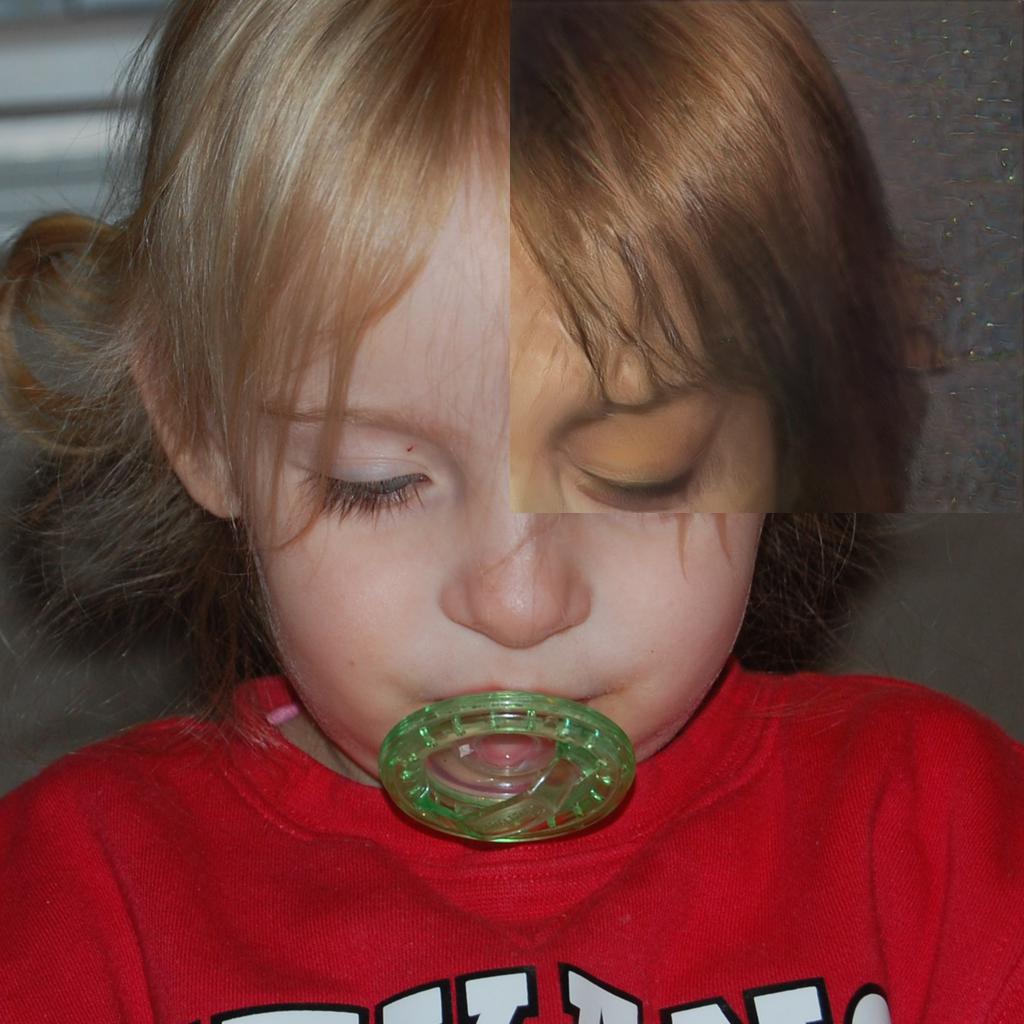} & \inc{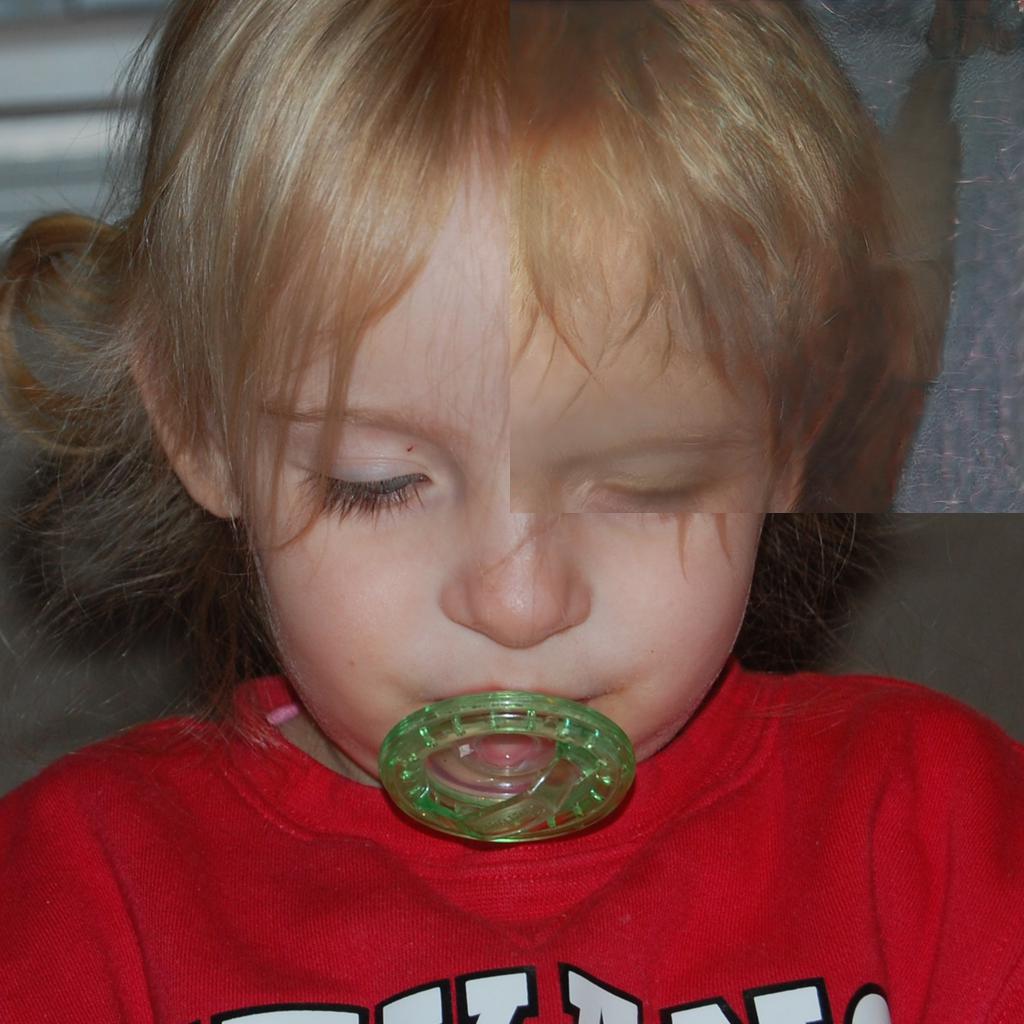} & \inc{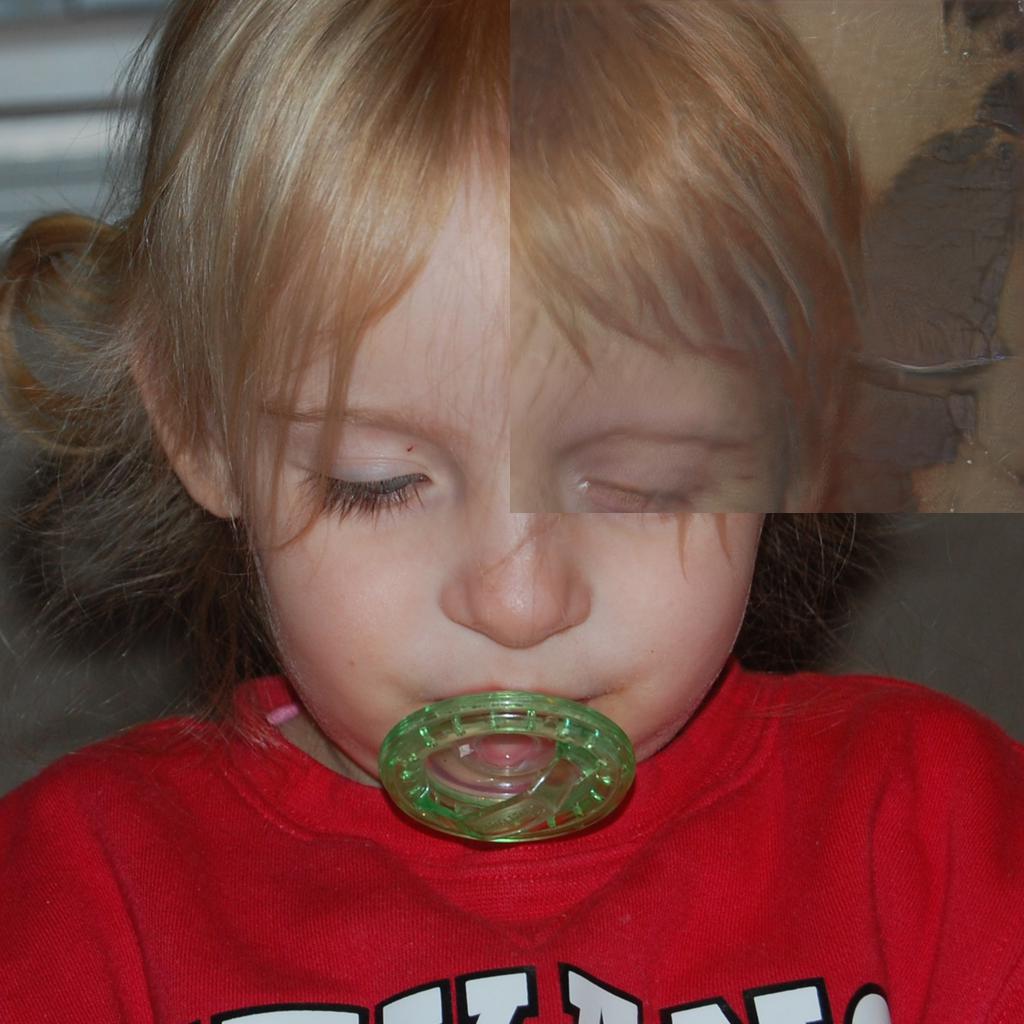}\\

 \end{tabular}
 \caption{\tc{Sampling using Variational Inference. Given the input image (left column), we show the estimated mean image $G(\mu_v)$ (third column), alongside samples around the mean $G(\mu_v + \sigma_v \epsilon)$.}}
 \label{figsampling}
\end{figure*}

% \subsection{Sampling}

In Fig. \ref{figsampling}, we show samples from the variational posterior $q(w|\theta)$, for both super-resolution and in-painting. For super-resolution, we show an extreme downsampling example (x256) going from 1024x1024 to 4x4, in order to clearly see the potential variability in the reconstructions. The Variational Inference method gives samples of reasonably high variability and fidelity, although in harder cases (Supp. Fig. \ref{samplingsupp} and \ref{samplinginpsupp}) it overfits the posterior.

In supplementary section \ref{ablation}, we show in an ablation study over the hyper-parameters $\lambda_{pixel}$, $\lambda_{percept}$ and $\lambda_{colin}$ that our method is not sensitive to the choice of these parameters, as there are multiple levels of magnitudes giving good results. In addition, in supplementary section \ref{downstream}, we show that BRGM-reconstructed images can be used for downstream processing, through an example of edema severity prediction on the Chest X-Ray images.

\subsection{Quantitative evaluation}

Table \ref{eval} (left) reports performance metrics of super-resolution on 100 unseen images at different resolution levels. At low $16^2$ input resolutions, our method outperforms all other super-resolution methods consistently on all three datasets. However, at resolutions of $32^2$ and higher, \srfbn\ achieves the lowest LPIPS\cite{zhang2018unreasonable} and root mean squared error (RMSE), albeit the qualitative results from this method showed that the reconstructions are overly smooth and lack detail. The performance degradation of our model is likely because the StyleGAN2 generator $G$ cannot easily generate these unseen images at high resolutions, although this is expected to change in the near future given the fast-paced improvements in such generator models. However, compared to those methods, our method is more generalisable as it is not specific to a particular type of corruption, and can increase the resolution by a factor higher than 4x. In Supplementary Table \ref{eval-super-resolution-supplementary}, we additionally provide PSNR, SSIM and MAE scores, which show a similar behavior to LPIPS and RMSE.

For quantitative evaluation on in-painting, we generated 7 masks similar to the setup of \cite{abdal2020image2stylegan}, and applied them in cyclical order to 100 unseen images from the test sets of each dataset. In Table \ref{eval} (top-right), we show that our method consistently outperforms \patchgan\ with respect to all performance measures. 

To account for human perceptual quality, we performed a forced-choice pairwise comparison test, which has been shown to be most sensitive and simple for users to perform \cite{mantiuk2012comparison}. Twenty raters were each shown 100 test pairs of the true image and the four reconstructed images by each algorithm, and raters were asked to choose the best reconstruction (see supplementary section \ref{supp-eval-humans} for more information on the design). We opted for this paired test instead of the mean opinion score (MOS) because it also accounts for fidelity of the reconstruction to the true image. This is important in our setup, because a method such as PULSE can reconstruct high-resolution faces that are nonetheless of a different person (see Fig. \ref{ffhq-super-resolution}). In Table \ref{eval} (bottom-right), the results confirm that out method is the best at low $16^2$ resolution and second-best at $32^2$ resolution, with lower performance at $64^2$ resolution.

\newcommand{\tb}[1]{\textbf{#1}}
\newcommand{\lpi}{\multicolumn{1}{c}{\small LPIPS$\downarrow$}}
\newcommand{\rms}{\multicolumn{1}{c}{\small RMSE$\downarrow$}}

\begin{table}[!t]
\setlength{\tabcolsep}{2.0pt}
\begin{subtable}[t]{.5\linewidth}
\centering
\fnt Super-resolution (LPIPS$\downarrow$ / RMSE$\downarrow$)\\
\resizebox{\columnwidth}{!}{%
\renewcommand{\arraystretch}{1.37}
\begin{tabular}{l|cccc}
\toprule
% Dataset &       \multicolumn{2}{c}{Ours}  &    \multicolumn{2}{c}{PULSE}    &   \multicolumn{2}{c}{ESRGAN}    &   \multicolumn{2}{c}{SRFBN}  \\
 Dataset &       \brgm  &    \pulse    &   \esrgan    &   \srfbn  \\
%         &       LP/RM  &    LP/RM    &   LP/RM    &   LP/RM  \\

\midrule
   FFHQ $16^2$ &  \tb{0.24}/25.66 &  0.29/27.14 &  0.35/29.32 &  0.33/\tb{22.07} \\
   FFHQ $32^2$ &  0.30/18.93 &  0.48/42.97 &  0.29/23.02 &  \tb{0.23}/\tb{12.73} \\
   FFHQ $64^2$ &  0.36/16.07 &  0.53/41.31 &  0.26/18.37 &   \tb{0.23}/\tb{9.40} \\
%  FFHQ $128^2$ &  0.34/15.84 &  0.57/34.89 &  0.15/15.84 &   \tb{0.09}/\tb{7.55} \\
   X-ray $16^2$ &  \tb{0.18}/\tb{11.61} &           - &  0.32/14.67 &  0.37/12.28 \\
   X-ray $32^2$ &  0.23/10.47 &           - &  0.32/12.56 &   \tb{0.21}/\tb{6.84} \\
   X-ray $64^2$ &  0.31/10.58 &           - &   0.30/8.67 &   \tb{0.22}/\tb{5.32} \\
%  X-ray $128^2$ &  0.27/10.53 &           - &   0.20/7.19 &   \tb{0.07}/\tb{4.33} \\
 Brains $16^2$ &  \tb{0.12}/\tb{12.42} &           - &  0.34/22.81 &  0.33/12.57 \\
 Brains $32^2$ &  \tb{0.17}/11.08 &           - &  0.31/14.16 &   0.18/\tb{6.80} \\

\bottomrule
\end{tabular}
}
\end{subtable}
\begin{subtable}[t]{.5\linewidth}
\centering
\fnt In-painting\\
\resizebox{\columnwidth}{!}{%
\begin{tabular}{l|cccc|cccc}
\toprule
 &  \multicolumn{4}{c|}{\brgm}    & \multicolumn{4}{c}{\patchgan} \\
Dataset & LPIPS & RMSE & PSNR & SSIM & LPIPS & RMSE & PSNR & SSIM\\
\midrule
   FFHQ &  \tb{0.19} & \tb{24.28} & \tb{21.33} & \tb{0.84} &  0.24 & 30.75 & 19.67 & 0.82 \\
  X-ray &  \tb{0.13} & \tb{13.55} & \tb{27.47} & \tb{0.91} &  0.20 & 27.80 & 22.02 & 0.86 \\
 Brains & \tb{0.09} & \tb{8.65} & \tb{30.94} & \tb{0.88} &  0.22 & 24.74 & 21.47 & 0.75 \\
\bottomrule
\end{tabular}}
\vspace{0.5em}

\fnt Human evaluation (proportion of votes for best image)\\
\resizebox{\columnwidth}{!}{%
\begin{tabular}{l|cccc}
\toprule
 Dataset &       \brgm  &    \pulse    &   \esrgan    &   \srfbn  \\

\midrule
  FFHQ $16^2$ &  \tb{42\%} &   32\% &    11\% &   15\% \\
  FFHQ $32^2$ &  39\% &   2\% &    12\% &   \tb{47\%} \\
  FFHQ $64^2$ &  14\% &   8\% &    32\% &   \tb{45\%} \\
% FFHQ $128^2$ &  14\% &   10\% &    \tb{39\%} &   38\% \\
\bottomrule
\end{tabular}
}
% \caption{Human evaluation}
\end{subtable}
\caption{(left) Evaluation on (x4) super-resolution at different input resolution levels. Reported are LPIPS/RMSE scores. (top-right) Evaluation of BRGM and SN-PatchGAN on in-painting. (bottom-right) Human evaluation showing the proportion of votes for the best super-resolution re-construction in the forced-choice pairwise comparison test. Bold numbers show best performance.}
\label{eval}
\end{table}

\subsection{Method limitations and potential negative societal impact}
\label{limitations}

In Supp. Fig. \ref{failure}, we show failure cases on the super-resolution task. The reason for the failures is likely due to the limited generalisation abilities of the StyleGAN2 generator to such unseen images. We particularly note that, as opposed to the simple inversion of Image2StyleGAN \cite{abdal2019image2stylegan}, which relies on latent variables at high resolution to recover the fine details, we cannot optimize these high-resolution latent variables, thus having to rely on the proper ability of StyleGAN2 to extrapolate from lower-level latent variables.  Another limitation of our method is the inconsistency between the downsampled input image and the given input image, which we exemplify in Supp. Figs \ref{imgdiff-ffhq} and \ref{imgdiff-medical}. We attribute this again to the limited generalisation of the generator to these unseen images. The same inconsistency also applies to in-painting, as shown in Fig. \ref{evolution}.

By leveraging models pre-trained on FFHQ, our methodology can be potentially biased towards images from people that are over-represented in the dataset. On the medical datasets, we also have biases in disease labels. For example, the MIMIC dataset contains both healthy and pneumonia lung images, but many other lung conditions are not covered, while for the brain dataset it contains healthy brains as well as Alzheimer's and Parkinson's, but does not cover rarer brain diseases. Before deployment in the real-world, further work is required to make the method robust to diverse inputs, in order to avoid negative impact on the users. 

\section{Conclusion}

We proposed a simple Bayesian framework for performing different reconstruction tasks using deep generative models such as StyleGAN2. We estimate the optimal reconstruction as the Bayesian MAP estimate, and use Variational Inference to sample from an approximate posterior of all possible solutions. We demonstrated our method on two reconstruction tasks, and on three distinct datasets, including two challenging medical datasets, obtaining competitive results in comparison with state-of-the-art models. Future work can focus on jointly optimizing the parameters of the corruption models, as well as extending to more complex corruption models.

% \section*{References}

{
\small

\bibliographystyle{unsrtnat}
\bibliography{bibliography}

}

\pagebreak

\appendix

\begin{center}
% \Large{\textbf{Bayesian Image Reconstruction using Deep Generative Models}} 
\Large{Supplementary Material} 
\end{center}

\section{Derivation of loss function for the Bayesian MAP estimate}
\label{supderiv}

We assume $w = [w_1, \dots, w_{18}] \in \mc{R}^{512 \times 18}$ is the StyleGAN2 latent vector, $I \in \mc{R}^{n \times n}$ is the corrupted input image, $G : \mc{R}^{512 \times 18} \to \mc{R}^{n_G \times n_G}$ is the StyleGAN2 generator network function, $f : \mc{R}^{n_G \times n_G} \to \mc{R}^{n_f \times n_f}$ is the corruption function, and $\phi : \mc{R}^{n_f \times n_f} \to  \mc{R}^{n_\phi \times n_\phi}$ is a function describing the perceptual network. ${n_G \times n_G}$, ${n_f \times n_f}$ and ${n_\phi \times n_\phi}$ are the resolutions of the clean image $G(w)$, corrupted image $f \ci G(w)$ and of the perceptual embedding $\phi \ci f \ci G(w)$. The full Bayesian posterior $p(w|I)$ of our model is proportional to:

\begin{equation}
\begin{aligned}
\label{fulldist}
p(w|I) \propto p(w)p(I|w) = & \prod_i \mathcal{N}(w_i|\mu, \sigma^2) \prod_{i,j} \mathcal{M}(cos^{-1}\frac{w_iw_j^T}{|w_i| |w_j|}|0, \kappa) \\
             & \mathcal{N}(I|f \ci G(w), \sigma_{pixel}^2 \mc{I}_{n_f^2})\ \mathcal{N}(\phi(I)|\phi \ci f \ci G(w), \sigma_{percept}^2 \mc{I}_{n_{\phi}^2}) 
\end{aligned}
\end{equation}

where $\mu \in \mc{R}$, $\sigma \in \mc{R}$ are means and standard deviations of the prior on $w_i$, $\mathcal{M}(.|0, \kappa)$ is the von Mises distribution\footnote{The von-Mises distribution is the analogous of the Gaussian distribution over angles $[0-2\pi]$. $\mathcal{M}(.|\mu, \kappa)$ is analogous to $\mathcal{N}(.|\mu, \sigma)$, where $\kappa^{-1}=\sigma^2$} with mean zero and scale parameter $\kappa$, and $\sigma_{pixel}^2 \mc{I}_{n_f^2}$ and $\sigma_{percept}^2 \mc{I}_{n_{\phi}^2}$ are identity matrices scaled by variance terms. %For notational simplicity, we will drop the dimensionality of the identity matrices, and just refer to them as $\mc{I}$. 

The Bayesian MAP estimate is the vector $w^*$ that maximizes Eq. \ref{fulldist}, and provides the most likely vector $w$ that could have generated input image $I$:

\begin{equation}
\begin{aligned}
\label{bmap}
& w^* = \argmax_w p(w)p(I|w) = \argmax_w \prod_i \mathcal{N}(w_i|\mu, \sigma^2) \prod_{i,j} \mathcal{M}(cos^{-1}\frac{w_iw_j^T}{|w_i| |w_j|}|0, \kappa) \\
& \mathcal{N}(I|f \ci G(w), \sigma_{pixel}^2 \mc{I}_{n_f^2}) \mathcal{N}(\phi(I)|\phi \ci f \ci G(w), \sigma_{percept}^2 \mc{I}_{n_{\phi}^2}) 
\end{aligned}
\end{equation}

Since logarithm is a strictly increasing function that won't change the output of the $\argmax_w$ operator, we take the logarithm to simplify Eq. \ref{bmap} to:

\begin{equation}
\begin{aligned}
\label{log}
& w^* = \argmax_w \sum_i log\  \mathcal{N}(w_i|\mu, \sigma^2) + \sum_{i,j} log\  \mathcal{M}(cos^{-1}\frac{w_iw_j^T}{|w_i| |w_j|}|0, \kappa) + \\
& log\ \mathcal{N}(I|f \ci G(w), \sigma_{pixel}^2 \mc{I}_{n_f^2}) + log\ \mathcal{N}(\phi(I)|\phi \ci f \ci G(w), \sigma_{percept}^2 \mc{I}_{n_{\phi}^2}) 
\end{aligned}
\end{equation}

We expand the probability density functions of each distribution to get:

\begin{equation}
\begin{aligned}
\label{constants}
& w^* = \argmax_w \sum_i C_1 - \frac{(w_i-\mu)^2}{2\sigma_i^2} + \sum_{i,j} C_2 + \kappa cos(cos^{-1}\frac{w_iw_j^T}{|w_i| |w_j|}) \\
& + C_1 - \frac{1}{2} (I - f \ci G(w))^T(\sigma_{pixel}^{-2} \mc{I}_{n_f^2})(I-f \ci G(w)) \\
& + C_2 - \frac{1}{2} (I - \phi \ci f \ci G(w))^T(\sigma_{percept}^{-2} \mc{I}_{n_{\phi}^2})(I-\phi \ci f \ci G(w)) \\
\end{aligned}
\end{equation}

where $C_1 = log\ (2 \pi \sigma_i^2)^{-\frac{1}{2}}$, $C_2 = log\ (-2 \pi I_0(\kappa))$, $C_3 = log\ ((2\pi)^{n^2} |\sigma_{pixel} \mc{I}_{n_f^2}| )^{-\frac{1}{2}}$ and $C_4 = log\ ((2\pi)^{m^2} | \sigma_{percept} \mc{I}_{n_{\phi}^2} | )^{-\frac{1}{2}} $ are constants with respect to $w$, so we can ignore them. 

We remove the constants, multiply by (-2), which requires switching to the $\argmin$ operator, to get:

\begin{equation}
\begin{aligned}
\label{bfinalappendix}
& w^* = \argmin_w \sum_i \left(\frac{w_i-\mu}{\sigma_i}\right)^2 - 2\kappa \sum_{i,j} \frac{w_iw_j^T}{|w_i| |w_j|} \\
& + \sigma_{pixel}^{-2} \norm{I - f \ci G(w)}_2^2 + \sigma_{percept}^{-2} \norm{I - \phi \ci f \ci G(w)}_2^2 \\
\end{aligned}
\end{equation}

This is equivalent to Eq. \ref{fullloss}, which finishes our proof.

\begin{figure}
 \centering
 \begin{tikzpicture}[scale=0.9, every node/.style={scale=0.9}]
    % Define nodes
\node[latent, label={$=f \ci G(w)$}]                               (Ilr) {$I_{COR}$}; % I_{LR}
\node[obs, right=of Ilr, xshift=0.2cm]                               (I) {$I$}; % I
\node[latent, left=of Ilr, xshift=-0.1cm, label={$=G(w)$}] (Ihr) {$I_{CLN}$}; % I_{HR}
\node[latent, left=of Ihr, xshift=-0.2cm] (w) {$w$}; % w

\node[obs, above=of w, xshift=-1cm] (mu) {$\mu$}; 
\node[obs, above=of w, xshift=0cm] (sigma) {$\sigma$}; 
\node[obs, above=of w, xshift=1cm] (kappa) {$\kappa$};  

% variational params
\node[latent, left=of w, xshift=-0.7cm] (muv) {$\mu_v$}; % \mu_v
\node[latent, above=of muv, yshift=-0.7cm] (sigmav) {$\sigma_v$}; % sigma_v
\node[latent, left=of sigmav, xshift=-0.0cm] (rhov) {$\rho_v$}; % rho_v

\node[obs, above=of I, xshift=-0.8cm] (sigmapix) {\small{$\sigma_{pixel}$}}; 
\node[obs, above=of I, xshift=0.8cm] (sigmaperc) {\small{$\sigma_{perc}$}};
% Connect the nodes
%\edge {w} -- node [text width=1.5cm,midway,above] {My very} {Ihr} ; %
% \draw[arrow] (w) -- node [text width=2.5cm,midway,above] {My very} (Ihr);
\path (w) edge [->, >={triangle 45}]   (Ihr)  ;%
\path (w) -- node[above] {G}  (Ihr)  ;%

\edge {Ihr} {Ilr} ; %
\path (Ihr) -- node[above] {f}  (Ilr)  ;%
\edge {Ilr} {I} ; %

\edge {muv} {w};
\edge {sigmav} {w};
\edge {rhov} {sigmav};

\edge {mu} {w} ; %
\edge {sigma} {w} ; %
\edge {kappa} {w} ; %

\edge {sigmapix} {I} ; %
\edge {sigmaperc} {I} ; %

  % Plates
% \plate {yx} {(x)(y)} {$N$} ;
\draw[red,thick,line width=0.15mm] (-9.5,-0.6)  rectangle (-6,1.7);
\node[above] at (-7.7,1.70) {\textcolor{red}{variational parameters}};

 \end{tikzpicture}
\caption{\tc{Graphical model of our method. In gray shade are known observations or parameters: the input corrupted image $I$, the parameters $\mu$, $\sigma$ and $\kappa$ defining the prior on latent vector $w$, and $\sigma_{pixel}$, $\sigma_{percept}$, the parameters defining the noise model over $I$. In the red box are the variational parameters $\mu_v$, $\sigma_v$ and $\rho_v$ defining an approximated Gaussian posterior over $w$ (section \ref{samplingmethod}). Unknown latent variables (in white), to be estimated, are $w$, the latent vectors of StyleGAN, $I_{CLN}$, the clean image, and $I_{COR}$, the corrupted image simulated through the pipeline. Transformation $G$ is modelled by the StyleGAN2 generator, while $f$  by a known corruption model (e.g. downsampling with a known kernel).} }
\label{modeldiagram}
\end{figure}
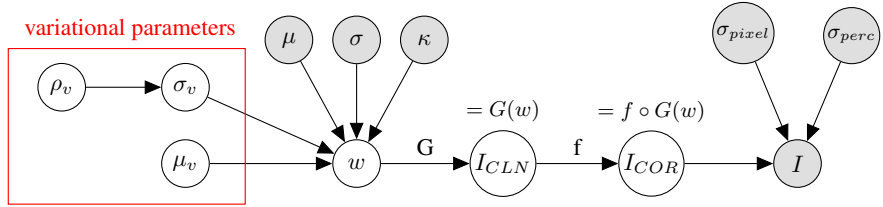

%%%% Training StyleGAN results

\section{Training StyleGAN2}
\label{training}

In Fig. \ref{generated}, we show uncurated images generated by the cross-validated StyleGAN2 trained on our medical datasets, along with a few real examples. For the high-resolution X-rays, we notice that the image quality is very good, although some artifacts are still present: some text tags are not properly generated, some bones and rib contours are wiggly, and the shoulder bones show less contrast. For the brain dataset, we do not notice any clear artifacts, although we did not assess distributional preservation of regional volumes as in \cite{tudosiu2020neuromorphologicaly}. For the cross-validated FFHQ model, we obtained an FID of 4.01, around 0.7 points higher than the best result of 3.31 reported for config-e \cite{karras2020analyzing}.

\FloatBarrier

\newcommand{\genF}{\fld/00610-generate-images-ffhq/}
\newcommand{\genX}{\fld/00618-generate-images-xray/}
\newcommand{\genB}{\fld/00619-generate-images-brains/}

\newcommand{\realX}{images}
%00068d26-8d583659-af7de1da-fc6c0476-d94aada1
%0006a816-5140e307-815c7b9f-4856cbc6-670b7e6d
%0007606b-8bbb964c-bd4a9a11-2c49927b-1a4425b7
%0007f9a9-e098be9b-81942dab-995b3d31-4b19f648
%000bda97-398156dd-880483aa-ccfbfad3-eebeed57
%000debcd-537f4956-6d8801d4-75c9e646-a922d57c

\newcommand{\realB}{images}
%ADNI_ADNI-3T-FS-5.3-Long_242881.long.027_S_2183_base_mri_talairach_norm
%OASIS_OAS1_0043_MR1_mri_talairach_norm
%OASIS_OAS1_0229_MR1_mri_talairach_norm
%ADNI_ADNI-1.5T-FS-5.3-Long_86304.long.002_S_1070_base_mri_talairach_norm
%GSP_101207_PD57VU_FS_mri_talairach_norm
%COBRE_0040073_mri_talairach_norm
%GSP_100423_NA42WH_FS_mri_talairach_norm

\renewcommand{\w}{2.7cm}

% results of standard StyleGAN training on X-ray and brains
\begin{figure*}
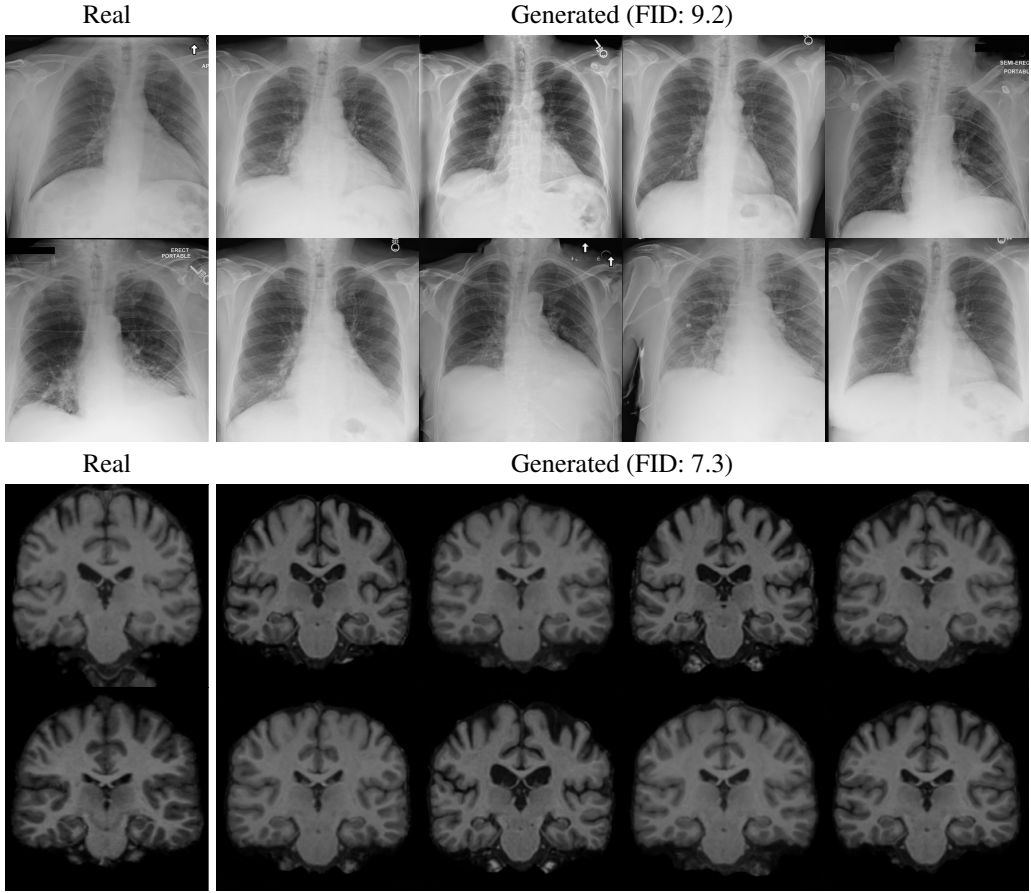

\centering
\setlength{\tabcolsep}{0pt}
\renewcommand{\arraystretch}{1.5}% Spread rows out...
\begin{tabular}{c||cccc}\arrayrulecolor{white}
Real & \multicolumn{4}{c}{Generated (FID: 9.2)}\\
\inc{\realX/00068d26-8d583659-af7de1da-fc6c0476-d94aada1}&\inc{\genX/seed0001}&\inc{\genX/seed0002}&\inc{\genX/seed0003}&\inc{\genX/seed0004}\\
\inc{\realX/0007f9a9-e098be9b-81942dab-995b3d31-4b19f648}&\inc{\genX/seed0005}&\inc{\genX/seed0006}&\inc{\genX/seed0007}&\inc{\genX/seed0008}\\

Real & \multicolumn{4}{c}{Generated (FID: 7.3)} \\

\inc{\realB/COBRE_0040073_mri_talairach_norm}&\inc{\genB/seed0001}&\inc{\genB/seed0002}&\inc{\genB/seed0003}&\inc{\genB/seed0004} \\
\inc{\realB/OASIS_OAS1_0043_MR1_mri_talairach_norm}&\inc{\genB/seed0005}&\inc{\genB/seed0006}&\inc{\genB/seed0007}&\inc{\genB/seed0008} \\

\end{tabular}
\caption{Uncurated images generated by our StyleGAN2 generator trained on the chest X-ray dataset (MIMIC III) (top) and the brain dataset (bottom). Left images are random examples of real images from the actual datasets, while the right-side images are generated. The image quality is relatively good, albeit some anatomical artifacts are still observed, such as incomplete labels, wiggly bones or discontinuous wires.} % TODO: add arrows pointing to artefacts. 
\label{generated}
\end{figure*}

\section{Inference times of our method}
\label{times}

% inference times
The inference time of our method is as follows: for MAP inference, it takes between 32-34 seconds for 500 iterations on a 1024x1024 image, while for fitting the variational posterior parameters ($\mu$, $\rho$) it takes approximately 2.5 minutes for 500 iterations. Once the variational posterior is fit, the model can generate any arbitrary number of samples instantaneously.

\renewcommand{\w}{2.9cm}

\begin{figure*}
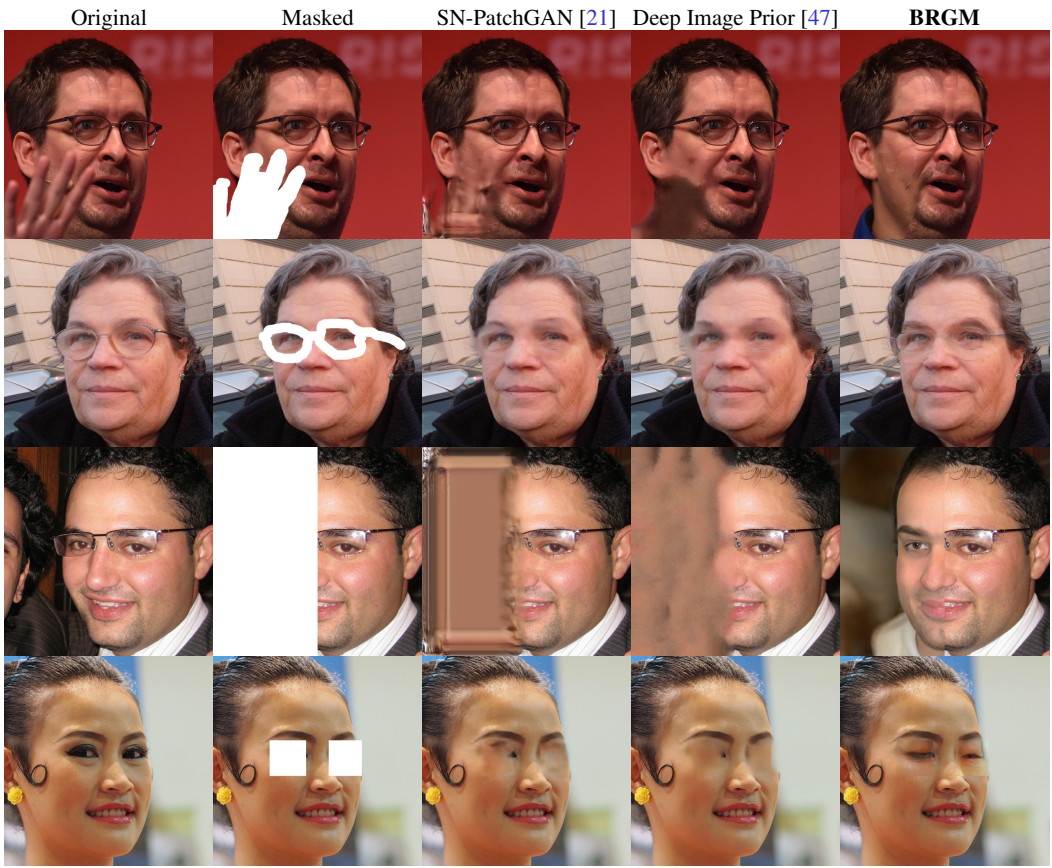

\begin{center}
%\begin{table}[]
%\centering
\setlength{\tabcolsep}{0pt}
\renewcommand{\arraystretch}{0.5}% Spread rows out...
\resizebox{\columnwidth}{!}{%
\begin{tabu}{ccccc}%\arrayrulecolor{LightGray}
\rowfont{\footnotesize} Original & Masked & \patchgan & \dip & \brgm \\

% examples using arbitrary masks
\inc{\inFH{\nrIthree}}&
\inc{\inFM{\nrIthree}}&
\inc{\inFP{\nrIthree}}&
\inc{\inFD{\nrIthree}}&
\inc{\inFO{\nrIthree}}\\

\inc{\inFH{\nrIfour}}&
\inc{\inFM{\nrIfour}}&
\inc{\inFP{\nrIfour}}&
\inc{\inFD{\nrIfour}}&
\inc{\inFO{\nrIfour}}\\

% examples using rectangular masks
\inc{\inFEH{\nrIfive}}&
\inc{\inFEM{\nrIfive}}&
\inc{\inFEP{\nrIfive}}&
\inc{\inFED{\nrIfive}}&
\inc{\inFEO{\nrIfive}}\\

\inc{\inFEH{\nrIsix}}&
\inc{\inFEM{\nrIsix}}&
\inc{\inFEP{\nrIsix}}&
\inc{\inFED{\nrIsix}}&
\inc{\inFEO{\nrIsix}}\\

\end{tabu}
}
\caption{Uncurated in-painting examples by BRGM on the FFHQ dataset, compared against \patchgan\ and \dip.}% \cite{karras2019style}.}
\label{ffhq-inpainting}
\end{center}
\end{figure*}

\begin{figure*}
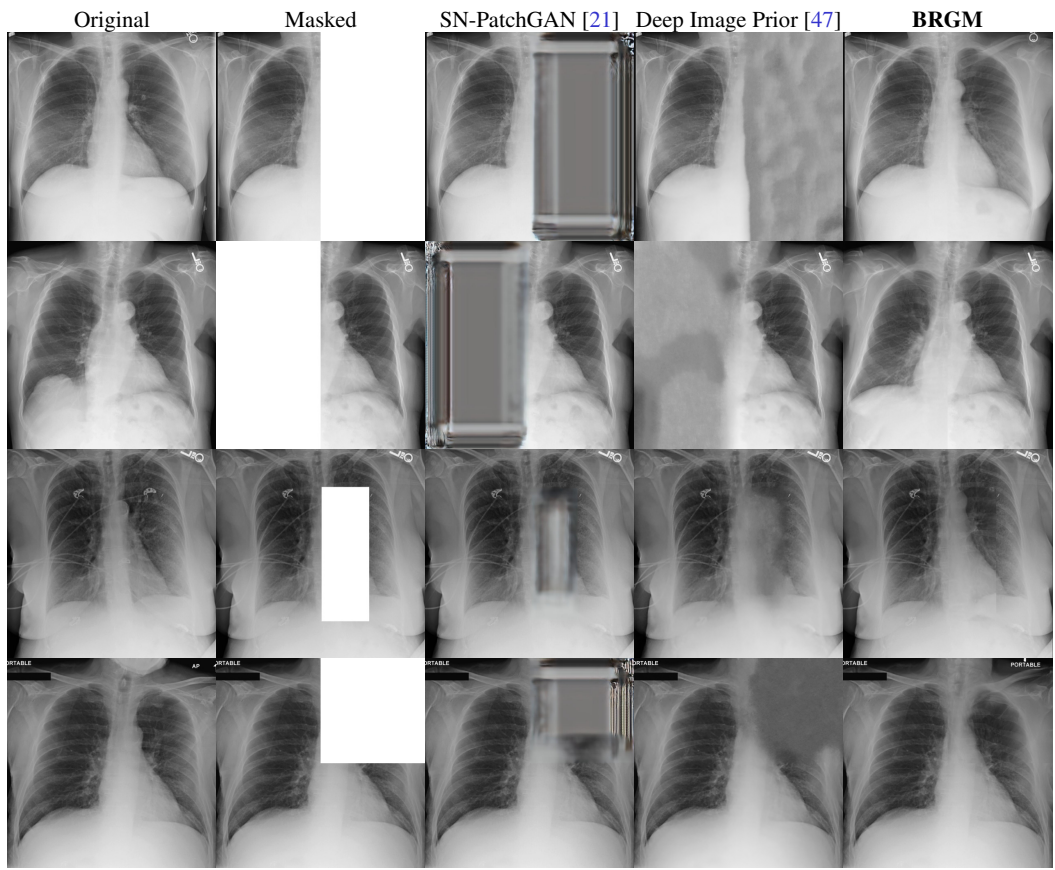

\begin{center}
%\begin{table}[]
%\centering
\setlength{\tabcolsep}{0pt}
\renewcommand{\arraystretch}{0.5}% Spread rows out...
\resizebox{\columnwidth}{!}{%
\begin{tabu}{ccccc}%\arrayrulecolor{LightGray}
\rowfont{\footnotesize} Original & Masked & \patchgan & \dip & \brgm \\

\inc{\inXH{\nrIM}}&
\inc{\inXM{\nrIM}}&
\inc{\inXP{\nrIM}}&
\inc{\inXD{\nrIM}}&
\inc{\inXO{\nrIM}}\\

\inc{\inXH{\nrIMtwo}}&
\inc{\inXM{\nrIMtwo}}&
\inc{\inXP{\nrIMtwo}}&
\inc{\inXD{\nrIMtwo}}&
\inc{\inXO{\nrIMtwo}}\\

\inc{\inXH{\nrIMthree}}&
\inc{\inXM{\nrIMthree}}&
\inc{\inXP{\nrIMthree}}&
\inc{\inXD{\nrIMthree}}&
\inc{\inXO{\nrIMthree}}\\

\inc{\inXH{\nrIMfour}}&
\inc{\inXM{\nrIMfour}}&
\inc{\inXP{\nrIMfour}}&
\inc{\inXD{\nrIMfour}}&
\inc{\inXO{\nrIMfour}}\\

\end{tabu}
}
\caption{Uncurated in-painting examples by BRGM on the Chest X-ray dataset, compared against \patchgan\ and \dip.}% \cite{johnson2016mimic}.}
\label{xray-inpainting}
\end{center}
\end{figure*}

\begin{figure*}
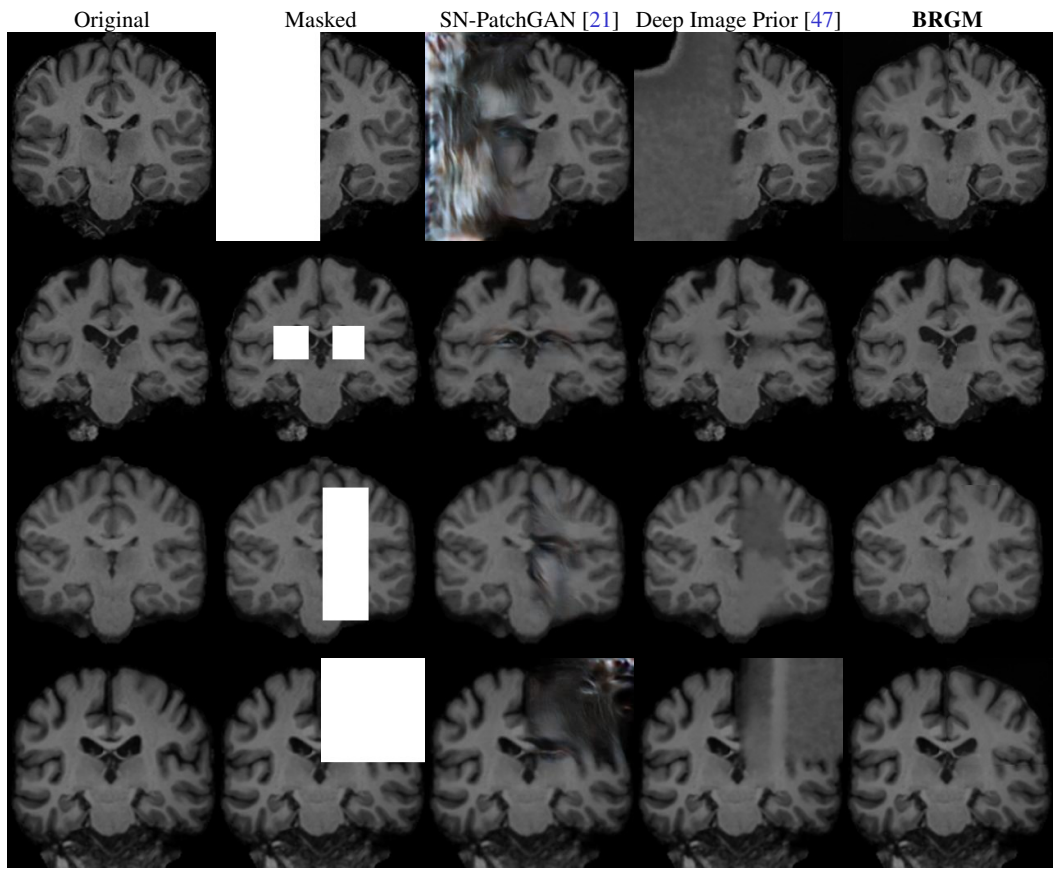

\begin{center}
%\begin{table}[]
%\centering
\setlength{\tabcolsep}{0pt}
\renewcommand{\arraystretch}{0.5}% Spread rows out...
\resizebox{\columnwidth}{!}{%
\begin{tabu}{ccccc}%\arrayrulecolor{LightGray}
\rowfont{\footnotesize} Original & Masked & \patchgan & \dip & \brgm \\

\inc{\inBH{\nrIMB}}&
\inc{\inBM{\nrIMB}}&
\inc{\inBP{\nrIMB}}&
\inc{\inBD{\nrIMB}}&
\inc{\inBO{\nrIMB}}\\

\inc{\inBH{\nrIMBtwo}}&
\inc{\inBM{\nrIMBtwo}}&
\inc{\inBP{\nrIMBtwo}}&
\inc{\inBD{\nrIMBtwo}}&
\inc{\inBO{\nrIMBtwo}}\\

\inc{\inBH{\nrIMBthree}}&
\inc{\inBM{\nrIMBthree}}&
\inc{\inBP{\nrIMBthree}}&
\inc{\inBD{\nrIMBthree}}&
\inc{\inBO{\nrIMBthree}}\\

\inc{\inBH{\nrIMBfour}}&
\inc{\inBM{\nrIMBfour}}&
\inc{\inBP{\nrIMBfour}}&
\inc{\inBD{\nrIMBfour}}&
\inc{\inBO{\nrIMBfour}}\\

\end{tabu}
}
\caption{Uncurated in-painting examples by BRGM on the brain dataset, compared against \patchgan\ and \dip.}% \cite{dalca2018anatomical}.}
\label{brain-inpainting}
\end{center}
\end{figure*}

% \FloatBarrier

% \section{Performance evaluation}
% \label{supp-eval}

\newcommand{\mcc}[1]{\multicolumn{1}{c}{#1}}

\begin{table*}
\centering
\setlength{\tabcolsep}{2pt}
\resizebox{0.9\columnwidth}{!}{%
\begin{tabular}{l|ccc|ccc|ccc|ccc}
\toprule
Dataset &       \multicolumn{3}{c}{\brgm}  &    \multicolumn{3}{c}{\pulse}    &   \multicolumn{3}{c}{\esrgan}    &   \multicolumn{3}{c}{\srfbn}  \\
% \midrule
  & PSNR$\uparrow$ & SSIM$\uparrow$ & MAE$\downarrow$  & PSNR$\uparrow$ & SSIM$\uparrow$ & MAE$\downarrow$ & PSNR$\uparrow$ & SSIM$\uparrow$ & MAE$\downarrow$ & PSNR$\uparrow$ & SSIM$\uparrow$ & MAE$\downarrow$\\
\midrule
   FFHQ $16^2$ &  20.13 & 0.74 & 17.46 &  19.51 & 0.68 & 19.20 &  18.91 & 0.69 & 20.01 &  21.43 & 0.76 & 15.01 \\
   FFHQ $32^2$ &  22.74 & 0.74 & 12.52 &  15.37 & 0.35 & 33.13 &  21.10 & 0.72 & 14.53 &   26.28 & 0.89 & 7.46 \\
   FFHQ $64^2$ &  24.16 & 0.70 & 10.63 &  15.74 & 0.37 & 31.54 &  23.14 & 0.72 & 10.94 &   28.96 & 0.90 & 5.21 \\
%   FFHQ $128^2$ &  24.29 & 0.65 & 10.53 &  17.23 & 0.45 & 25.87 &   24.53 & 0.70 & 9.20 &   30.98 & 0.90 & 4.06 \\
   X-ray $16^2$ &   27.14 & 0.91 & 7.45 &             - & - & - &  25.17 & 0.87 & 10.14 &   26.88 & 0.92 & 7.63 \\
   X-ray $32^2$ &   27.84 & 0.84 & 6.77 &             - & - & - &   26.44 & 0.81 & 8.36 &   31.80 & 0.95 & 3.71 \\
   X-ray $64^2$ &   27.62 & 0.79 & 6.63 &             - & - & - &   29.47 & 0.87 & 5.33 &   33.91 & 0.95 & 2.47 \\
%   X-ray $128^2$ &   27.34 & 0.78 & 6.74 &             - & - & - &   30.78 & 0.87 & 4.28 &   35.59 & 0.96 & 1.77 \\
 Brains $16^2$ &   26.33 & 0.84 & 7.29 &             - & - & - &  21.06 & 0.60 & 14.27 &   26.21 & 0.77 & 8.62 \\
 Brains $32^2$ &   27.30 & 0.81 & 6.54 &             - & - & - &   25.23 & 0.78 & 8.35 &   31.60 & 0.93 & 3.86 \\

\bottomrule
\end{tabular}
}
\caption{Additional performance metrics (PSNR, SSIM and MAE) for the super-resolution evaluation.}
\label{eval-super-resolution-supplementary}
\end{table*}

\begin{table*}
\centering
\setlength{\tabcolsep}{3pt}
\resizebox{\columnwidth}{!}{%
\begin{tabular}{l|cc|cc|rr|rr}
\toprule
Dataset &       \multicolumn{2}{c}{\brgm}  &    \multicolumn{2}{c}{\pulse}    &   \multicolumn{2}{c}{\esrgan}    &   \multicolumn{2}{c}{\srfbn}  \\
%  Dataset &       \brgm  &    \pulse    &   \esrgan    &   \srfbn  \\
%         &       LP/RM  &    LP/RM    &   LP/RM    &   LP/RM  \\

 & \lpi & \rms  & \lpi & \rms & \lpi & \rms & \lpi & \rms\\
\midrule
  FFHQ $16^2$ & 0.24 $\pm$ 0.07 & 25.66 $\pm$ 6.13 & 0.29 $\pm$ 0.07 & 27.14 $\pm$ 4.08 & 0.35 $\pm$ 0.07 & 29.32 $\pm$ 6.52 & 0.33 $\pm$ 0.05 & 22.07 $\pm$ 4.47 \\
  FFHQ $32^2$ & 0.30 $\pm$ 0.07 & 18.93 $\pm$ 3.95 & 0.48 $\pm$ 0.08 & 42.97 $\pm$ 3.78 & 0.29 $\pm$ 0.05 & 23.02 $\pm$ 5.48 & 0.23 $\pm$ 0.04 & 12.73 $\pm$ 3.08 \\
  FFHQ $64^2$ & 0.36 $\pm$ 0.06 & 16.07 $\pm$ 3.21 & 0.53 $\pm$ 0.07 & 41.31 $\pm$ 3.57 & 0.26 $\pm$ 0.05 & 18.37 $\pm$ 5.06 &  0.23 $\pm$ 0.04 & 9.40 $\pm$ 2.48 \\
 FFHQ $128^2$ & 0.34 $\pm$ 0.05 & 15.84 $\pm$ 3.23 & 0.57 $\pm$ 0.06 & 34.89 $\pm$ 2.21 & 0.15 $\pm$ 0.05 & 15.84 $\pm$ 4.83 &  0.09 $\pm$ 0.02 & 7.55 $\pm$ 2.30 \\
 X-ray $16^2$ & 0.18 $\pm$ 0.05 & 11.61 $\pm$ 3.22 &                              - & - & 0.32 $\pm$ 0.07 & 14.67 $\pm$ 4.48 & 0.37 $\pm$ 0.04 & 12.28 $\pm$ 4.39 \\
 X-ray $32^2$ & 0.23 $\pm$ 0.05 & 10.47 $\pm$ 2.04 &                              - & - & 0.32 $\pm$ 0.05 & 12.56 $\pm$ 3.34 &  0.21 $\pm$ 0.03 & 6.84 $\pm$ 2.01 \\
 X-ray $64^2$ & 0.31 $\pm$ 0.04 & 10.58 $\pm$ 1.81 &                              - & - &  0.30 $\pm$ 0.03 & 8.67 $\pm$ 1.86 &  0.22 $\pm$ 0.02 & 5.32 $\pm$ 1.44 \\
X-ray $128^2$ & 0.27 $\pm$ 0.03 & 10.53 $\pm$ 1.91 &                              - & - &  0.20 $\pm$ 0.02 & 7.19 $\pm$ 1.34 &  0.07 $\pm$ 0.01 & 4.33 $\pm$ 1.30 \\
Brains $16^2$ & 0.12 $\pm$ 0.03 & 12.42 $\pm$ 1.71 &                              - & - & 0.34 $\pm$ 0.04 & 22.81 $\pm$ 3.26 & 0.33 $\pm$ 0.03 & 12.57 $\pm$ 1.51 \\
Brains $32^2$ & 0.17 $\pm$ 0.03 & 11.08 $\pm$ 1.29 &                              - & - & 0.31 $\pm$ 0.03 & 14.16 $\pm$ 2.36 &  0.18 $\pm$ 0.03 & 6.80 $\pm$ 1.14 \\

\bottomrule
\end{tabular}
}
\caption{Performance metrics for super-resolution as in Table \ref{eval} (left), but additionally including the standard deviation of scores across the 100 test images.}
\label{eval-super-resolution-supplementary-lpips-std}
\end{table*}

\FloatBarrier

\section{Additional Evaluation Results}
\label{supp-eval-humans}

\begin{figure*}
\begin{center}
\includegraphics[width=0.8\textwidth]{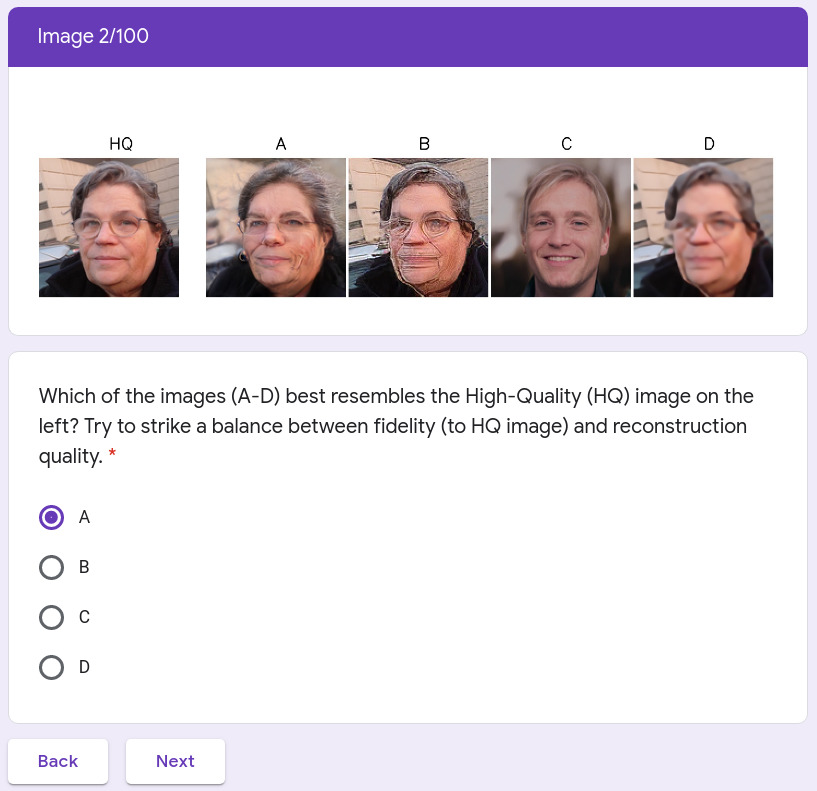}
\caption{Setup of our human study, using a forced-choice pairwise comparison design. Each rater is shown a true, high-quality image on the left, and four potential reconstructions (A-D) by different algorithms. They have to select which reconstruction best resembled the HQ image. }
\label{human-study}
\end{center}
\end{figure*}

To evaluate human perceptual quality, we performed a forced-choice pairwise comparison test as shown in Fig. \ref{human-study}. Each rater is shown a true, high-quality image on the left, and four potential reconstructions they have to choose from.  For each input resolution level ($16^2$, $32^2$, \dots), we ran the human evaluation on 20 raters using 100 pairs of 5 images each (total of 500 images per experiment shown to each rater). We launched all human evaluations on Amazon Mechanical Turk. We paid \$346 for the crowdsourcing effort, which gave workers an hourly wage of approximately \$9. We obtained IRB approval for this study from our institution.

% \section{Additional Sampling Results}

\renewcommand{\w}{1.8cm}

\renewcommand{\is}{52116}
\renewcommand{\step}{step700}

\begin{figure*}
\setlength{\tabcolsep}{0pt}
\renewcommand{\arraystretch}{1.5}% Spread rows out...
\begin{tabular}{c|c||c||ccccc}\arrayrulecolor{white}
& True & Est. Mean & Sample 1 & Sample 2 & Sample 3 & Sample 4 & Sample 5\\
HR & \inc{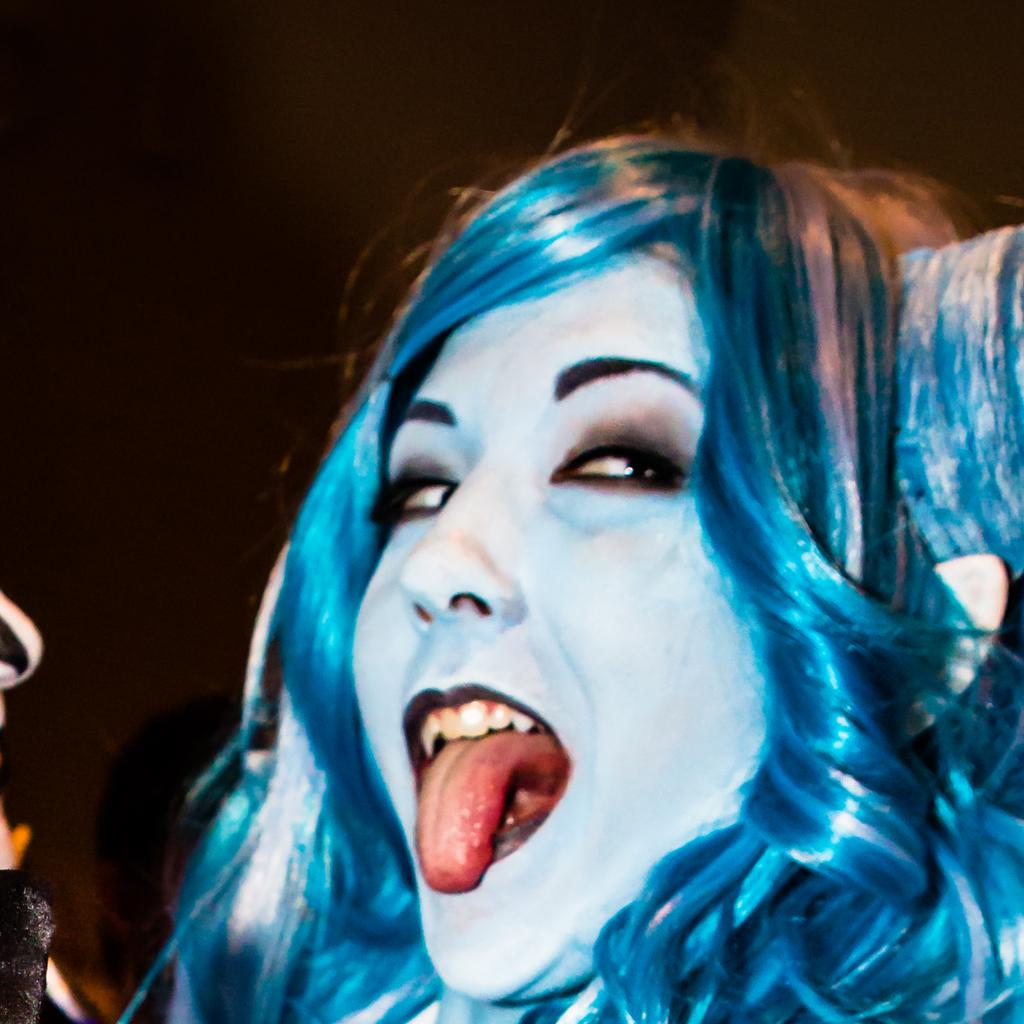} & \inc{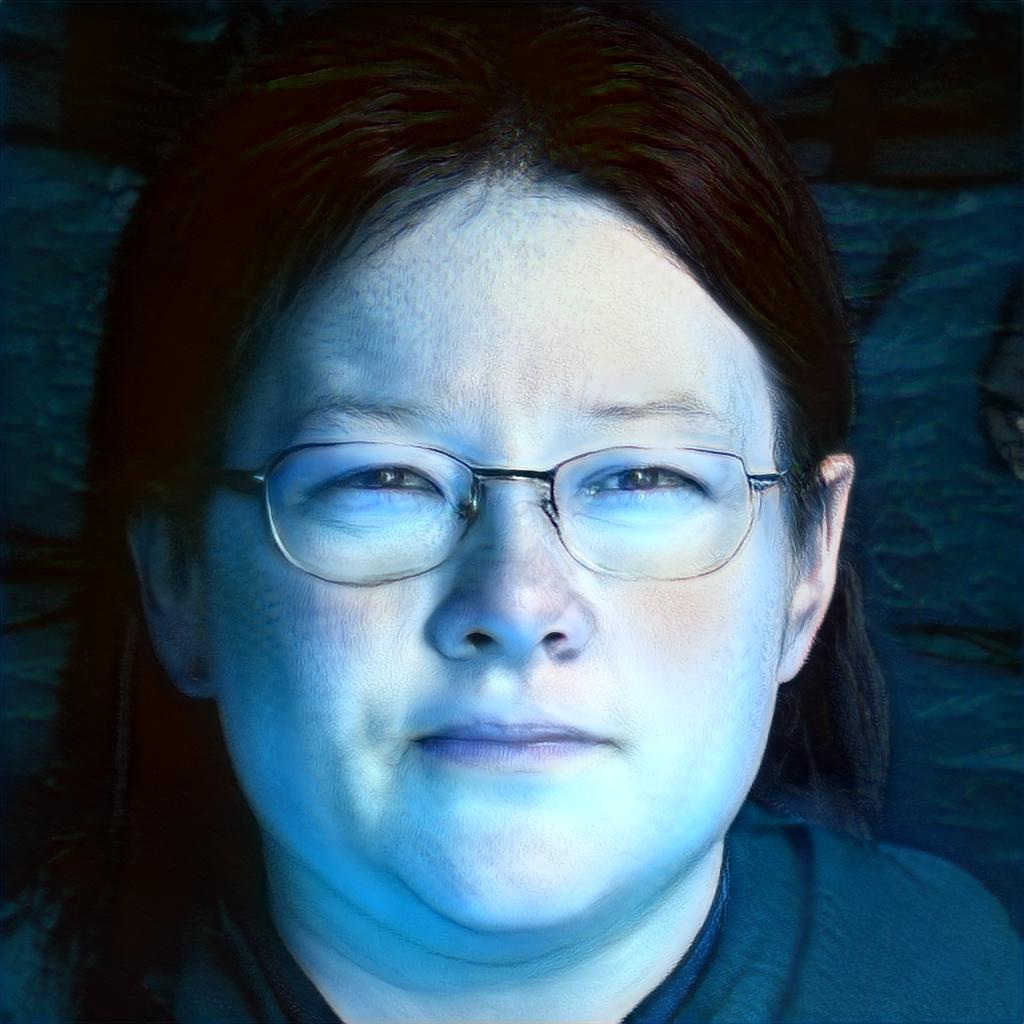} & \inc{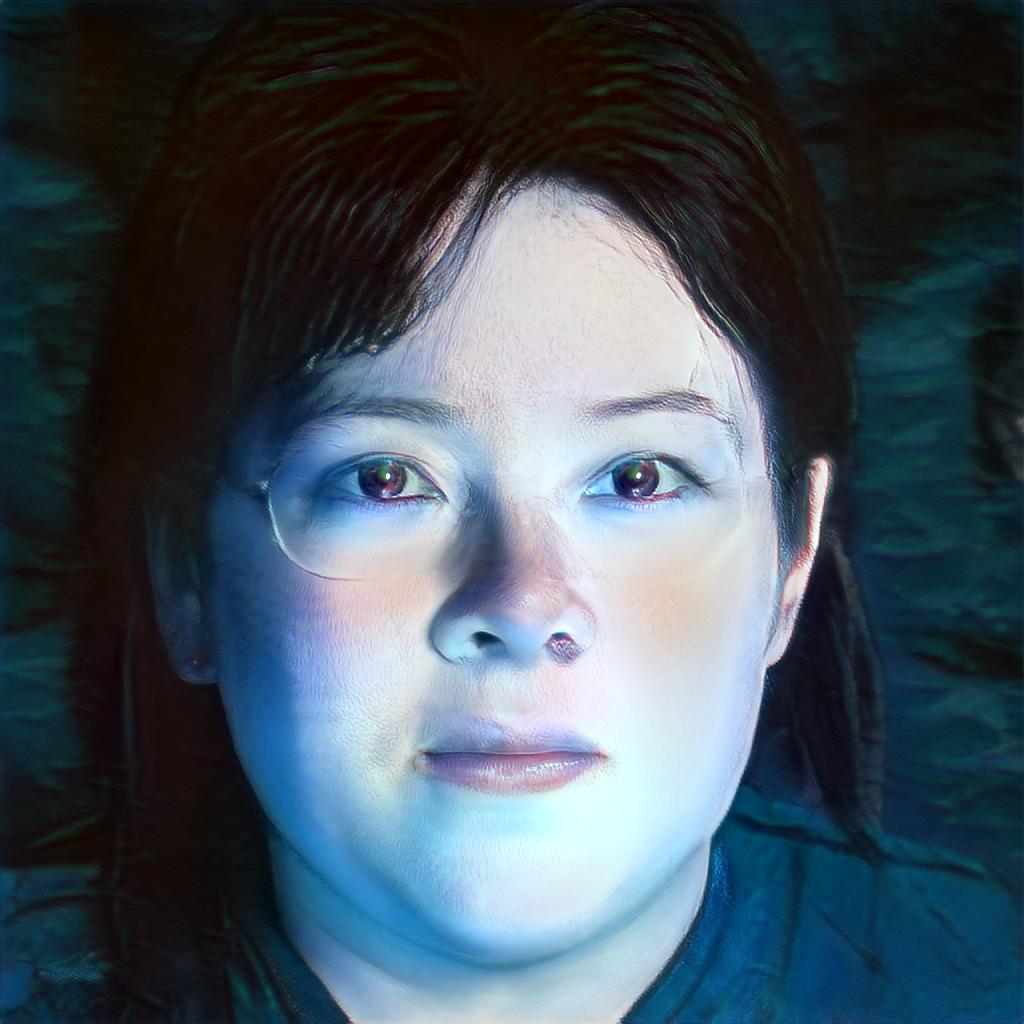} & \inc{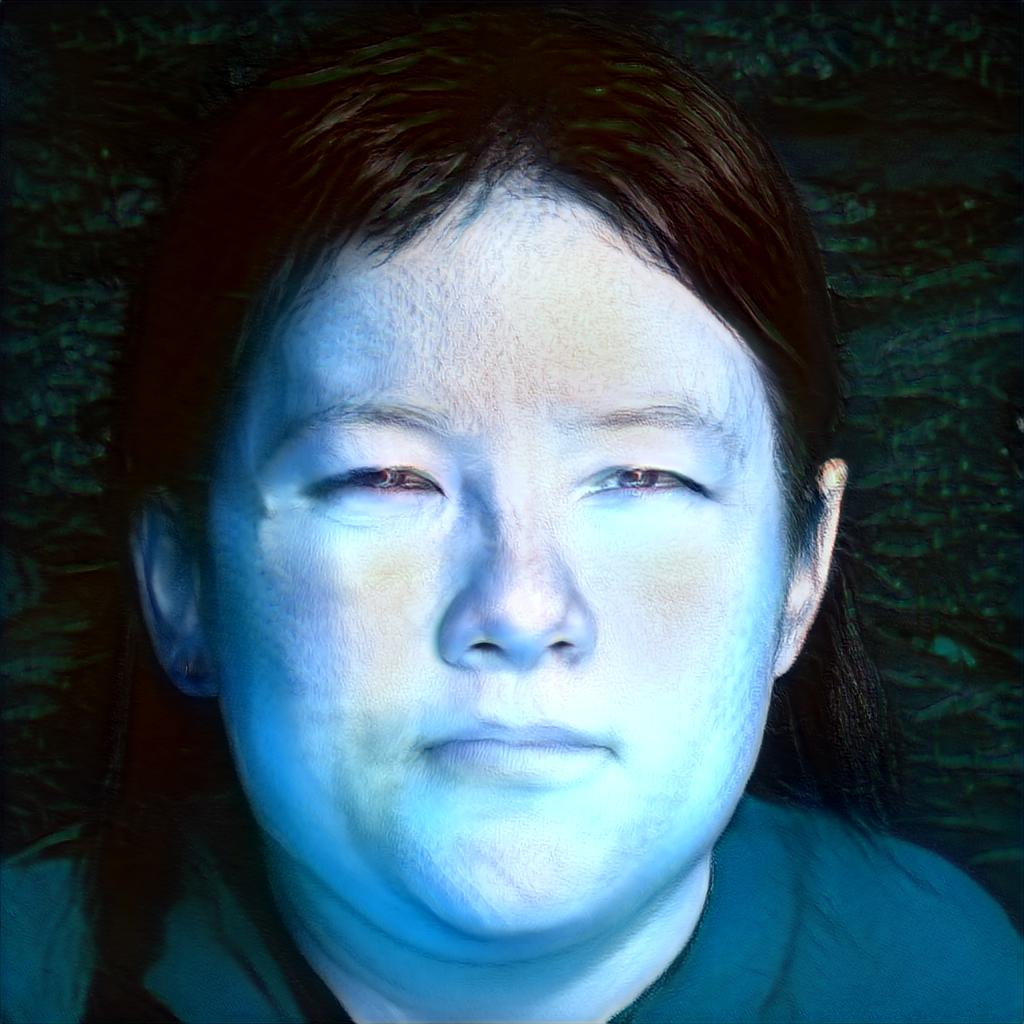} & \inc{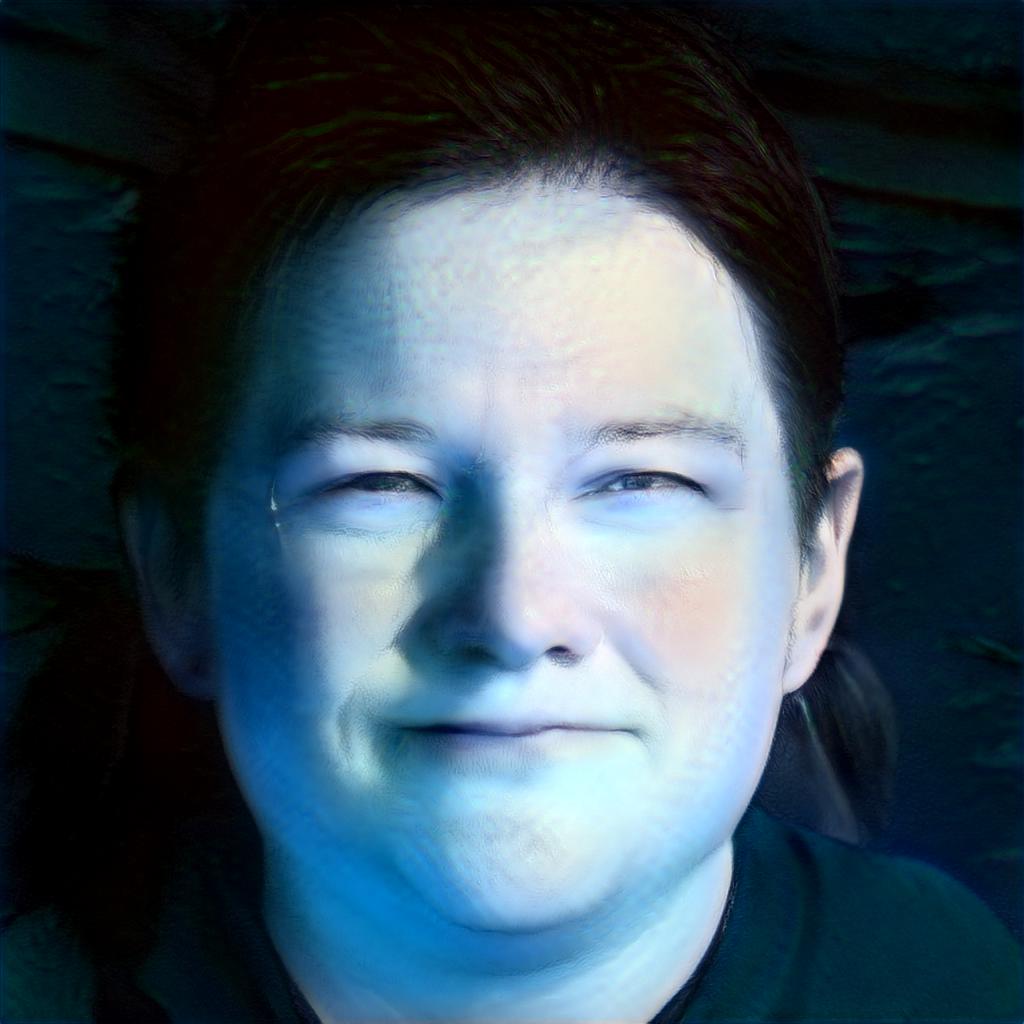} & \inc{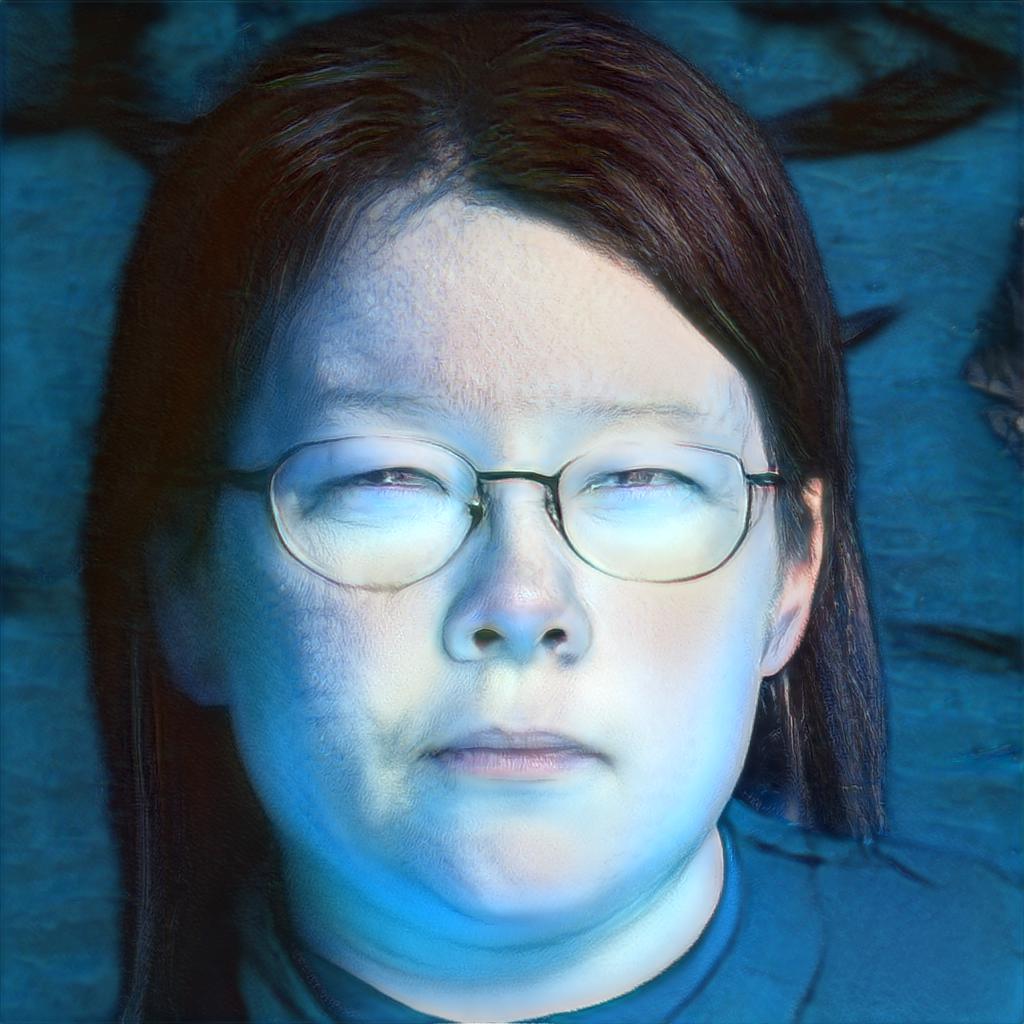} & \inc{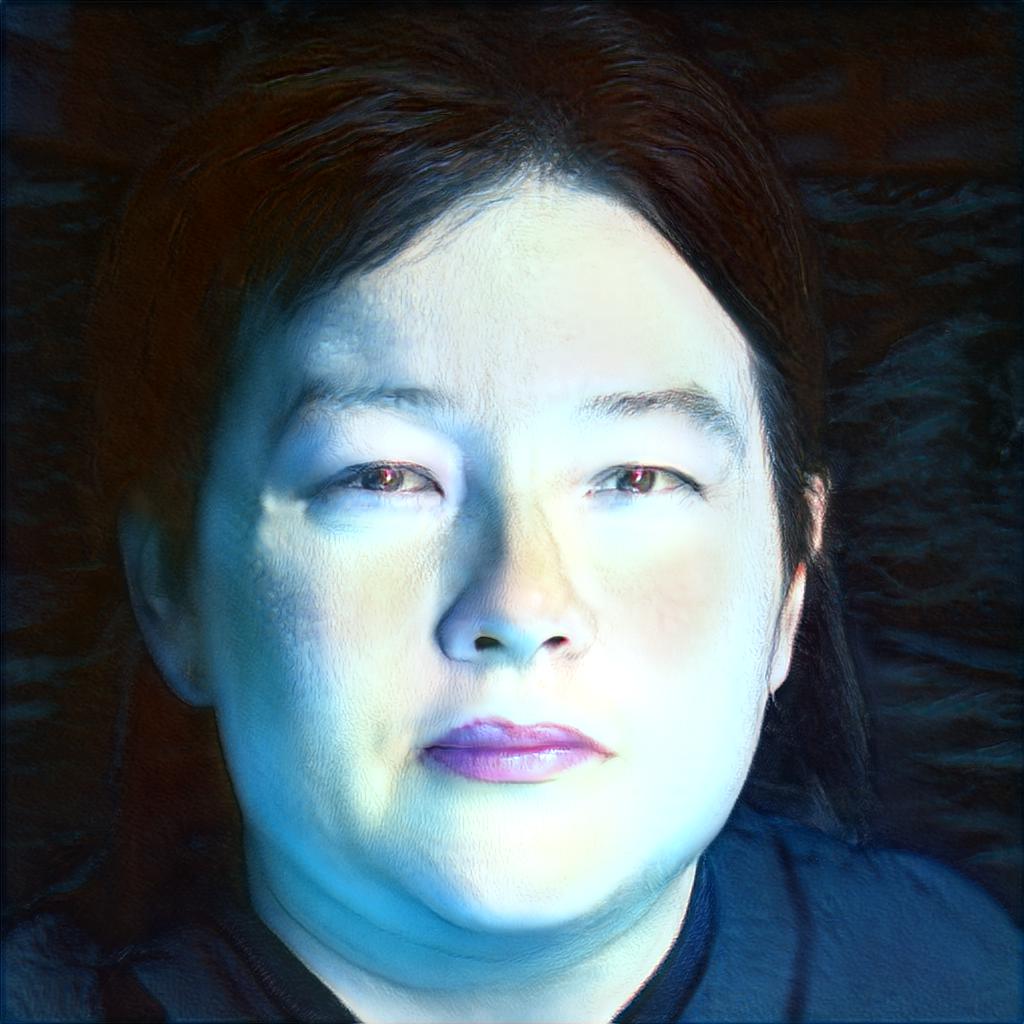}\\
 
LR & \inc{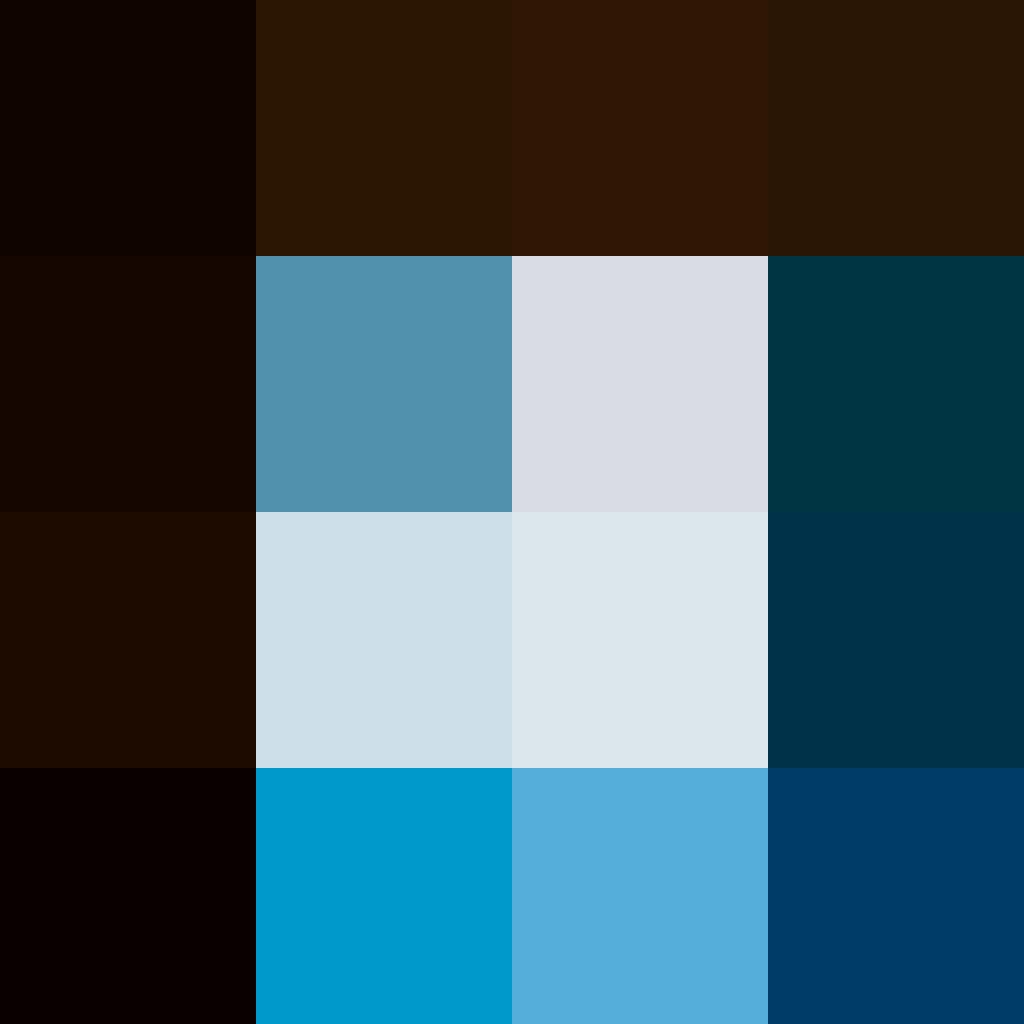} & \inc{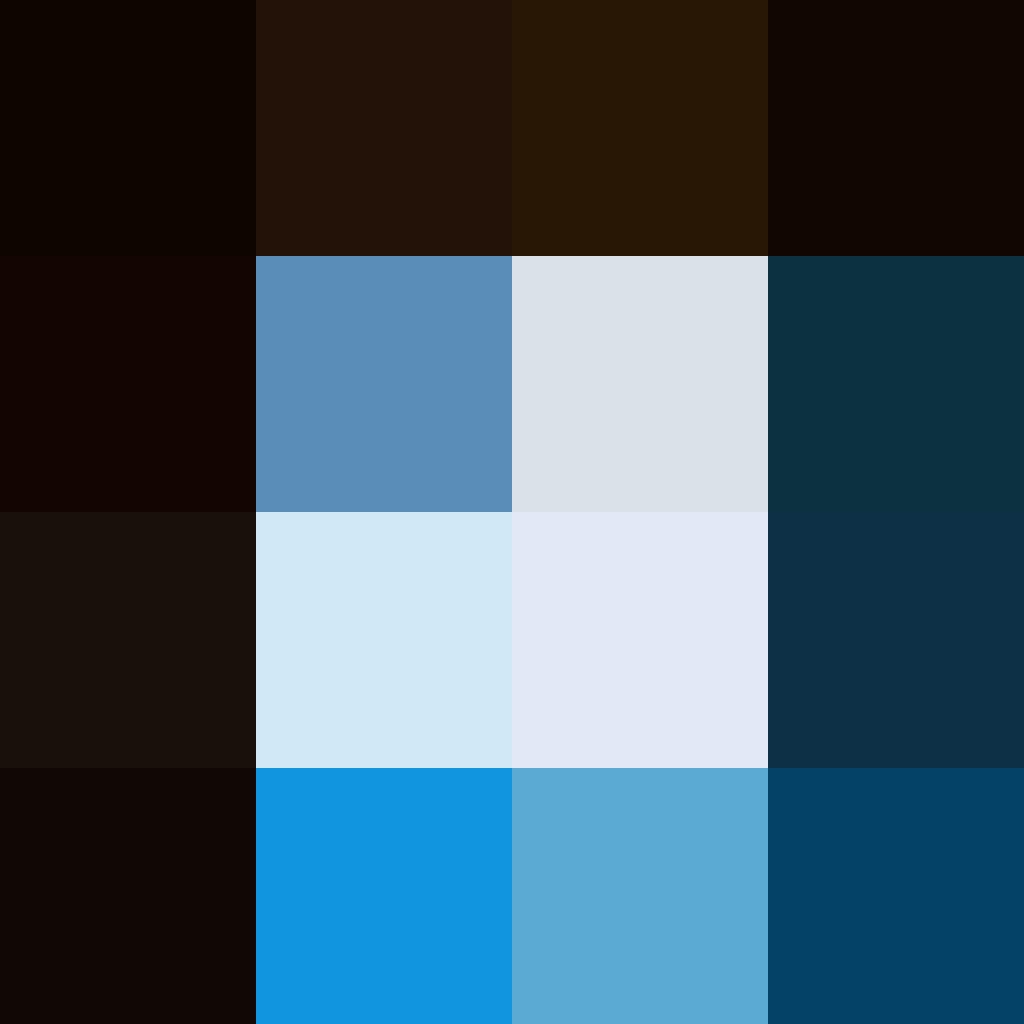} & \inc{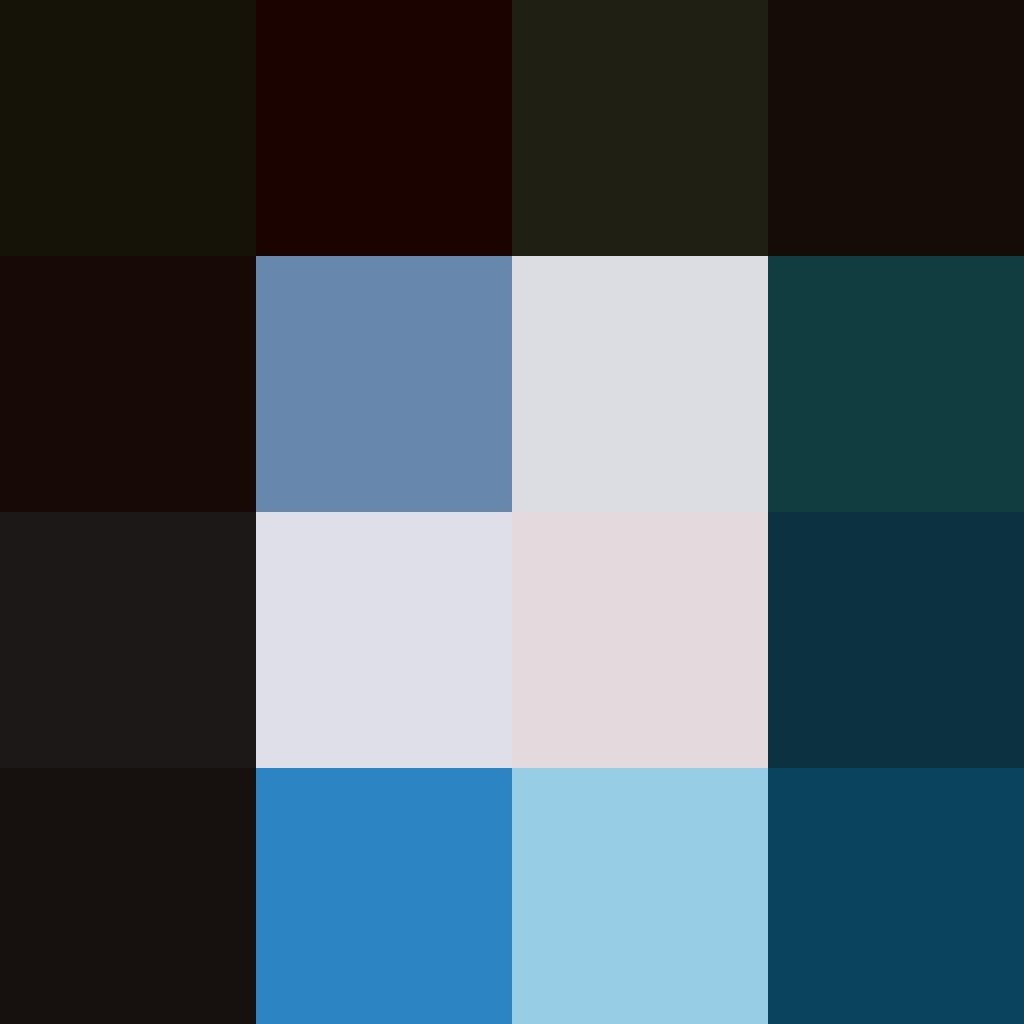} & \inc{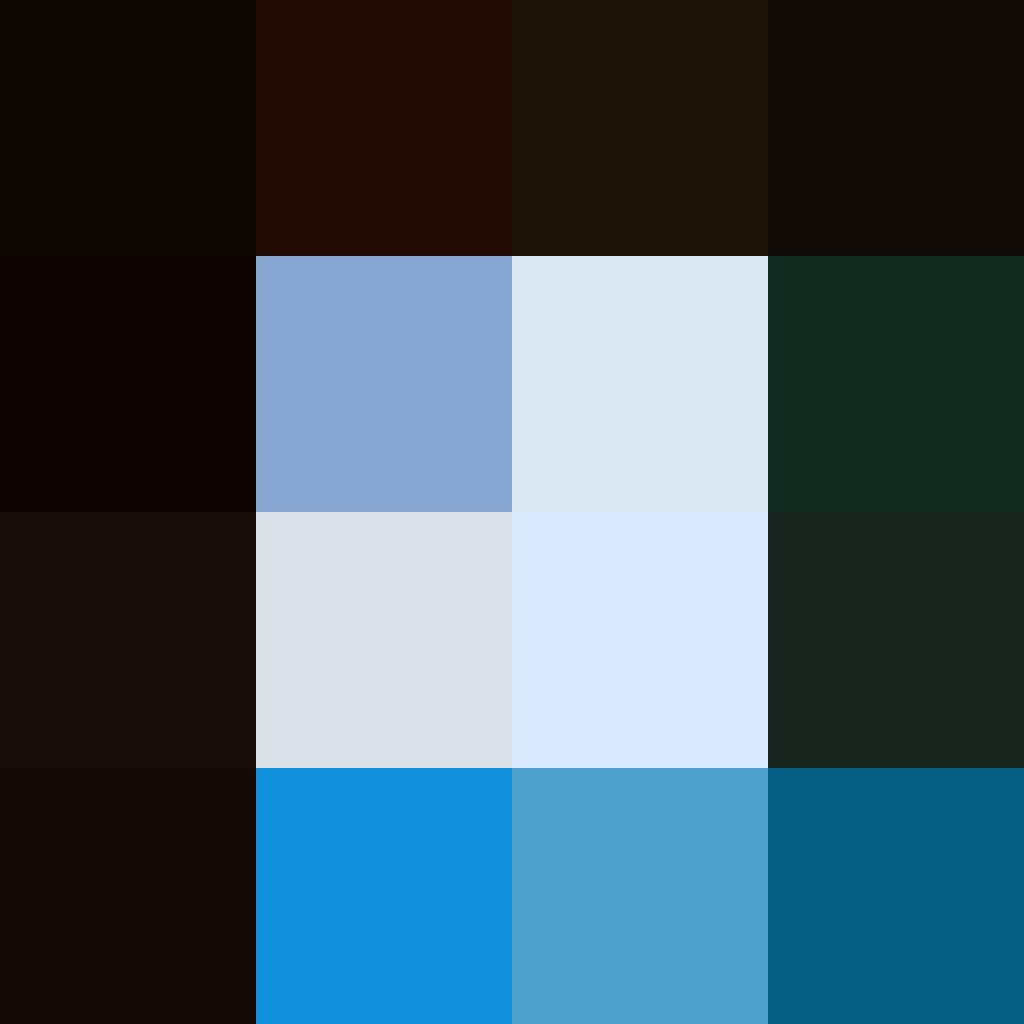} & \inc{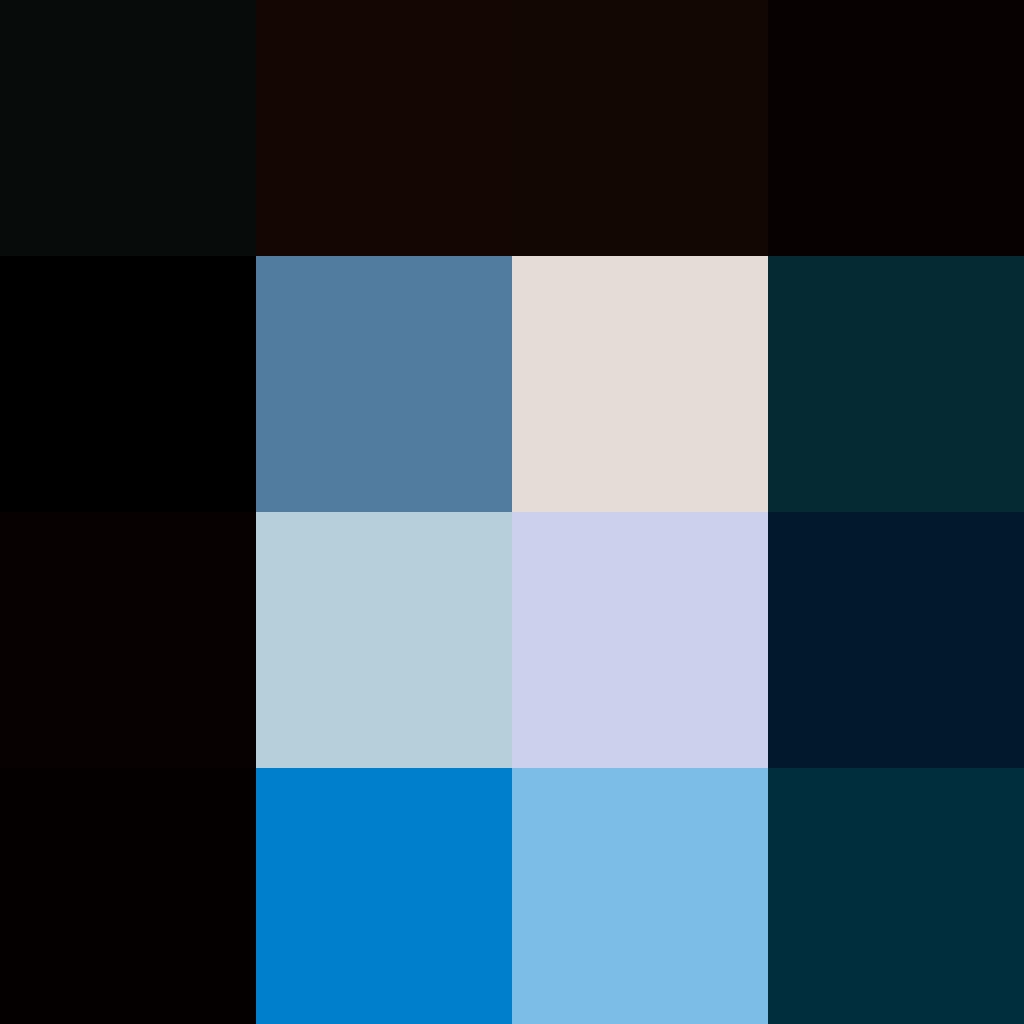} & \inc{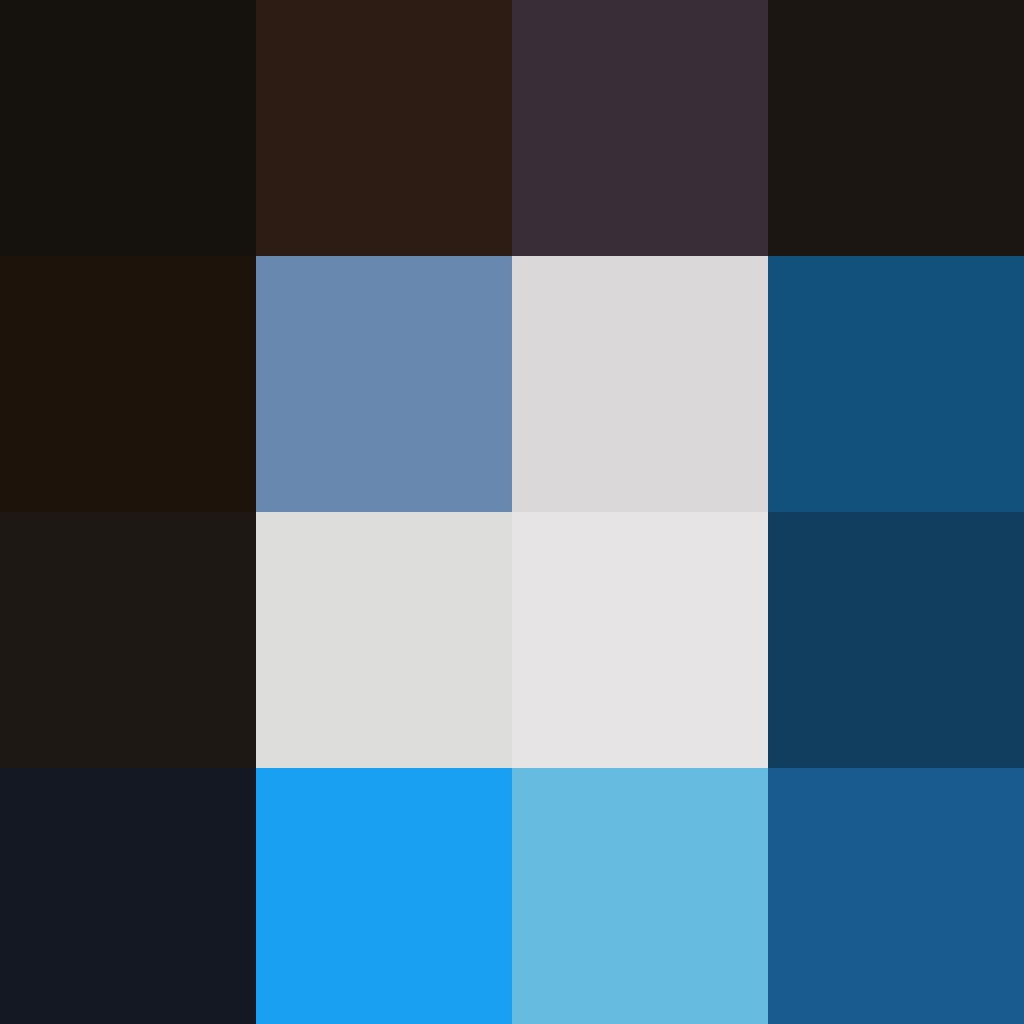} & \inc{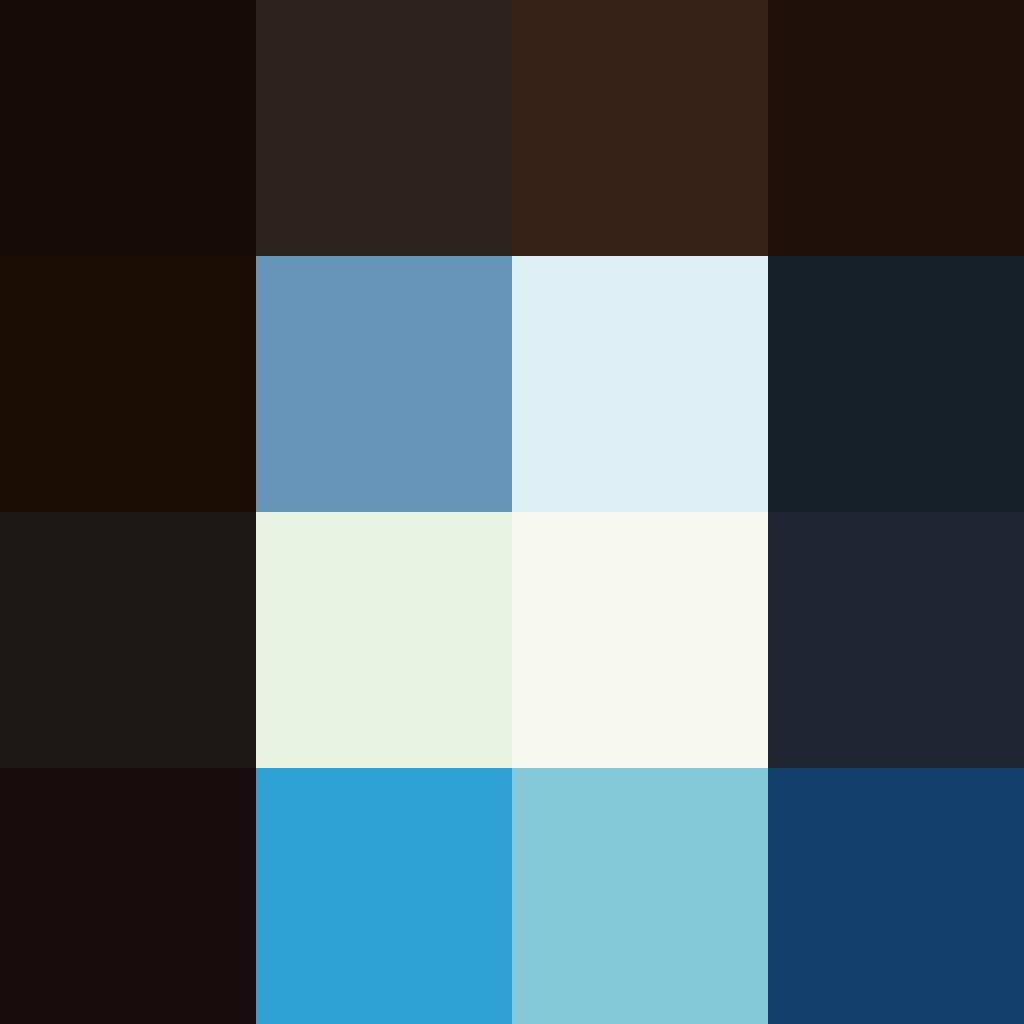}\\
 \end{tabular}

\renewcommand{\is}{68319}
\renewcommand{\step}{step490}

\begin{tabular}{c|c||c||ccccc}\arrayrulecolor{white}
% & True & Mean & Sample 1 & Sample 2 & Sample 3 & Sample 4 & Sample 5\\
HR & \inc{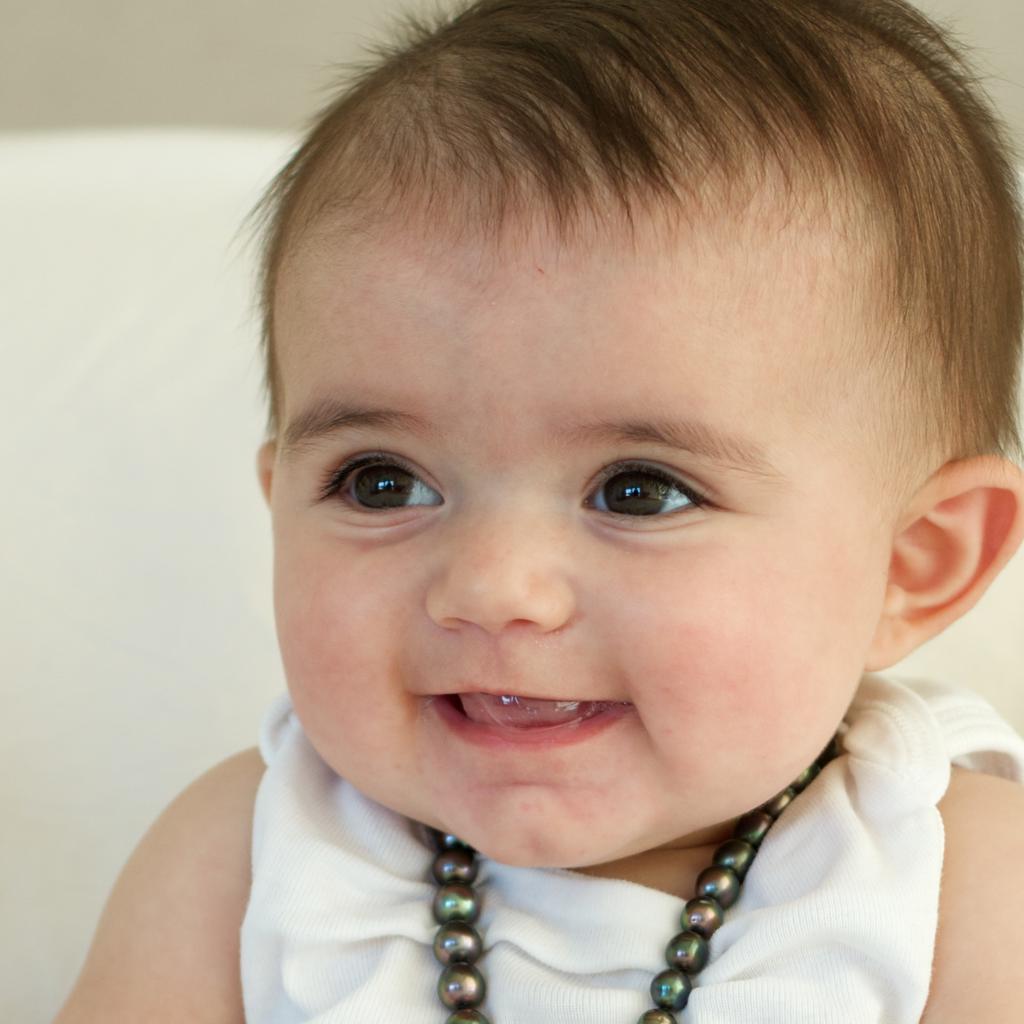} & \inc{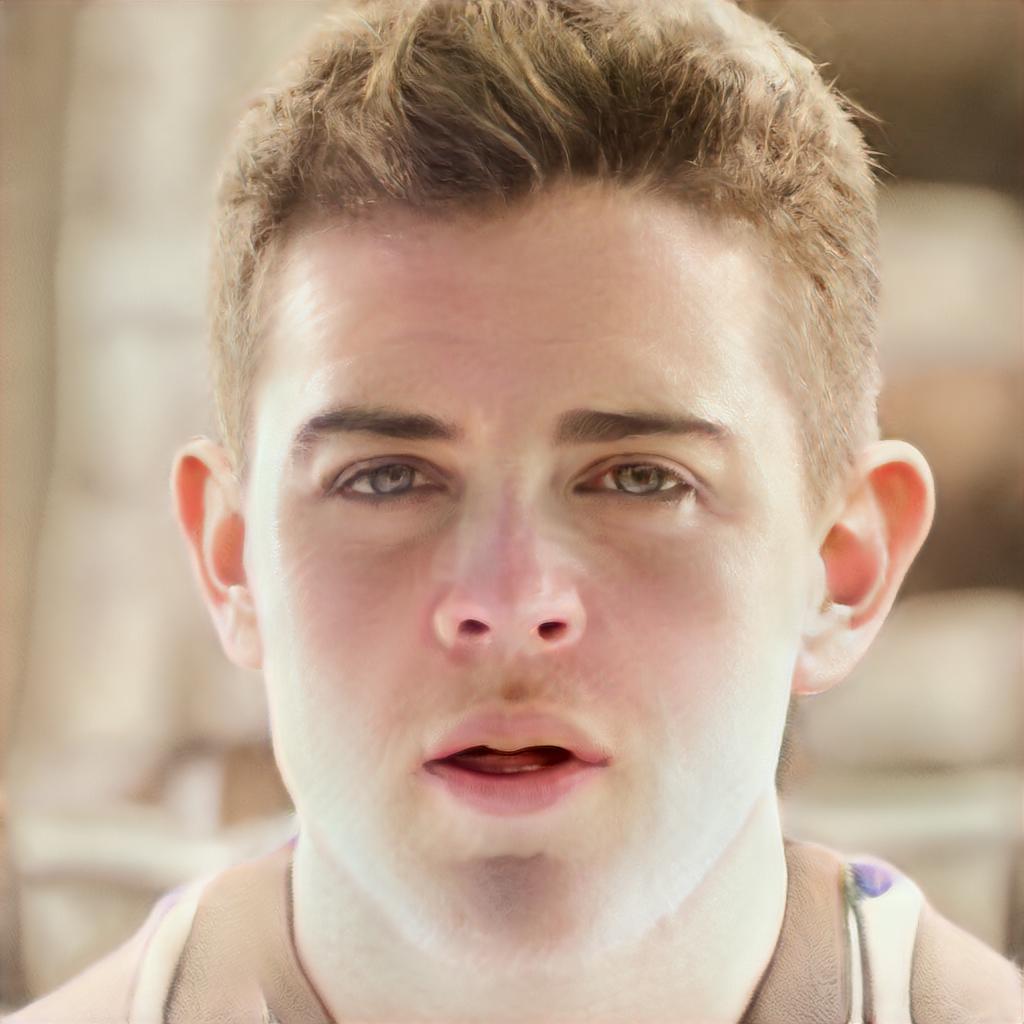} & \inc{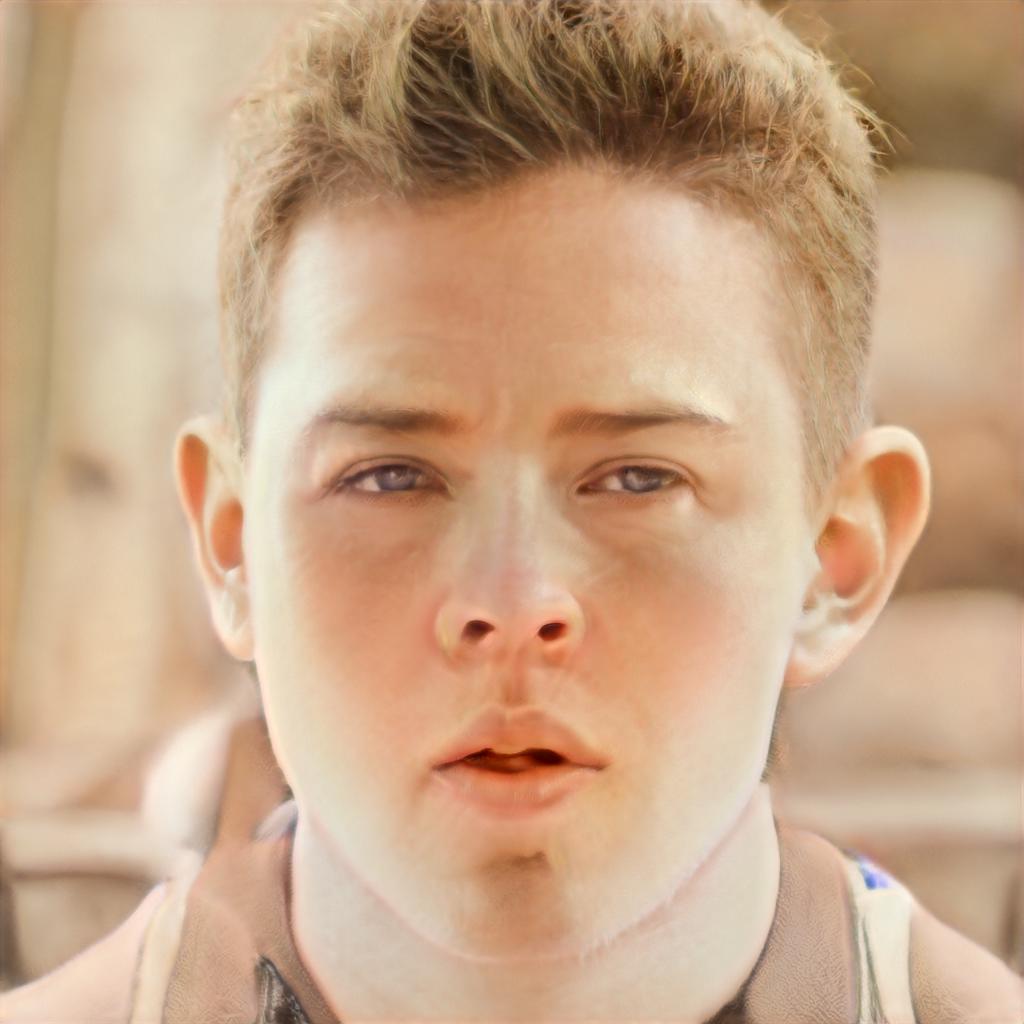} & \inc{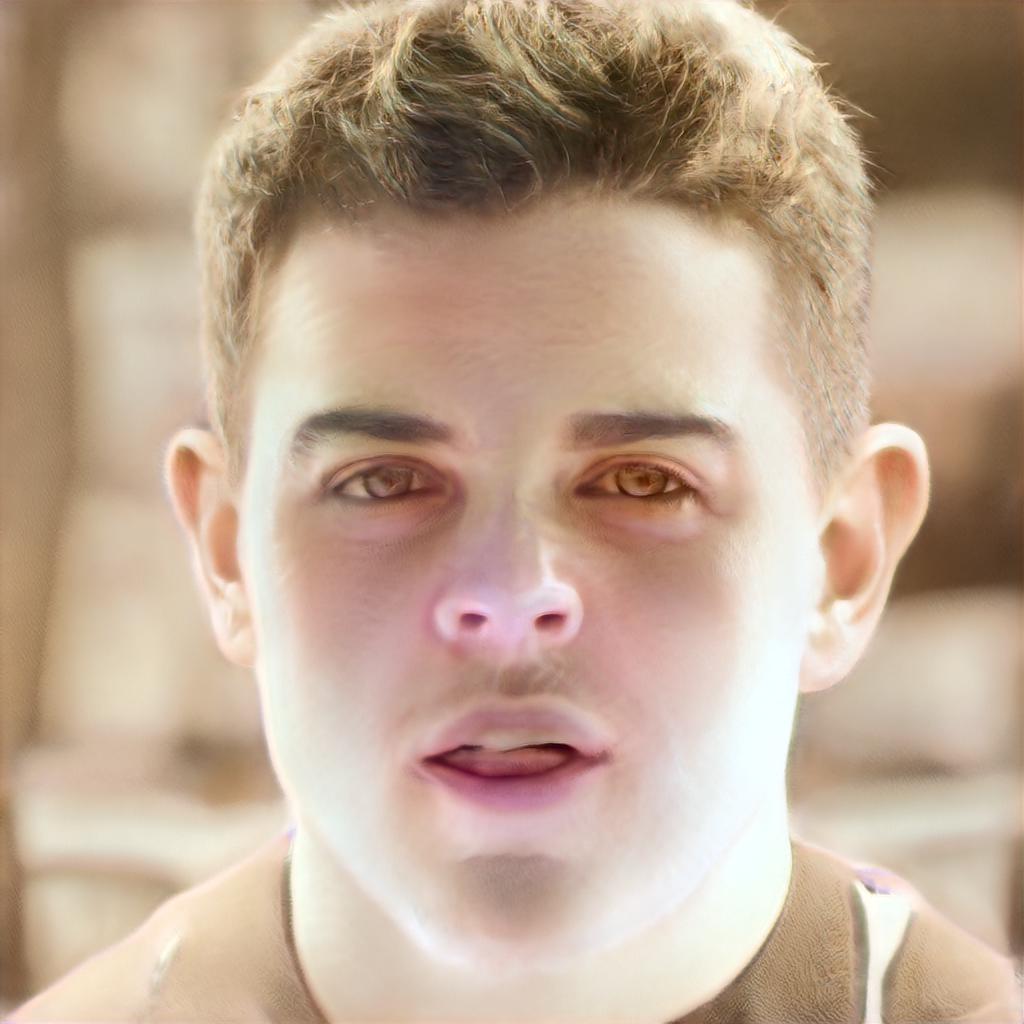} & \inc{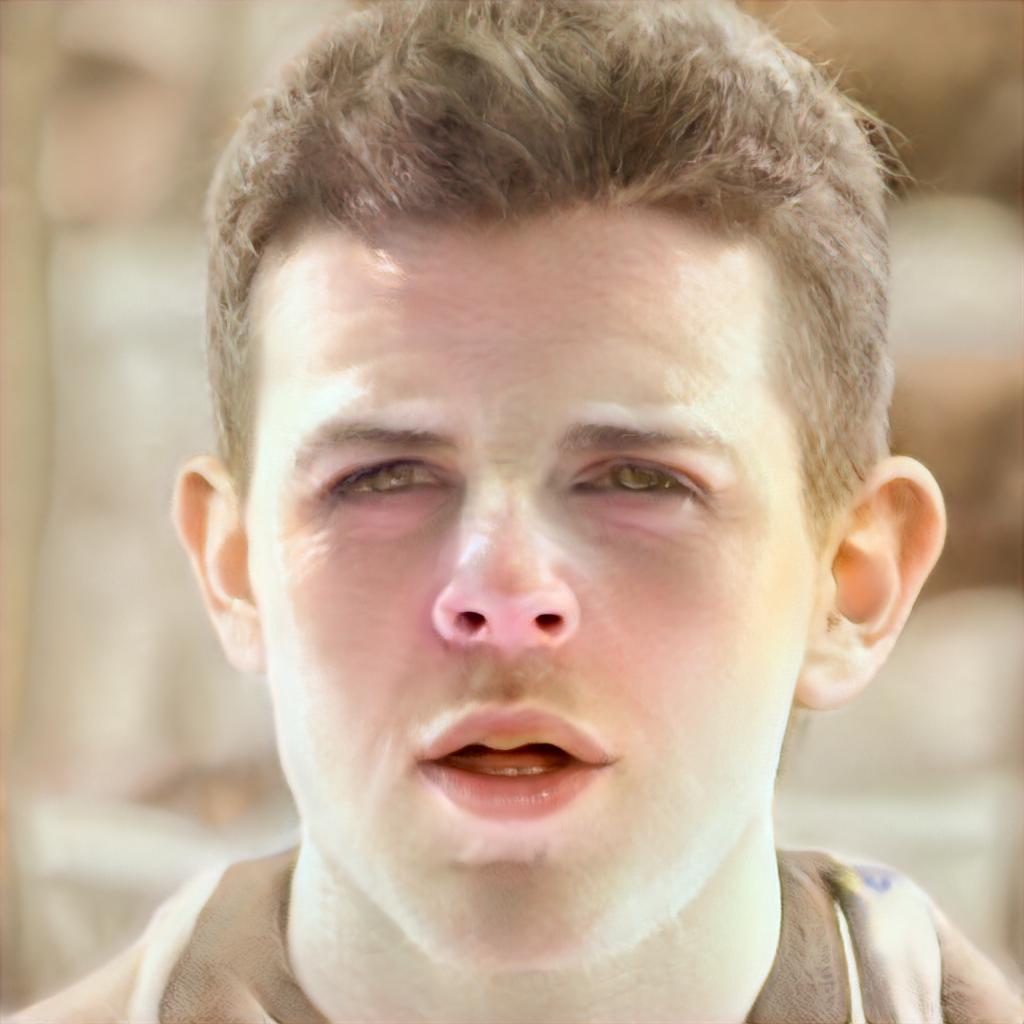} & \inc{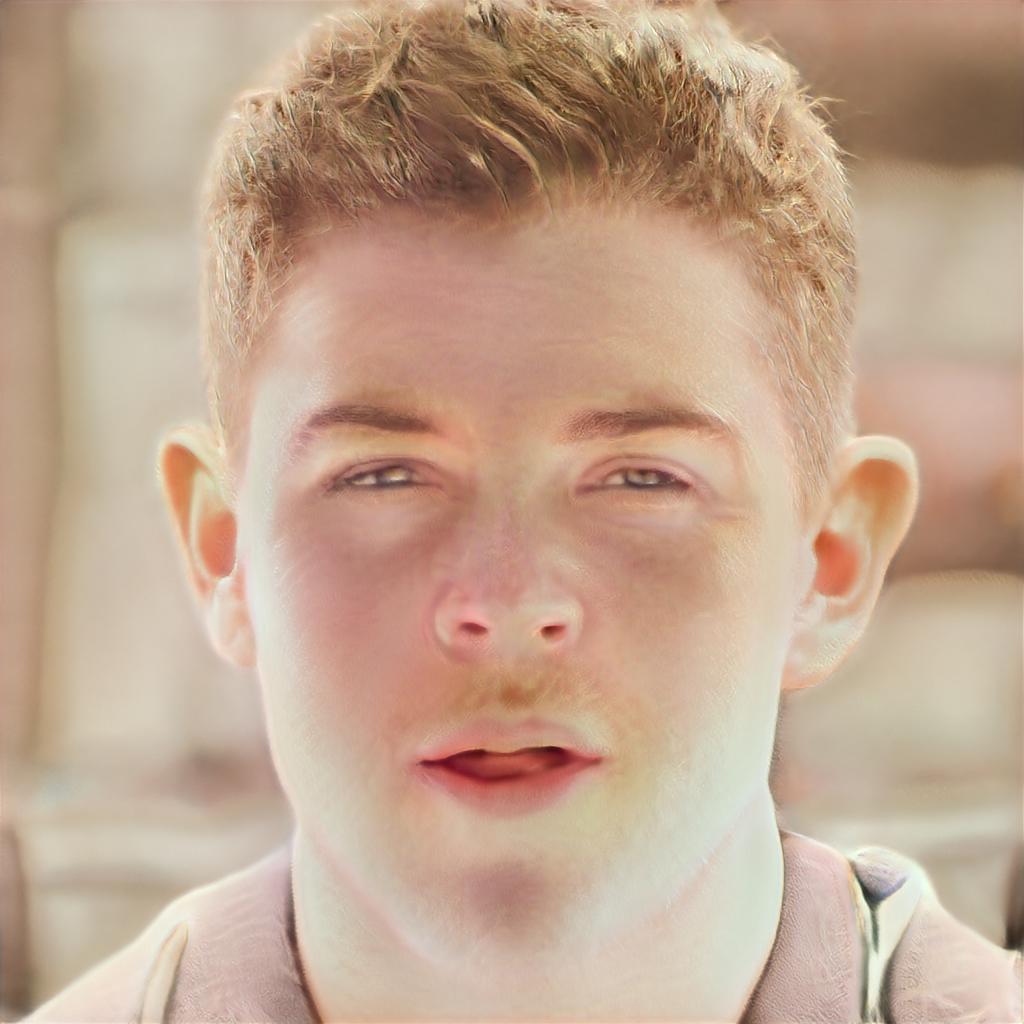} & \inc{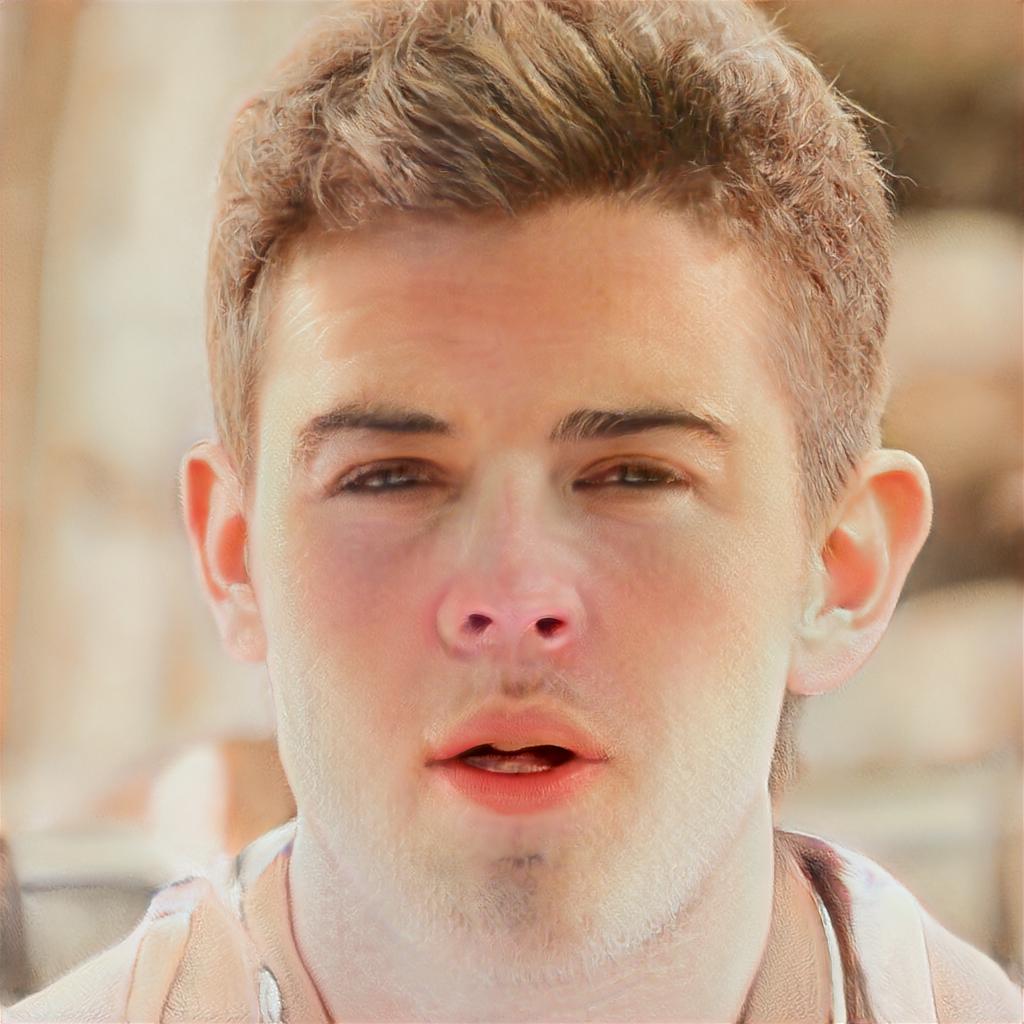}\\
 
LR & \inc{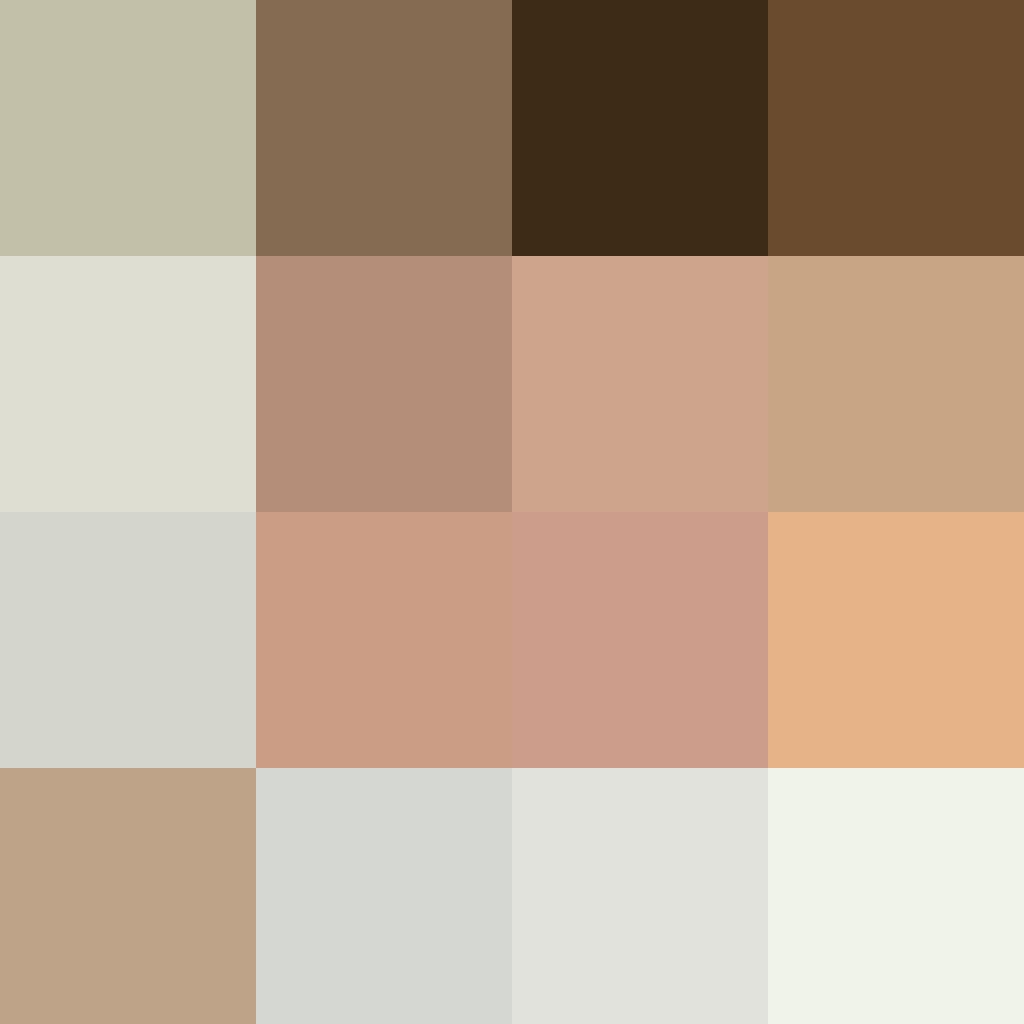} & \inc{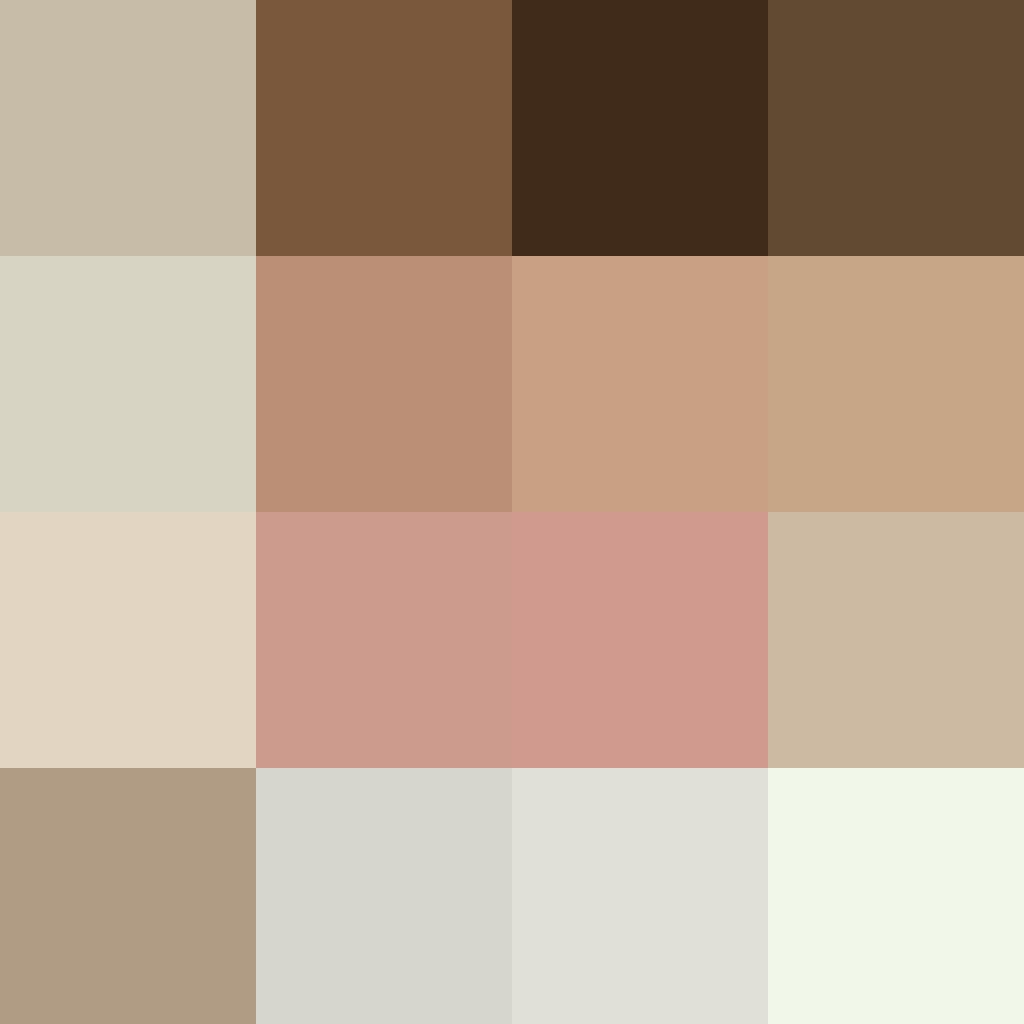} & \inc{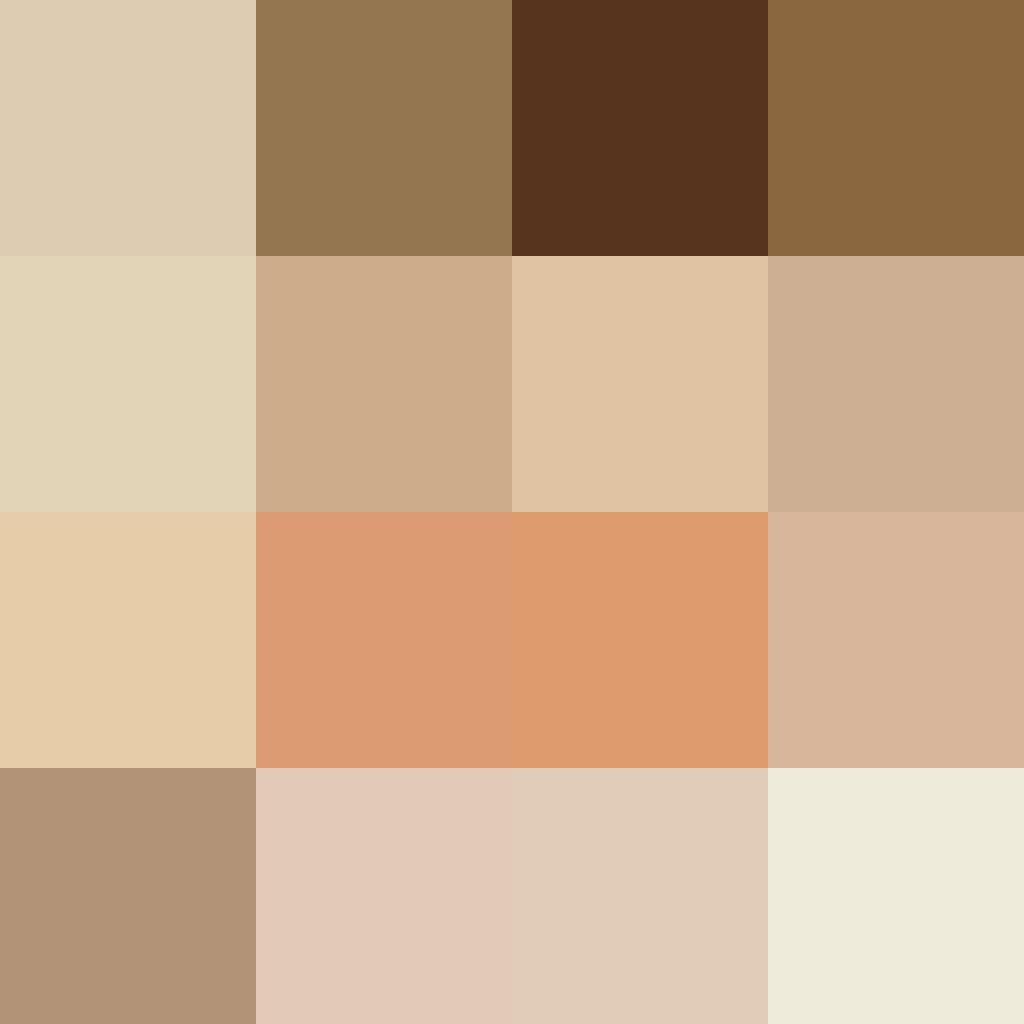} & \inc{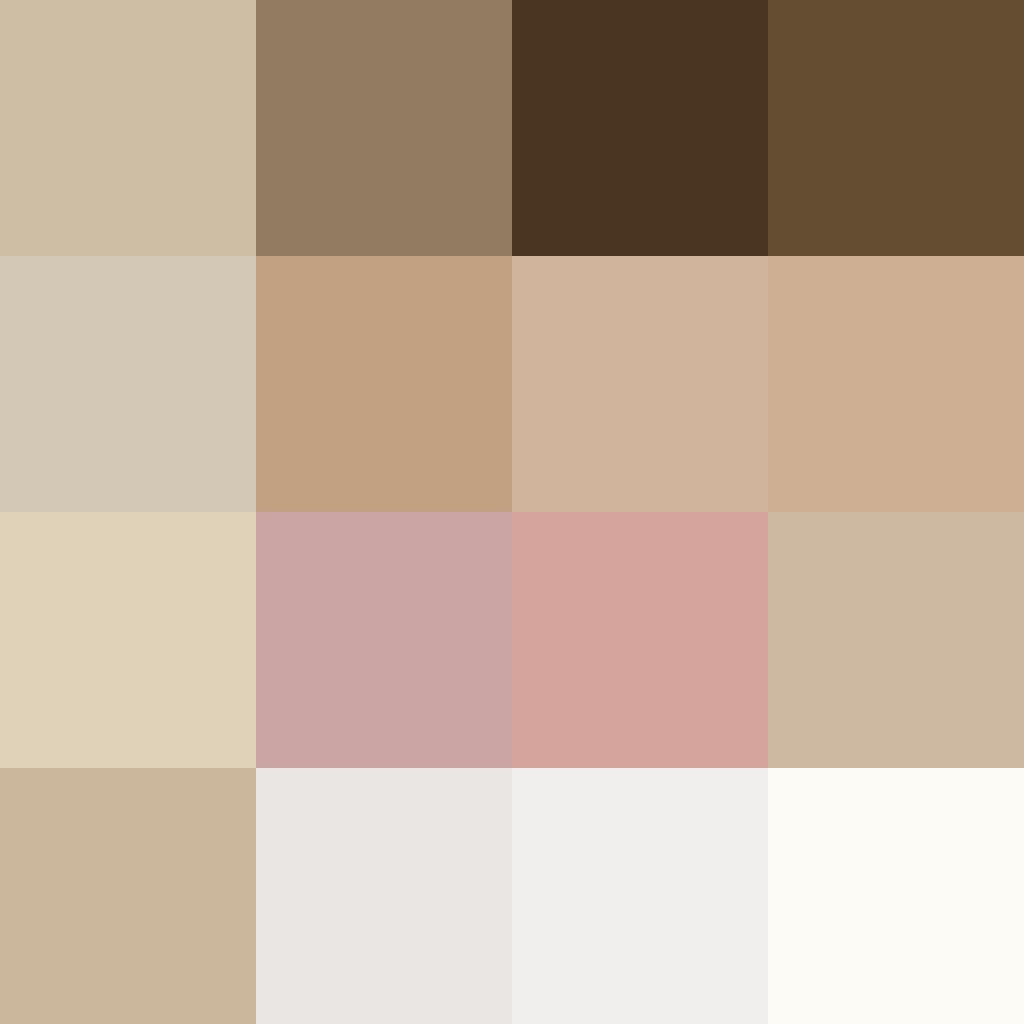} & \inc{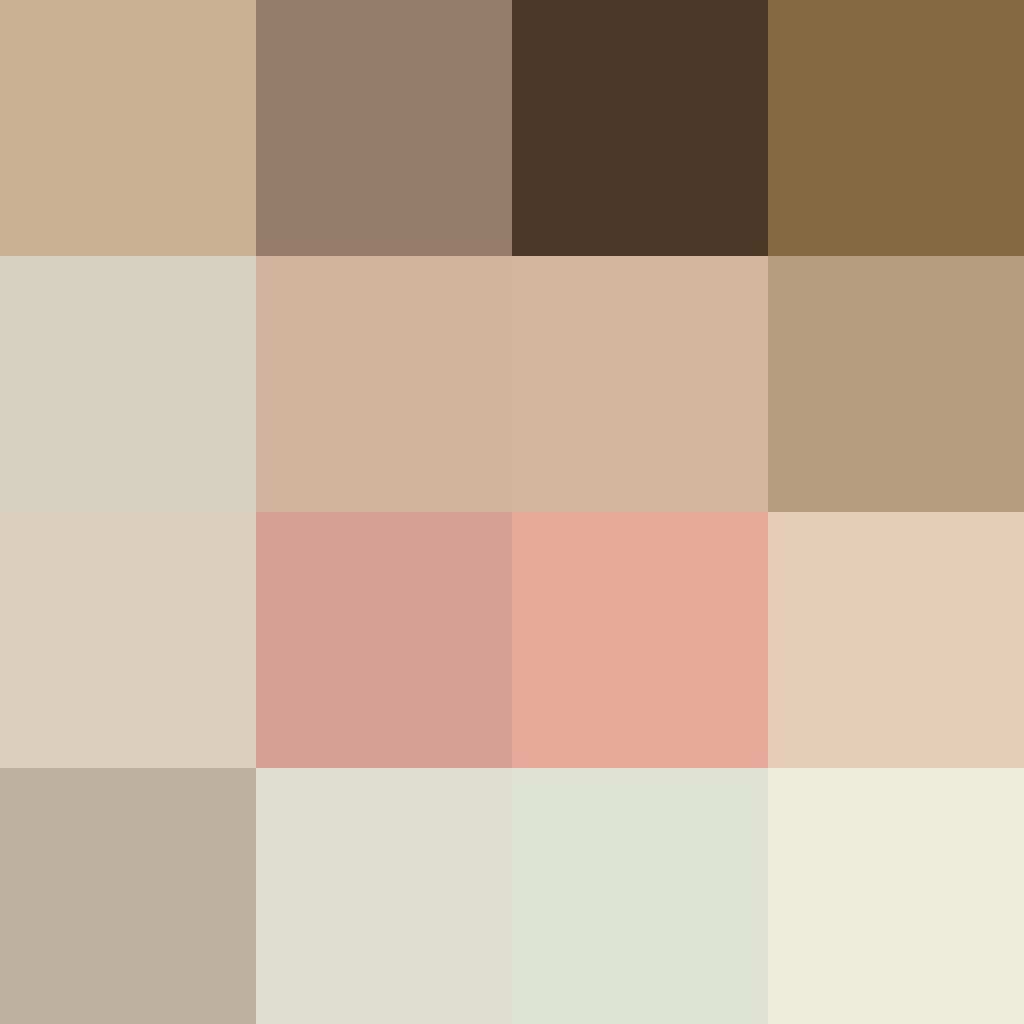} & \inc{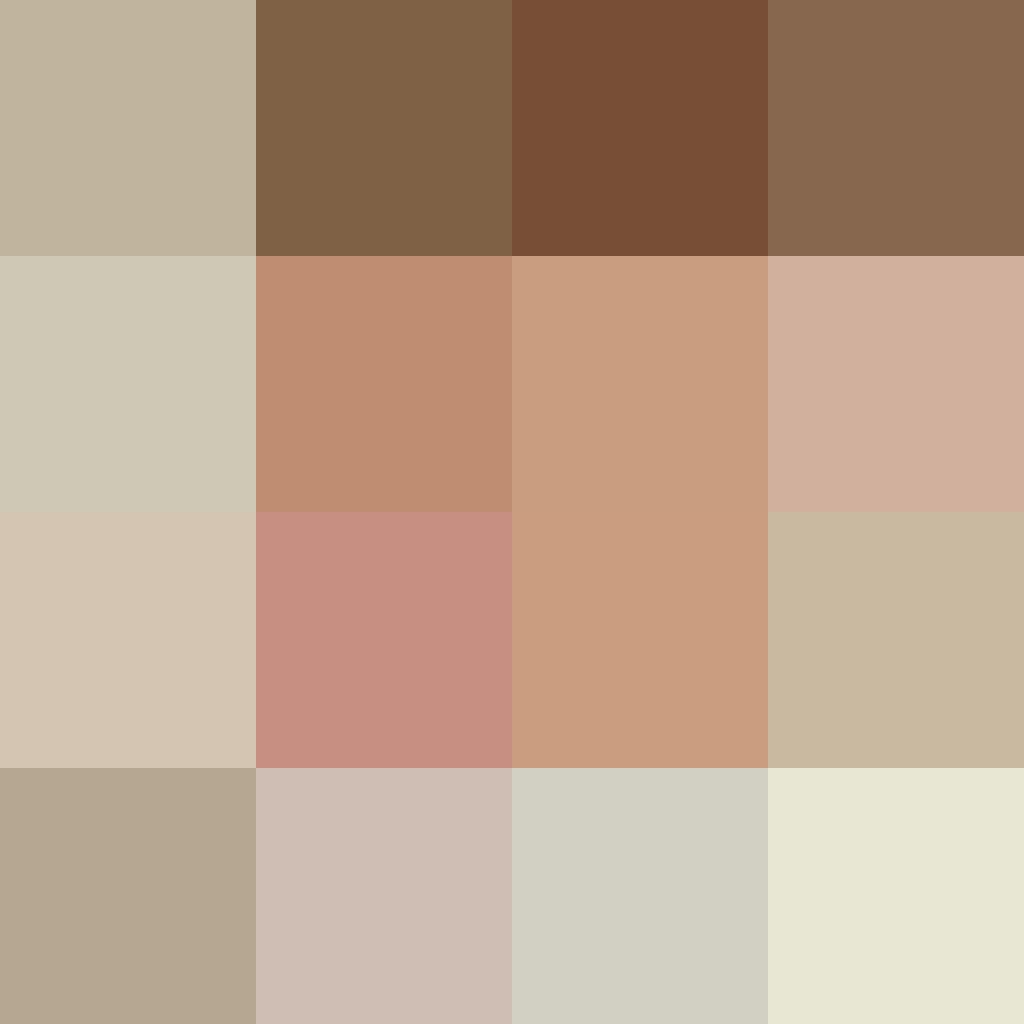} & \inc{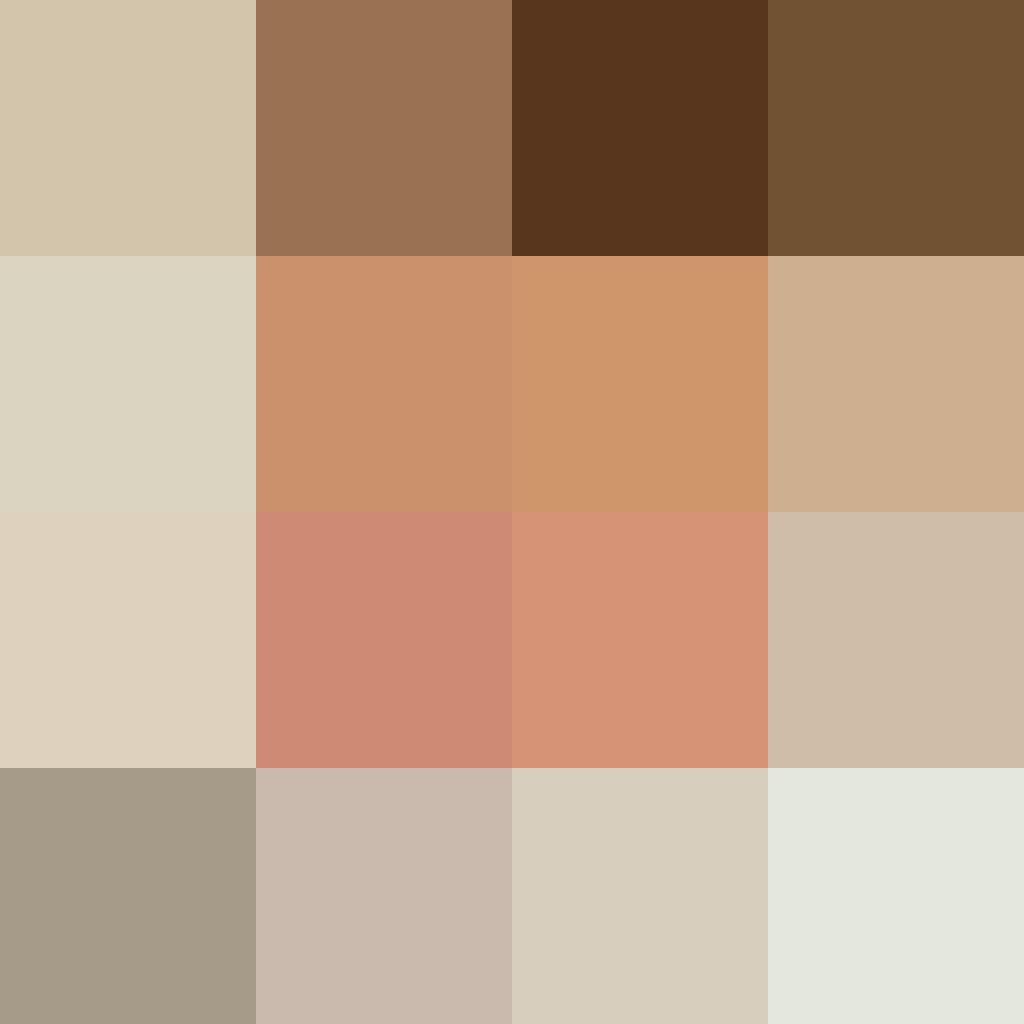}\\
 \end{tabular}
 
\renewcommand{\flds}{results/samXRAY}
\renewcommand{\is}{img00000664}
\renewcommand{\step}{step400}

\begin{tabular}{c|c||c||ccccc}\arrayrulecolor{white}
% & True & Mean & Sample 1 & Sample 2 & Sample 3 & Sample 4 & Sample 5\\
HR & \inc{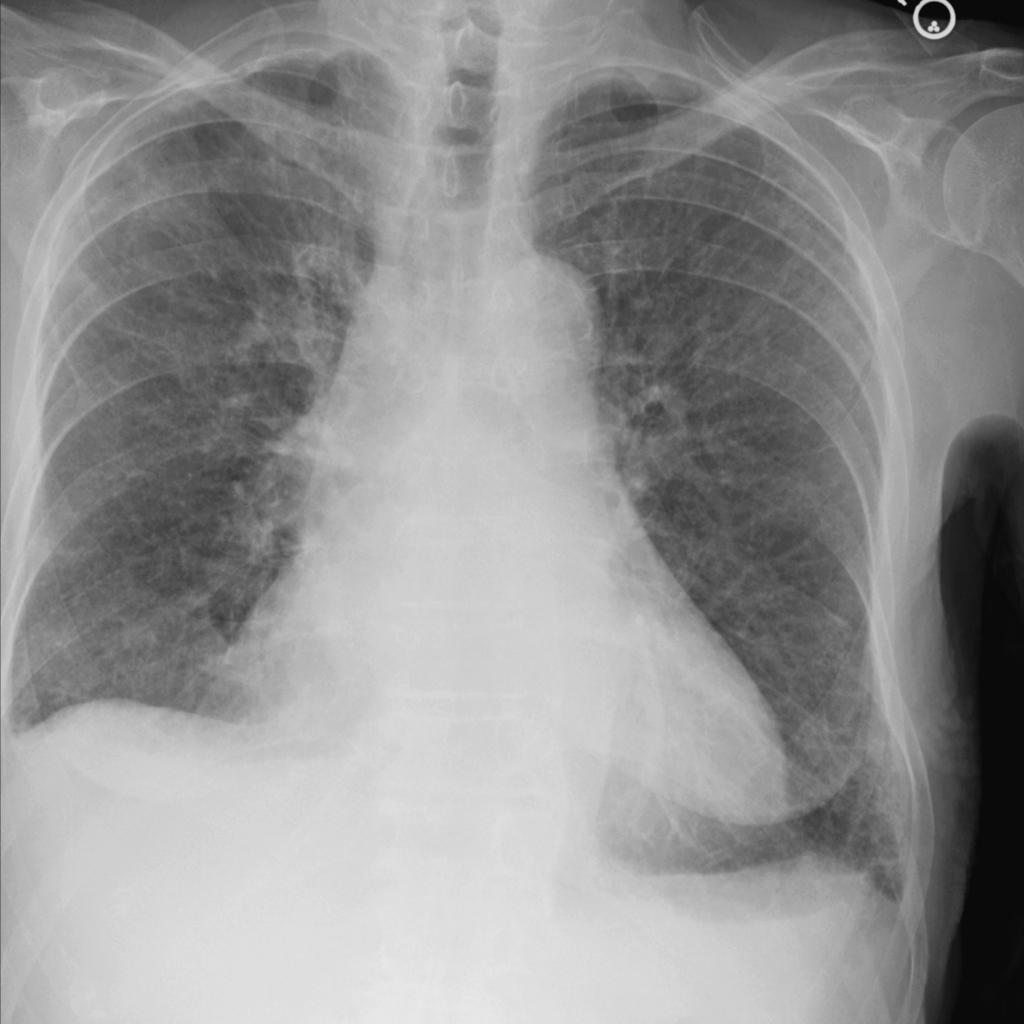} & \inc{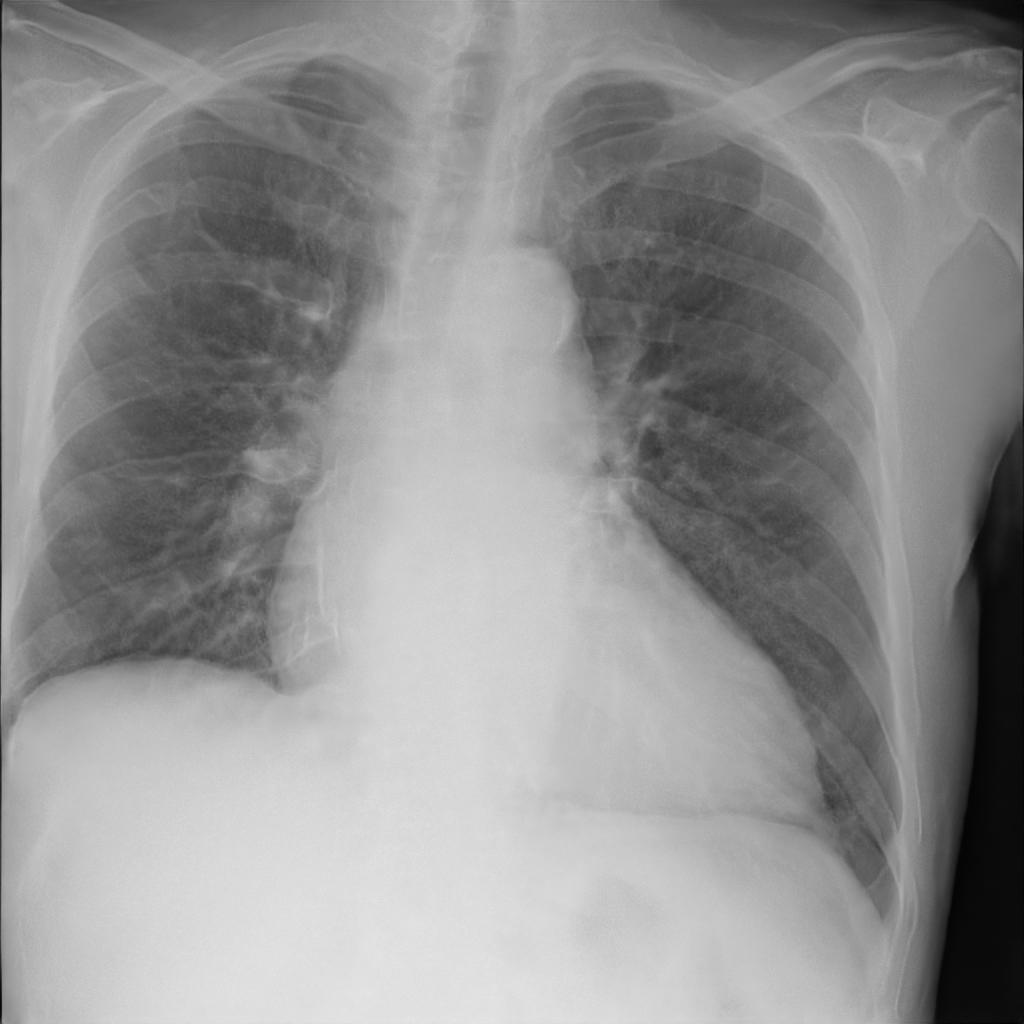} & \inc{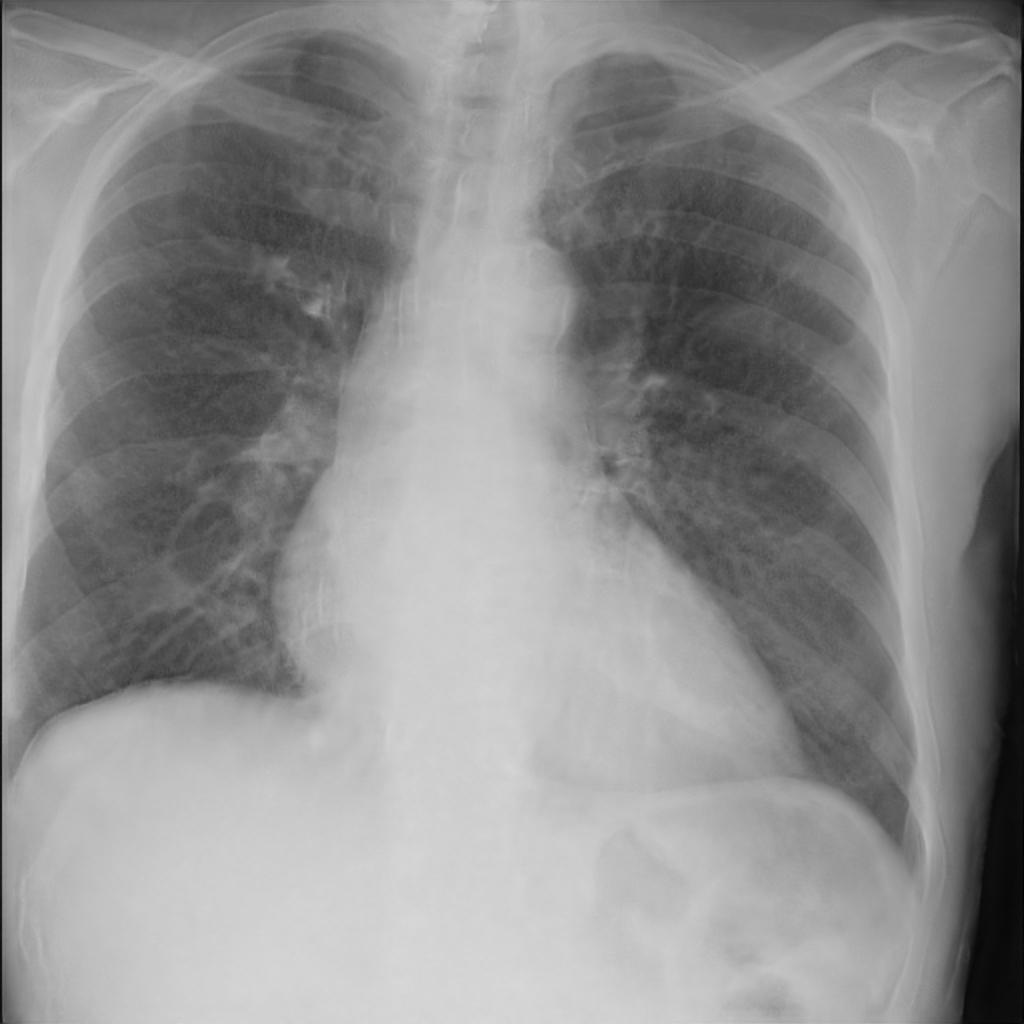} & \inc{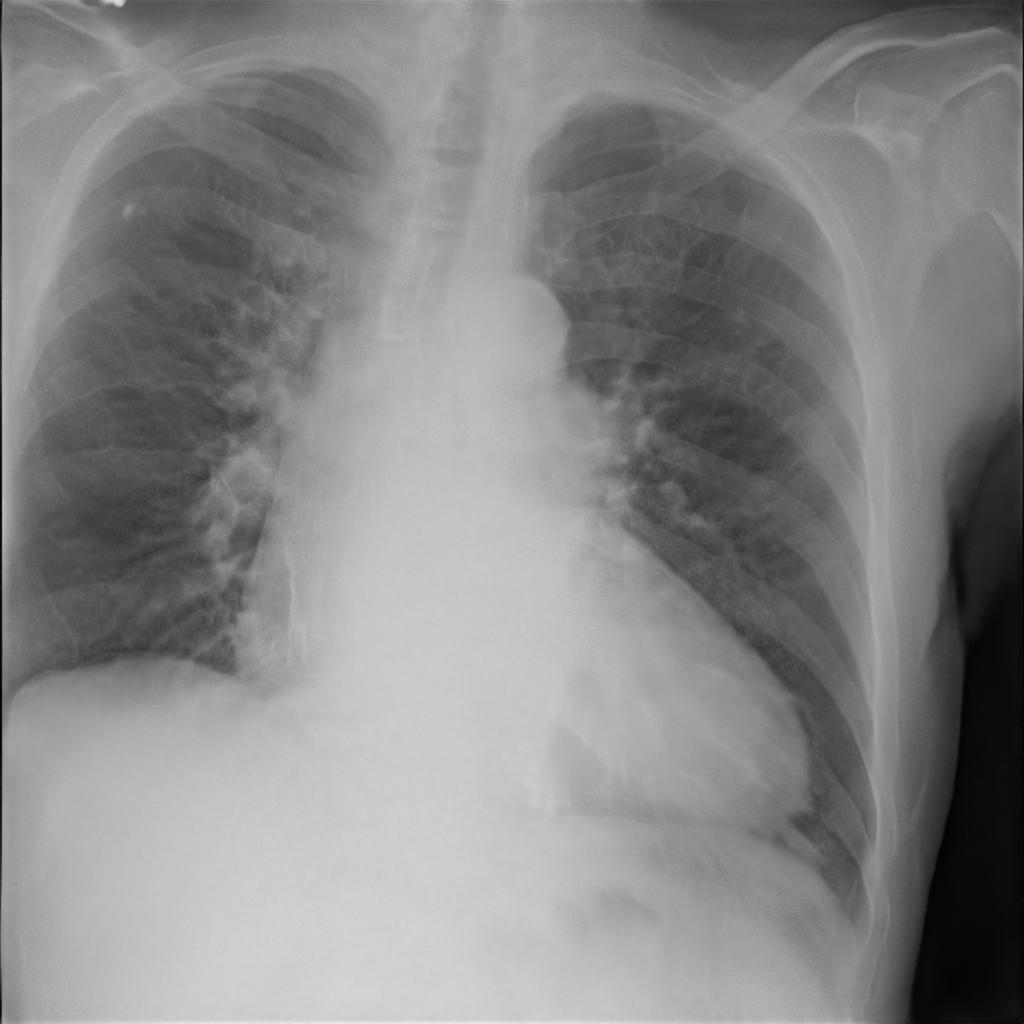} & \inc{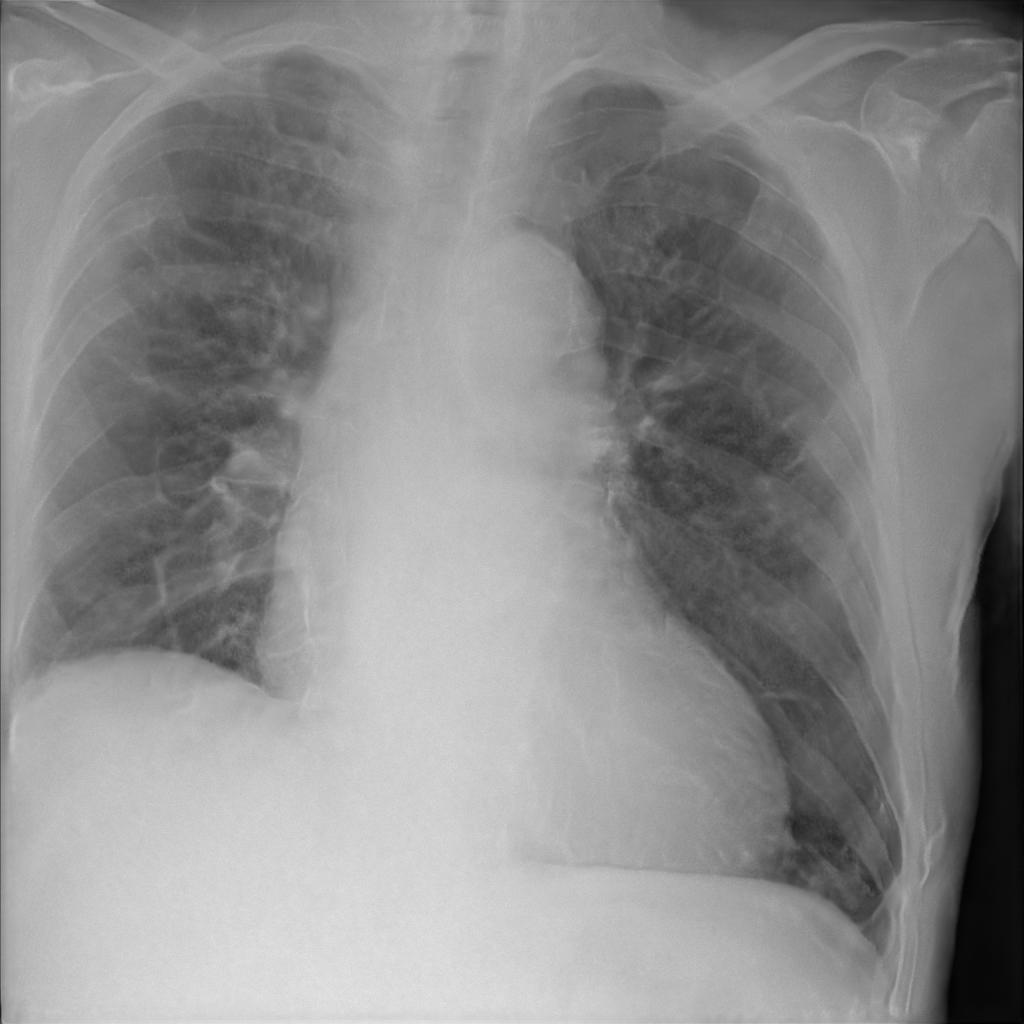} & \inc{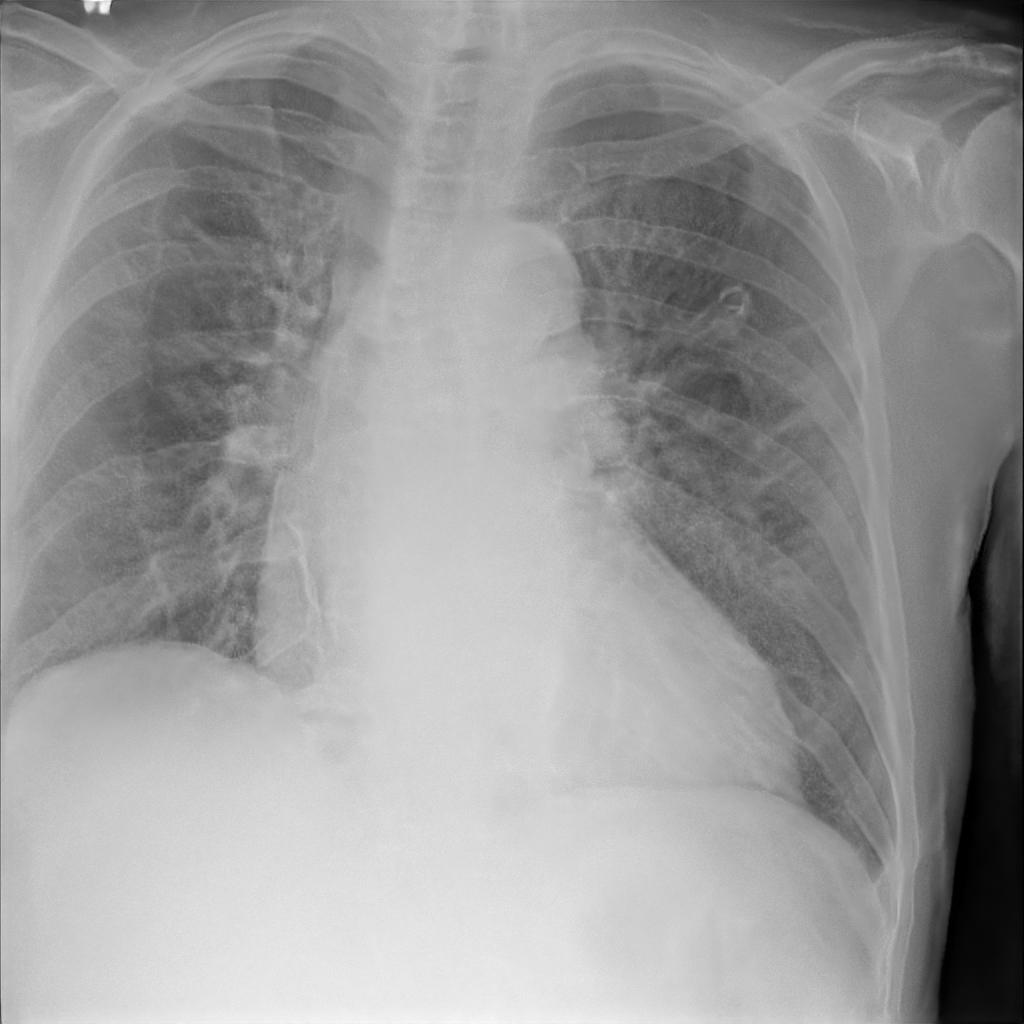} & \inc{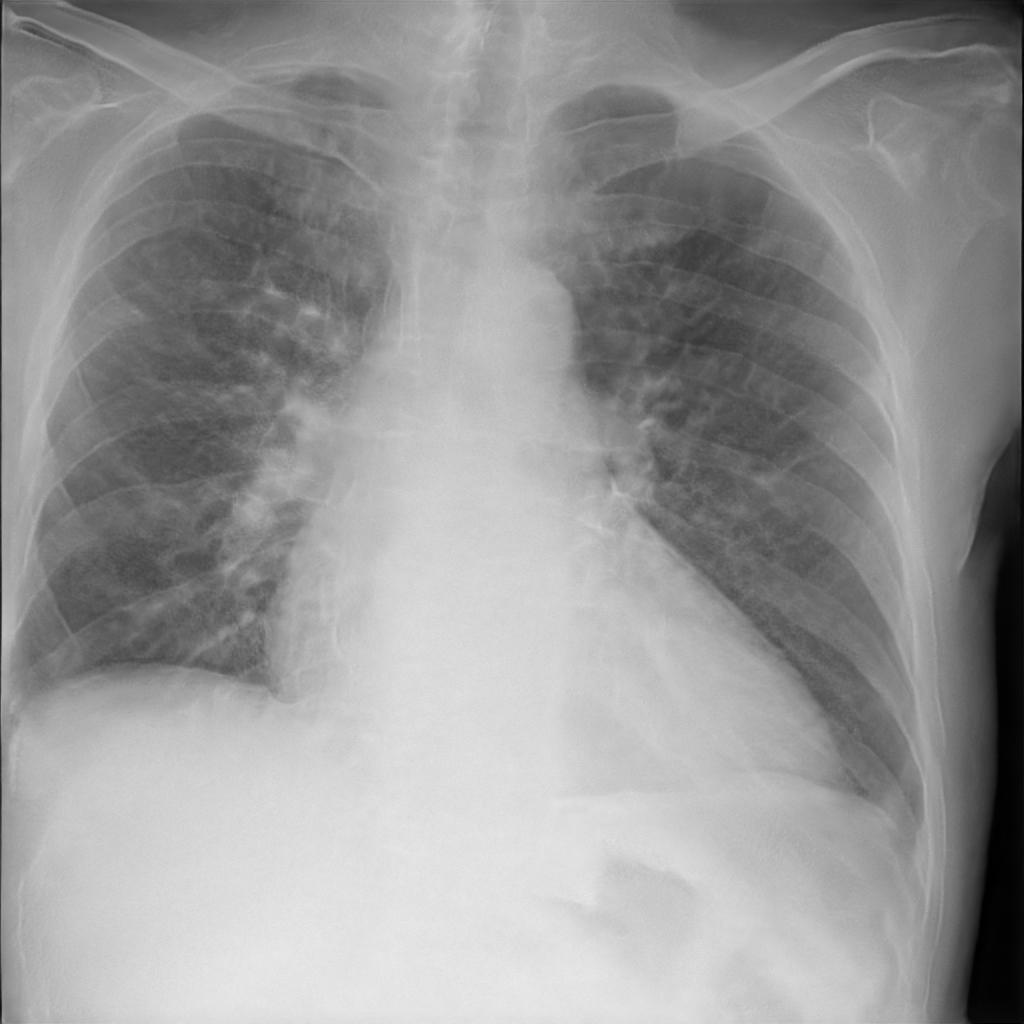}\\
 
LR & \inc{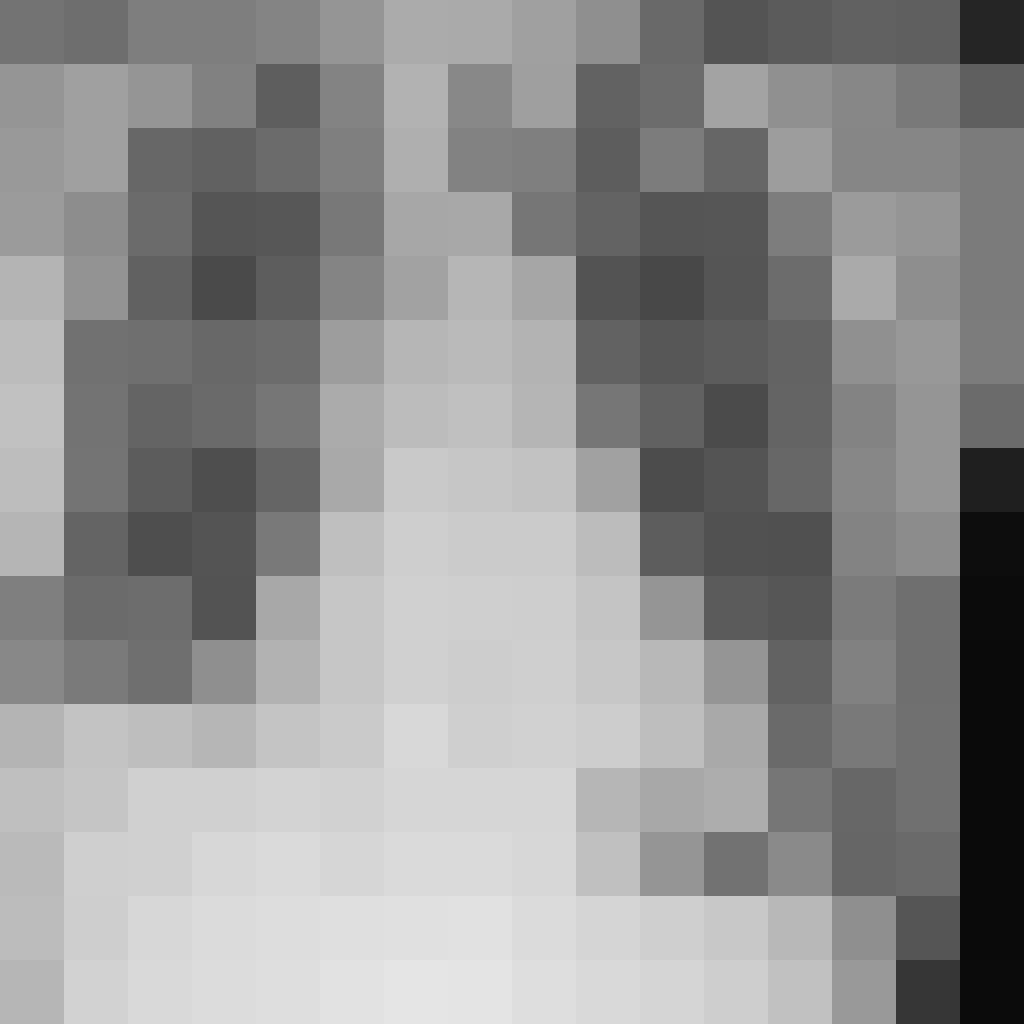} & \inc{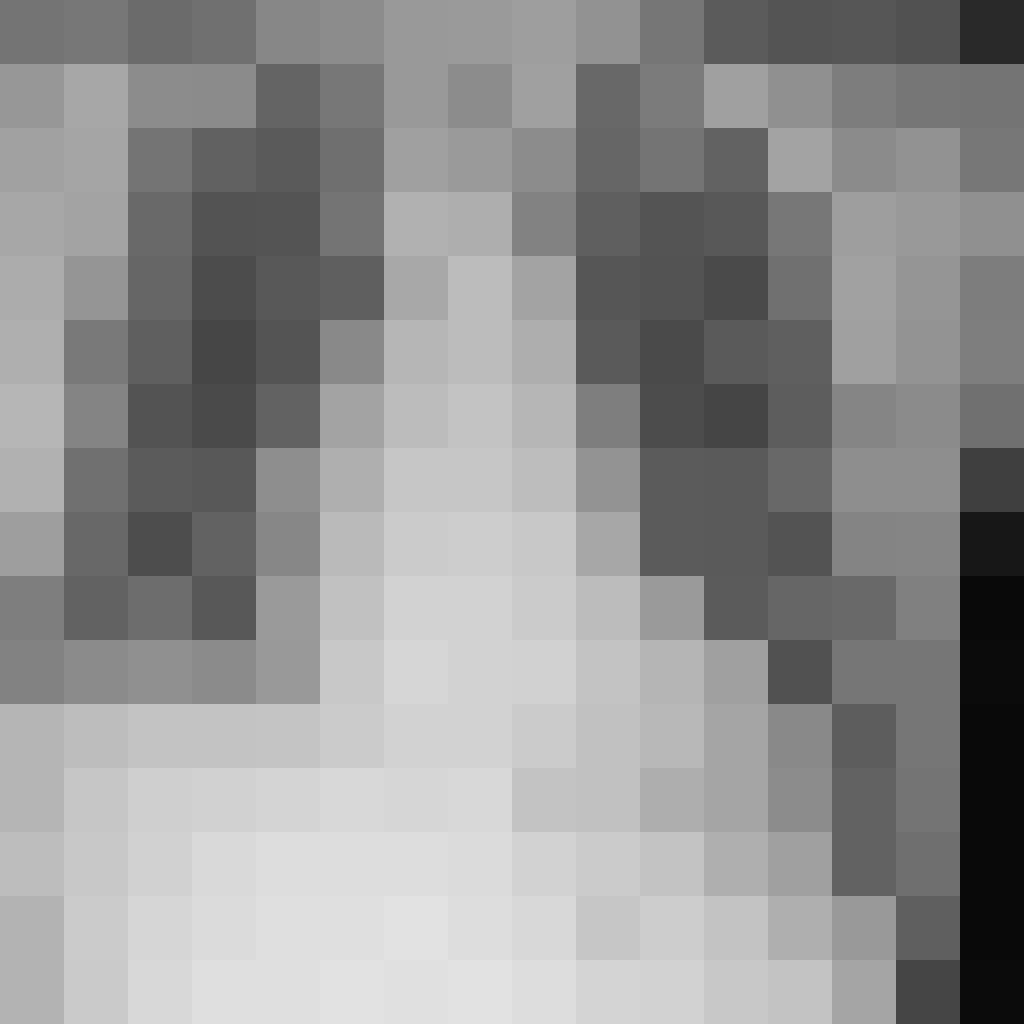} & \inc{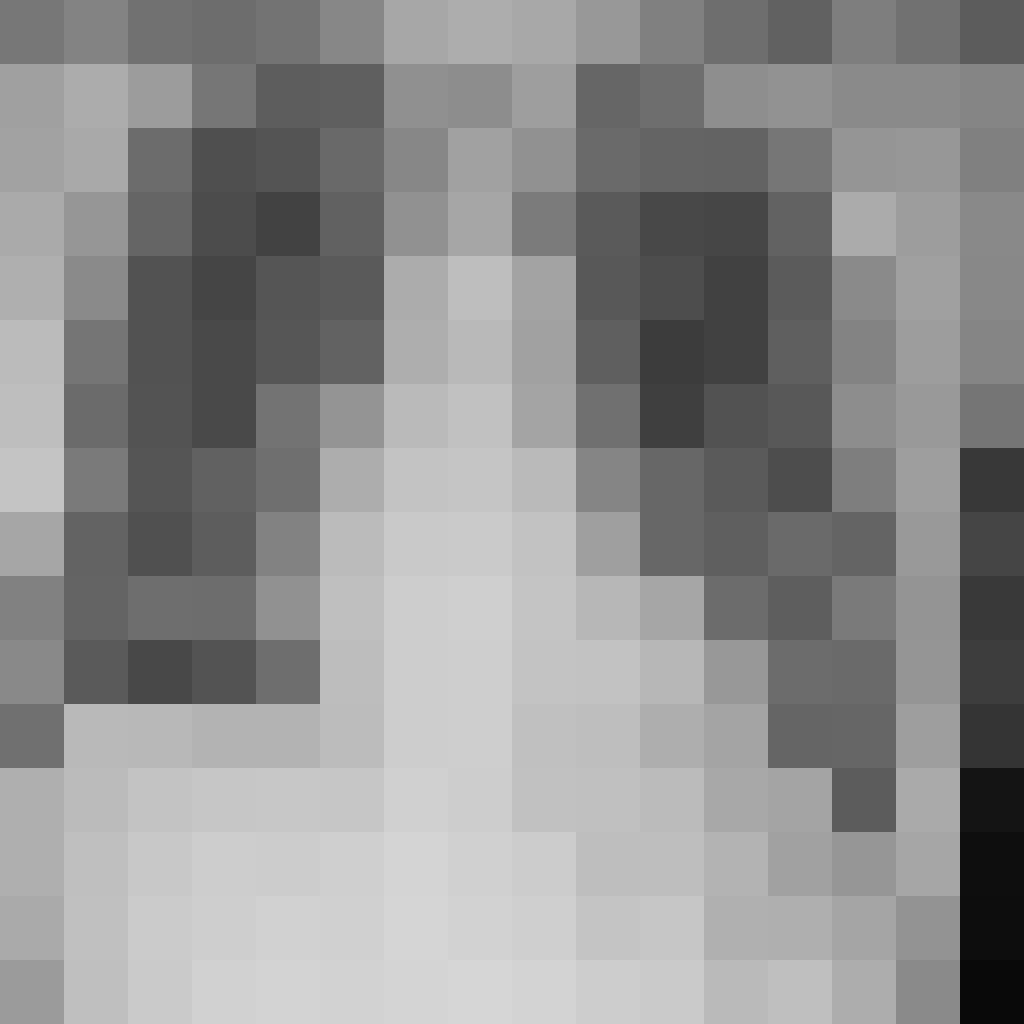} & \inc{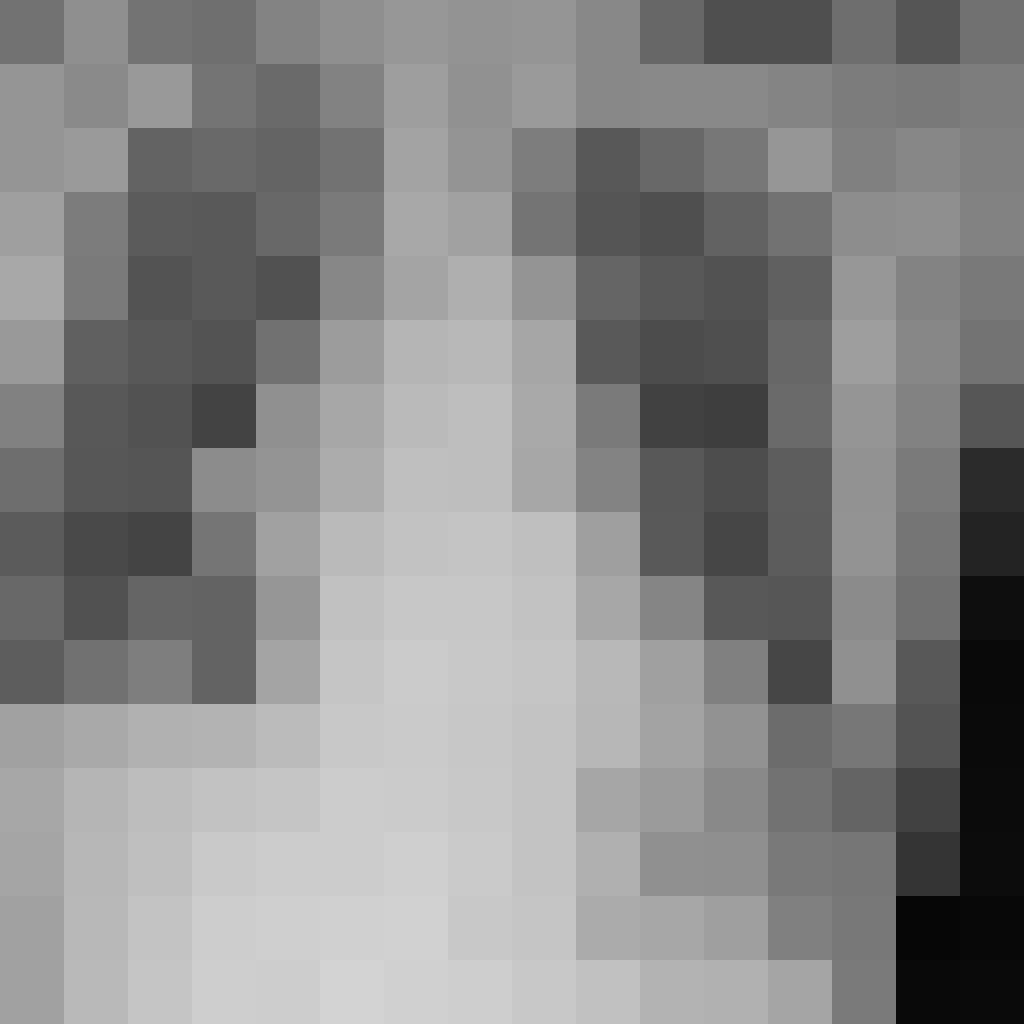} & \inc{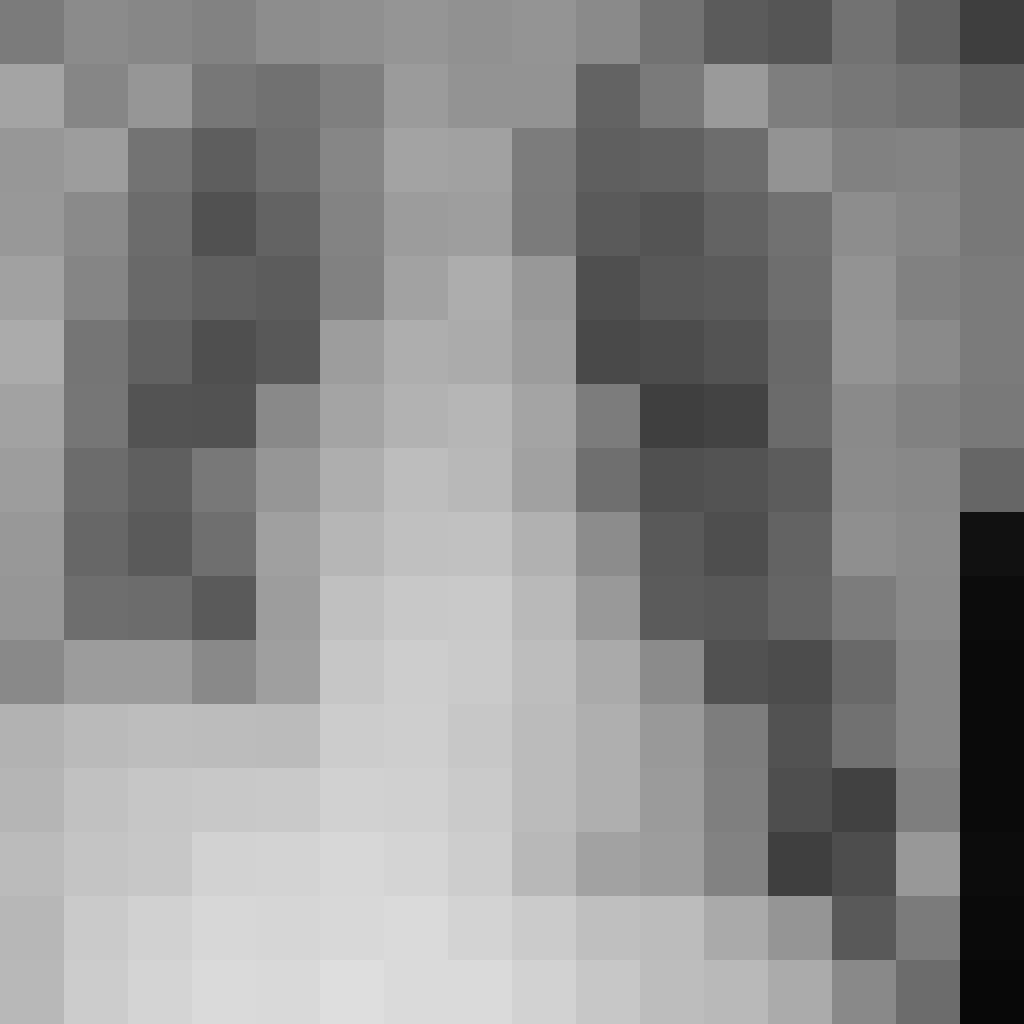} & \inc{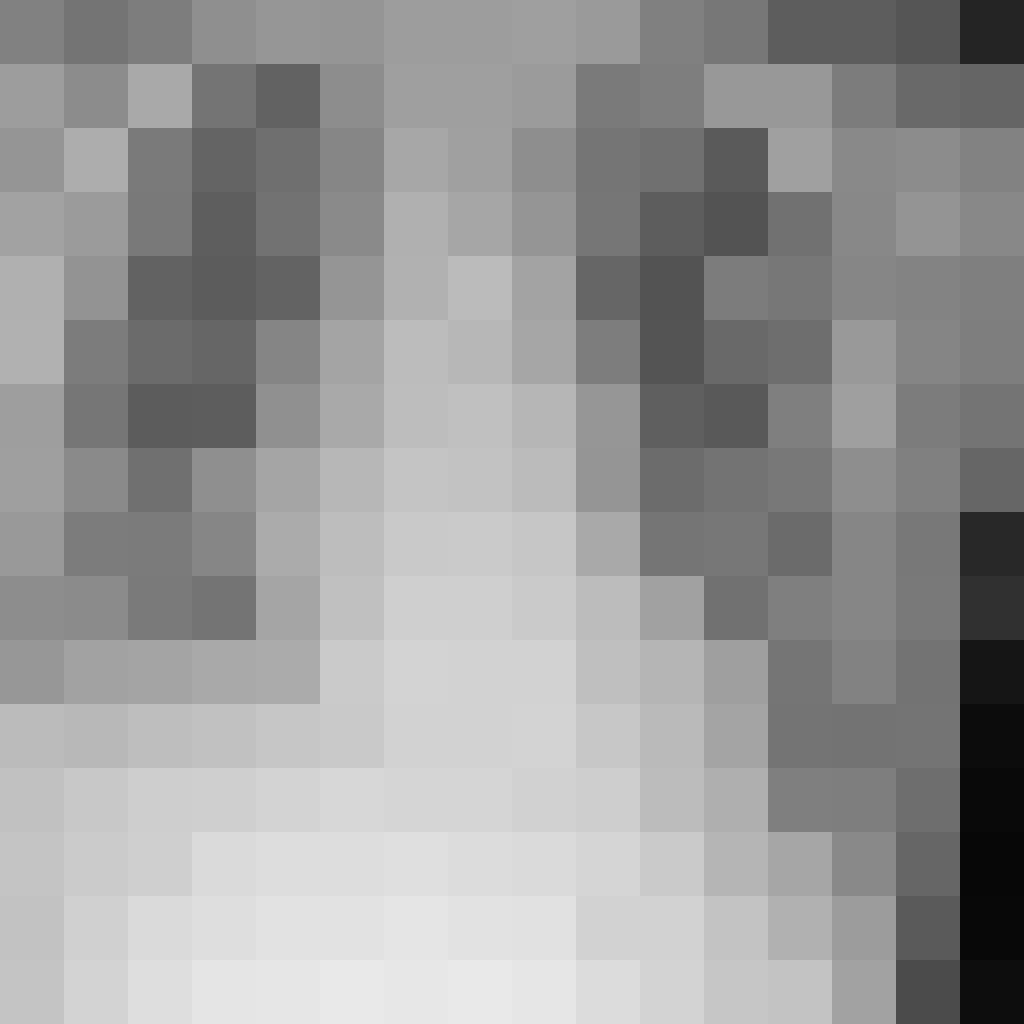} & \inc{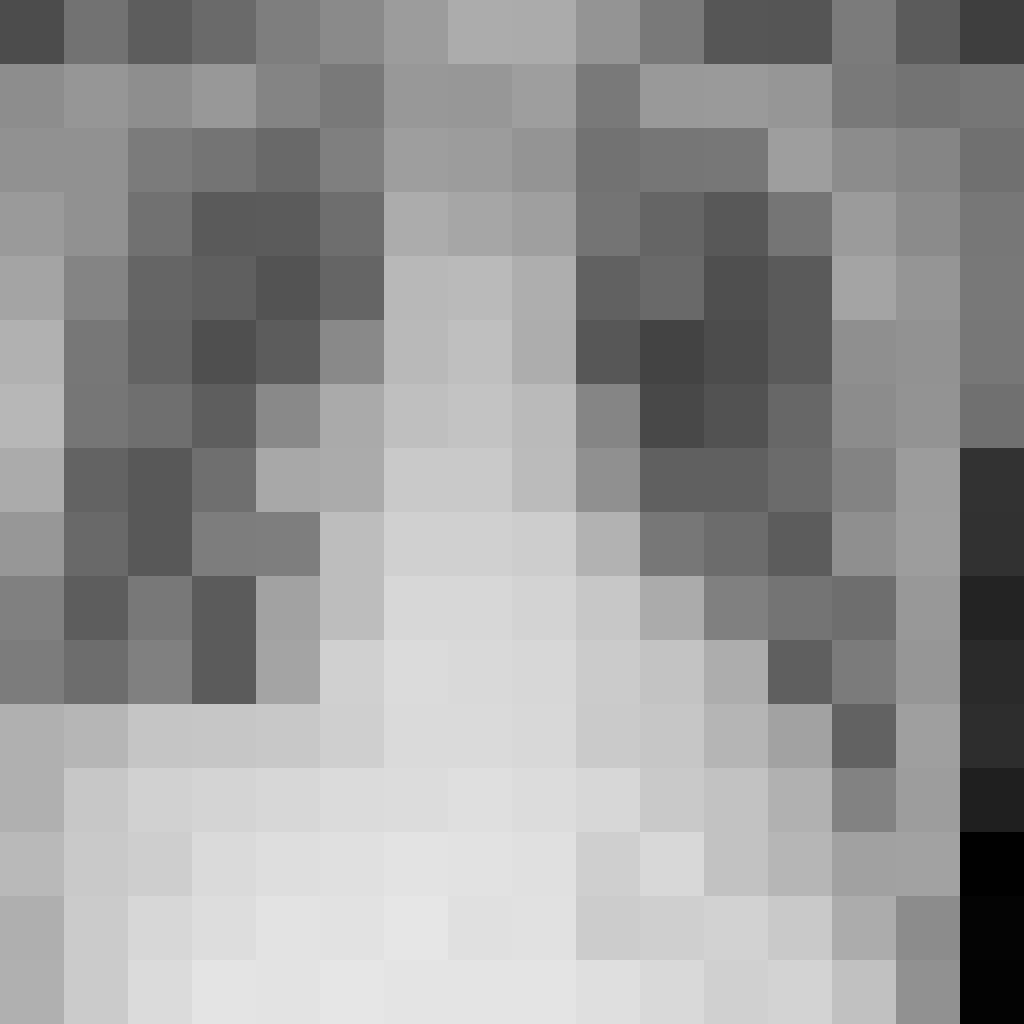}\\
 \end{tabular}

\renewcommand{\flds}{results/samBrains}
\renewcommand{\is}{img00000034}
\renewcommand{\step}{step400}

\begin{tabular}{c|c||c||ccccc}\arrayrulecolor{white}
% & True & Mean & Sample 1 & Sample 2 & Sample 3 & Sample 4 & Sample 5\\
HR & \inc{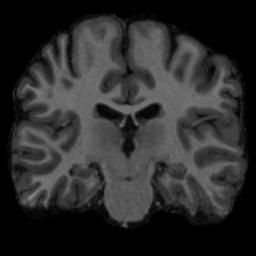} & \inc{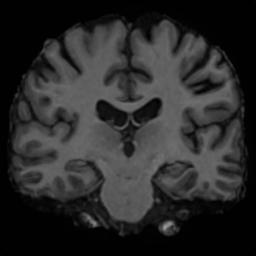} & \inc{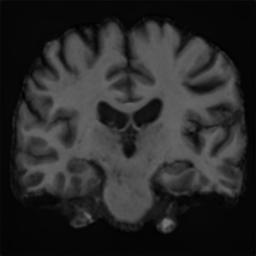} & \inc{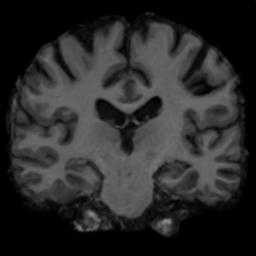} & \inc{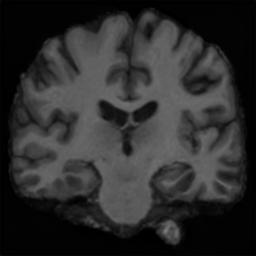} & \inc{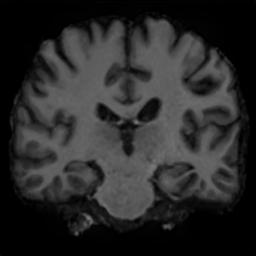} & \inc{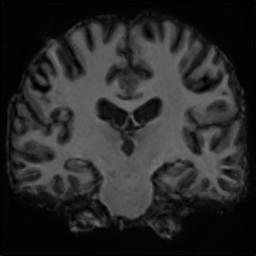}\\
 
LR & \inc{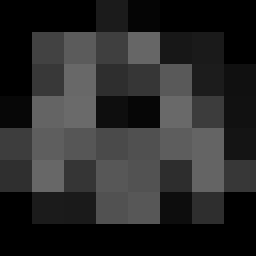} & \inc{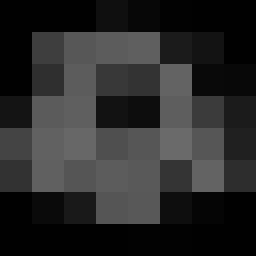} & \inc{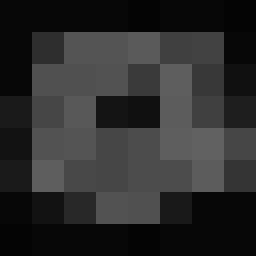} & \inc{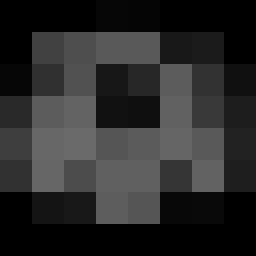} & \inc{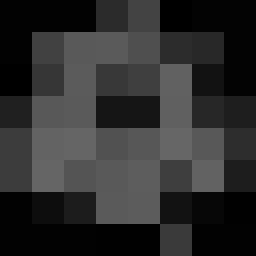} & \inc{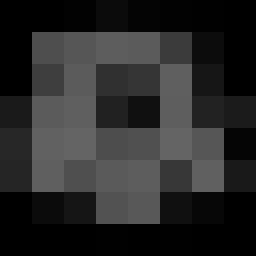} & \inc{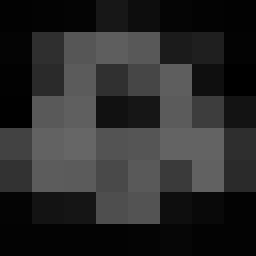}\\

 \end{tabular}
\caption{Sampling of multiple reconstructions using Variational Inference on super-resolution tasks with varying factors. From left, we show the true image, the estimated variational mean, alongside five random samples around that mean. For each high-resolution (HR) image, we show the corresponding low-resolution (LR) image below. While for some of the images, the reconstructions don't match the true image, the downsampled low-resolution images do match with the true image. We chose such extreme super-resolution in order to obtain a wide posterior distribution.}
\label{samplingsupp}
\end{figure*}

% samFFHQinp/01014_mergedsample5_step290.jpg  samFFHQinp/35529_mergedsample5_step290.jpg   samFFHQinp/61041_mergedsample5_step290.jpg
% samFFHQinp/03456_mergedsample5_step290.jpg  samFFHQinp/41697_mergedsample5_step290.jpg   samFFHQinp/61356_mergedsample5_step290.jpg
% samFFHQinp/07129_mergedsample5_step290.jpg  samFFHQinp/42025_mergedsample5_step290.jpg   samFFHQinp/62243_mergedsample5_step290.jpg
% samFFHQinp/10306_mergedsample5_step290.jpg  samFFHQinp/49049_mergedsample5_step290.jpg   samFFHQinp/64408_mergedsample5_step290.jpg
% samFFHQinp/15489_mergedsample5_step290.jpg  samFFHQinp/52116_mergedsample5_step290.jpg   samFFHQinp/66859_mergedsample5_step290.jpg
% samFFHQinp/22706_mergedsample5_step290.jpg  samFFHQinp/56008_mergedsample5_step2900.jpg  samFFHQinp/68319_mergedsample5_step290.jpg
% samFFHQinp/22863_mergedsample5_step290.jpg  samFFHQinp/56008_mergedsample5_step290.jpg
% samFFHQinp/27058_mergedsample5_step290.jpg  samFFHQinp/60260_mergedsample5_step290.jpg

\renewcommand{\flds}{results/samFFHQinp}
\renewcommand{\is}{01014}
\renewcommand{\step}{step270}

\begin{figure*}
\setlength{\tabcolsep}{0pt}
\renewcommand{\arraystretch}{1.5}% Spread rows out...
\begin{tabular}{p{1.1cm}|c||c||ccccc}\arrayrulecolor{white}
& True & Est. Mean & Sample 1 & Sample 2 & Sample 3 & Sample 4 & Sample 5\\
Clean & \inc{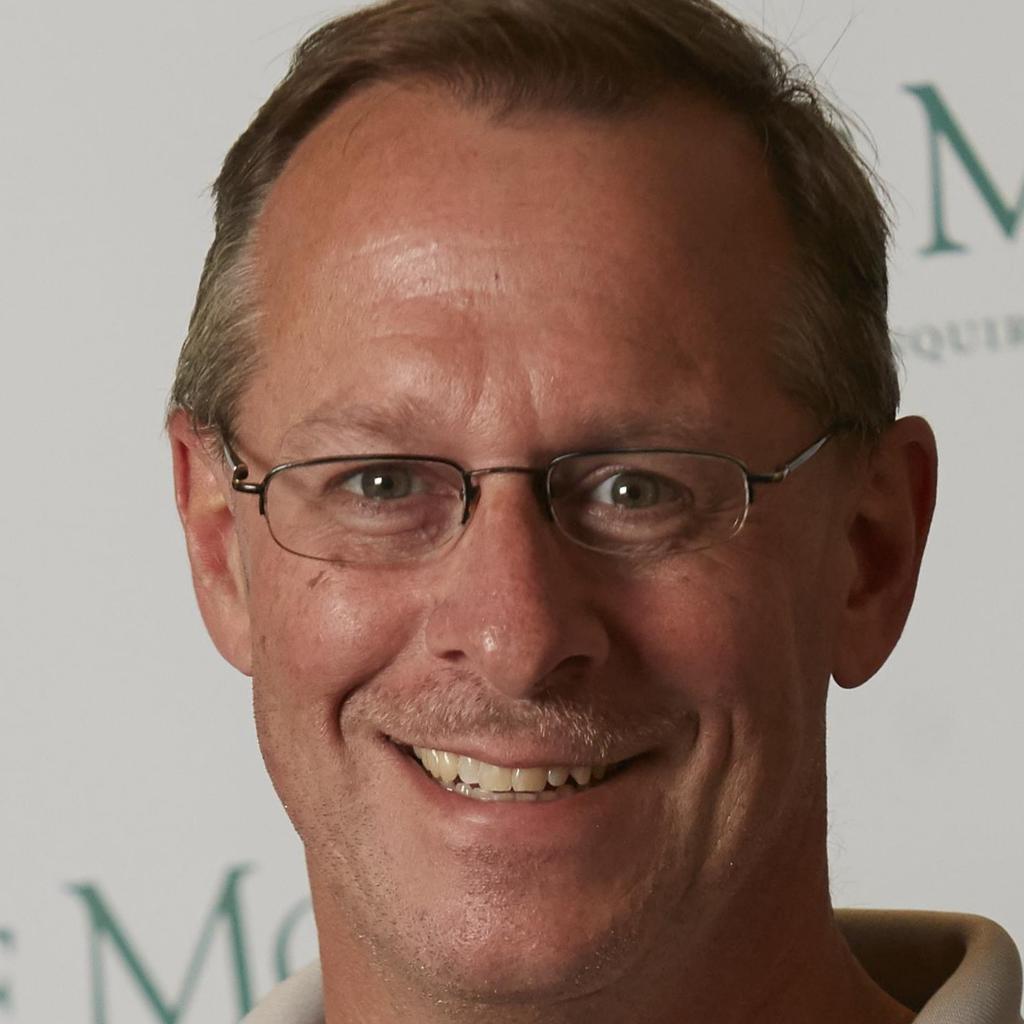} & \inc{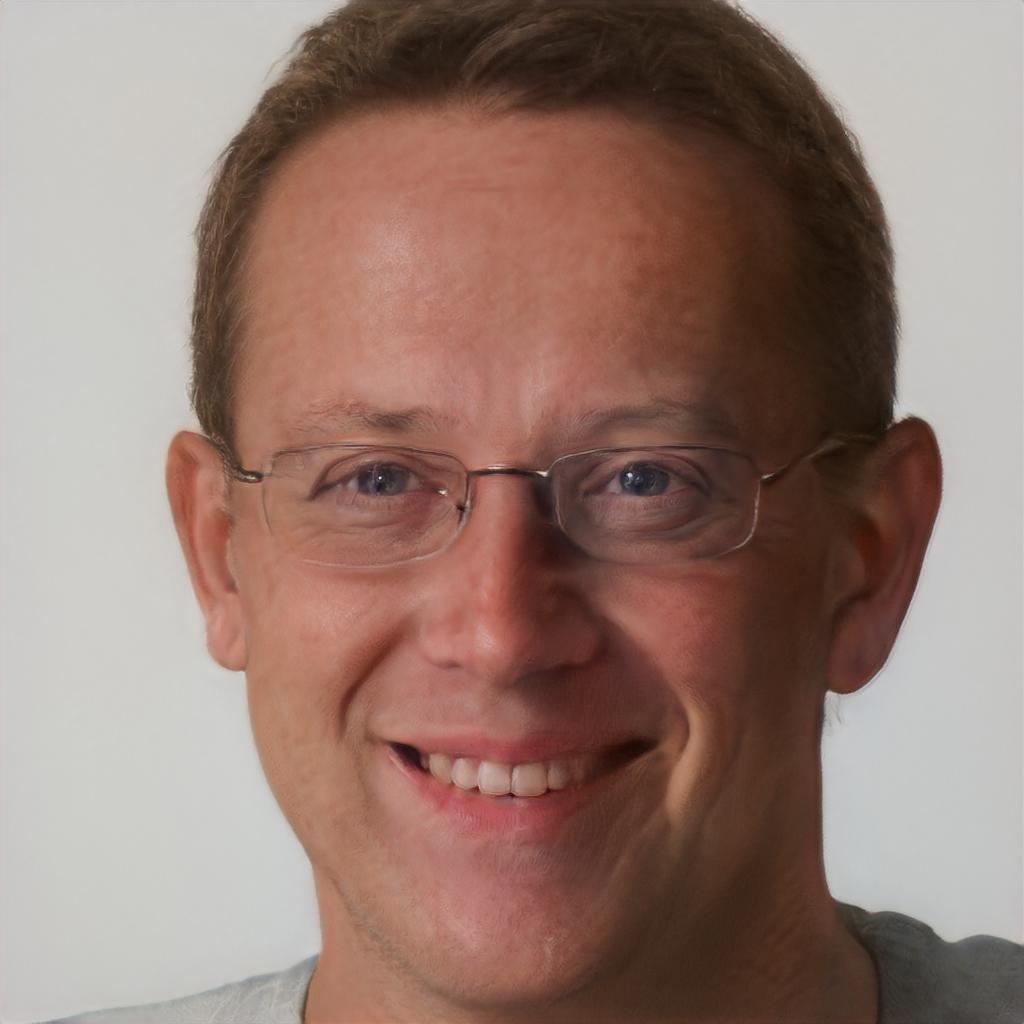} & \inc{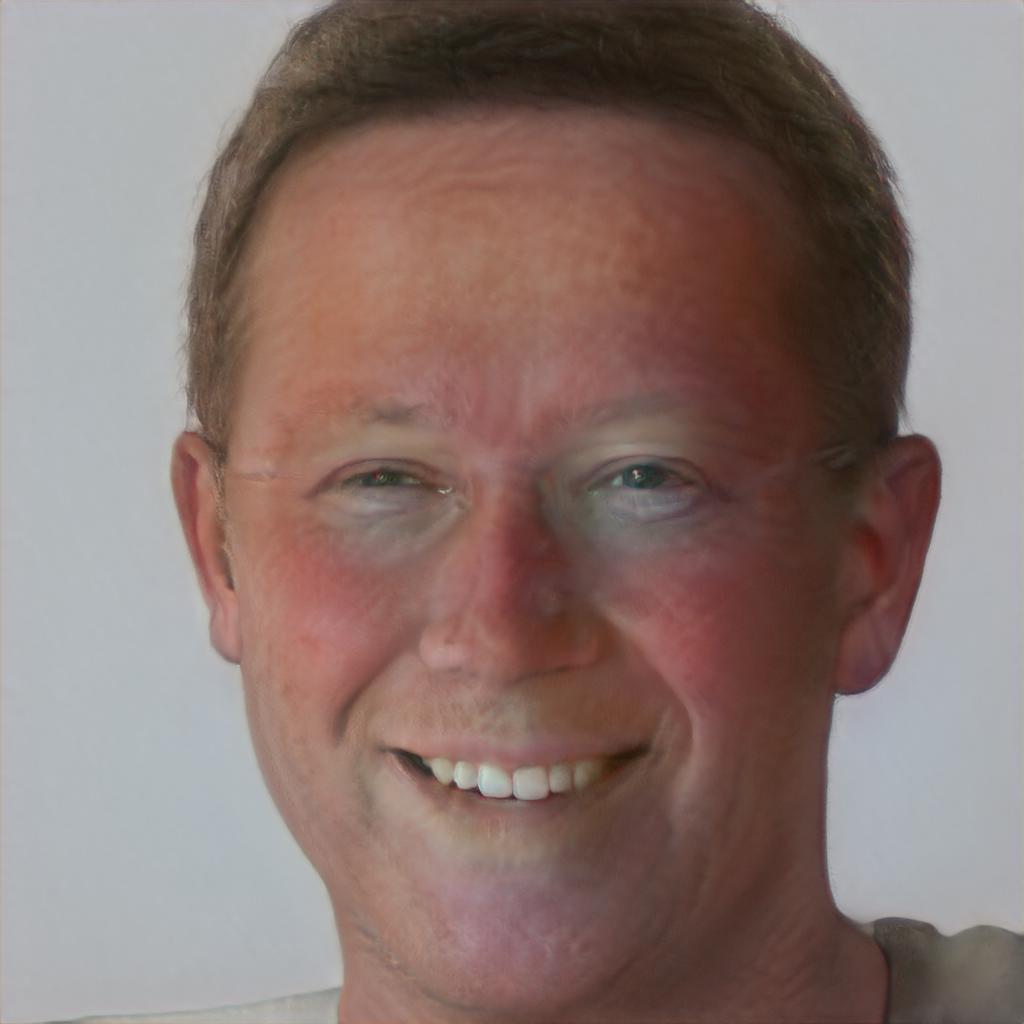} & \inc{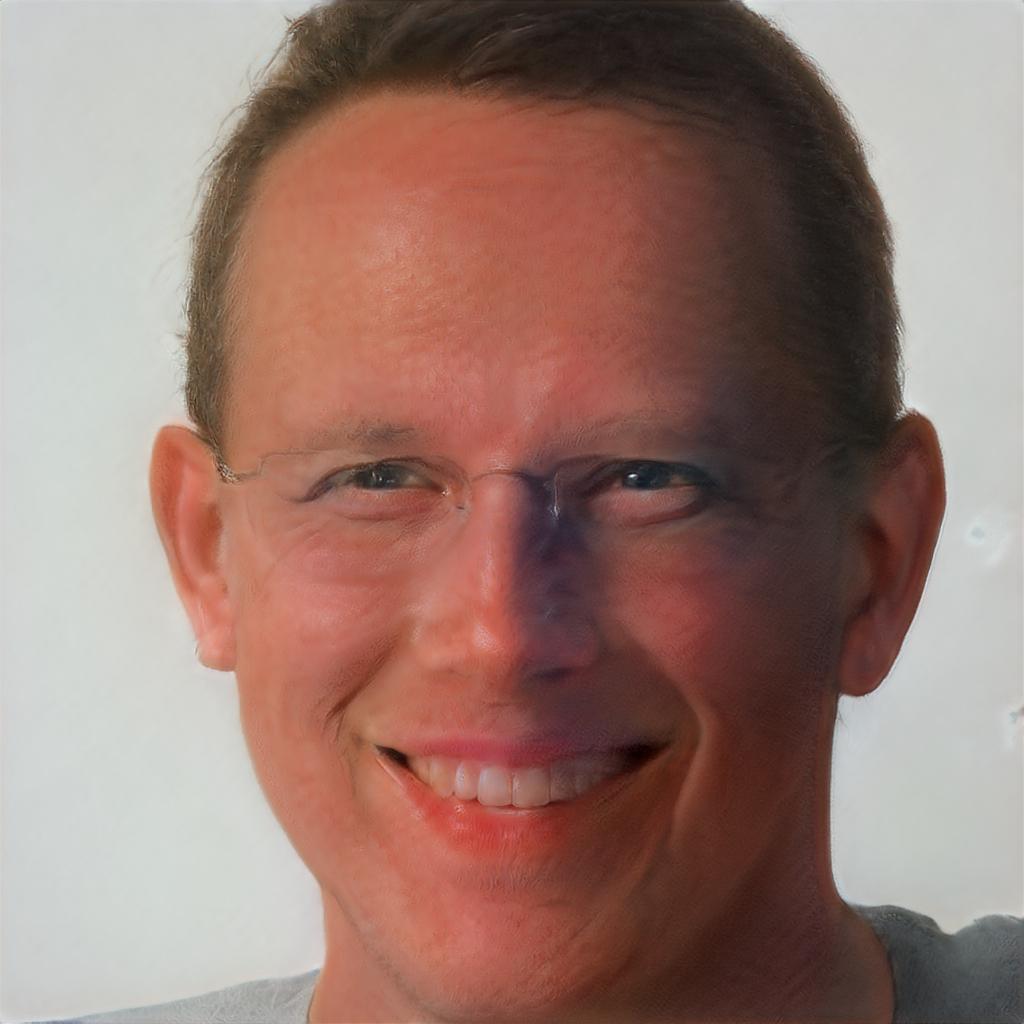} & \inc{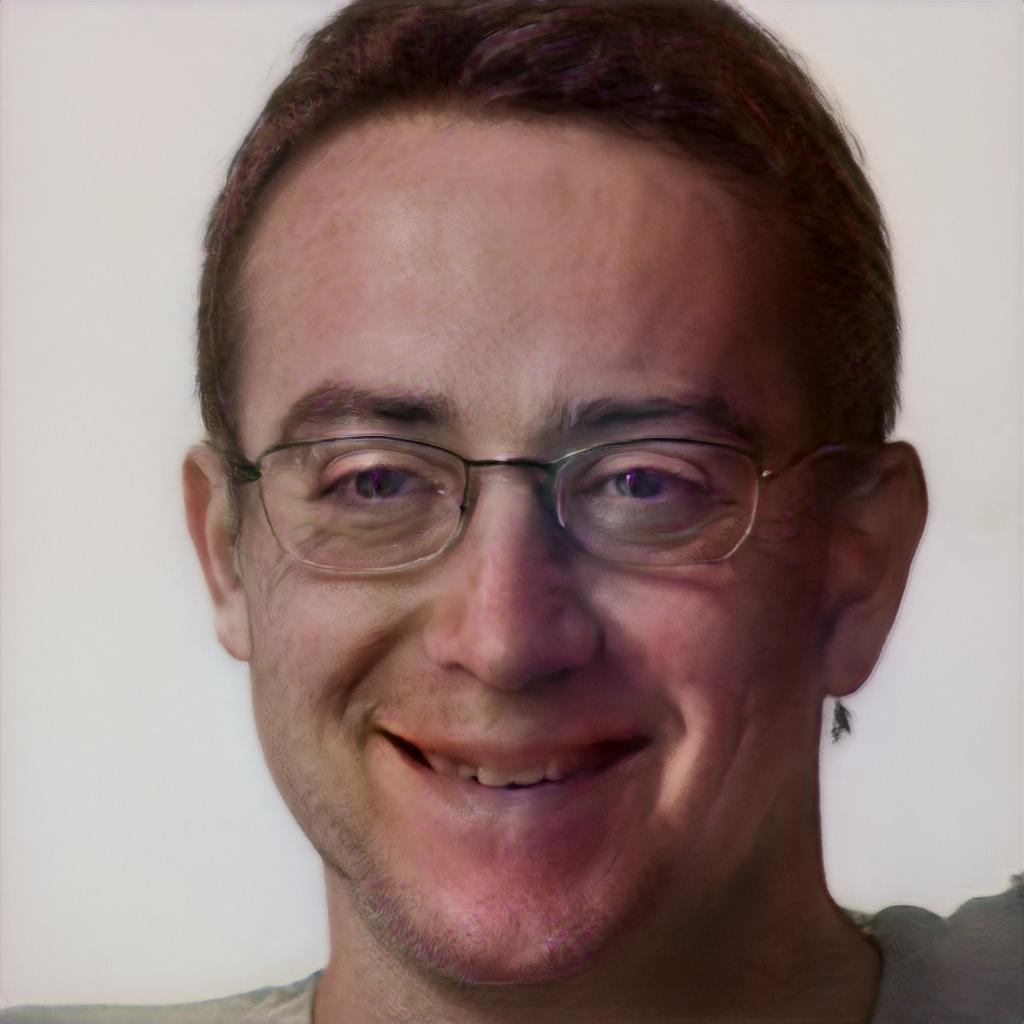} & \inc{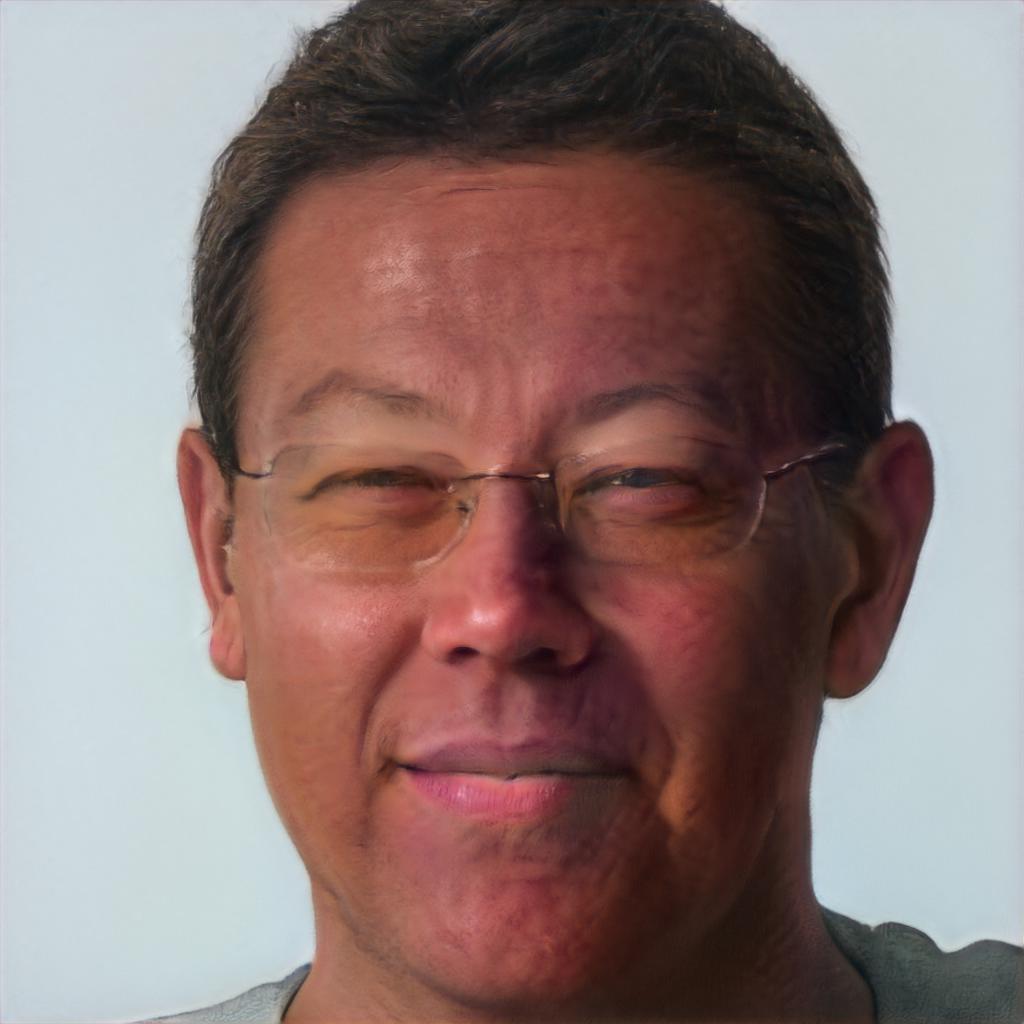} & \inc{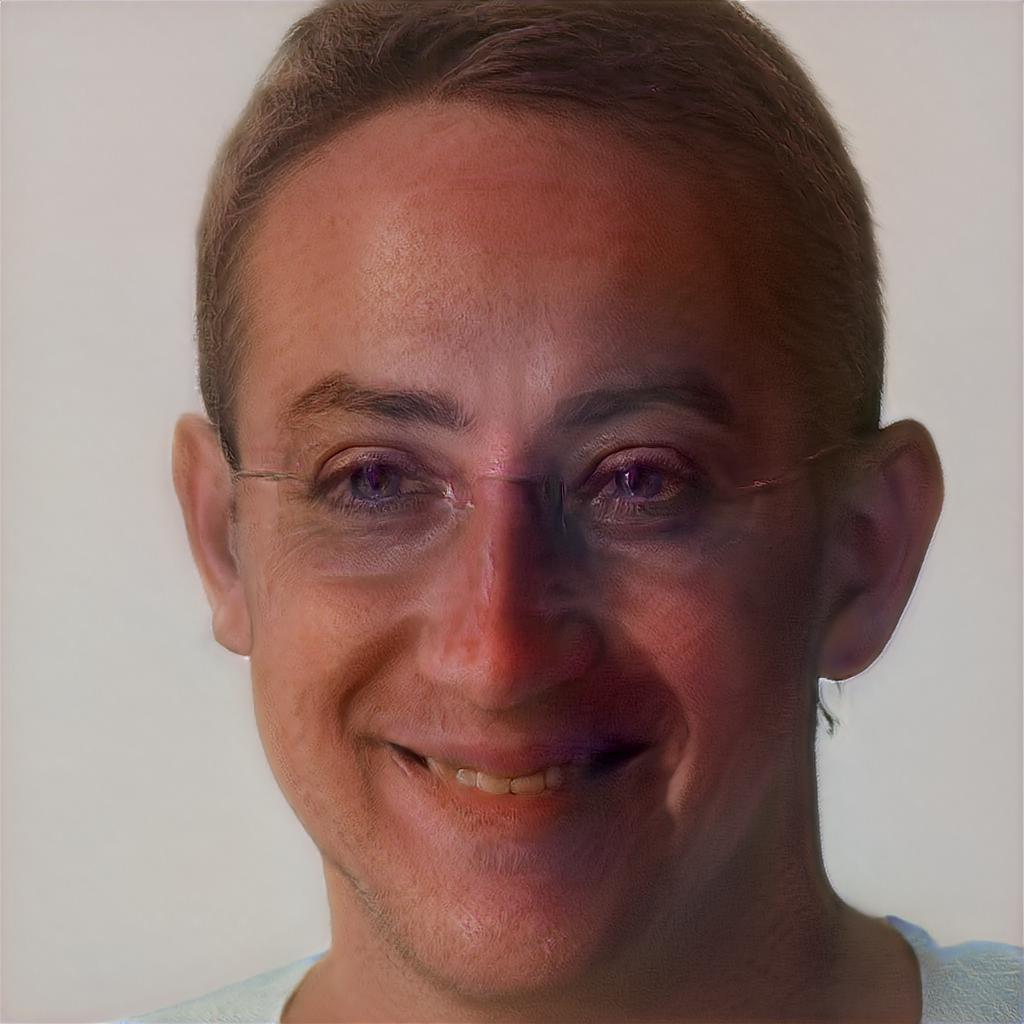}\\
 
Mask/ Merged & \inc{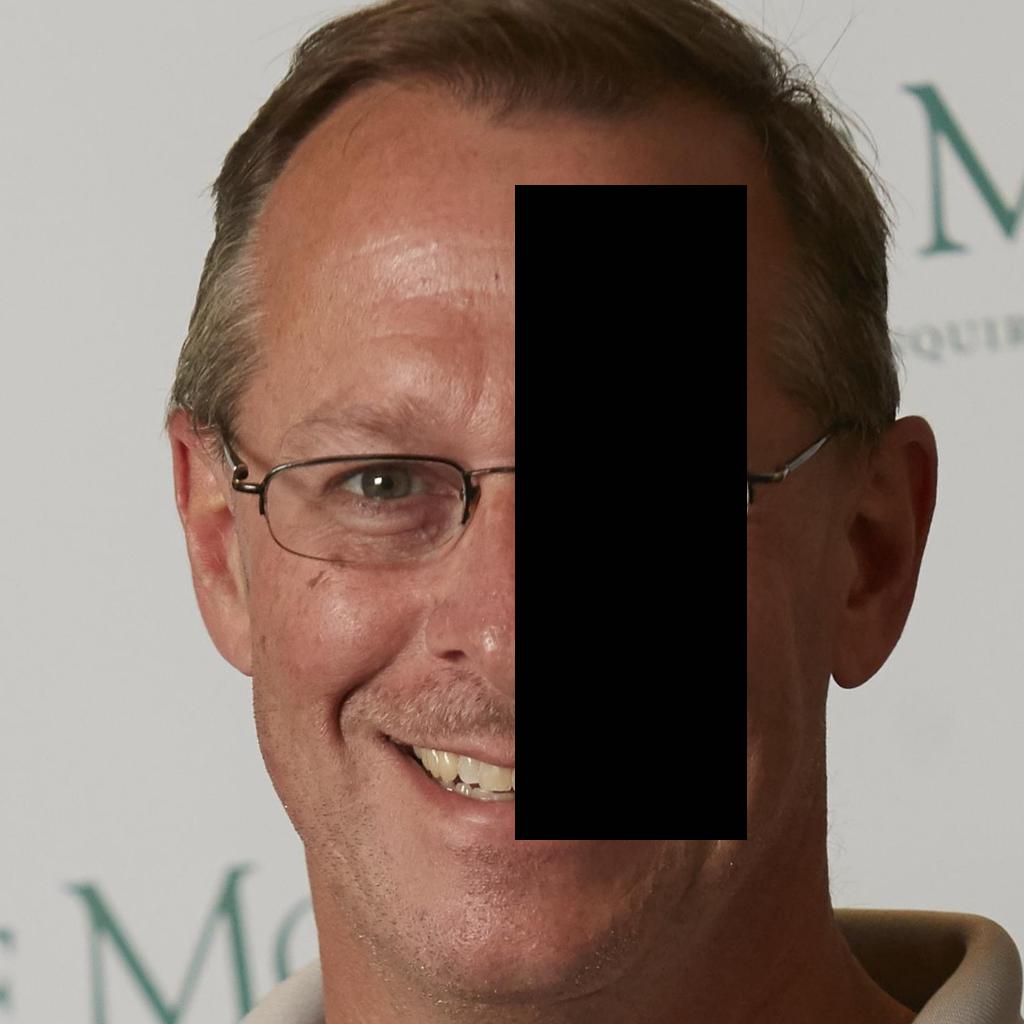} & \inc{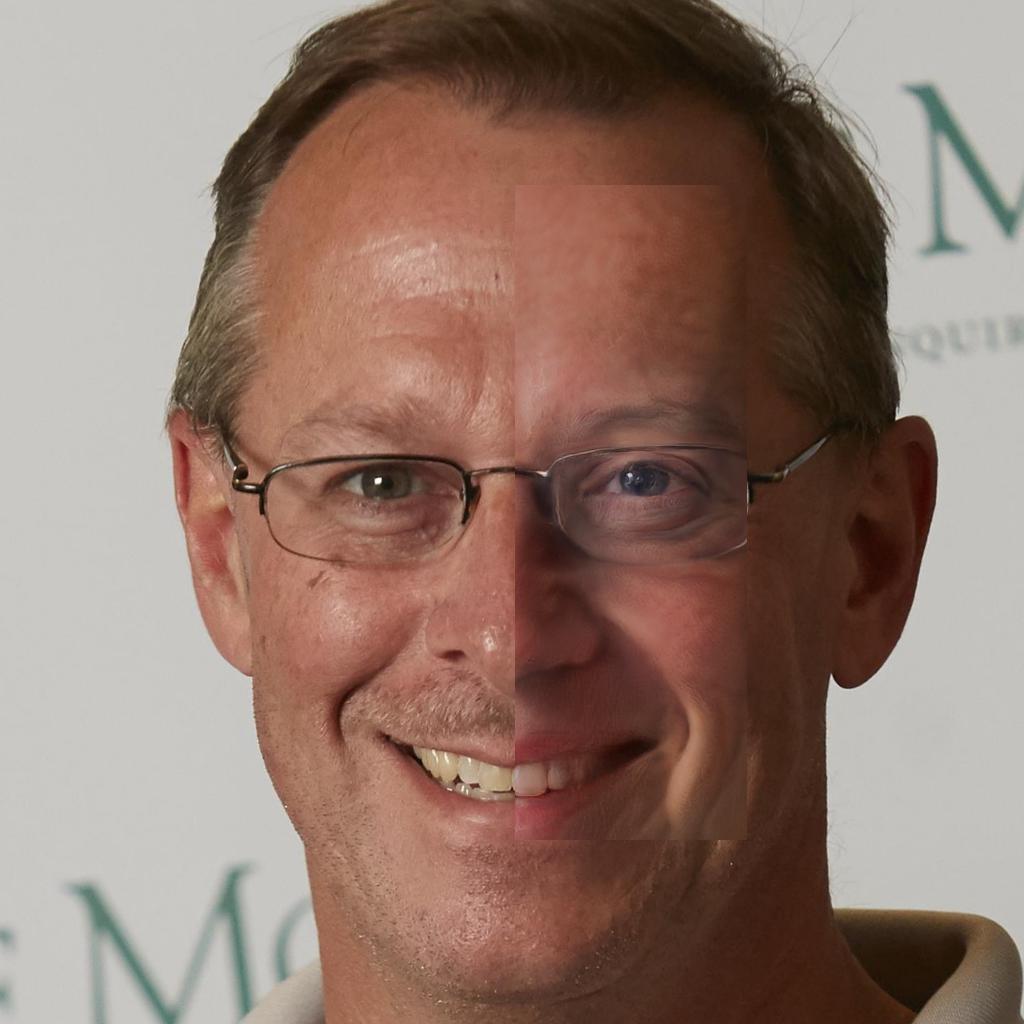} & \inc{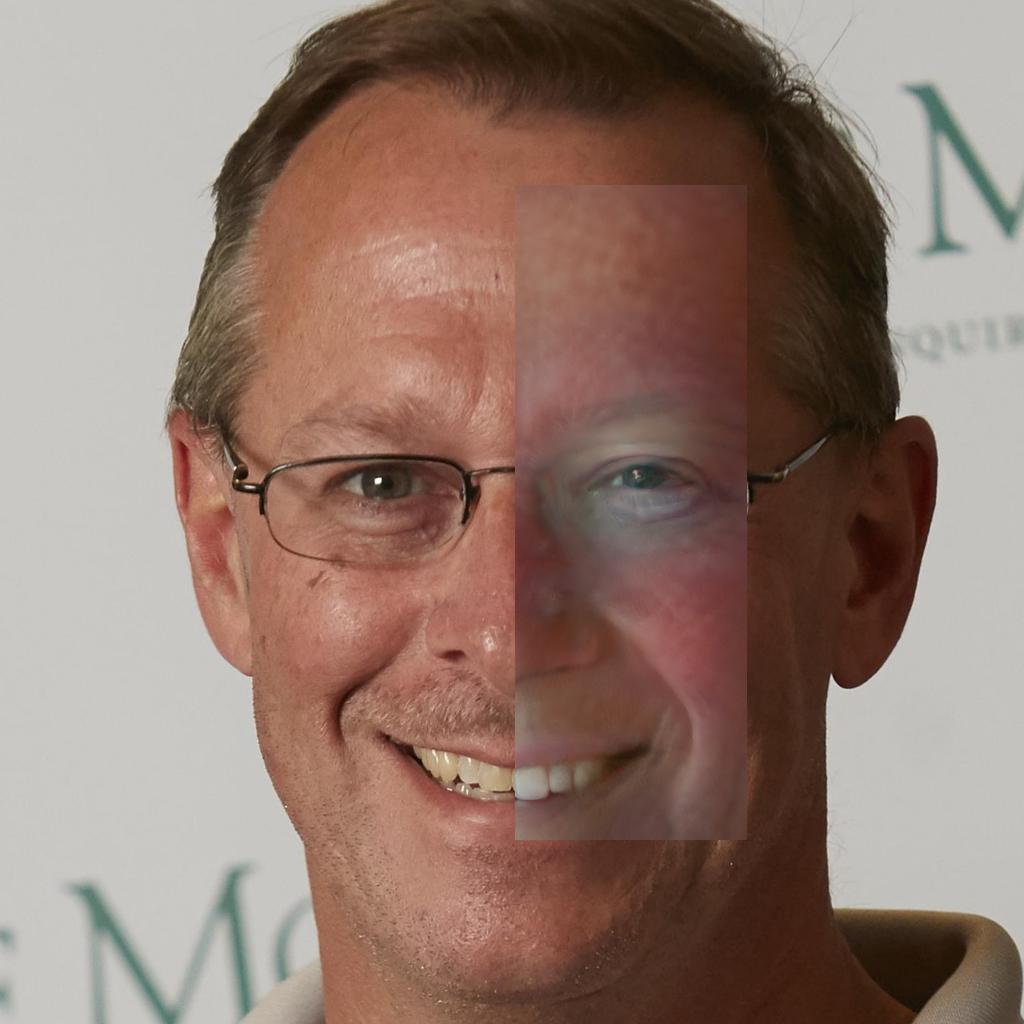} & \inc{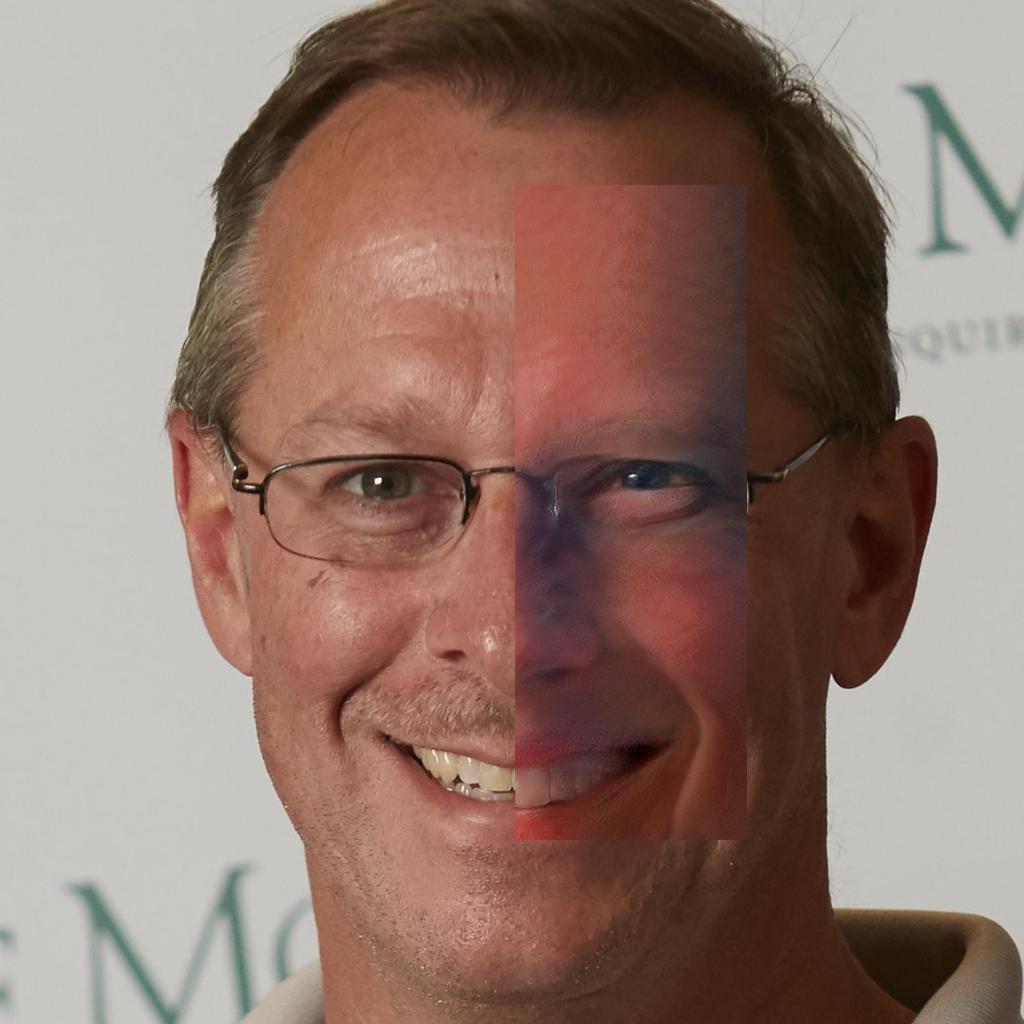} & \inc{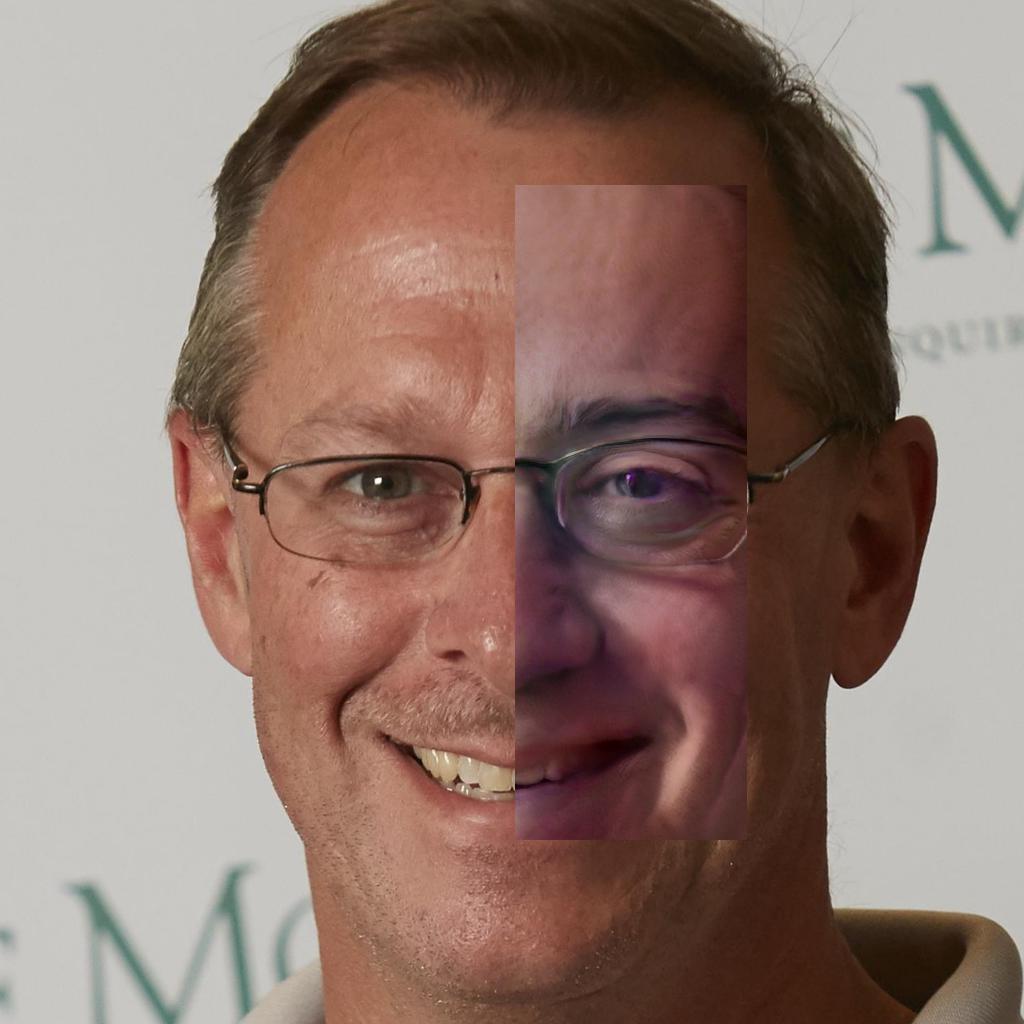} & \inc{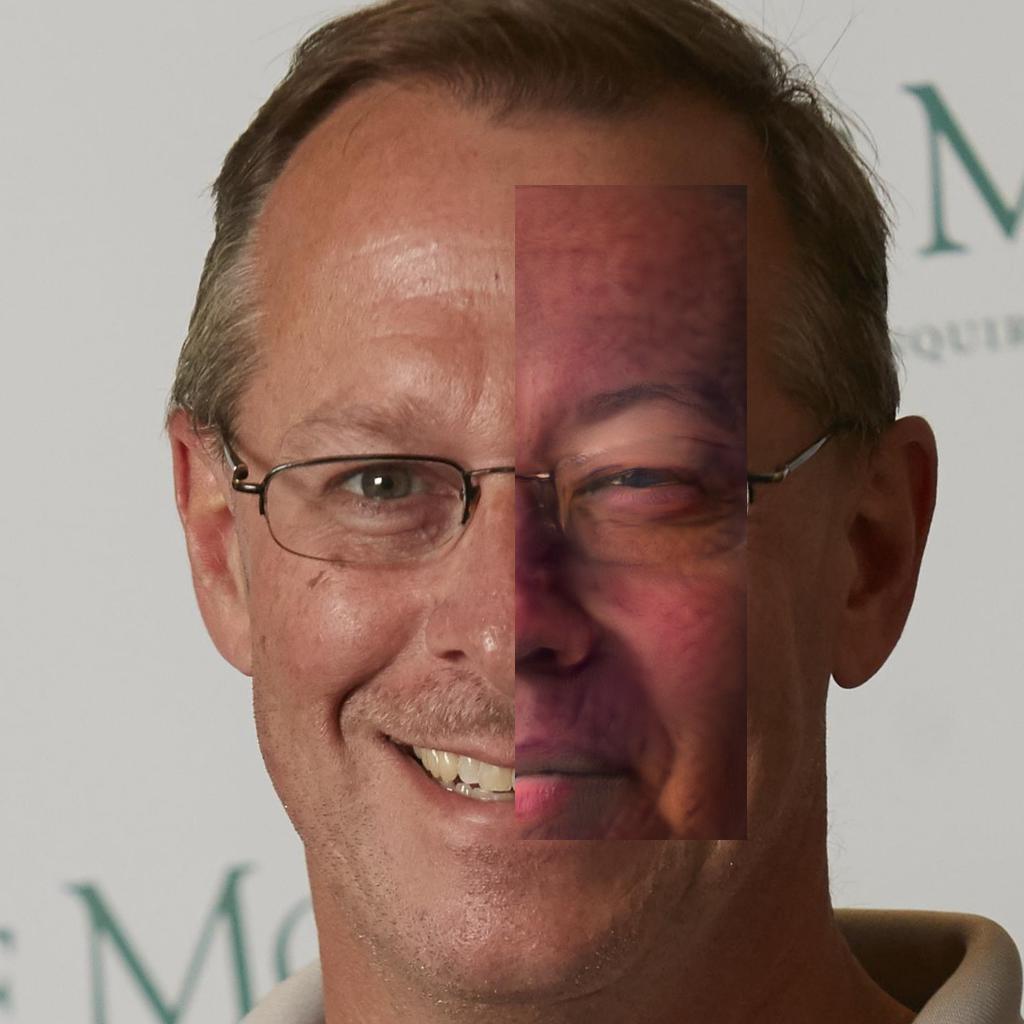} & \inc{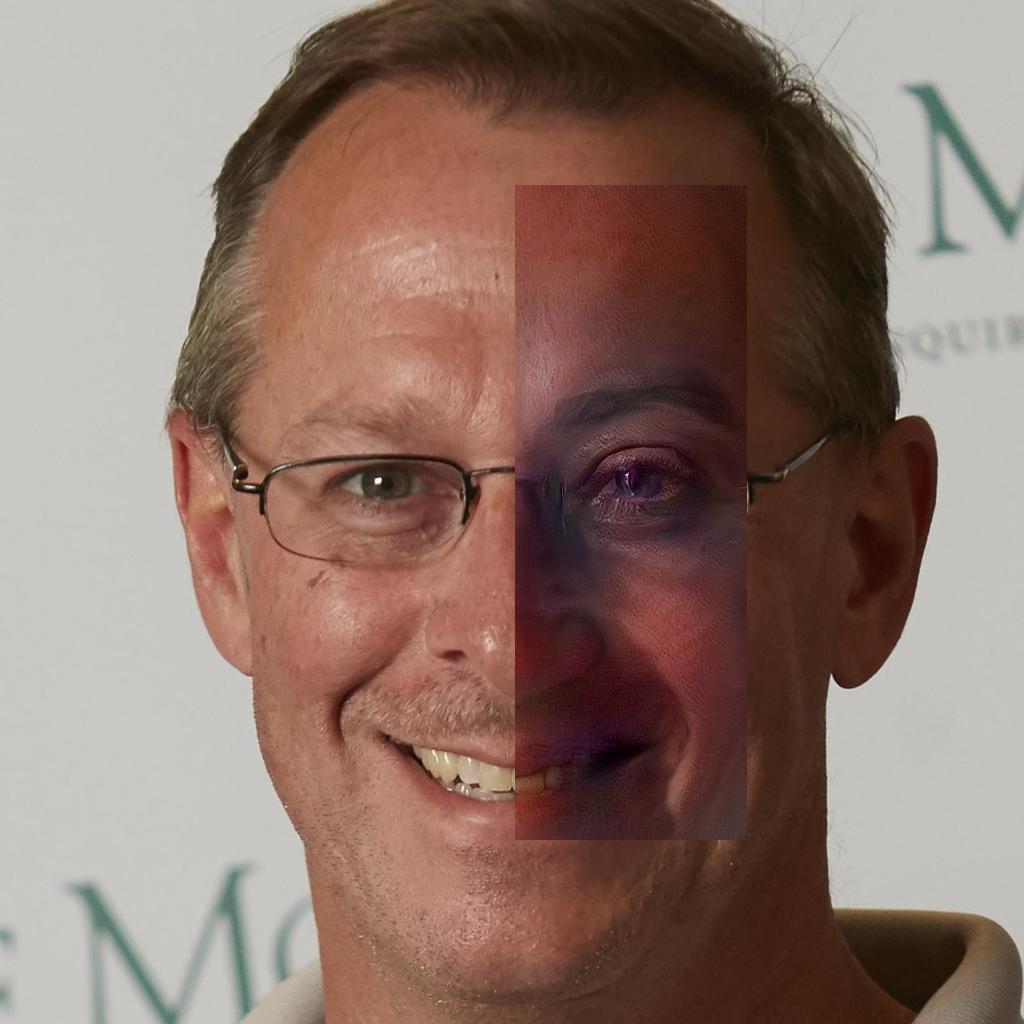}\\
 \end{tabular}

 \renewcommand{\is}{03456}
\renewcommand{\step}{step290}
 
 \begin{tabular}{p{1.1cm}|c||c||ccccc}\arrayrulecolor{white}
% & True & Mean & Sample 1 & Sample 2 & Sample 3 & Sample 4 & Sample 5\\
Clean & \inc{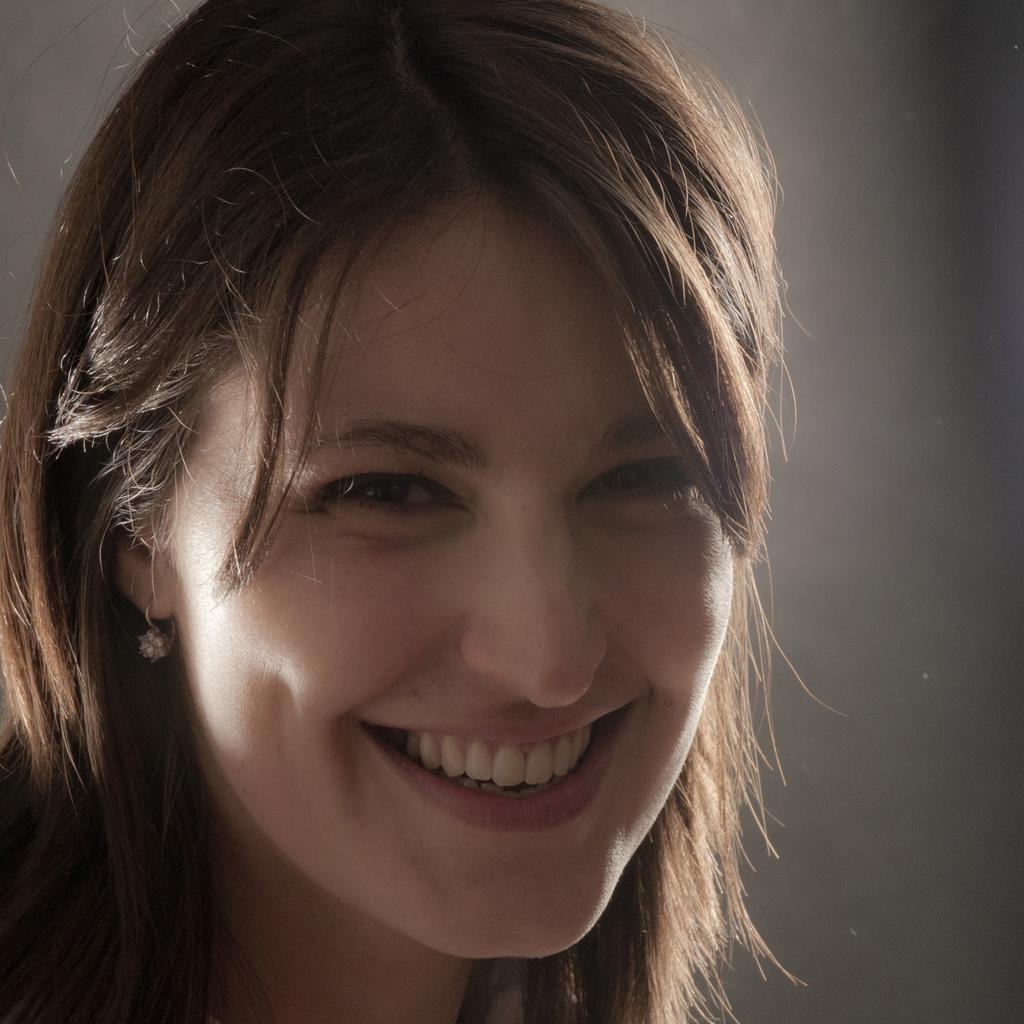} & \inc{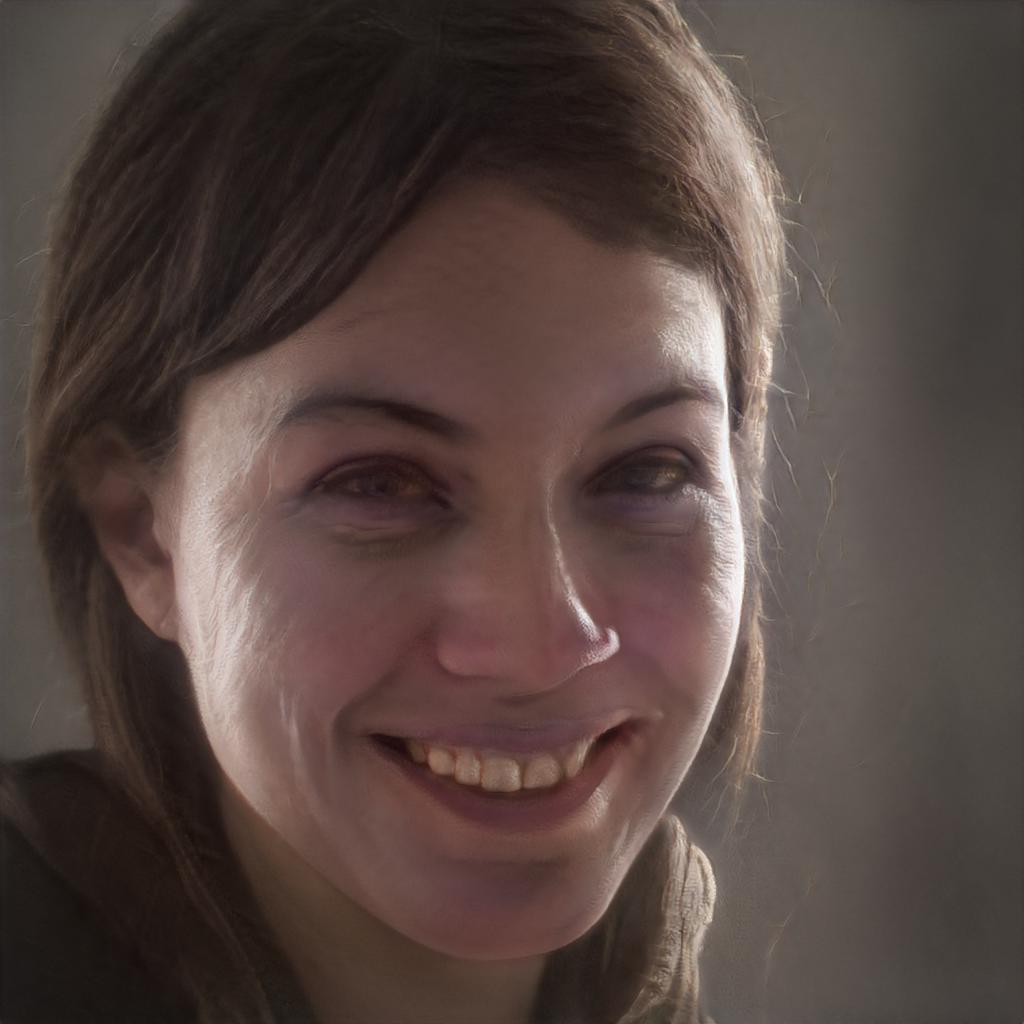} & \inc{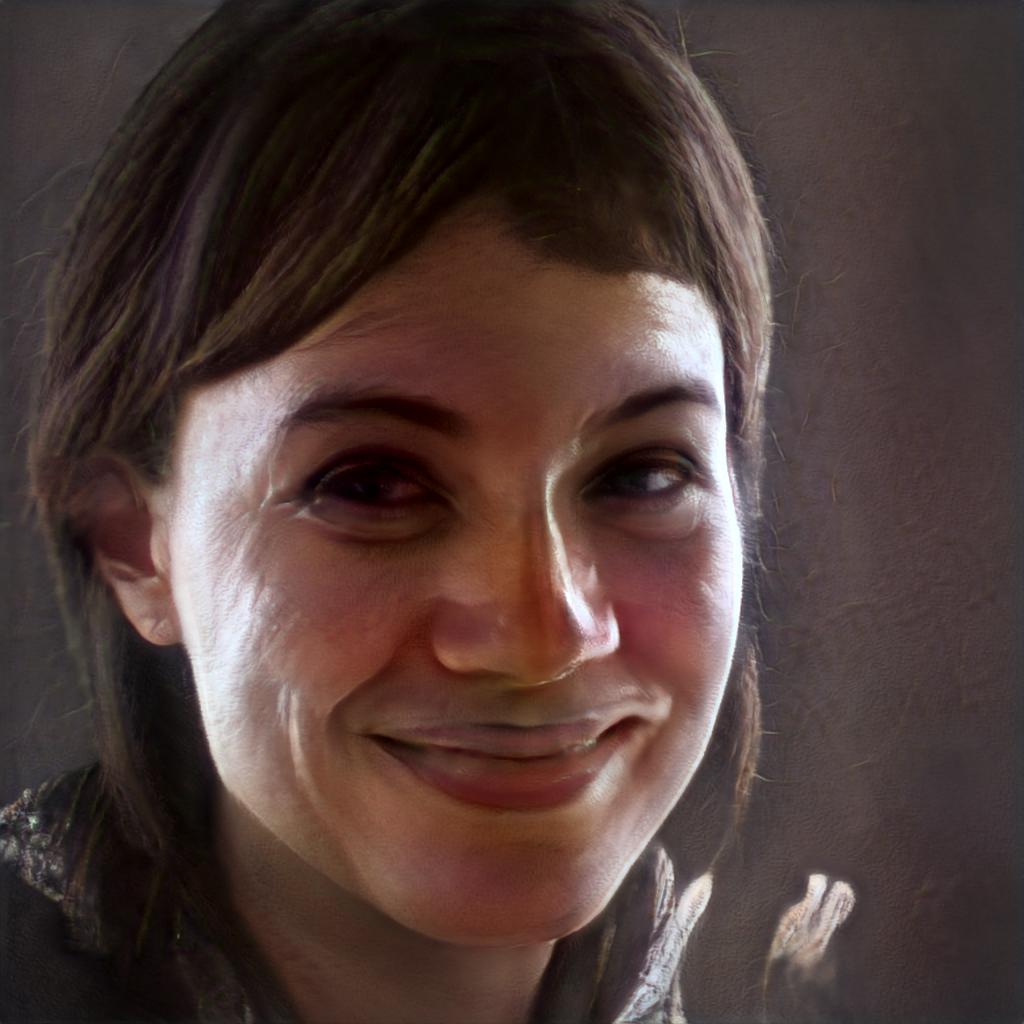} & \inc{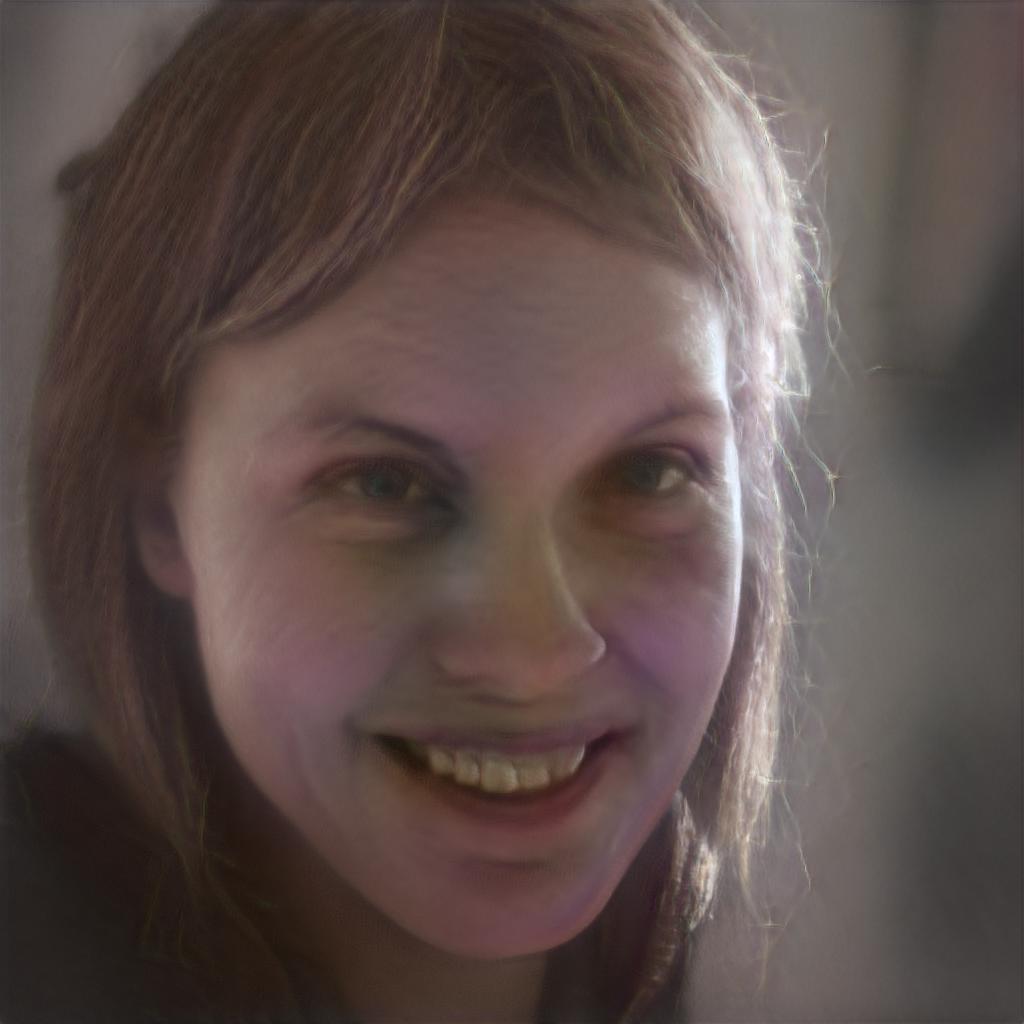} & \inc{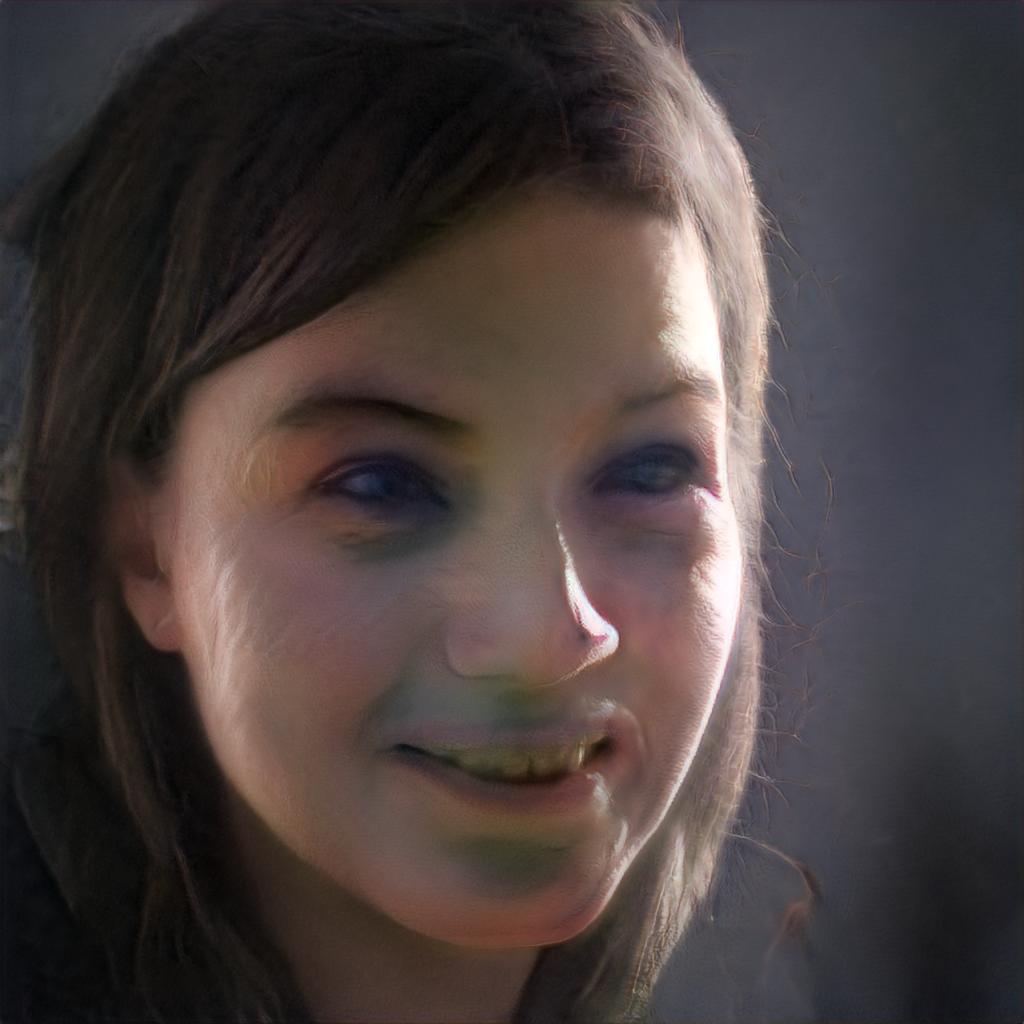} & \inc{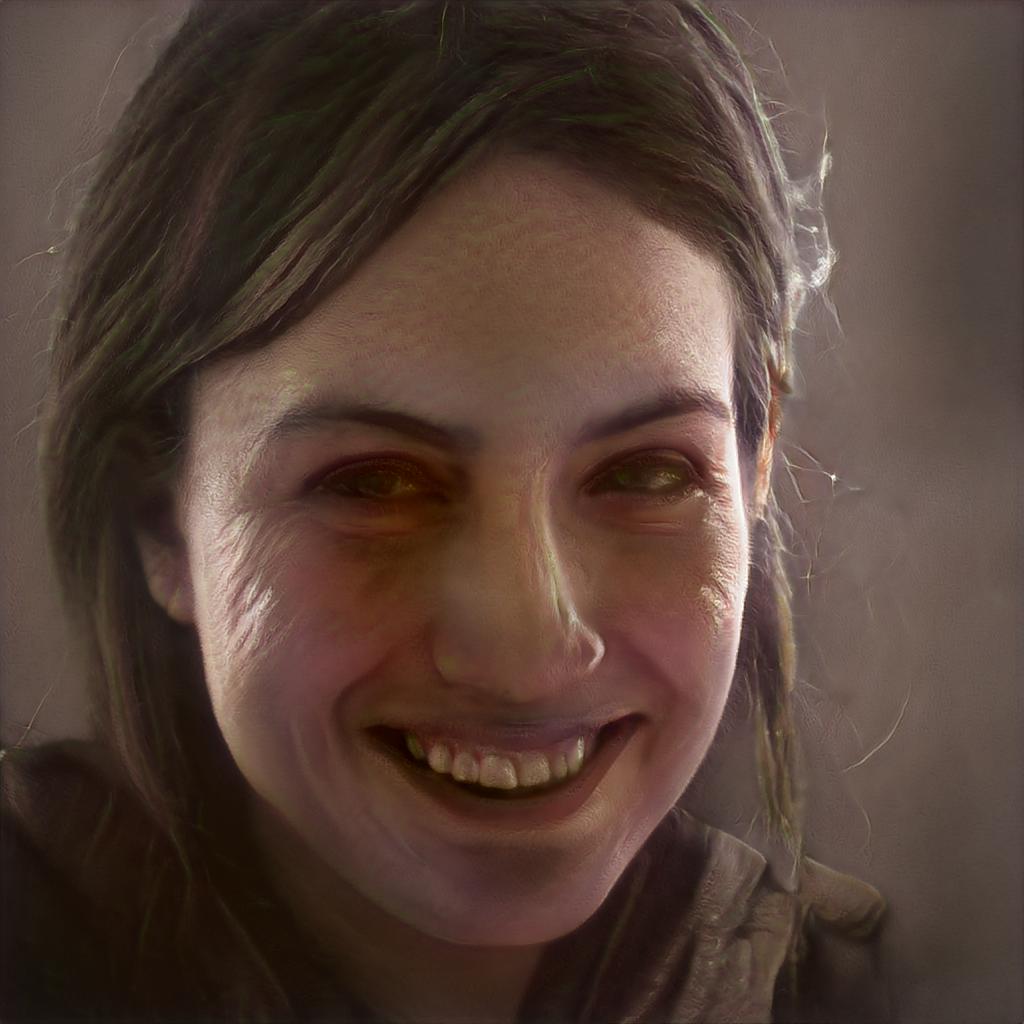} & \inc{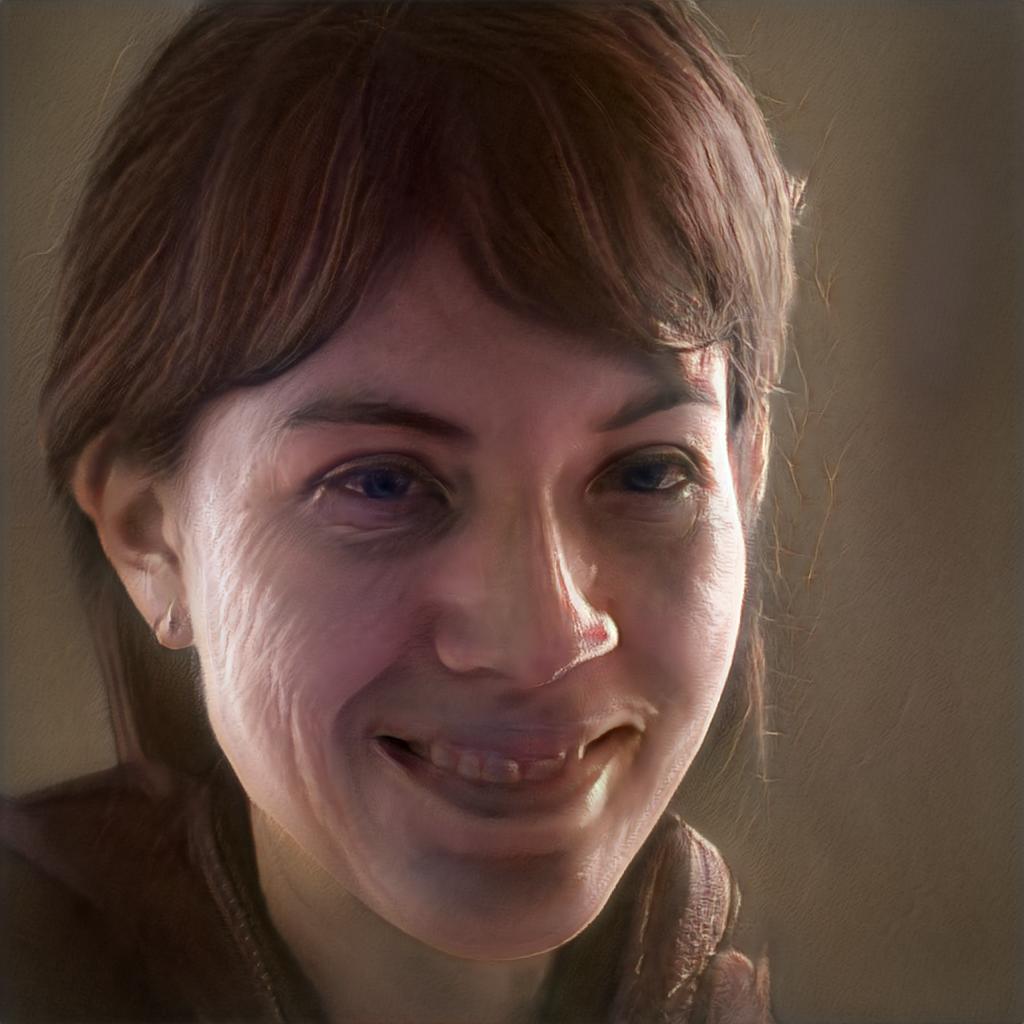}\\
 
Mask/ Merged & \inc{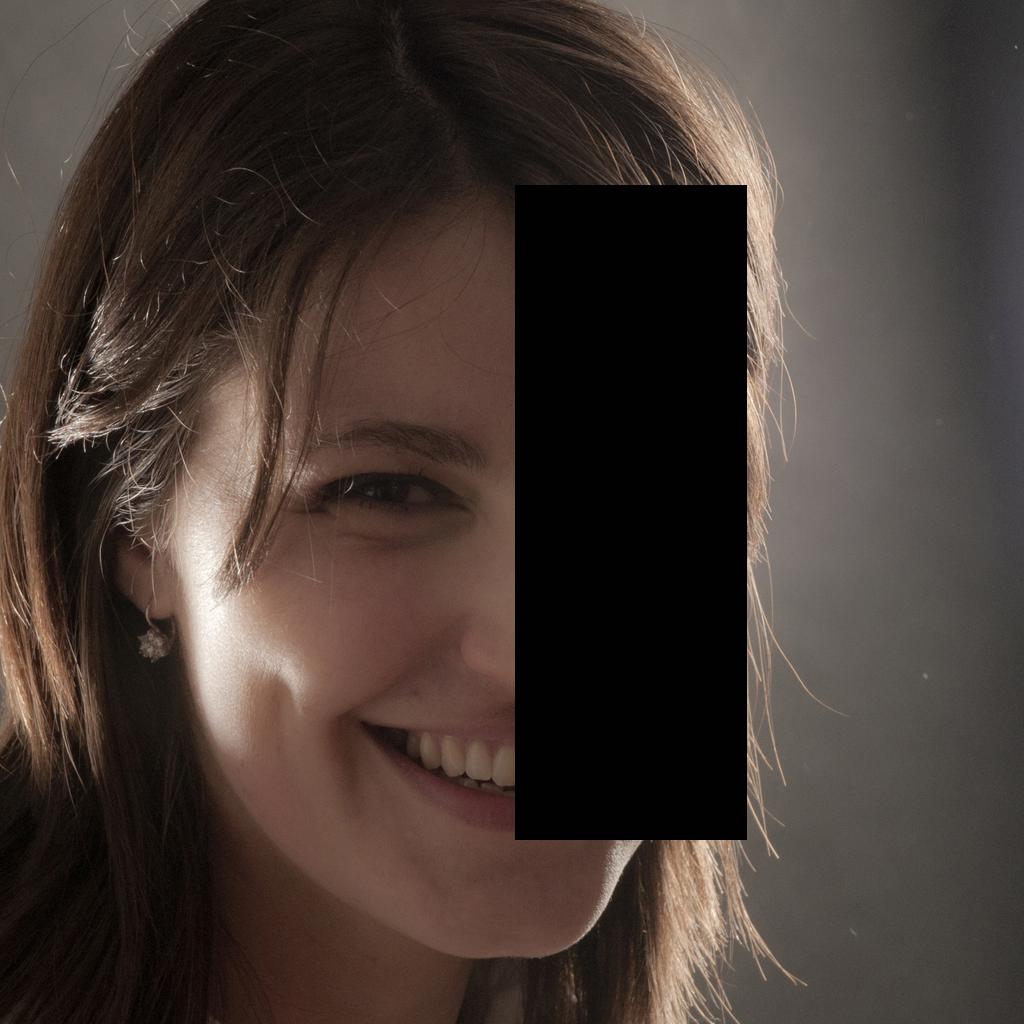} & \inc{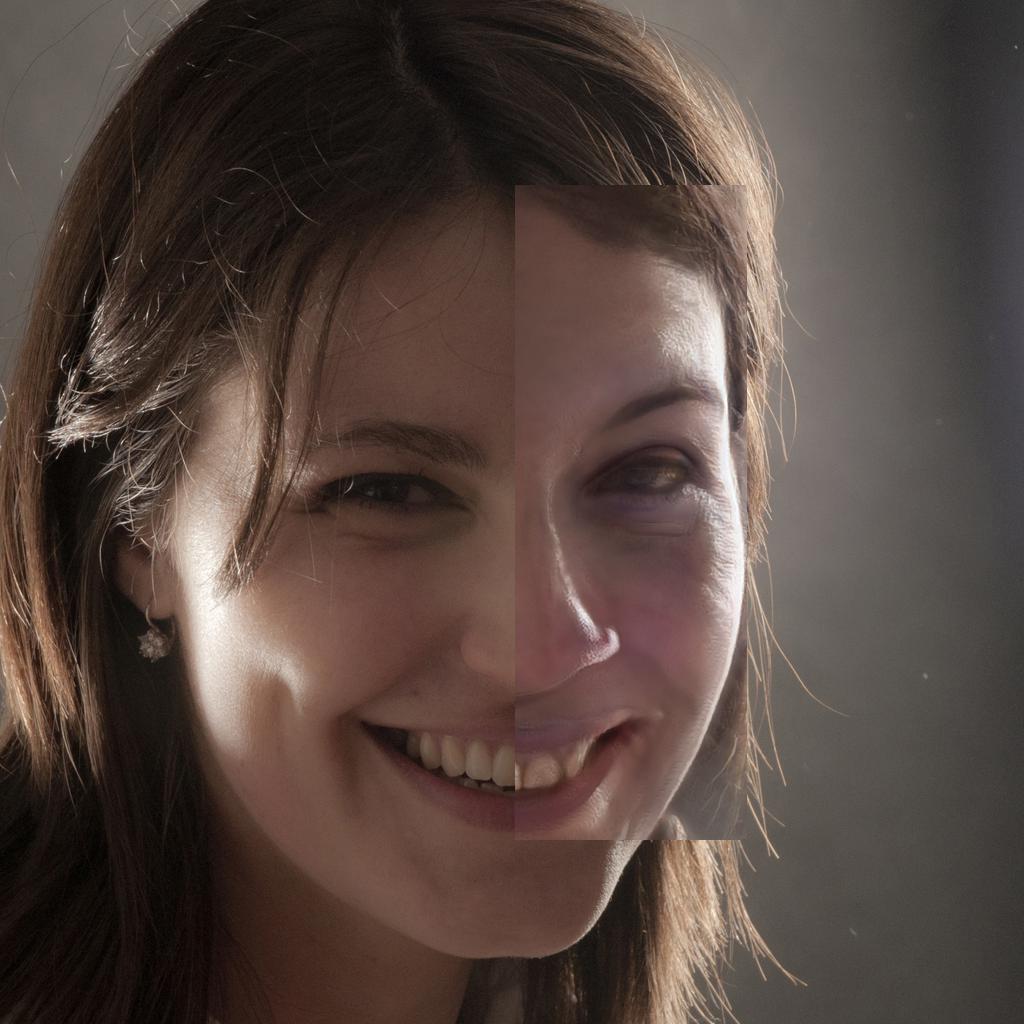} & \inc{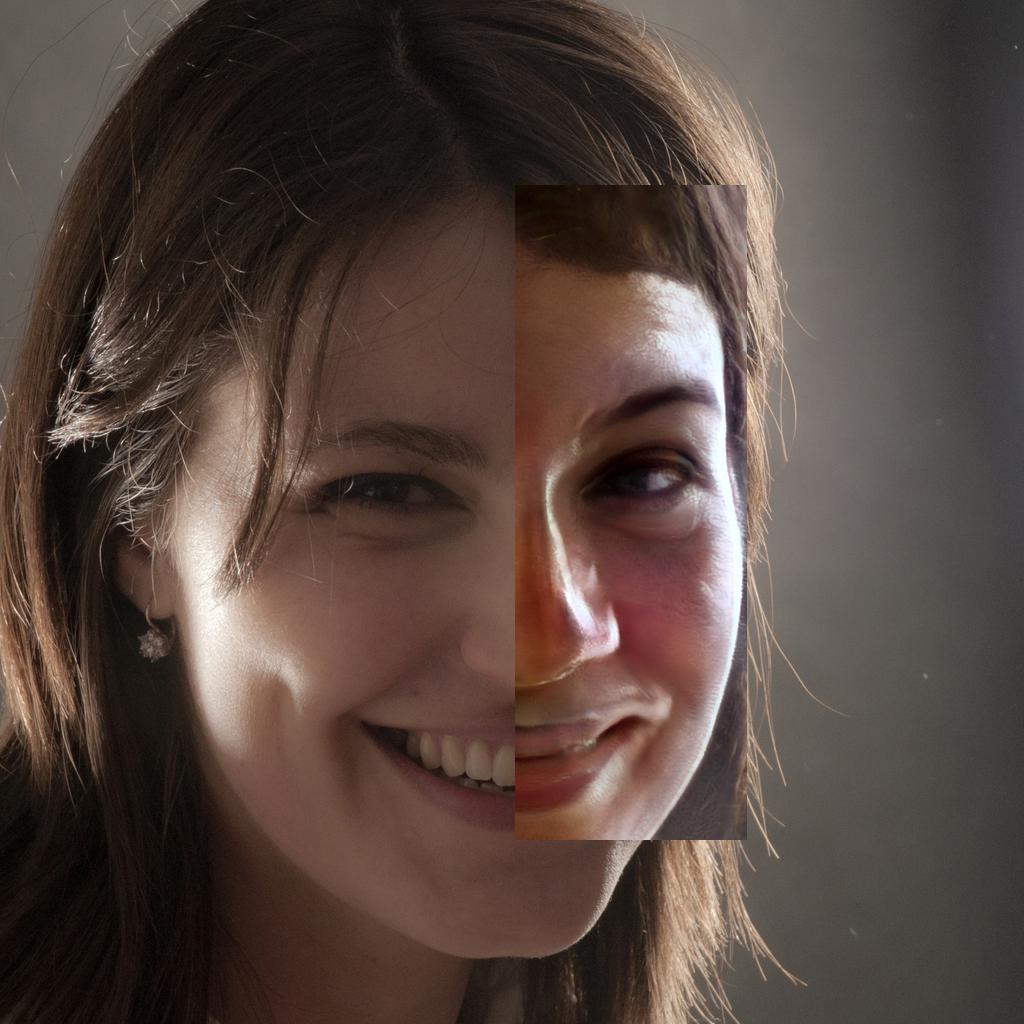} & \inc{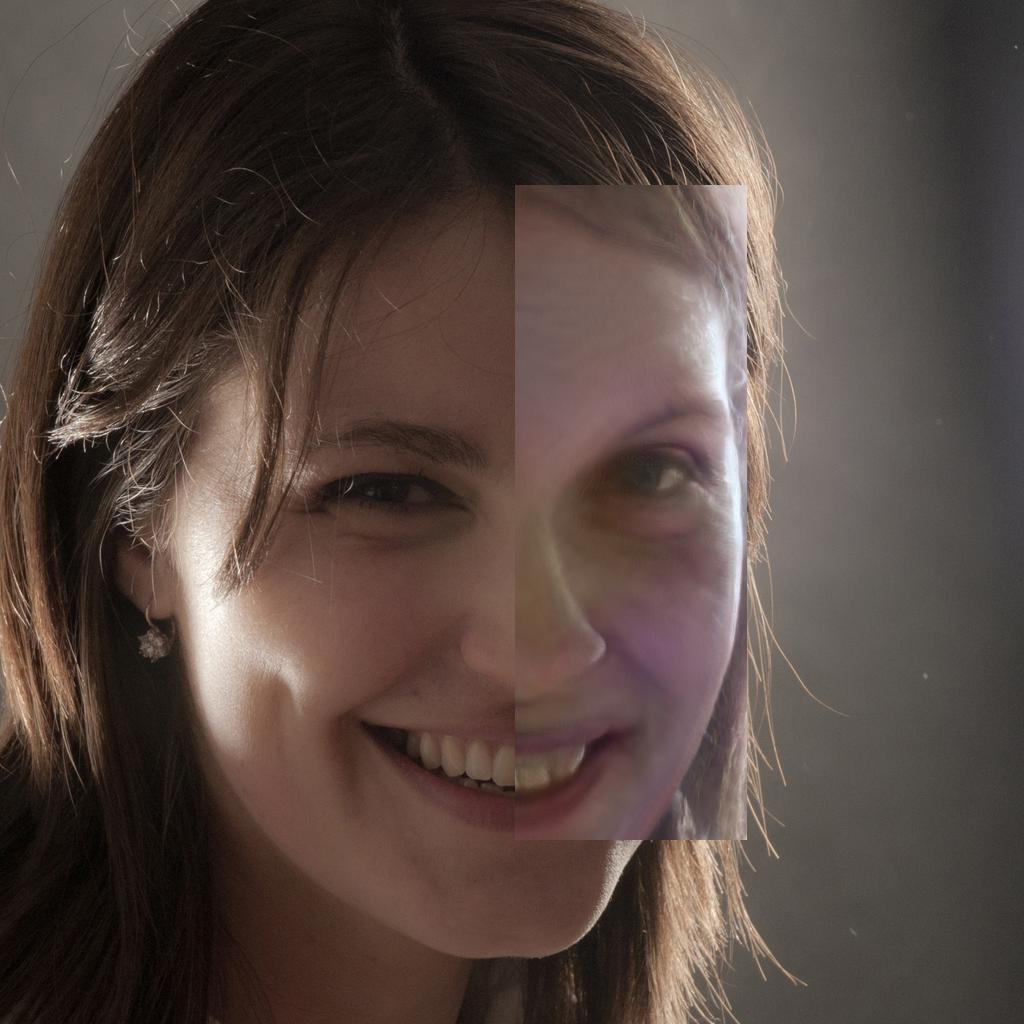} & \inc{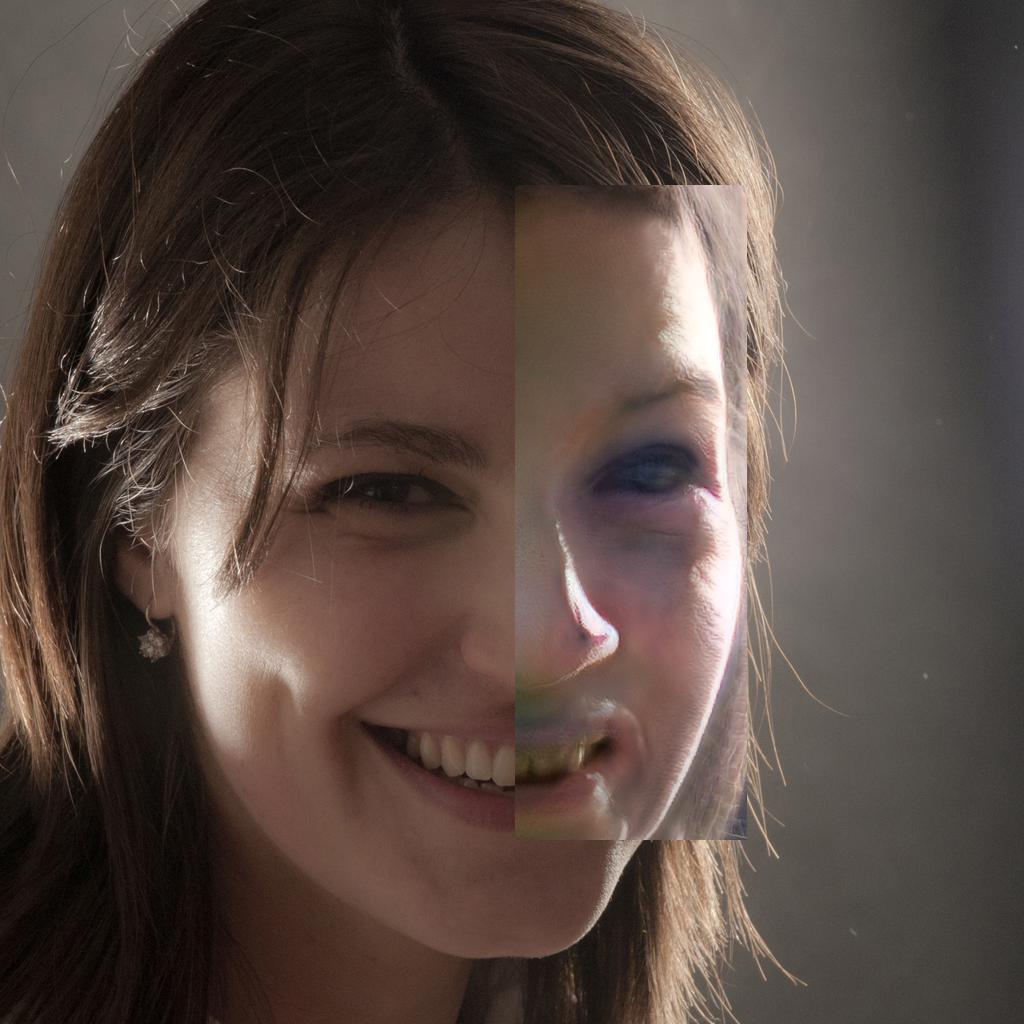} & \inc{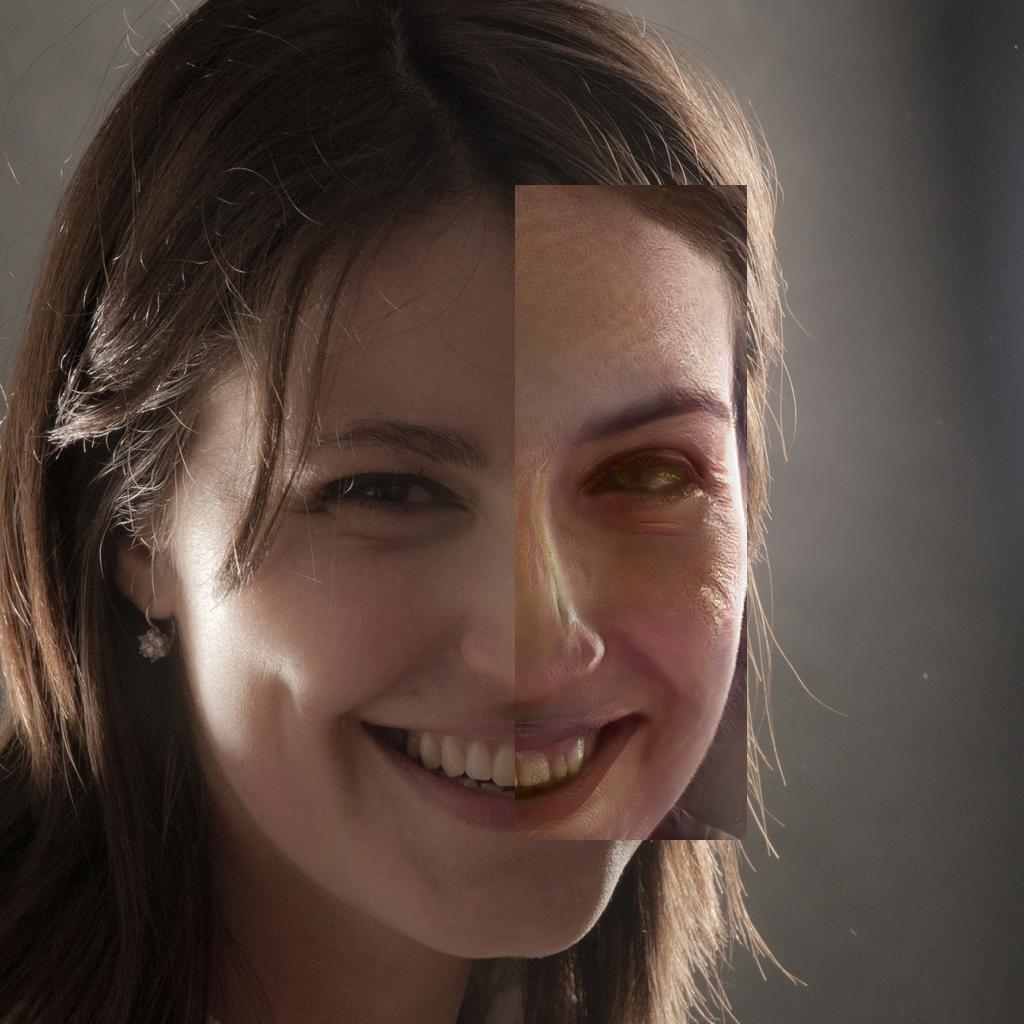} & \inc{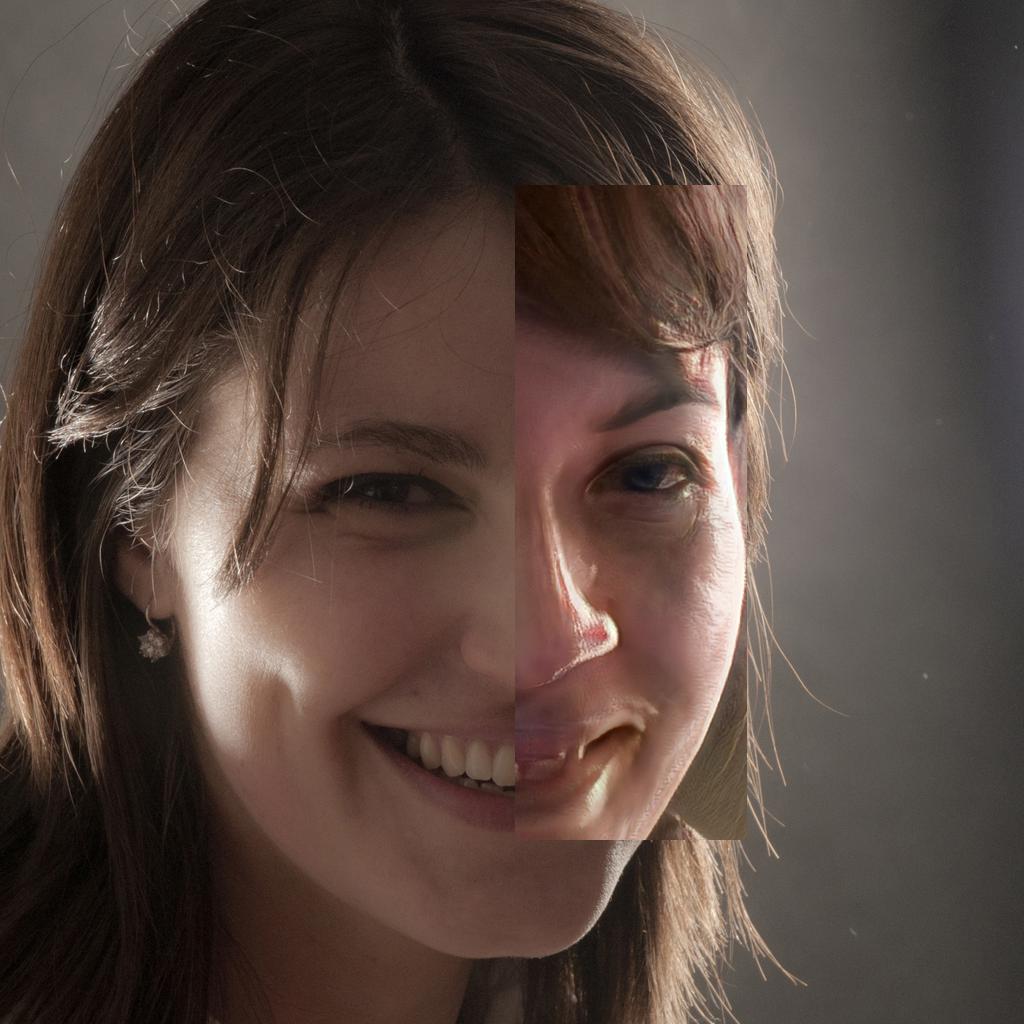}\\
 \end{tabular}

\renewcommand{\flds}{results/samXRAYinp}
\renewcommand{\is}{img00000664}
\renewcommand{\step}{step440}

\begin{tabular}{p{1.1cm}|c||c||ccccc}\arrayrulecolor{white}
% & True & Mean & Sample 1 & Sample 2 & Sample 3 & Sample 4 & Sample 5\\
Clean & \inc{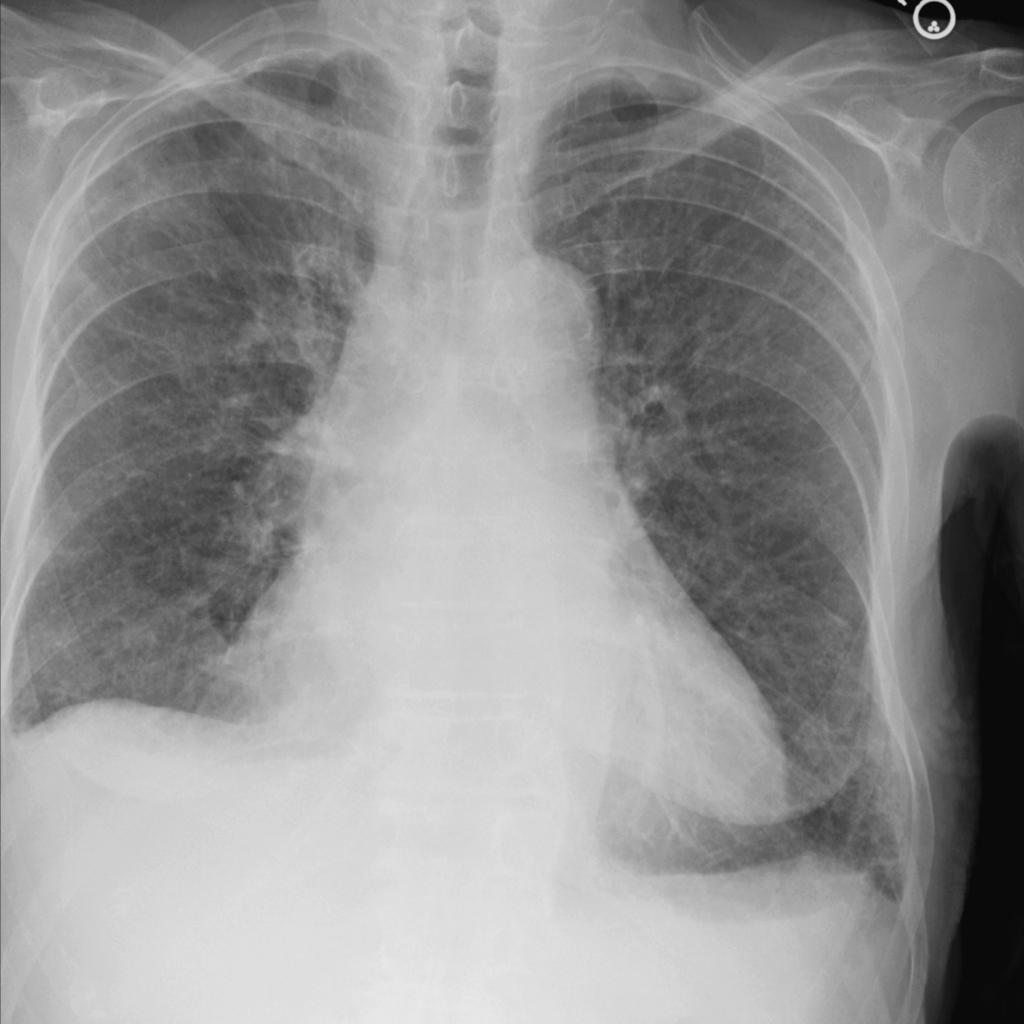} & \inc{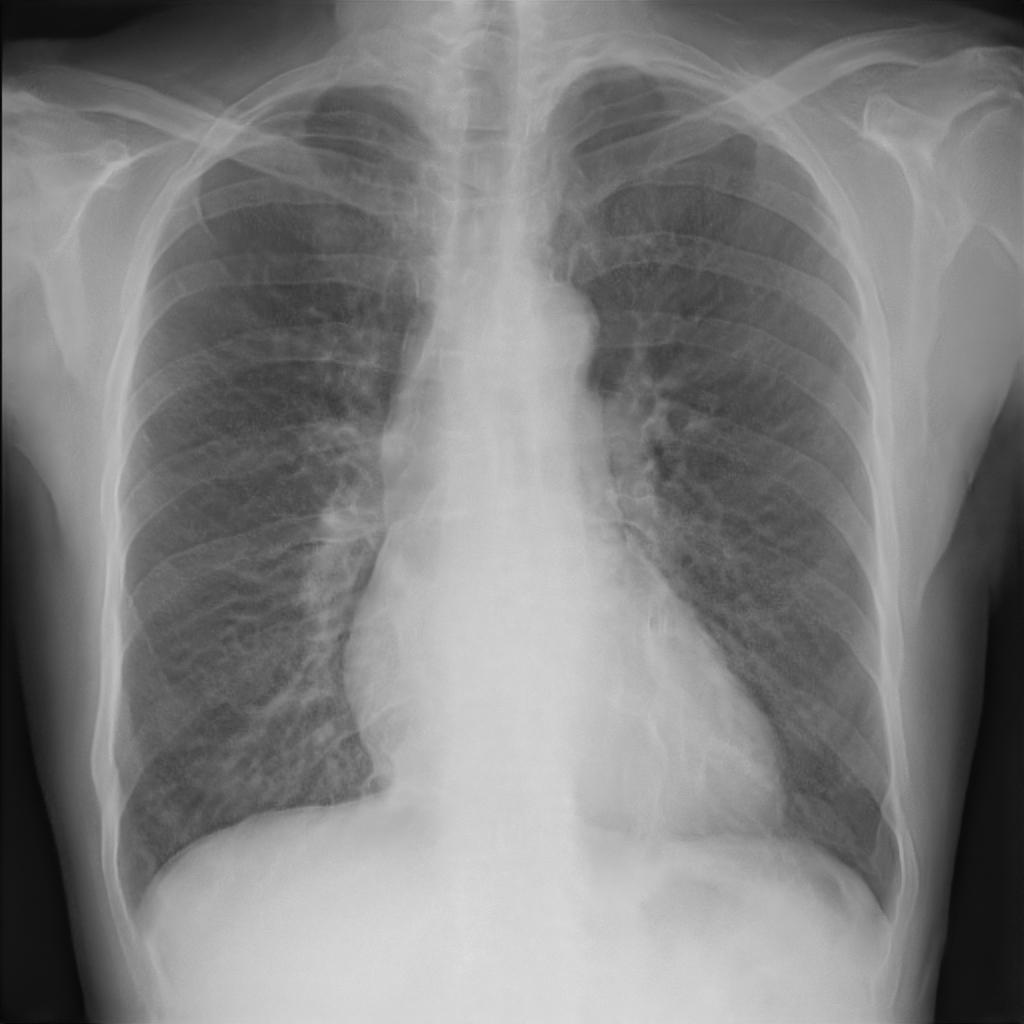} & \inc{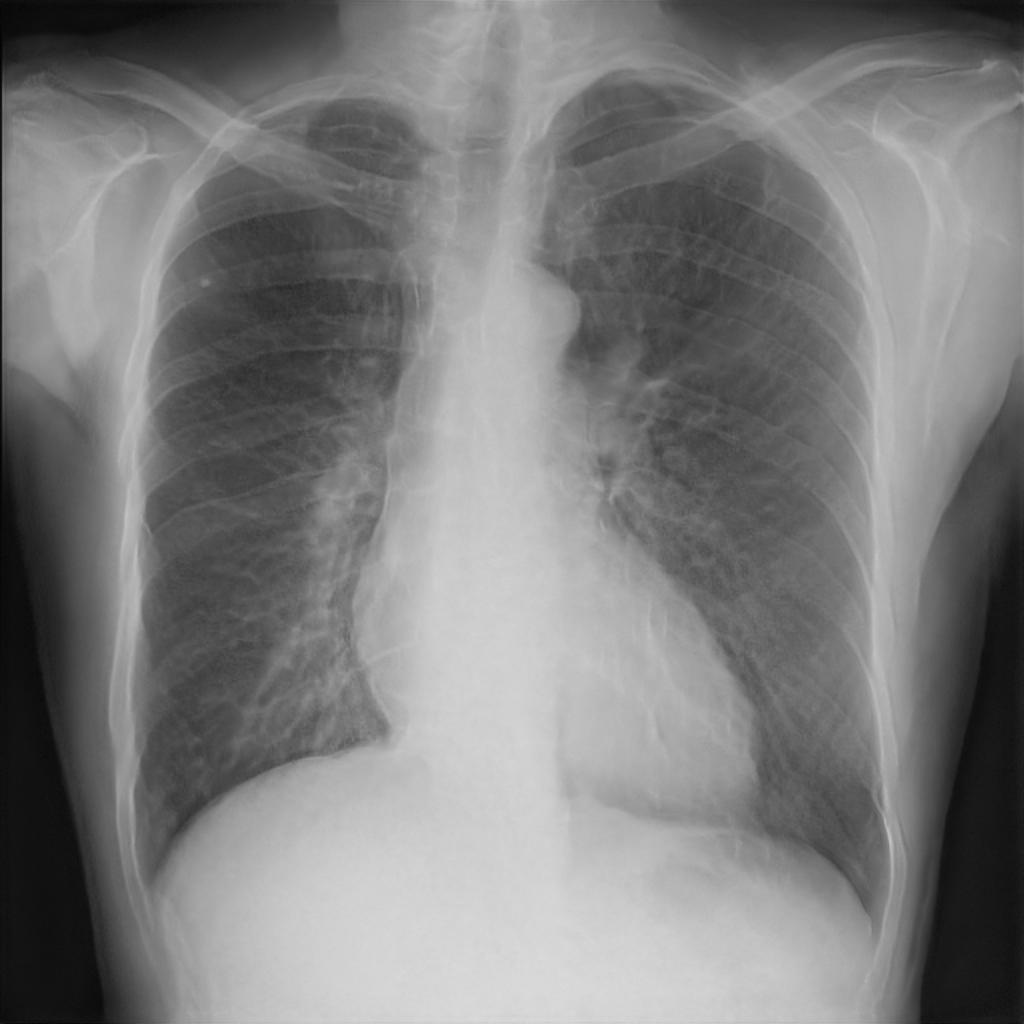} & \inc{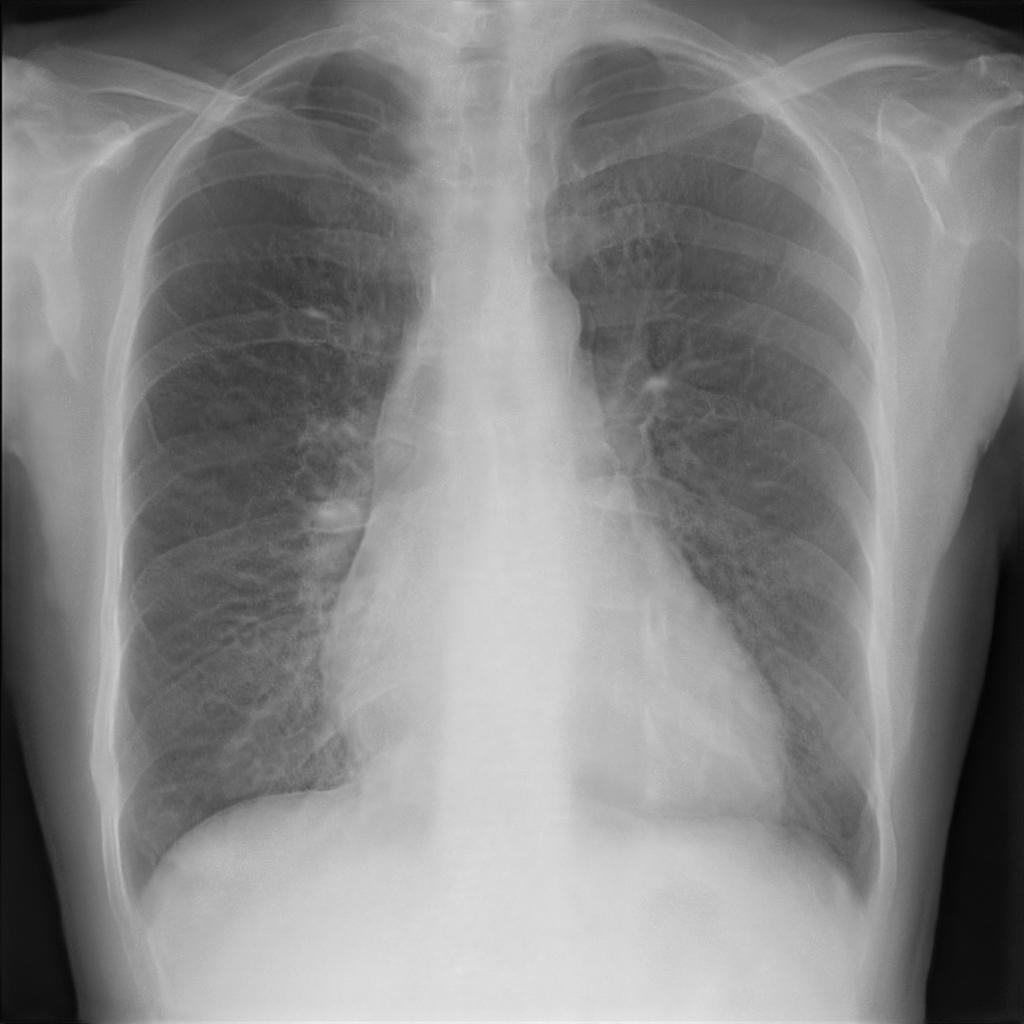} & \inc{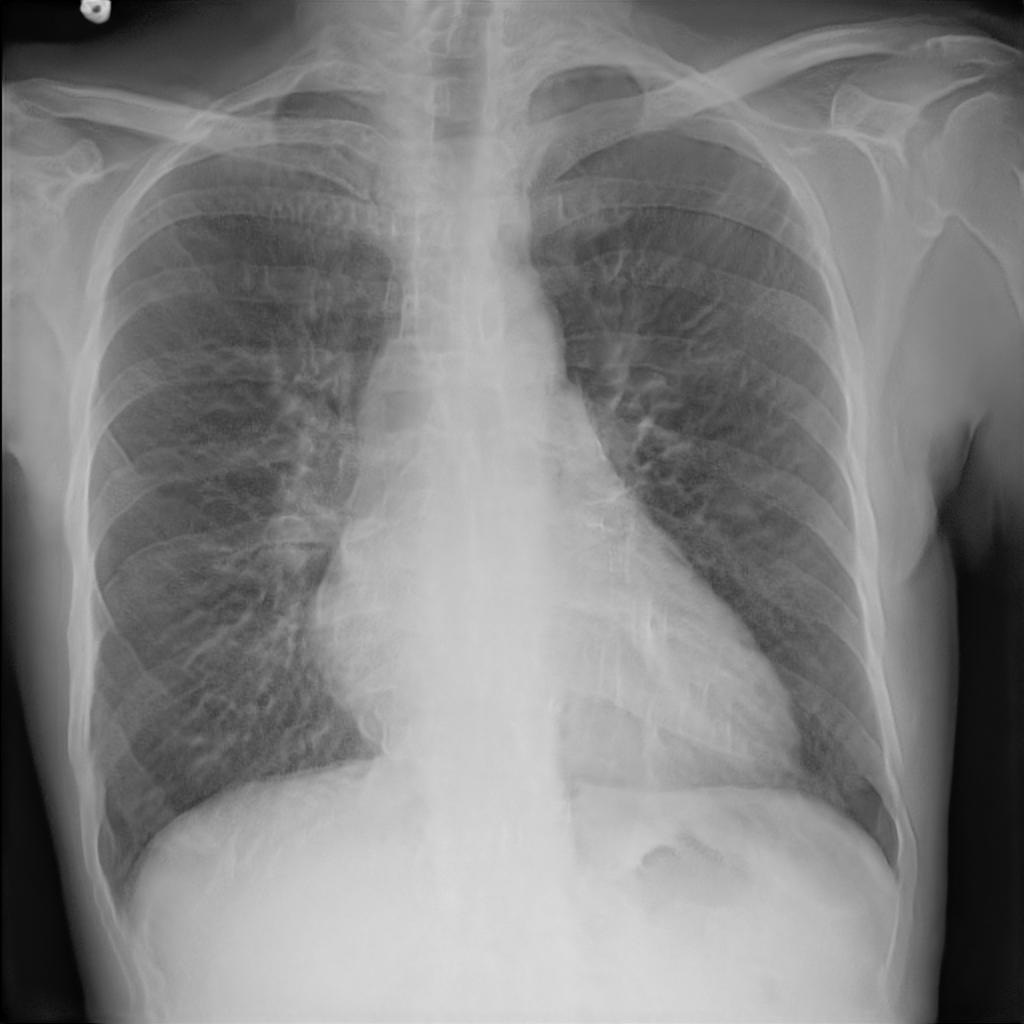} & \inc{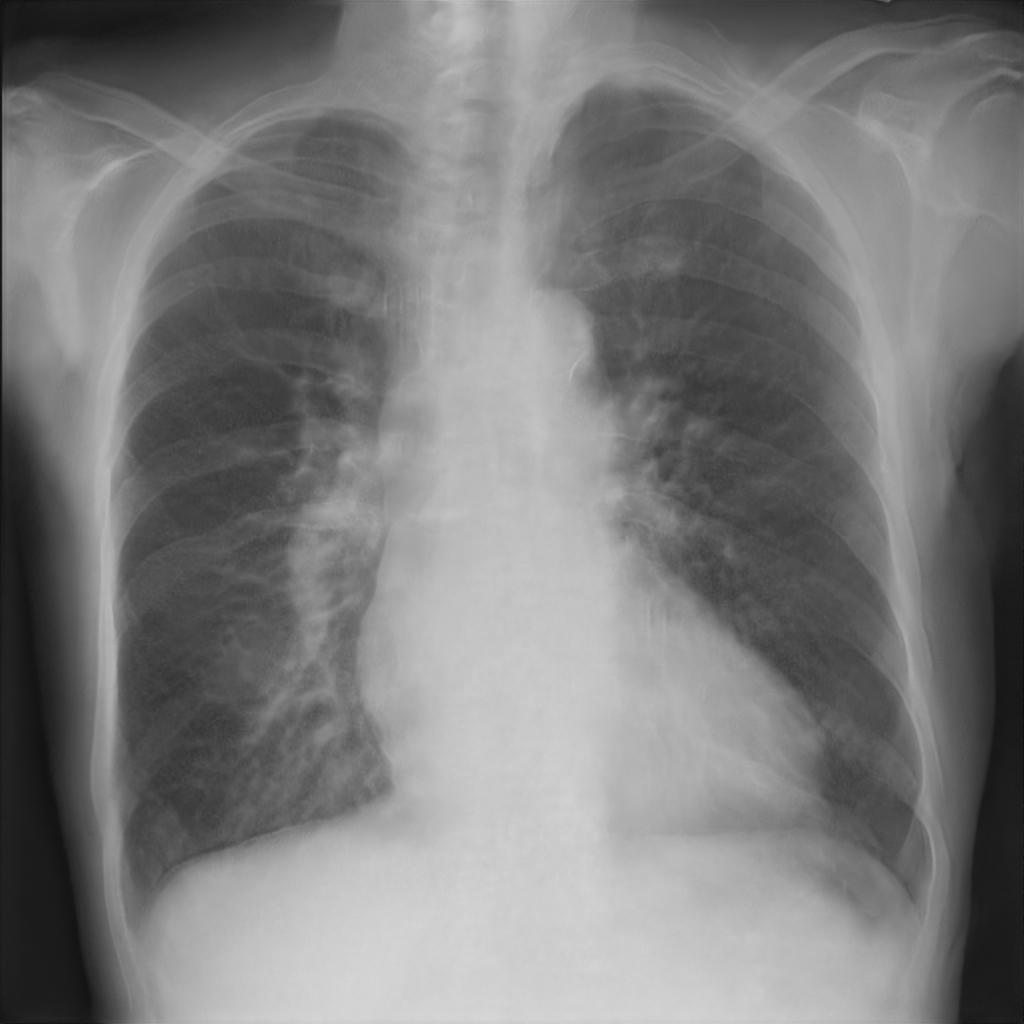} & \inc{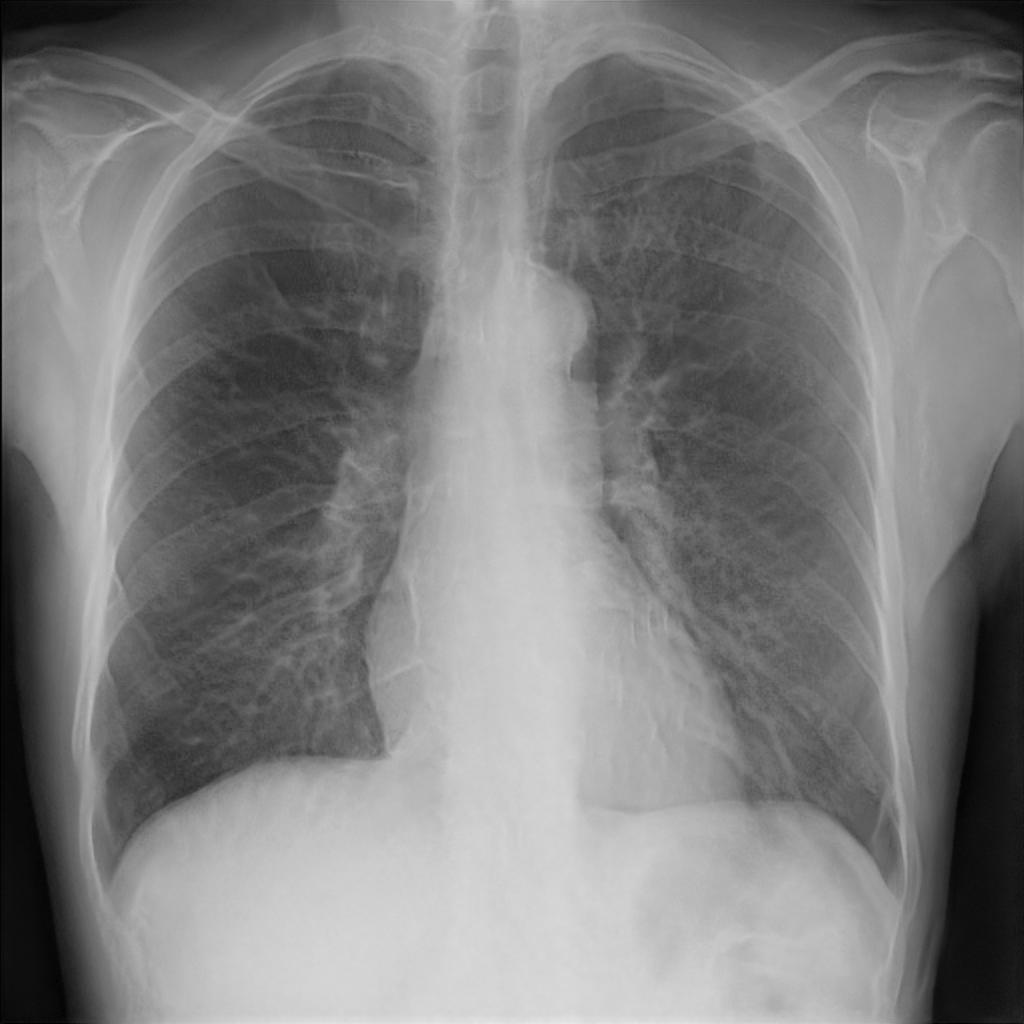}\\

Mask/ Merged & \inc{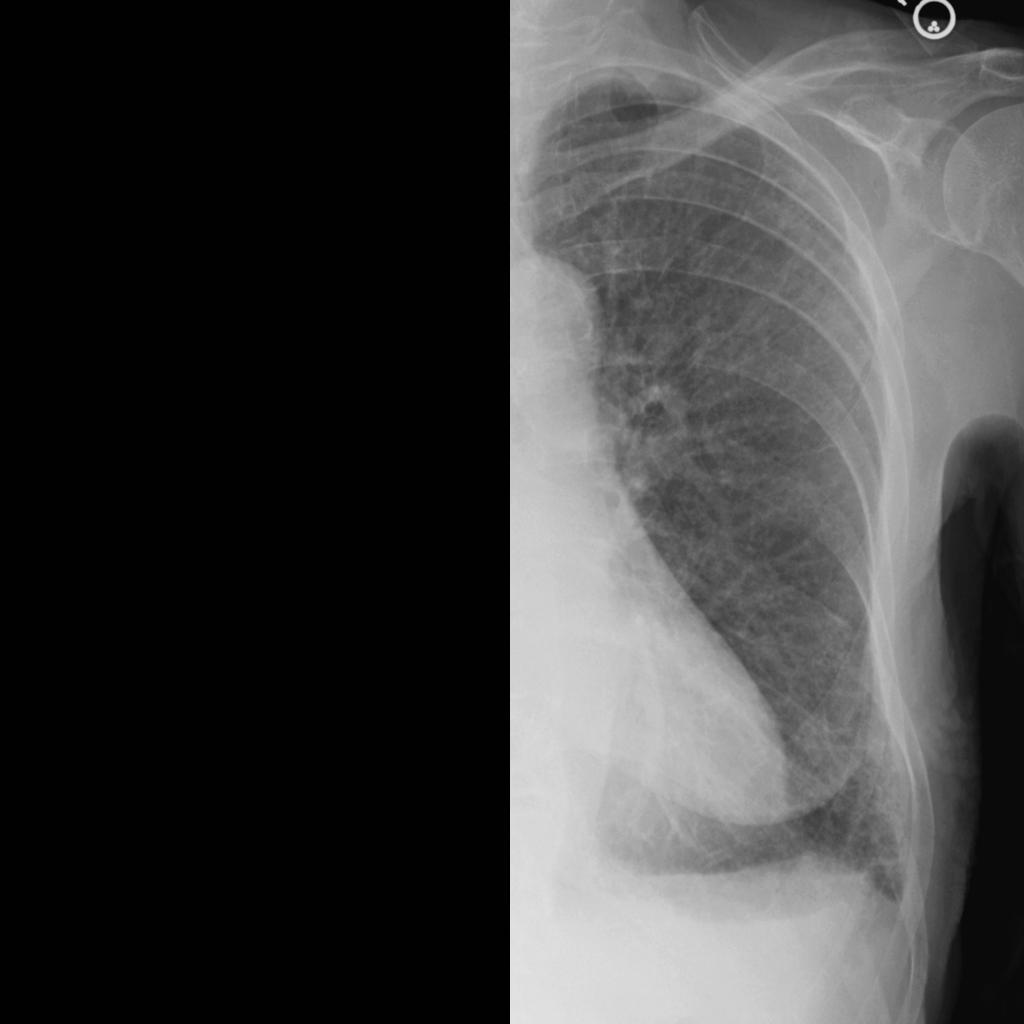} & \inc{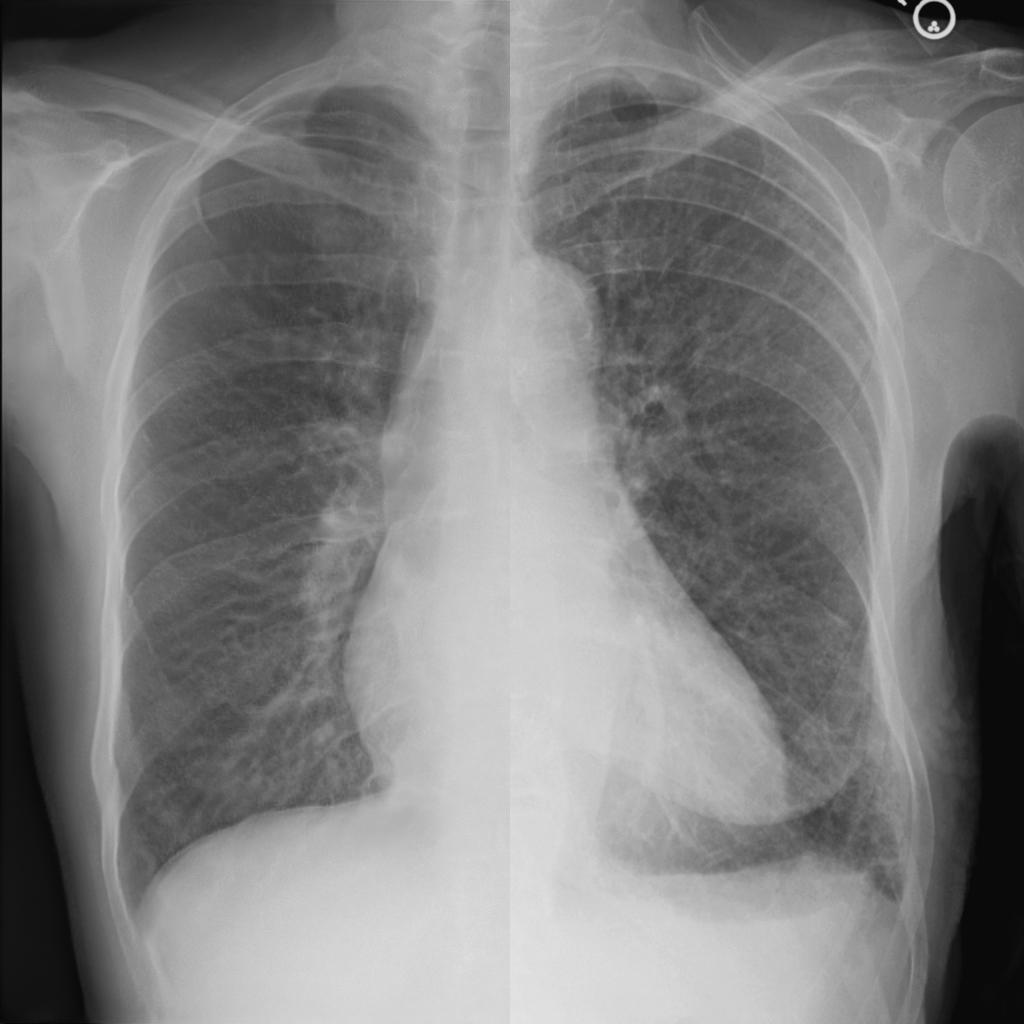} & \inc{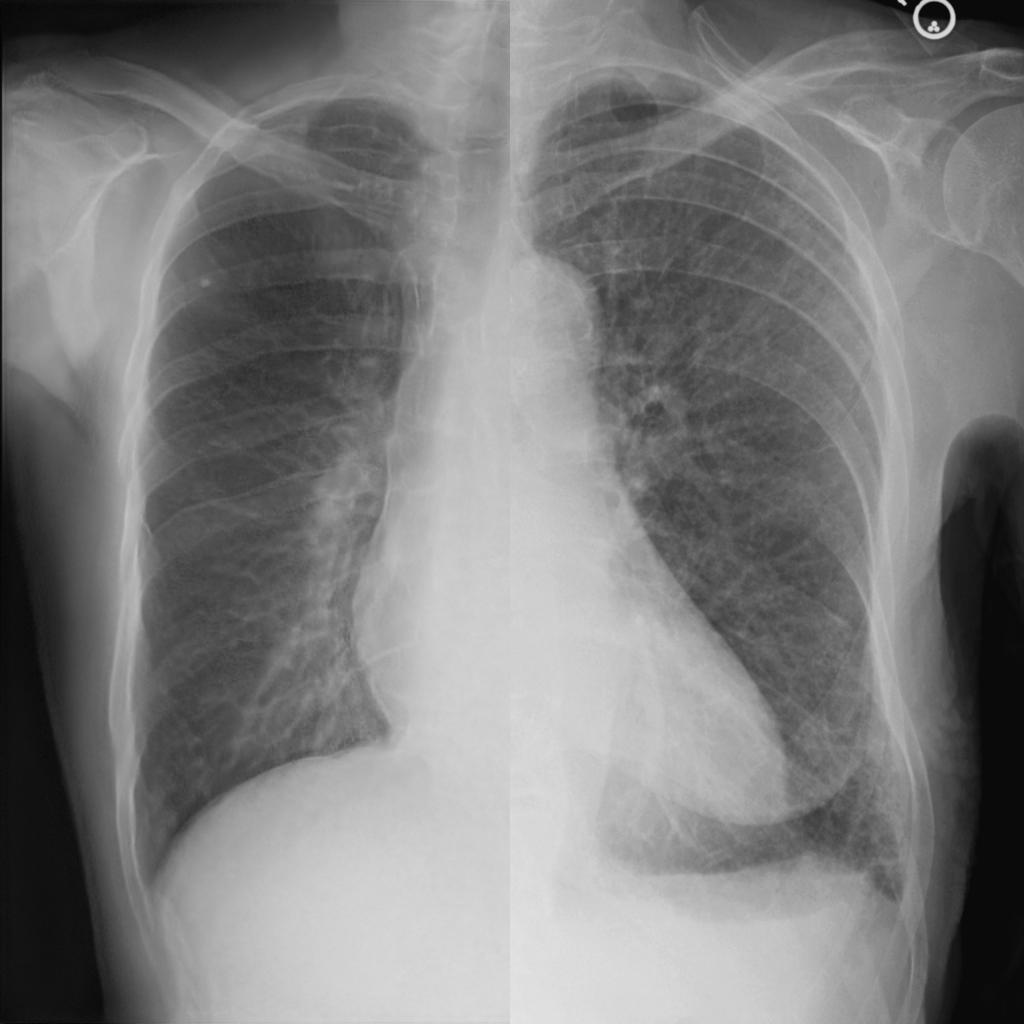} & \inc{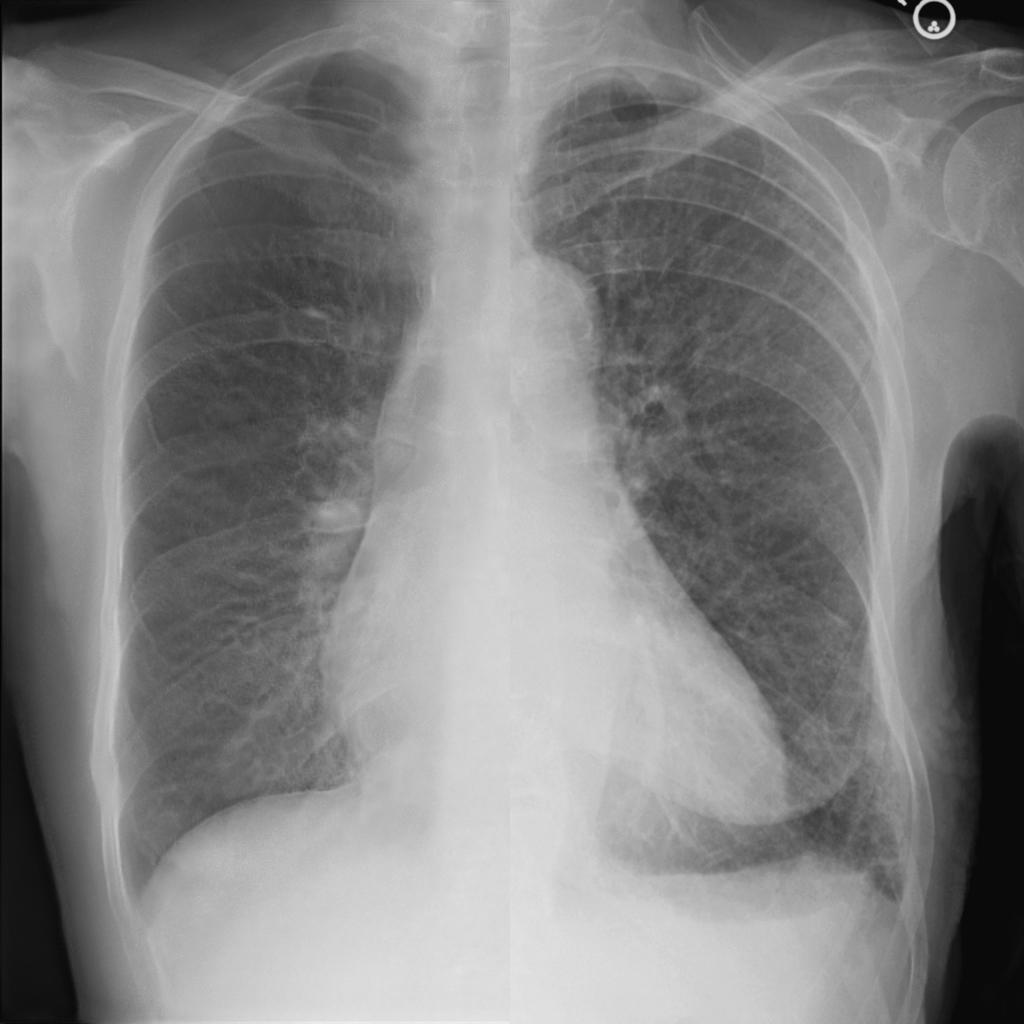} & \inc{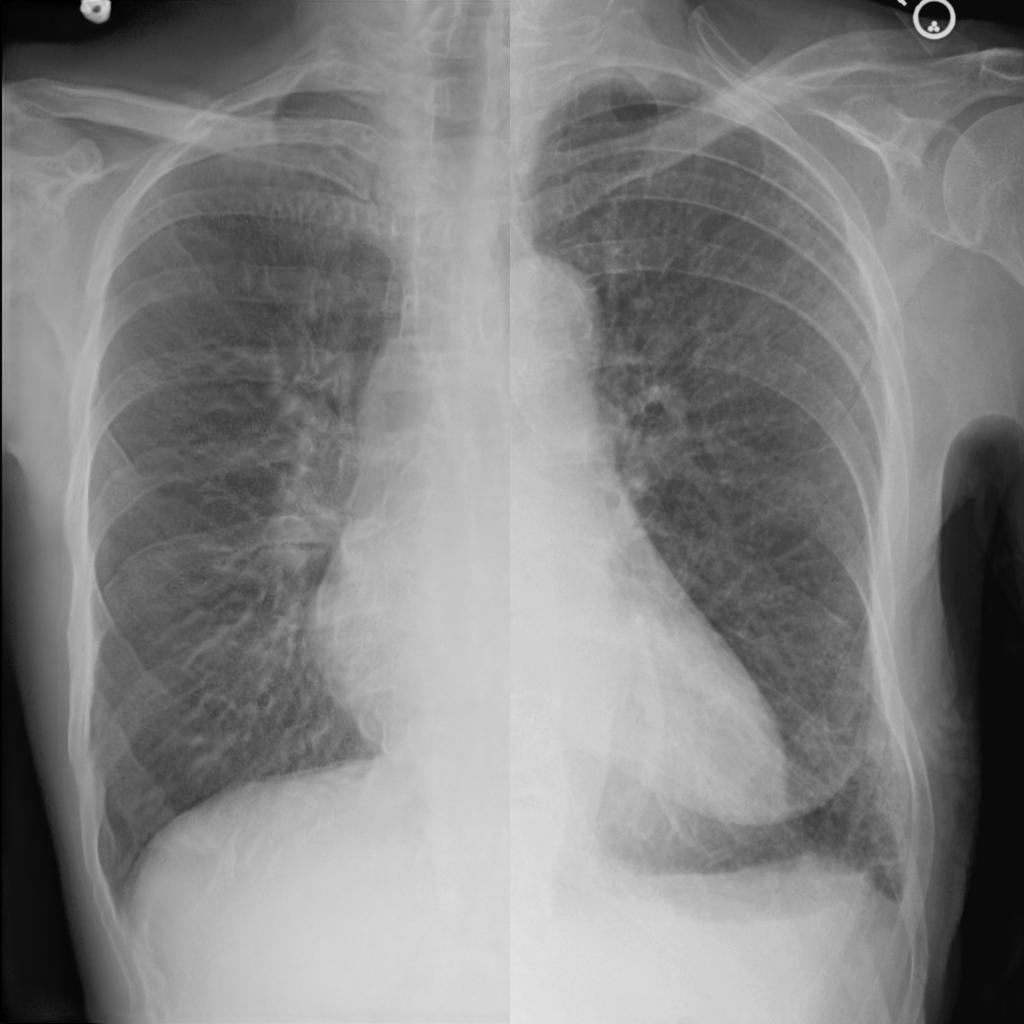} & \inc{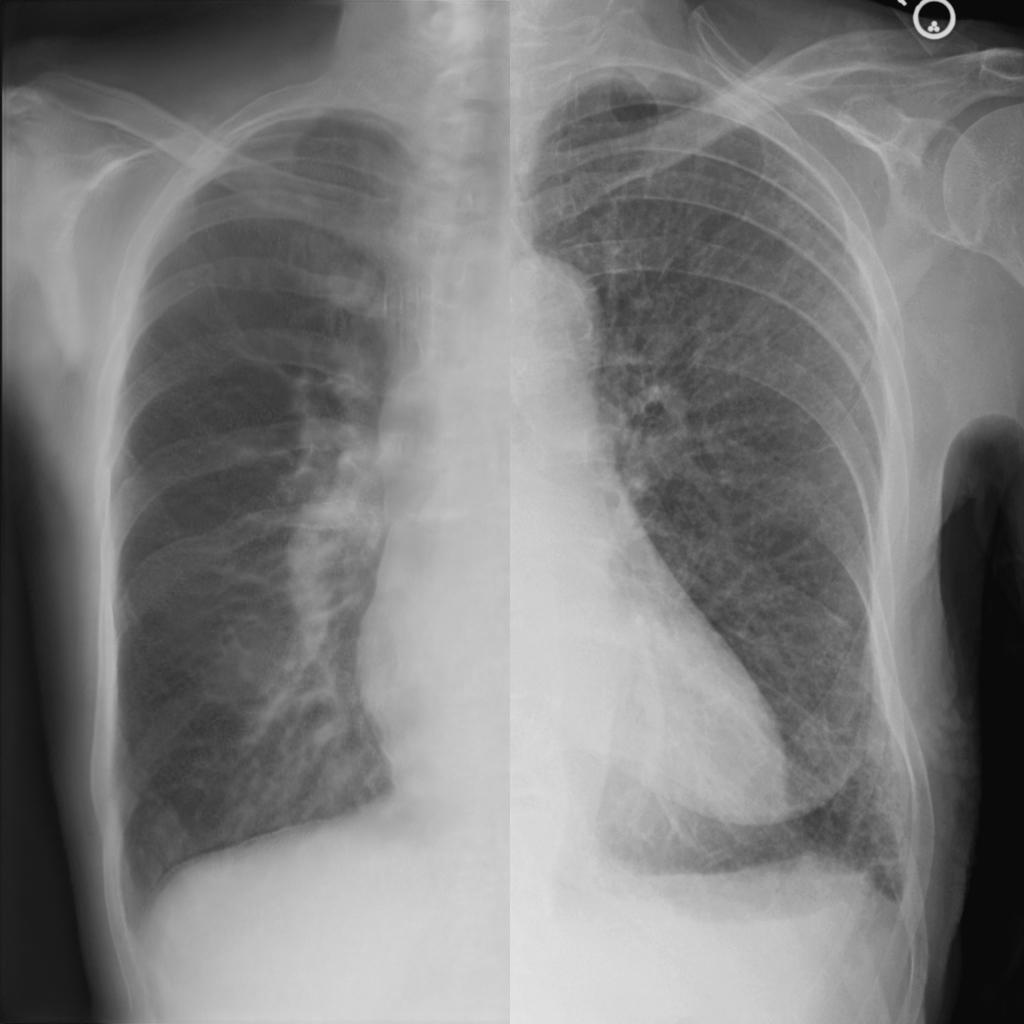} & \inc{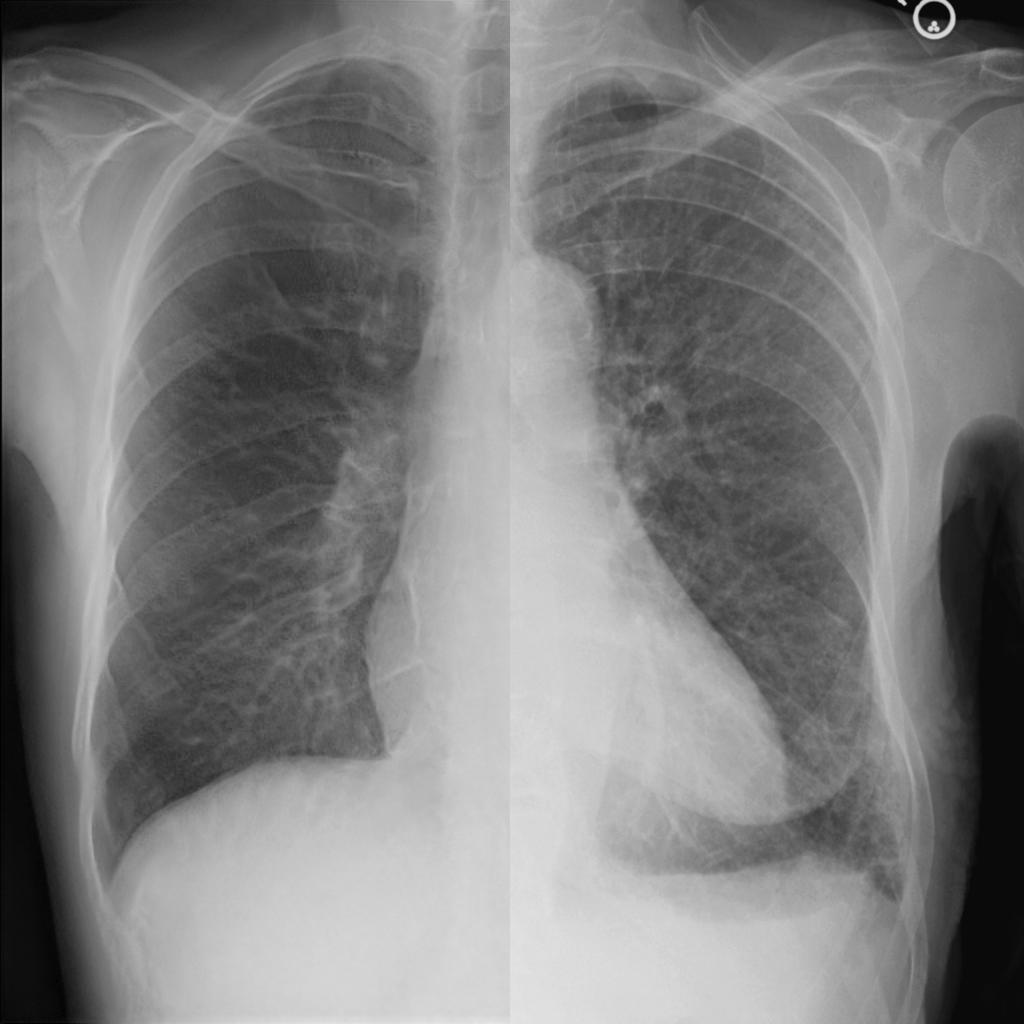}\\
 \end{tabular}

\renewcommand{\flds}{results/samBrainsInp}
\renewcommand{\is}{img00000034} %img00000099 img00000398
\renewcommand{\step}{step490}

\begin{tabular}{p{1.1cm}|c||c||ccccc}\arrayrulecolor{white}
% & True & Mean & Sample 1 & Sample 2 & Sample 3 & Sample 4 & Sample 5\\
Clean & \inc{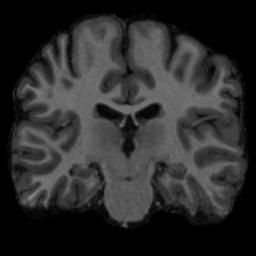} & \inc{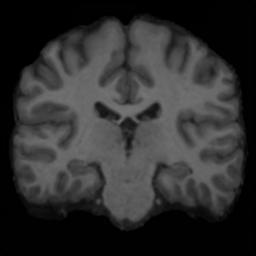} & \inc{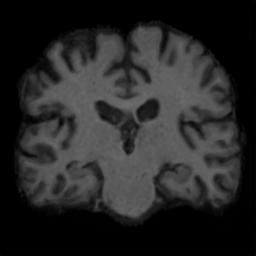} & \inc{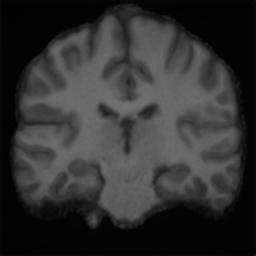} & \inc{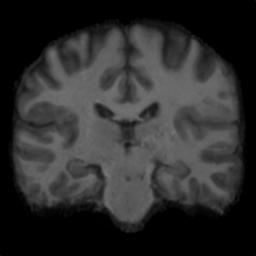} & \inc{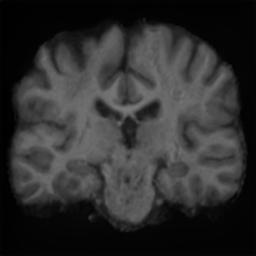} & \inc{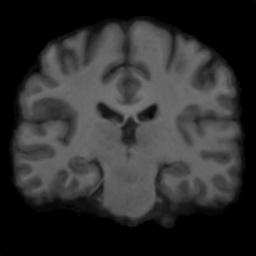}\\

Mask/ Merged & \inc{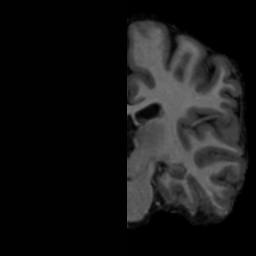} & \inc{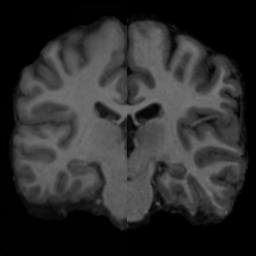} & \inc{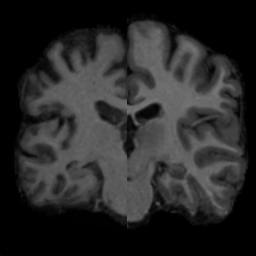} & \inc{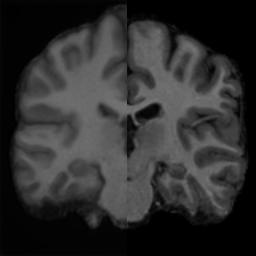} & \inc{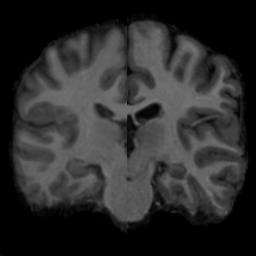} & \inc{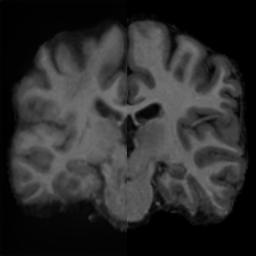} & \inc{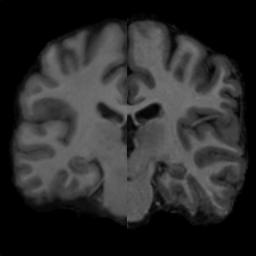}\\

 \end{tabular}
 \caption{Sampling of multiple reconstructions using Variational Inference on in-painting tasks. From left, we show the true image, the estimated variational mean, alongside five random samples around that mean. For the mean and for each sample, we show both the clean image, as well as the true image with the in-painted area from the sample.}
\label{samplinginpsupp}
\end{figure*}

\section{Sensitivity to hyper-parameters}
\label{ablation}

To understand how sensitive our model is to the hyper-parameters $\lambda_{pixel}$, $\lambda_{percept}$ and $\lambda_{colin}$, we performed in Table \ref{tab:ablation} an ablation analysis and computed the perceptual distance (LPIPS) and root mean squared error (RMSE) between our reconstructions and the true images. We performed this ablation on 64x super-resolution on 5 FFHQ test images, where we varied only one parameter at a time, keeping the others fixed to the following values: $\lambda_{pixel} = 10^{-5}$, $\lambda_{percept} = 10^5$ and $\lambda_{colin} = 10^{-2}$. The results show that our method is not overly-sensitive to the choice of the lambda hyper-parameters, as there are multiple values over which results are satisfactory: $\lambda_{pixel} \ge 10^{-6}$, $10^{5} \le \lambda_{percept} \le 10^6$, and $\lambda_{colin} \le 10^{-2}$.

\begin{table}
\centering
\renewcommand{\arraystretch}{1.5}% Spread rows out...
\begin{tabular}{c|ccccc}
\hline
$\lambda_{pixel}$ & $10^{-7}$ & $10^{-6}$ & $10^{-5}$ & $10^{-4}$ & $10^{-3}$\\
LPIPS/RMSE & 0.65/49.48 & 0.57/36.33 & 0.59/36.73 & 0.58/35.87 & 0.58/36.77\\
\hline
$\lambda_{percept}$ & $10^{3}$ & $10^{4}$ & $10^{5}$ & $10^{6}$ & $10^{7}$\\
LPIPS/RMSE & 0.60/36.42 & 0.59/35.19 & 0.57/34.67 & 0.56/33.56 & 0.60/38.86\\
\hline
$\lambda_{colin}$ & $10^{-4}$ & $10^{-3}$ & $10^{-2}$ & $10^{-1}$ & $10^{0}$\\
LPIPS/RMSE & 0.56/31.57 & 0.57/34.38 & 0.58/35.35 & 0.64/44.83 & 0.68/63.95\\
\hline
\end{tabular}
\vspace{0.25em}
\caption{Results of ablation study over hyper-parameters $\lambda_{pixel}$, $\lambda_{percept}$ and $\lambda_{colin}$. Reported are perceptual distance (LPIPS) and pixelwise root mean squared errors (RMSE) between the true images and the reconstructed images.}
\label{tab:ablation}
\end{table}

\section{Downstream processing of BRGM-restored images}
\label{downstream}

In order to show that the BRGM-reconstructed images are useful for downstream tasks, we ran a ResNet-16 that predicts lung-edema severity on BRGM-restored Chest X-Ray images. Edema severity was predicted as either (1) no edema, (2) mild -- vascular congestion, (3) moderate -- interstitial edema and (4) severe -- alveolar edema. The confusion matrix is shown in Table \ref{tab:downstream} (top table). We ran the model on 100 normal images (denoted as True), as well as the same 100 images but with the left-half masked and then reconstructed by our BRGM method (denoted as Reconstructed). We obtained a precision of 0.58 and a recall of 0.58. Most (73/100) of the severity scores are either preserved in the reconstructed images (diagonal), or change to an adjacent score (just above/below the diagonal). We note that, in general, the classification between adjacent severity levels (e.g. No edema vs Mild, Mild vs Moderate, etc ...) is a very hard problem, where the ResNet itself has an  score of 45.03\% (random guessing would give 25\% on a balanced dataset).

We also note that these results depend on how much of the image is reconstructed. If we re-run the analysis with a smaller mask (only a quadrant masked and reconstructed instead of half of the image -- see Table \ref{tab:downstream} bottom), we obtain significantly better results in the confusion matrix, and a precision of 0.75 and recall of 0.75. In particular, we note that the true severe cases improved their diagnosis.

\begin{table}
\centering
\renewcommand{\arraystretch}{1.5}% Spread rows out...
Mask 50\% of image (left half of image)\\
\begin{tabular}{c|ccccc}
True Reconstructed & No edema	& Mild & Moderate & Severe\\
\hline
No edema & 53 & 2 & 11 & 1 \\ 
Mild & 5 & 0 & 1 & 0 \\
Moderate & 9 & 3 & 5 & 0 \\
Severe & 3 & 3 & 4 & 0 \\
\end{tabular}
\vspace{2.5em}

Mask 25\% of image (top-right quadrant only)\\
\begin{tabular}{c|ccccc}
True Reconstructed & No edema	& Mild & Moderate & Severe\\
\hline
No edema & 66 & 2 & 8 & 1\\
Mild & 3 & 1 & 3 & 0\\
Moderate & 5 & 0 & 4 & 1\\
Severe & 0 & 1 & 1 & 4\\
\end{tabular}
\caption{Results of downstream processing on BRGM-reconstructed images vs true images. (Top table) We show the confusion matrix of a ResNet-16 at edema prediction on BRGM-inpainted images vs True images. For the inpainted images, the mask covered 50\% of the pixels (left-half). (Bottom table) As above, but only 25\% of image is masked and then reconstructed. }
\label{tab:downstream}
\end{table}

\section{Method Inconsistency}
\label{inconsistency}

One caveat of our method is that it can create reconstructions that are inconsistent with the input data. We highlight this in Figs. \ref{imgdiff-ffhq} and \ref{imgdiff-medical}. This is because our method relies on the ability of a pre-trained generator to generate any potential realistic image as input. In addition to that, in Eq. (\ref{fullloss}), our method optimizes the pixelwise and perceptual loss terms between the input image and the downsampled reconstruction. As we show in Figs. \ref{imgdiff-ffhq} and \ref{imgdiff-medical}, while there are little differences between the input and the reconstruction at low 16x16 resolutions, at higher 128x128 resolutions, these differences become larger and more noticeable. Another aspect that contributes to this issue is the extra prior term $\loss_{cosine}$, which is however required to ensure better reconstructions (see Fig \ref{evolution}). Nevertheless, we believe that the inconsistency is fundamentally caused by limitations of the generator $G$, that will be solved in the near future with better generator models that offer improved generalisability to unseen images.

% difference images

% 16x16
\newcommand{\diffFT}[1]{\fld/00607-recon-real-imagesffhq_test-super-resolution/image#1-target} % Target
\newcommand{\diffFO}[1]{\fld/00607-recon-real-imagesffhq_test-super-resolution/image#1-corrupted-step5000} % Ours
\newcommand{\diffFD}[1]{\fld/00607-recon-real-imagesffhq_test-super-resolution/image#1-diff} % Diff

\newcommand{\diffXT}[1]{\fld/00608-recon-real-imagesxray_frontal_test-super-resolution/image#1-target}
\newcommand{\diffXO}[1]{\fld/00608-recon-real-imagesxray_frontal_test-super-resolution/image#1-corrupted-step5000}
\newcommand{\diffXD}[1]{\fld/00608-recon-real-imagesxray_frontal_test-super-resolution/image#1-diff}

\newcommand{\diffBT}[1]{\fld/00626-recon-real-imagesbrains_test_mono-super-resolution/image#1-target}
\newcommand{\diffBO}[1]{\fld/00626-recon-real-imagesbrains_test_mono-super-resolution/image#1-corrupted-step5000}
\newcommand{\diffBD}[1]{\fld/00626-recon-real-imagesbrains_test_mono-super-resolution/image#1-diff}

% 32x32
\newcommand{\diffFTtwo}[1]{\fld/00620-recon-real-imagesffhq_test-super-resolution/image#1-target} % Target
\newcommand{\diffFOtwo}[1]{\fld/00620-recon-real-imagesffhq_test-super-resolution/image#1-corrupted-step5000} % Ours
\newcommand{\diffFDtwo}[1]{\fld/00620-recon-real-imagesffhq_test-super-resolution/image#1-diff} % Diff

\newcommand{\diffXTtwo}[1]{\fld/00622-recon-real-imagesxray_frontal_test-super-resolution/image#1-target}
\newcommand{\diffXOtwo}[1]{\fld/00622-recon-real-imagesxray_frontal_test-super-resolution/image#1-corrupted-step5000}
\newcommand{\diffXDtwo}[1]{\fld/00622-recon-real-imagesxray_frontal_test-super-resolution/image#1-diff}

\newcommand{\diffBTtwo}[1]{\fld/00604-recon-real-imagesbrains_test_mono-super-resolution/image#1-target}
\newcommand{\diffBOtwo}[1]{\fld/00604-recon-real-imagesbrains_test_mono-super-resolution/image#1-corrupted-step5000}
\newcommand{\diffBDtwo}[1]{\fld/00604-recon-real-imagesbrains_test_mono-super-resolution/image#1-diff}

% 64x64
\newcommand{\diffFTthree}[1]{\fld/00624-recon-real-imagesffhq_test-super-resolution/image#1-target} % Target
\newcommand{\diffFOthree}[1]{\fld/00624-recon-real-imagesffhq_test-super-resolution/image#1-corrupted-step5000} % Ours
\newcommand{\diffFDthree}[1]{\fld/00624-recon-real-imagesffhq_test-super-resolution/image#1-diff} % Diff

\newcommand{\diffXTthree}[1]{\fld/00625-recon-real-imagesxray_frontal_test-super-resolution/image#1-target}
\newcommand{\diffXOthree}[1]{\fld/00625-recon-real-imagesxray_frontal_test-super-resolution/image#1-corrupted-step5000}
\newcommand{\diffXDthree}[1]{\fld/00625-recon-real-imagesxray_frontal_test-super-resolution/image#1-diff}

% 128x128
\newcommand{\diffFTfour}[1]{\fld/00598-recon-real-imagesffhq_test-super-resolution/image#1-target} % Target
\newcommand{\diffFOfour}[1]{\fld/00598-recon-real-imagesffhq_test-super-resolution/image#1-corrupted-step5000} % Ours
\newcommand{\diffFDfour}[1]{\fld/00598-recon-real-imagesffhq_test-super-resolution/image#1-diff} % Diff

\newcommand{\diffXTfour}[1]{\fld/00601-recon-real-imagesxray_frontal_test-super-resolution/image#1-target}
\newcommand{\diffXOfour}[1]{\fld/00601-recon-real-imagesxray_frontal_test-super-resolution/image#1-corrupted-step5000}
\newcommand{\diffXDfour}[1]{\fld/00601-recon-real-imagesxray_frontal_test-super-resolution/image#1-diff}

\newcommand{\nrD}{0016}

\newcommand{\nrDtwo}{0017}
\newcommand{\nrDthree}{0000}
\newcommand{\nrDfour}{0000}

\renewcommand{\w}{3.2cm}

% results of standard StyleGAN training on X-ray and brains
\begin{figure*}
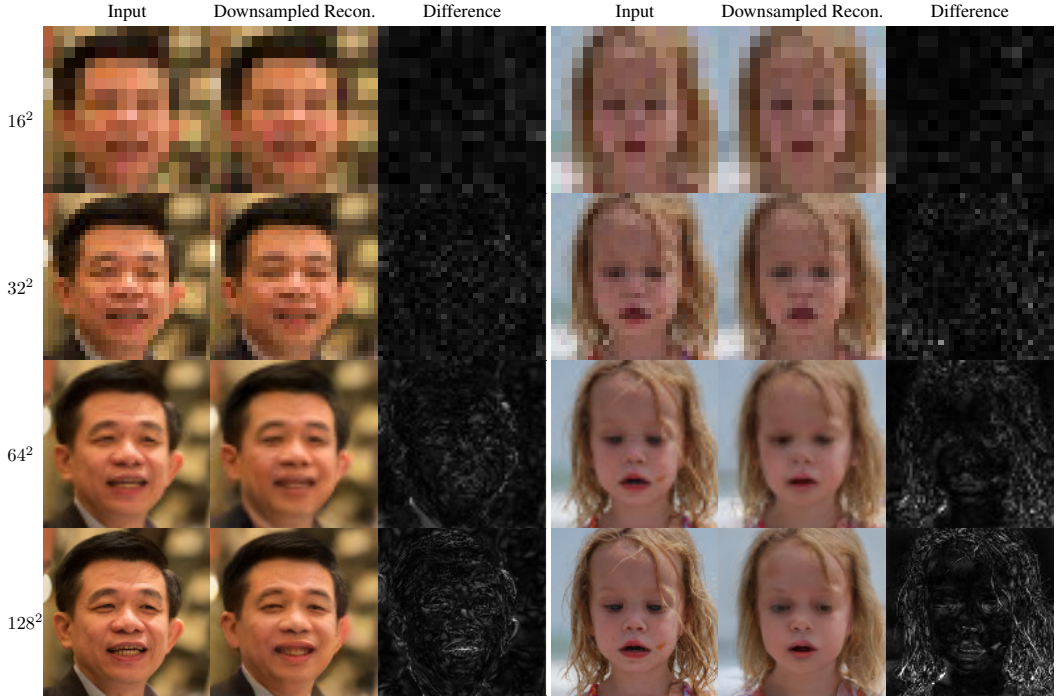

\centering
\setlength{\tabcolsep}{0pt}
\renewcommand{\arraystretch}{1.5}% Spread rows out...
\resizebox{\columnwidth}{!}{%
\begin{tabular}{lccc||ccc}\arrayrulecolor{white}
& Input & Downsampled Recon. & Difference & Input & Downsampled Recon. & Difference\\

$16^2$   & \inc{\diffFT{\nrD}} & \inc{\diffFO{\nrD}} & \inc{\diffFD{\nrD}} & \inc{\diffFT{\nrDtwo}} & \inc{\diffFO{\nrDtwo}} & \inc{\diffFD{\nrDtwo}}\\
$32^2$   & \inc{\diffFTtwo{\nrD}} & \inc{\diffFOtwo{\nrD}} & \inc{\diffFDtwo{\nrD}} & \inc{\diffFTtwo{\nrDtwo}} & \inc{\diffFOtwo{\nrDtwo}} & \inc{\diffFDtwo{\nrDtwo}}\\ 
$64^2$   & \inc{\diffFTthree{\nrD}} & \inc{\diffFOthree{\nrD}} & \inc{\diffFDthree{\nrD}} & \inc{\diffFTthree{\nrDtwo}} & \inc{\diffFOthree{\nrDtwo}} & \inc{\diffFDthree{\nrDtwo}}\\
$128^2$ & \inc{\diffFTfour{\nrD}} & \inc{\diffFOfour{\nrD}} & \inc{\diffFDfour{\nrD}} & \inc{\diffFTfour{\nrDtwo}} & \inc{\diffFOfour{\nrDtwo}} & \inc{\diffFDfour{\nrDtwo}}\\
\end{tabular}
}
\caption{Inconsistency of our method on FFHQ, across different resolution levels, using uncurated example pictures. Left columns shows input images, and the middle columns show downsampled reconstructions (i.e. $f \ci G (w^*)$) where images were 4x super-resolved, then downsampled by 4x to match again the input. Right columns show difference between input and the downsampled reconstructions. For higher resolution inputs (128x128), the method cannot accurately reconstruct the input image, likely because the generator has limited generalisability to such unseen faces from FFHQ (our method was trained not on the entire FFHQ, but on a training subset). The difference maps, representing x3 scaled mean absolute errors, show that certain regions in particular are not well reconstructed, such as the hair of the girl on the right. }
\label{imgdiff-ffhq}
\end{figure*}

\begin{figure*}
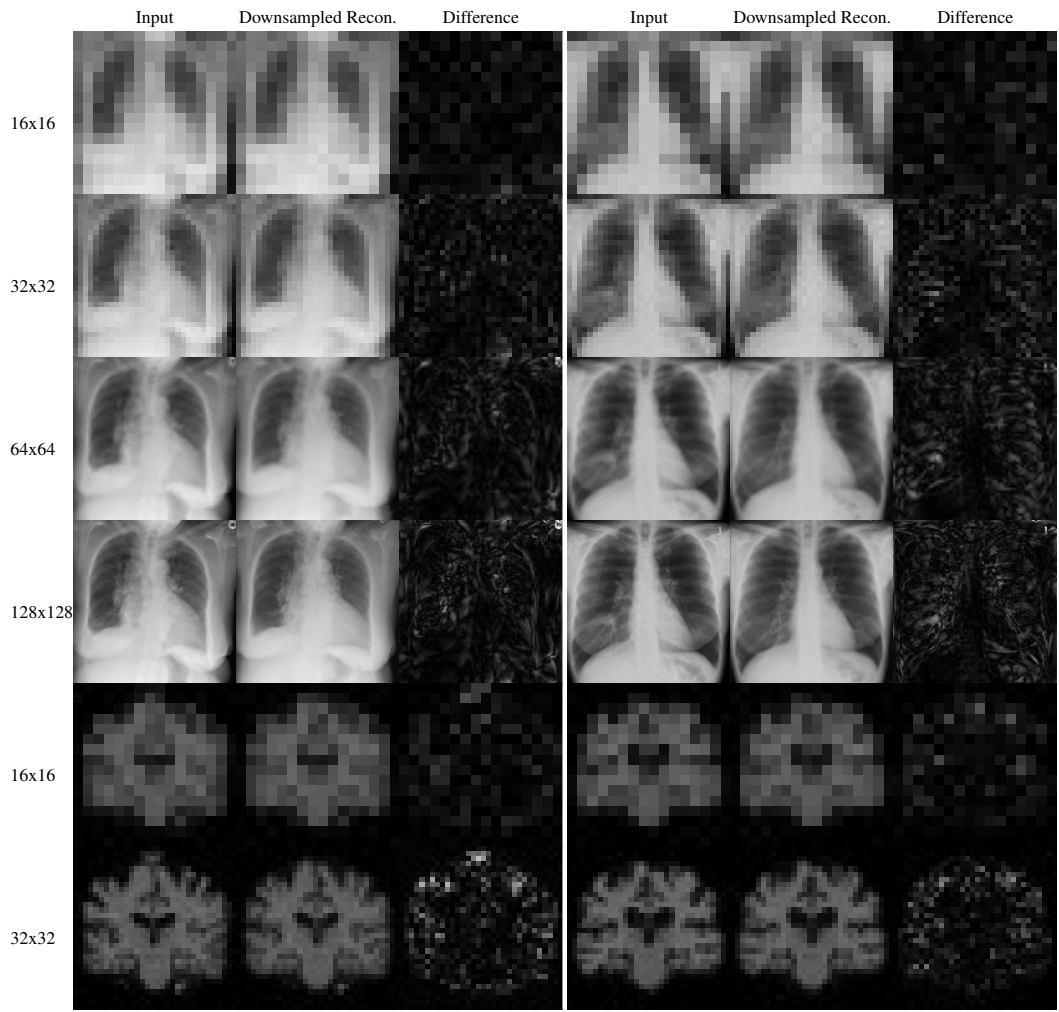

\centering
\setlength{\tabcolsep}{0pt}
\renewcommand{\arraystretch}{1.5}% Spread rows out...
\resizebox{\columnwidth}{!}{%
\begin{tabular}{lccc||ccc}\arrayrulecolor{white}
& Input & Downsampled Recon. & Difference & Input & Downsampled Recon. & Difference\\

16x16   & \inc{\diffXT{\nrD}} & \inc{\diffXO{\nrD}} & \inc{\diffXD{\nrD}} & \inc{\diffXT{\nrDtwo}} & \inc{\diffXO{\nrDtwo}} & \inc{\diffXD{\nrDtwo}}\\
32x32   & \inc{\diffXTtwo{\nrD}} & \inc{\diffXOtwo{\nrD}} & \inc{\diffXDtwo{\nrD}}& \inc{\diffXTtwo{\nrDtwo}} & \inc{\diffXOtwo{\nrDtwo}} & \inc{\diffXDtwo{\nrDtwo}}\\
64x64   & \inc{\diffXTthree{\nrD}} & \inc{\diffXOthree{\nrD}} & \inc{\diffXDthree{\nrD}} & \inc{\diffXTthree{\nrDtwo}} & \inc{\diffXOthree{\nrDtwo}} & \inc{\diffXDthree{\nrDtwo}}\\
128x128 & \inc{\diffXTfour{\nrD}} & \inc{\diffXOfour{\nrD}} & \inc{\diffXDfour{\nrD}} & \inc{\diffXTfour{\nrDtwo}} & \inc{\diffXOfour{\nrDtwo}} & \inc{\diffXDfour{\nrDtwo}}\\

16x16   & \inc{\diffBT{\nrD}} & \inc{\diffBO{\nrD}} & \inc{\diffBD{\nrD}} & \inc{\diffBT{\nrDtwo}} & \inc{\diffBO{\nrDtwo}} & \inc{\diffBD{\nrDtwo}}\\
32x32   & \inc{\diffBTtwo{\nrD}} & \inc{\diffBOtwo{\nrD}} & \inc{\diffBDtwo{\nrD}} & \inc{\diffBTtwo{\nrDtwo}} & \inc{\diffBOtwo{\nrDtwo}} & \inc{\diffBDtwo{\nrDtwo}}\\

\end{tabular}
}
\caption{Inconsistency of our method on the medical datasets, using uncurated examples. Same setup as in Fig. \ref{imgdiff-ffhq}. }
\label{imgdiff-medical}
\end{figure*}

\renewcommand{\fld}{results}

\newcommand{\nrFA}{0096}
\newcommand{\nrFAtwo}{0095}
\newcommand{\nrFAth}{0090}
\newcommand{\nrFAfo}{0093}
\newcommand{\nrFAfi}{0080}
\newcommand{\nrFAsi}{0082}

\newcommand{\fai}[1]{\fld/00620-recon-real-imagesffhq_test-super-resolution/image#1-target}
\newcommand{\fao}[1]{\fld/00620-recon-real-imagesffhq_test-super-resolution/image#1-clean-step5000}
\newcommand{\fat}[1]{\fld/00620-recon-real-imagesffhq_test-super-resolution/image#1-true}

% inconsistency difference in main paper
\newcommand{\diffFTmain}[1]{\fld/00598-recon-real-imagesffhq_test-super-resolution/image#1-target} % Target
\newcommand{\diffFOmain}[1]{\fld/00598-recon-real-imagesffhq_test-super-resolution/image#1-corrupted-step5000} % Ours
\newcommand{\diffFDmain}[1]{\fld/00598-recon-real-imagesffhq_test-super-resolution/image#1-diff} % Diff

\newcommand{\nrDmain}{0017}

\renewcommand{\w}{2.4cm}

\begin{figure}
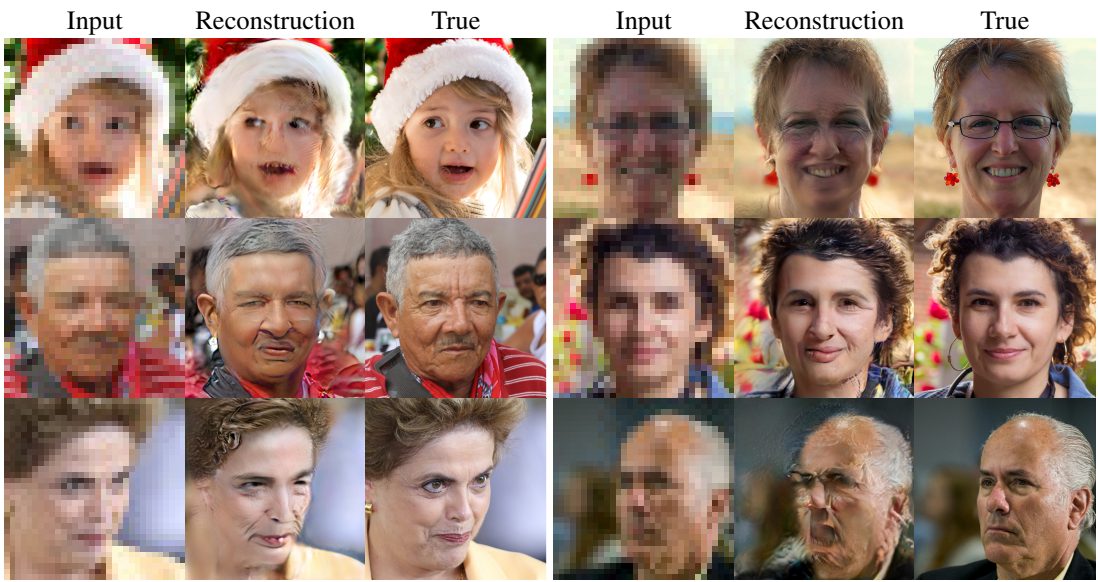

\centering
\setlength{\tabcolsep}{0pt}
%  \resizebox{\columnwidth}{!}{%
\begin{tabular}{ccc||ccc}\arrayrulecolor{white}
Input & Reconstruction & True & Input & Reconstruction & True\\
 \inc{\fai{\nrFA}} & \inc{\fao{\nrFA}} & \inc{\fat{\nrFA}} &\inc{\fai{\nrFAtwo}} & \inc{\fao{\nrFAtwo}} & \inc{\fat{\nrFAtwo}}\\
 
 \inc{\fai{\nrFAth}} & \inc{\fao{\nrFAth}} & \inc{\fat{\nrFAth}} & \inc{\fai{\nrFAfo}} & \inc{\fao{\nrFAfo}} & \inc{\fat{\nrFAfo}}\\
 
 \inc{\fai{\nrFAfi}} & \inc{\fao{\nrFAfi}} & \inc{\fat{\nrFAfi}} & \inc{\fai{\nrFAsi}} & \inc{\fao{\nrFAsi}} & \inc{\fat{\nrFAsi}}\\

\end{tabular}
% }
\caption{Failure cases of our method.}
\label{failure}
\end{figure}

% \section{Models Card}
% \label{modelscard}

\end{document}